\RequirePackage{filecontents}

\documentclass[preprint,12pt,authoryear]{elsarticle}

\usepackage{amssymb}
\usepackage{bm}
\usepackage{subcaption}
\usepackage{float}
\usepackage{xcolor}
\usepackage{soul}
\usepackage{hyperref}
\hypersetup{colorlinks,allcolors=black}

\newcommand{\bff}{\bm f}
\newcommand{\bg}{\bm g}

\newcommand{\bM}{\bm M}

\newcommand{\bn}{\bm n}

\newcommand{\bx}{\bm x}

\newcommand{\bu}{\bm u}

\newcommand{\bv}{\bm v}
\newcommand{\bV}{\bm V}

\newcommand{\btau}{\bm \tau}

\journal{Expert Systems with Applications}

\begin{document}

\begin{frontmatter}



\title{Forecasting through deep learning and modal decomposition in two-phase concentric jets}


\author[1]{Le{\'o}n Mata}
\ead{leonmatacervera@gmail.com}
\author[1]{Rodrigo Abad{\'i}a-Heredia\corref{cor1}}
\ead{sr.abadia@upm.es}
\author[1]{Manuel Lopez-Martin}
\ead{mlopezm@acm.org}
\author[1]{Jos{\'e} M. P{\'e}rez}
\ead{josemiguel.perez@upm.es}
\author[1]{Soledad Le Clainche}
\ead{soledad.leclainche@upm.es}

\cortext[cor1]{Corresponding author}

\affiliation[1]{organization={ETSI Aeronáutica y del Espacio, Universidad Politécnica de Madrid},
	addressline={Plaza Cardenal Cisneros, 3},
	city={Madrid},
	postcode={28040},
	state={Madrid},
	country={Spain}}
\date{\today}

\begin{abstract}
This work aims to improve fuel chamber injectors' performance in turbofan engines, thus implying improved performance and reduction of pollutants. This requires the development of models that allow real-time prediction and improvement of the fuel/air mixture. However, the work carried out to date involves using experimental data (complicated to measure) or the numerical resolution of the complete problem (computationally prohibitive). The latter involves the resolution of a system of partial differential equations (PDE). These problems make difficult to develop a real-time prediction tool. Therefore, in this work, we propose using machine learning in conjunction with (complementarily cheaper) single-phase flow numerical simulations in the presence of tangential discontinuities to estimate the mixing process in two-phase flows. In this meaning we study the application of two proposed neural network (NN) models\protect\footnote{The code of one model and two data sets are available at \href{https://github.com/RAbadiaH/forecasting-through-deep-learning-and-modal-decomposition-in-multi-phase-concentric-jets.git}{https://github.com/RAbadiaH/forecasting-through-deep-learning-and-modal-decomposition-in-multi-phase-concentric-jets.git}} as PDE surrogate models. Where the future dynamics is predicted by the NN, given some preliminary information. We show the low computational cost required by these models, both in their training and inference phases. We also show how NN training can be improved by reducing data complexity through a modal decomposition technique called higher order dynamic mode decomposition (HODMD), which identifies the main structures inside flow dynamics and reconstructs the original flow using only these main structures. This reconstruction has the same number of samples and spatial dimension as the original flow, but with a less complex dynamics and preserving its main features. The core idea of this work is to test the limits of applicability of deep learning models to data forecasting in complex fluid dynamics problems. Generalization capabilities of the models are demonstrated by using the same NN architectures to forecast the future dynamics of four different two-phase flows.
\end{abstract}


\end{frontmatter}





\section{Introduction} 

In turbofan engines, the fuel injectors inject the fuel into the combustion chamber, usually in a liquid phase, while the air is in the chamber at high pressures and temperatures. The injection must be correct regarding fuel quantity and injection time \citep{Baumgarten}, pp. 5-46, to ensure efficient fuel combustion. The injection consists of several steps. The first process that must occur is atomization. A common strategy is to inject the flow with rotation (swirly injection) instead of the case where it does not (axial injection). The flow rotation facilitates the atomization of the liquid (droplet formation) by hydrodynamic rupture (thanks to surface tension). If rotation is not introduced, the fuel is discharged as a jet into the combustion chamber, usually through a converging nozzle. In this case, the flow can be broken into droplets downstream (atomization) thanks to the effect of the surface tension (which depends on the Webber number) and the relative velocity between the fuel and the surrounding gas \citep{Baumgarten}. After droplet generation, processes of collision and coalescence of the droplets (coalescence) can occur. As the jet advances, it opens, and the evaporation phase begins, favored by the pressure drop, the increase in temperature, and the previous atomization. The vaporization process is of utmost importance since droplets that do not vaporize quickly have a high probability of not being burned during combustion, increasing pollutant emissions and reducing engine performance.

As can be seen, rupturing the separation surface between fuel and air (upstream) is important in the mixing process. At this point, combustion has not yet occurred. The direct way to obtain information on this process in real-time would be to take measurements through sensors or perform numerical simulations that consider both phases. This information would be beneficial when optimizing combustion. However, taking measures is complicated or even impossible in the first case, while the computational cost is very high in the second case. This work aims to optimize the mixture process (in real-time) by using simplified forecasting models. For this purpose, an analogy is made between the generation and growth of instabilities in shear layers in single-phase flows with the rupture of the interphase between two fluids; fuel and air.

 To this end, this work focuses on data-driven Reduced Order Models (ROMs) based on both deep learning and modal decomposition techniques. Where the single-phase flow simulations are the input of the neural networks, and the simulations obtained from the equivalent two-phase flow (with the same dimensionless parameters) are the output. Thus, the network can predict the mixing process before atomization without solving the equations corresponding to a two-phase flow, in literature this is known as a PDE surrogate model \citep{gupta2022}. There are other works where machine learning models have been used for industrial applications \citep{Momenitabar2022}. More specifically, in the field of fluid dynamics it can be found \citep{fukami_fukagata_taira_2019}, \citep{guo2016}. Where in the latter, a deep learning model has been proposed to develop a real-time application, this model estimated the drag of a vehicle from numerical simulations of simple geometries. In this work, we do something similar but applied in a very different context, we compare the performance of recurrent (RNN) and convolutional (CNN) deep learning models facing a complex two-phase flow forecasting problem, and also show how modal decomposition techniques like higher order dynamic mode decomposition (HODMD), used in some way to reduce data complexity, may be utilised to improve the training performance of CNN and RNN models. The limitations of both models are also studied. In particular, we observe how the CNN model provides good and stable performance using, for training, both raw and complexity-reduced data set (HODMD), while RNN model has difficulties on both raw and complexity-reduced data set. This could be because recurrent architectures have difficulties when dealing with high-dimensional time series as input (in this work we deal with a complexity problem, not a dimensionality one as in Ref. \citep{Abadia22}, where the RNN model obtains the best results), running into a bias problem by constantly predicting the mean of the input values. This bias problem cannot be explained by their reduced number of parameters, since RNN models have more than 4 times the number of weights than their corresponding convolutional models. The root of the problem must be sought in the lack of spatial inductive bias of RNN models \citep{LopezMartin21,cohen,mitchell}, which becomes critical as the complexity of the fluid dynamics forecasting problem increases.

The article is organized as follows. Section \ref{sec: model_description} describes the two-phase flow problem and the simulations performed, Section \ref{sec: methodology} describes the methodology used to construct both deep learning models (RNN and CNN) and the modal decomposition technique (HODMD), which will be used in combination with the two artificial neural networks. Section \ref{sec: simulation results} shows the results obtained from simulations, Section \ref{sec: results hodmd} shows the results obtained from HODMD and Section \ref{sec: results ann} compares the predictions obtained by the NNs. Finally, the main conclusions are presented in Section \ref{sec: conclusions}.

The main contributions of this work are:
\begin{itemize}
	\item[-] Show how deep learning models can be applied to reduce computing time of fluid simulations by predicting future dynamics.
	\item[-] Show how the combination of modal decomposition techniques, to reduce data complexity, with deep learning models can be used in complex flows, to improve the training performance of both NNs.
	\item[-] Compare the prediction performance of several deep learning architectures with and without data complexity reduction (modal decomposition) when applied to a complex fluid dynamics two-phase forecasting problem.
	\item[-] Explore the limits of deep learning forecasting models trying to predict the future behavior of complex high-dimensional flows.
\end{itemize}

\subsection{Flow instabilities in two liquid jets}

This work considers mixing two liquid jets, consisting of two incompressible, viscous, and immiscible fluids. Both jets arise from two nozzles separated by a gap, whose length can be zero. The limit case (with a gap length equal to zero) generates a mixing layer near the nozzle, in which different physical instability processes can arise.

This type of configuration appears recurrently in many engineering problems. For example, atomization (gas-liquid), see for example Ref. \citep{Lefebvre1989} and Ref. \citep{LasherasAndHopfinger2000}), in fuel injection systems (liquid-liquid), in mixing process in combustion chambers (gas-liquid or liquid-liquid jets) or in pipelines with oil-water mixtures. These processes increase the contact area between both phases (interface area), facilitating different physical processes involving both fluids, for example, combustion. Despite the great importance of the liquid-liquid case, it has not been studied as extensively in the literature as the gas-liquid case.

A relatively small density ratio characterizes the liquid-liquid case compared to the gas-liquid case. This implies that the effect of gravity is negligible (which is proportional to the difference in densities) compared to the effect of surface tension. Although the ratio between densities is usually low, this need not be the case for the ratio of viscosities.

Various mechanisms can destabilize mixing layers. Based on linear stability theory, these mechanisms can be viscous or non-viscous. For example, the Rayleigh-Taylor (RT) inviscid instability occurs when there is a density difference between the two fluids \citep{Sharp1984} while the non-viscous Kelvin-Helmholtz (KH) instability occurs when there is a difference in tangential velocities across the interface between both fluids (see \citep{HoytAndTaylor1977} and \citep{ChigierAndEroglu1989}). The latter instability can be generalized to the non-viscous case by considering the discontinuity in viscosity across the interface. This difference can introduce amplification mechanisms at the interface \citep{Yih1967,Hinch1984} and instabilities not directly related to the interface that develop in each phase separately \citep{YeckoZaleskiFullana2002}. These are viscous instabilities of Tollmien-Schlichting wave-type that occur in each fluid separately, above and below the discontinuity. Finally, the KH instability can be destabilized again by mechanisms such as Plateau-Rayleigh instability.  

Since both densities are very similar, the RT instability will be negligible in first approximation. This implies that the main instability mechanisms depend mainly on the surface tension and viscosity difference. Machine learning techniques allow modeling these types of effects using neural networks, based on the results obtained from direct numerical simulations of the flow under study.

\section{Model description and numerical simulations}\label{sec: model_description}
As mentioned above, multiphase flows are important in many industrial processes of high complexity, such as combustion processes. To model multiphase flows using numerical simulations, there are different approaches depending on their own nature. For example, in the study of particle dispersion in air, it is reasonable to use an Eulerian-Lagrangian model when particles' volume fraction is small, and the particles' size is large. In such cases, the air is solved using the Navier-Stokes equations (Eulerian model) and the particles using Newton's second law (Lagrangian model). Both equations are coupled by a force that models the fluid-particle and particle-particle interactions. However, when the volume fraction grows, the Lagrangian tracking of all particles can be computationally prohibitive. This problem can be solved for small particle sizes by applying the volume average theory \citep{DrewAndPassman}. This leads to an Eulerian-Eulerian model of two fluids, where the new densities are the fluid and particle densities weighted by the volume fractions of each phase.

The volume average model can be viewed as a particular case of ensemble average theory. The latter presents difficulties in closing the average equations, giving both models similar results in the range of validity of the volume averaging model.

In the case of two immiscible flows, the problem can be solved by using the volume of fluid method (VOF). This method was introduced in Ref. \citep{HirtAndNichols1981}, and is one of the most efficient methods for solving the interface between two incompressible and immiscible fluids. Typical problems studied with this method are: the raising of bubbles in vertical cylinders \citep{ChenEtAl1999}, bubble rupture \citep{LawsonEtAl1999}, rupture of droplets in the flow with shear \citep{LiAndRenardy2000}, and the formation of a bubble in the flow with shear \citep{LiAndRenardy2000}. In this method, the equations from the Eulerian-Eulerian model are combined to obtain a system of equations for the mixture; conservation of mass and momentum. Additionally, the interface's position is part of the solution to the problem, for which an additional equation is needed. In the case of two-fluid flows where an interface separates both phases, the VOF method is more appropriate than the two-fluid model when the spatial resolution is good enough, because in the latter empirical closures are required for the averaged equations.

\subsection{Navier-Stokes equations for single-phase flow}

The Navier-Stokes equations for an incompressible flow are given by,
\begin{equation}
	\nabla \cdot \bv = 0
\end{equation}

\begin{equation}
	\frac{\partial (\rho \bv)}{\partial t} + \nabla \cdot (\rho  \bv  \bv) = - \nabla \left( p \right)+ \nabla \cdot (\btau)
\end{equation}
\par\noindent
where $\rho$, $\bv$ and $p$ are the density, velocity and pressure fields. $\btau$ is the viscous stress tensor, which for an incompressible flow is given by $\btau = \frac{1}{2} \mu \left( \nabla \bv + \nabla \bv^T \right)$ being $\mu$ the dynamic viscosity. $p/\rho$ is also known as kinematic pressure.


\subsection{Navier-Stokes equations for two-phase flow}

The governing equations for a system of two incompressible viscous fluids is given by,
\begin{equation} \label{eq:massTwoFluids}
	\frac{\partial (\rho_k\alpha_k)}{\partial t} + 
	\nabla \cdot \left(\rho_k \alpha_k \bv_k \right) = 0
\end{equation}
and 
\begin{equation} \label{eq:momentumTwoFluids}
	\frac{\partial \left(\rho_k \alpha_k \bv_k \right) }{\partial t} +
	\nabla \cdot \left( \rho_k \alpha_k  \bv_k  \bv_k \right)  =
	- \nabla (\alpha_k p) + \nabla \cdot (\alpha_k \btau_k)
	+ \alpha_k \rho_k \bg + \bff_k,
\end{equation}
with $k=1$ for fluid 1 and $k=2$ for fluid 2. In the previous equations $\alpha_k$ is the volume fraction of phase $k$. In the case of $k=1$ this value is defined as the volume average of the indicator function, that is, as,
$$
\alpha_1 = \frac{1}{\delta V} \displaystyle\int_{\delta V} f(\bx,t) dV,
$$
where the indicator function is given by,
$$
f(\bx,t) = 
\left\{
\begin{array}{ll}
	1 & \hbox{ if } \bx  \hbox{ is occupied by fluid } 1 \hbox{ at time } t\\
	0 & \hbox{ if } \bx  \hbox{ is occupied by fluid } 2 \hbox{ at time } t
\end{array}
\right.
$$
$\btau_k$ is the viscous stress tensor of phase $k$, being its expression the same as in the single-phase flow but particularized for the fluid $k$, i.e. $\btau_k = \frac{1}{2} \mu_k \left( \nabla \bv_k + \nabla \bv_k^T \right)$. Given this definition, at any location of the computational domain it holds that $\alpha_1 + \alpha_2 = 1$.

In this model, $\bff_k$ represents the momentum transfer across the interface between the fluids. The sum of these two terms is balanced with the contribution due to the surface tension.

\subsection{Multiphase flow equations: VOF model}

The VOF method is based on the equations of the mixture and the evolution equation of the interface between the two fluids. Defining the mixture density and velocity as,
\begin{equation} \label{eq:rhoMix}
	\rho = \alpha_1 \rho_1 + \alpha_2 \rho_2
\end{equation}
\begin{equation} \label{eq:vMix}
	\bv = \frac{1}{\rho} 
	\left( \alpha_1 \rho_1 \bv_{1} + \alpha_2 \rho_2 \bv_2 \right)
\end{equation}
and adding eq. (\ref{eq:massTwoFluids}) for $k=1$ and 2, we obtain the mass conservation equation of the mixture,
\begin{equation} \label{eq:VOFcontinuidad}
	\frac{\partial \rho}{\partial t} + \nabla \cdot \left( \rho \bv \right) = 0.
\end{equation}
Remembering that both flows are incompressible and that $\alpha_1 + \alpha_2 = 1$, then
\begin{equation} \label{eq: VOFcontinuidad2}
	\nabla \cdot \bv = 0.
\end{equation}
Similarly, the momentum conservation equation of the mixing is given by,
\begin{equation} \label{eq:VOFECdM}
	\frac{\partial}{\partial t} ( \rho \bv) + \nabla \cdot ( \rho \bv \bv) = - \nabla p + \nabla \cdot \btau + \rho \bg + \bff_{\sigma},
\end{equation}
where
\begin{equation} \label{eq:pMix}
	p = \alpha_1 p_1 + \alpha_2 p_2,
\end{equation}
$\btau$ is the stress tensor of the mixture defined as follows,
$$
\btau = \frac{1}{2} \mu \left( \nabla \bv + \nabla \bv^T \right)
$$
and
\begin{equation}
	\mu = \alpha_1 \mu_{1} + \alpha_2 \mu_{2}
\end{equation}
is the mixture' dynamic viscosity.

As far as the momentum equation is concerned, in this study the gravitational force will be disregarded (because $\rho_1 \approx \rho_2$), so there will be no mass forces. The other force appearing in the equations is the surface tension force, which is of great importance in multiphase flow problems. This interaction is modeled as follows:
\begin{equation}
	\bff_{\sigma} = \sigma \kappa \nabla \alpha, 
\end{equation}
where $\kappa$ is the curvature radius of the surface interphase,
\begin{equation}
	\kappa = - \nabla \cdot \bn = - \frac{\partial}{\partial x_i} \left( \frac{\partial \alpha / \partial x_i}{ \| \partial \alpha / \partial x_i \|} \right),
\end{equation}
and $\sigma$ is the surface tension, which depends on the two fluids in contact. 
Note that this term is only non-zero in the interface of the two fluids, being zero in the interior of each fluid separately.

Therefore, to close the system of equations, it is necessary to obtain the evolution equation of the interface. Defining $\alpha=\alpha_1$, then,
\begin{equation} \label{eq: VOFtransporte}
	\frac{\partial \alpha}{\partial t} +  \nabla \cdot (\alpha \bv) = 0
\end{equation}
The numerical solutions were computed using OpenFOAM\citep{OpenFoam}, which uses the Finite Volume Method and the specific solvers were \textbf{interFoam} (multiphase) and \textbf{icoFoam} (single-phase).


\subsection{Geometry}
The problem under study consists of two concentric jets separated by a circular plate through which two different fluids are injected into a main chamber. It is in this chamber where the interaction between the two fluids takes place, resulting in a mixing process. The injection tubes as well as the main chamber have the shape of a cylinder. Given the nature of the problem, a simplified case will be considered in which it is assumed that the problem is axisymmetric, thus reducing the complexity of the problem.

With regard to geometry, two cases are to be considered. The first is called simple configuration, in which the injection tubes are cylinders of a given diameter. The second case is known as modified configuration and includes a bluff body, which is an obstacle placed at the end of an injector which causes atomization of the jet thus enhancing interaction between both phases. For specific flow conditions, this second case  has a wide range of industrial applications, the most common one being its use in fuel injectors to improve the mixing of combustion products and air, improving flame stability and allowing to control the combustion process \citep{tong2017experimental}.

\begin{figure}[H]
	\centering
	\begin{subfigure}[b]{0.495\textwidth}
		\centering
		\includegraphics[trim=35 14 15 10, clip,width=\textwidth]{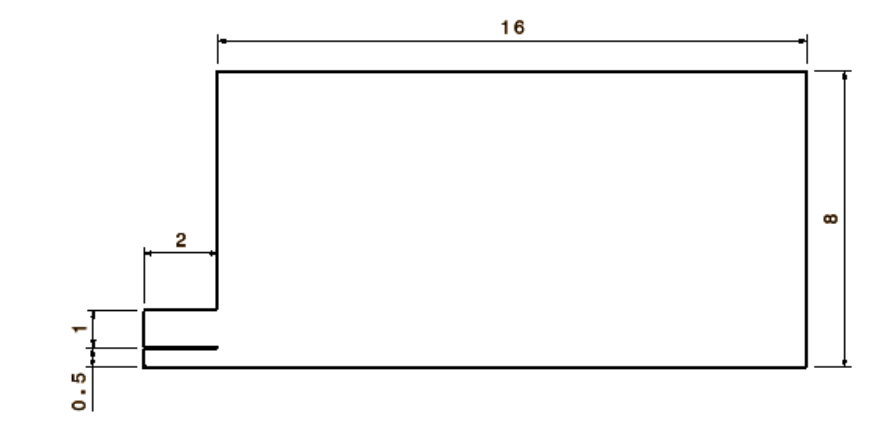}
		\label{fig:ampliacionmallalimpia}
	\end{subfigure}
	\hfill
	\begin{subfigure}[b]{0.495\textwidth}
		\centering
		\includegraphics[trim=35 14 15 10, clip,width=\textwidth]{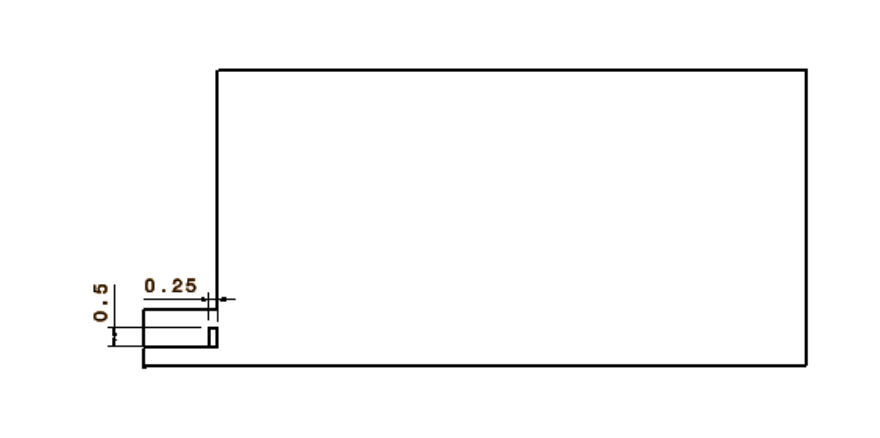}
		\label{fig:ampliacionmallamodificada}
	\end{subfigure}
	\caption{2D sections of the two cases of study: Simple configuration (left) and modified configuration (right). Dimensions are specified as multiples of a characteristic length h. Notice that the bottom horizontal line corresponds to the symmetry axis of the problem.}
	\label{fig: Geometría}
\end{figure}

Figure \ref{fig: Geometría} shows a 2D section of both geometries of the problem. The whole study developed in this paper is based on this geometry since, given the assumption that it is an axisymmetric problem, it is enough to solve for the flow in a plane containing the symmetry axis with any azimuth angle, rather than treating the whole 3D problem. Additionally, only half of the plane will be considered by adding the corresponding symmetry conditions over the axis. By doing so, the computational complexity will be extremely lower.

The present configuration (shape and dimensions) is extracted from  Ref. \citep{ling2019two}. As shown by these authors, it should be clarified that the separation between the two inlet tubes is small enough so that the separator plate does not have an impact on the type of flow instability (this is only true for the simple configuration, since the presence of the bluff body changes the flow in the modified geometry).  It is remarkable that the separation distance between jets can have a significant impact on the type of instability \citep{MartinetalSOCO20} depending on whether the value is larger or smaller than the vorticity layer thickness. By taking a sufficiently small separation, the results do not depend on the value of the separation \citep{ling2019two}.

A wedge-type structured mesh will be used, with a total of approximately 86000 cells. The cell size is highly dependent on the region of the geometry: it is critical to maintain a high resolution in the lower zone of the domain since there are important changes in the flow and it is necessary to capture correctly the interaction between both jets. In contrast, the upper zone of the main chamber has a large height to try to make the study independent of the boundary conditions, which typically occurs in reality due to the large difference in size between the mixing zone and the whole combustion chamber, so the flow near the injector is barely affected by the far field. As a consequence, there is little interest in the results in the upper zone, therefore a large cell size has been used. The different cell size allows, on the one hand, to obtain high resolution in the results by using small cell sizes in areas of interest; while, on the other hand, it reduces the computational cost by using large cell sizes in regions of lower relevance. A grid independence study was carried out based on the identification of the main patterns using the methodology presented in Section \ref{sec: hodmd}, showing that the presented configuration is optimal to identify the main patterns driving the main flow dynamics, which are suitable to develop a ROM using neural networks, as presented in the following sections.

\subsection{Initial conditions and boundary conditions} \label{Initial conditions and boundary conditions}
The equations require imposing initial and boundary conditions in order to solve the problem. On the one hand, the equations of the VOF model are solved using the following initial and boundary conditions:

\begin{itemize}
	\item \textbf{Volume fraction}: As an initial condition, the entire computational domain is filled with phase 2. Regarding the boundary conditions, phase 1 flows out of the lower inlet and phase 2 out of the upper inlet, in addition to imposing zero volume fraction gradient in all the other boundaries that define the geometry.
	\item \textbf{Pressure}: Initially, pressure is set equal to zero in the entire domain. As boundary conditions, pressure equals zero at both inlets and zero pressure gradient is imposed on all walls of the geometry, including the outlet.
	\item \textbf{Velocity}: Initial velocity is set equal to zero throughout the whole computational domain. Regarding the boundary conditions, unitary horizontal velocity is imposed on both inlets, slip condition on the upper wall of the main chamber and no slip on the rest of the boundaries except for the exit, where inlet-outlet condition is imposed.
\end{itemize}

On the other hand, the conditions imposed in the single-phase cases are:

\begin{itemize}
	\item \textbf{Kinematic pressure}: Same conditions as in the multiphase case but divided by the density.
	\item \textbf{Velocity}: Same as in the multiphase case except for the velocity at the lower inlet, which is set to 1/20 times the velocity at the upper inlet for reasons that will be explained in Section \ref{Fluid properties and dimensionless numbers}.
\end{itemize}

\subsection{Fluid properties and dimensionless numbers} \label{Fluid properties and dimensionless numbers}
The properties of the fluids used in this study are not intended to be those of any real fluid, but have been selected to obtain specific values of certain dimensionless numbers. Table \ref{tab:Propiedades fisicas} summarizes the values of the physical properties for the different multiphase cases. It should be noted that subscript 1 corresponds to fluid 1 and subscript 2 refers to fluid 2. 

\vspace{0.02\textwidth}

\begin{table}[H]   
	\centering
	\begin{tabular}{|c|c|c|c|c|} 
		\hline
		\textbf{$\rho_{1}$} & \textbf{$\rho_{2}$} & \textbf{$\nu_{1}$} &  \textbf{$\nu_{2}$} &  \textbf{$\sigma$} \\
		$(kg/m^3)$ & $(kg/m^3)$ & $(m^2/s)$ & $(m^2/s)$ & $(N/m)$ \\
		\hline
		1 & 1 & 1/30 & 1/600 & 0 or 1/80\\
		\hline
	\end{tabular}
	\caption{Physical properties of both fluids. Notice two different cases are considered regarding surface tension, since the numerical simulations will be done both with and without surface tension.}
	\label{tab:Propiedades fisicas}
\end{table}

Before understanding the choice of values for densities and kinematic viscosities it is important to define the two dimensionless numbers that govern the behavior of the problem, Reynolds number and Weber number as

\begin{equation}
	Re = \frac{\rho U h}{\mu},
\end{equation}

\begin{equation}
	We = \frac{\rho U^2 h}{\sigma},
\end{equation}

where $U$ and $h$ represent characteristic values for velocity ($U=1$, injection velocity) and length ($d$ = 1 , diameter of the inner jet).
Since two phases are present in the flow, there will be two different Reynolds numbers, whose values are $Re_1=30$ and $Re_2=600$, considering that the ratio between the two is equal to 20. Regarding the Weber number, its value will be $We=\infty$ when surface tension equals zero and $We=80$ otherwise. 
Considering the additional single-phase problem to be solved, the same ratio in the Reynolds numbers is used, which explains the selected values for inlet velocities.


Regarding the time evolution of the problem, the total simulation time is 500 time units, with a time step of $\Delta t = 0.005$. The simulation time must be large enough to eliminate the initial transient stage, and to generate enough data for the subsequent analysis.

Finally, Table \ref{tab: Nomenclatura} summarizes all the simulations carried out and the nomenclature used in the remaining of article for each case. Let us remember that simple and modified geometries refer to the two-concentric jet case without and with the presence of a bluff body, respectively.

Note that all the analysis presented are done in terms of dimensionless quantities, meaning that the results of a certain field are scaled by the corresponding characteristic value. As a result of this, any field computed from the equations will not have units and therefore will be dimensionless.
\begin{table}[H]   
	\centering
	\begin{tabular}{|c|c|c|c|} 
		\hline
		Geometry & Phases & Surface tension &  Nomenclature \\
		\hline
		Simple & Single-phase & - & S1 \\
		Simple & Multiphase & No & S2 \\
		Simple & Multiphase & Yes & S3 \\
		Modified & Single-phase & - & M1 \\
		Modified & Multiphase & No & M2 \\
		Modified & Multiphase & Yes & M3 \\
		\hline
	\end{tabular}
	\caption{Summary of all the computed cases, using DNS, and their nomenclature, which will be used in further sections to refer to them.}
	\label{tab: Nomenclatura}
\end{table}


\section{Methodology} \label{sec: methodology}
\begin{figure}[H]
	\centering
	\includegraphics[height=0.45\columnwidth]{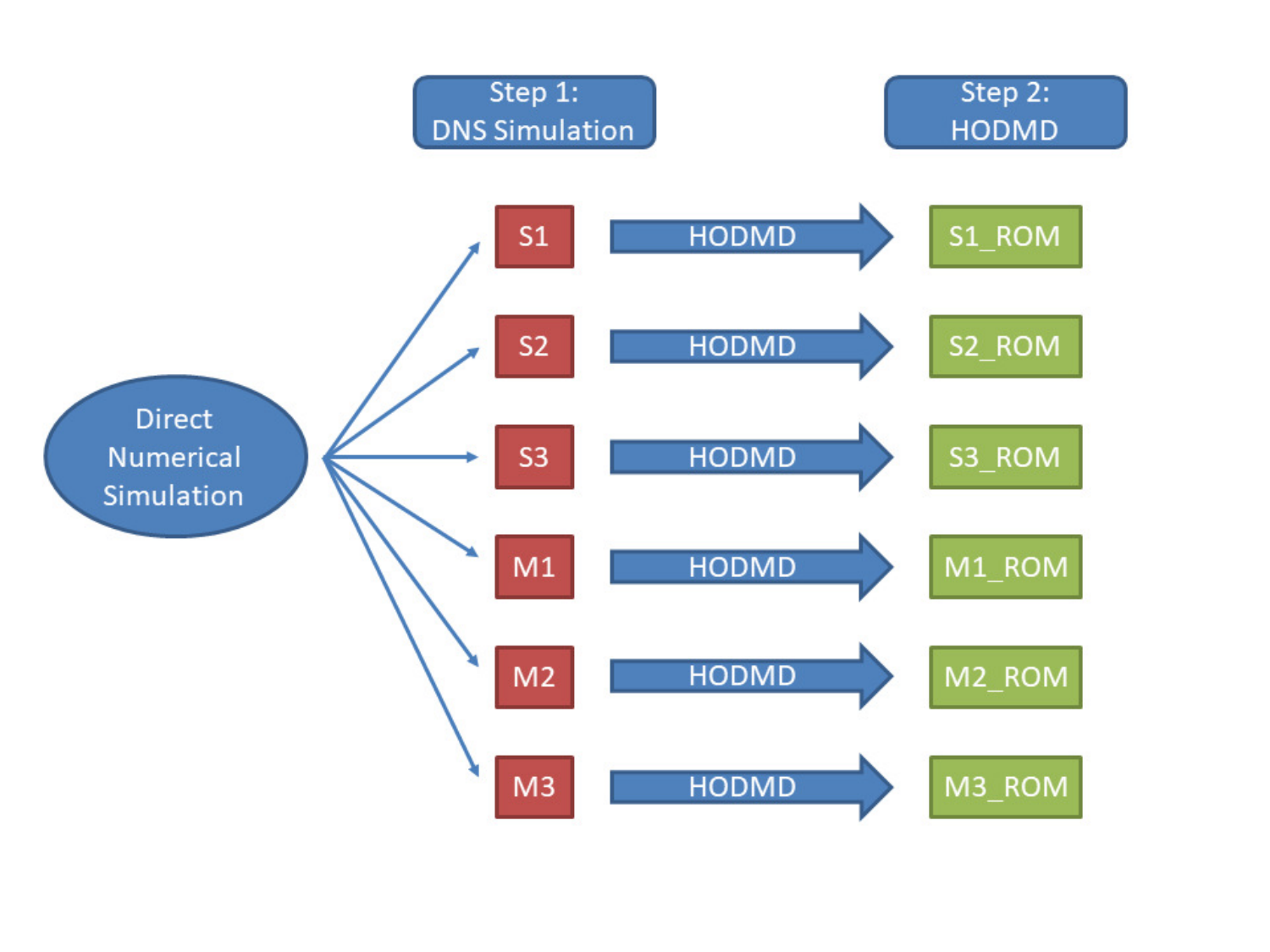}
	\includegraphics[height=0.45\columnwidth]{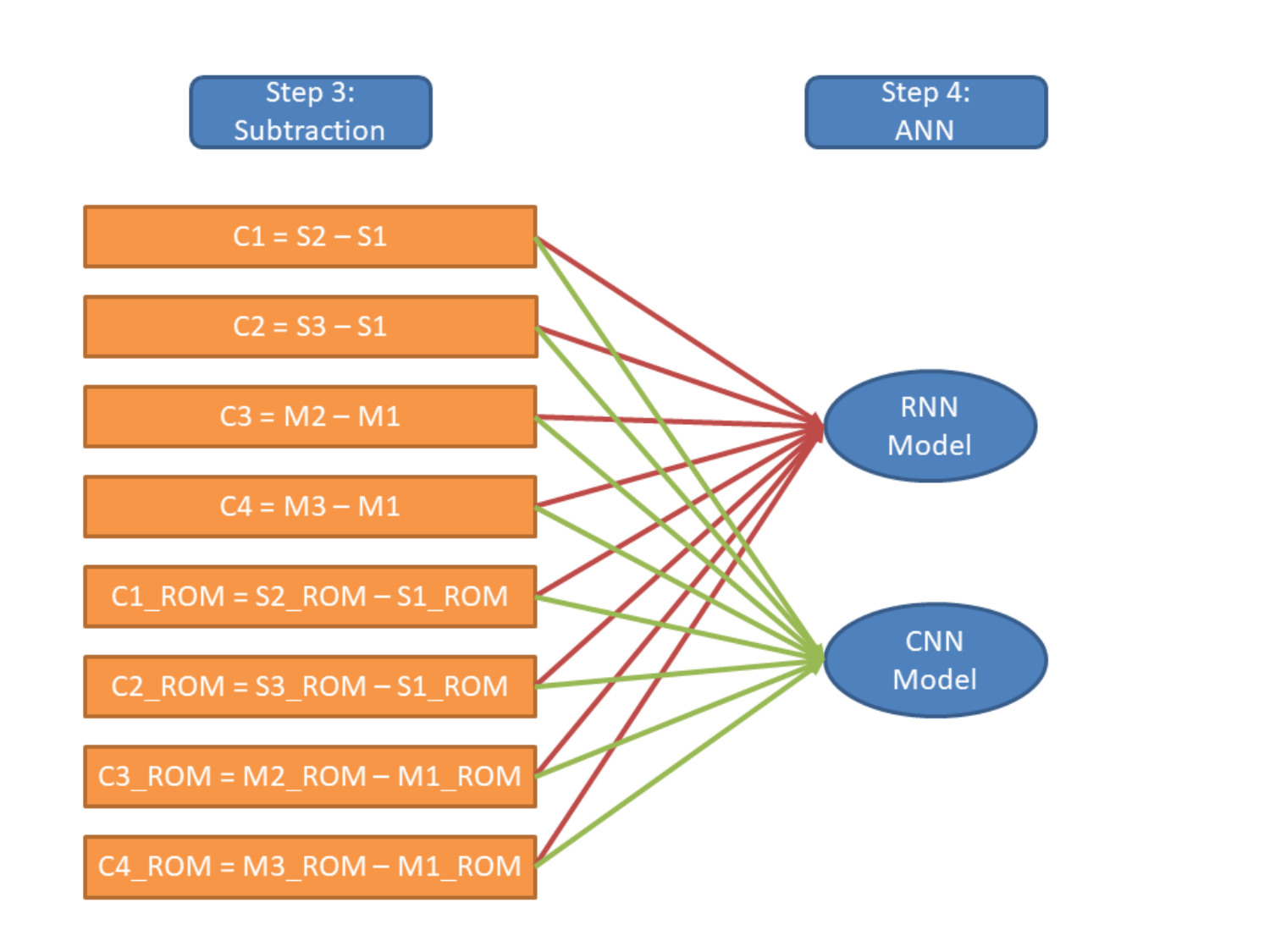}
	\caption{This graph represents the methodology used in this work. Each step is properly explained in Section \ref{sec: methodology}. First of all, we start by carrying out a DNS of all cases listed in Table \ref{tab: Nomenclatura}. In Step $2$, we use HODMD to create new data sets, which have a reduced complexity (Sec. \ref{sec: hodmd}). Next, we subtract the single-phase flow from the two-phase one (Sec. \ref{sec: preprocessing_data}). Finally, the eight data sets obtained from Step $3$ are used to train, validate and test the two artificial neural network models developed in this work (Sec. \ref{sec: Neural Networks}). In total, $16$ experiments are performed. Nomenclature used in this graph can be found in Tables \ref{tab: Nomenclatura}, \ref{tab: orig_hodmd_nomenclatura}, \ref{tab: train_cases_simple_geometry} and \ref{tab: train_cases_modified_geometry}.}
	\label{fig: methodology}
\end{figure}

As is listed in Table \ref{tab: Nomenclatura}, and Figure \ref{fig: methodology} at step $1$, there are six different data sets. Each one of these are organized in a set of $K$ time-equidistant samples, where each sample is a snapshot, $\bv_{k}$ in matrix form (dimension $J \times K$) as,
\begin{equation}
	\bV^{K} = [\bv_{1}, \bv_{2}, \dots, \bv_{k}, \bv_{k+1}, \dots, \bv_{K-1}, \bv_{K}],
	\label{eq: data structure}
\end{equation}
where $J$ is the total number of grid points defining the spatial domain. In this work, as we only predict the streamwise velocity component and we are dealing with a 2D spatial problem, $J = M_{x} \times M_{y}$, where $M_{x}$, $M_{y}$ are the number of cells along the streamwise ($x-$axis) and normal ($y-$axis) directions. Therefore $\bv_{i} \in \mathbb{R}^{M_{x}} \times \mathbb{R}^{M_{y}} \ \forall i \in [1,K]$.

\subsection{Higher order dynamic mode decomposition} \label{sec: hodmd}

Before talking about the neural networks (NN), Section \ref{sec: Neural Networks}, it is important to note that the six data sets (Figure \ref{fig: methodology}, step $1$) define the streamwise velocity flow field of a two-phase or single-phase flow. Note that small flow scales are part of these data coming straight away from the numerical simulations, which are sometimes connected to high frequencies or even incoherent events that increase the complexity of the flow dynamics, which could increase the difficulty in the training and subsequent predictions carried out by the NN. In this meaning we propose to develop a model based on the physical patterns driving the flow dynamics. For such aim we use Higher Order Dynamic Mode Decomposition (HODMD), a tool suitable to identify the main patterns and frequencies leading the flow dynamics (Figure \ref{fig: methodology}, step $2$). This method allows the flow reconstruction using only a few selected {\it DMD} modes and frequencies. Using this technique, it is possible to simplify the complexity of the flow without losing relevant information about its dynamics.

HODMD  \citep{LeClaincheVega17} is an extension of Dynamic Mode Decomposition (DMD) \citep{Schmid10} introduced for the analysis of complex flows \citep{LeClaincheVegaSoria17,LeClaincheFerrer18,LeClaincheHanFerrer19,LeClaincheetalJFM20}. Similarly to DMD, HODMD decomposes  spatio-temporal data $\bv(x,y,z,t_{k})$, as an expansion of $\bM$ DMD modes $\bu_{m}$, which are weighted by an amplitude $a_{m}$ as follows,
\begin{equation}\label{eq:dmdT}
	\bv(x,y,z,t_{k}) = \bv_k \simeq \sum_{m=1}^{M} a_{m}\bu_{m}(x,y,z)e^{(\delta_{m}+i\omega_{m})t_{k}},
\end{equation}
for $k = 1, \dots, K$. The real scalars $a_{m}\geq0$ are the {\it mode amplitudes}  and $\delta_{m}$ and $\omega_{m}$ are the temporal growth rates and frequencies, respectively. The modes $\bu_{m}$ (generally complex) are normalized to exhibit unit root mean square (RMS) norm. The dimension of the vector space spanned by them is called the {\it spatial complexity}, $N$, while the number of DMD modes retained in the expansion, $M$, is called the {\it spectral complexity}.

These two expressions, spectral complexity and spatial complexity are important because when it happens that $N<M$ (spatial complexity $<$ spectral complexity), the standard DMD does not give reliable results making it necessary to use HODMD instead, which is the case presented in this article. A more complete explanation is found in Ref. \citep{LeClaincheVega17}.

HODMD has the capability to filter out the flow structures with small amplitude (generally connected with noise or uncorrelated events), retaining only the large flow structures needed to describe the most relevant flow dynamics. This is done by finding the right values for three tunable parameters:  $d$ and two tolerances $\varepsilon$ and $\varepsilon_1$. The values chosen in this work, for the six data sets, are found in Section \ref{sec: results hodmd}. A more precise explanation of how to properly choose these values can be found in Ref. \citep{LeClaincheVega20}.

As was stated above, HODMD is capable to identify the flow structures which describe the most relevant flow dynamics. Each one of these main structures are associated to a mode which are used to reconstruct the original data set. Therefore, we end up with two data sets: the original one, $\bV^{K}$ (\ref{eq: data structure}), and the reconstructed, $\hat{\bV}^{K}$, which have same dimensions as the original $(M_{x} \times M_{y} \times K)$.

Given that there are six original data sets, after applying HODMD we finish with 12: six original and six reconstructed, as listed in Table \ref{tab: orig_hodmd_nomenclatura}.

\begin{table}[ht]   
	\centering
	\begin{tabular}{|c|c|} 
		\hline
		Original Nomenclature & Reconstructed Nomenclature\\
		\hline
		S1 & S1$_{ROM}$ \\
		S2 & S2$_{ROM}$ \\
		S3 & S3$_{ROM}$ \\
		M1 & M1$_{ROM}$ \\
		M2 & M2$_{ROM}$ \\
		M3 & M3$_{ROM}$ \\
		\hline
	\end{tabular}
	\caption{Expanded nomenclature of data sets used in this article, to make reference to the reconstructed data sets through HODMD. The original nomenclature was taken from Table \ref{tab: Nomenclatura}.}
	\label{tab: orig_hodmd_nomenclatura}
\end{table}


\subsection{Preprocessing data}\label{sec: preprocessing_data}
The main interest of using NNs in this work is to show how these \textit{self-learning} models are promising tools to reduce the required computational cost, by predicting the dynamics of a two-phase flow instead of performing the full direct numerical simulation (DNS), since solving a single-phase flow is computationally much less demanding than solving a two-phase one (Table \ref{tab: Times}). In this paper, we decided to proceed by performing simulations of both the single-phase and the two-phase flow, to try to predict the dynamics of the latter. Even if we have performed two simulations (single-phase and two-phase), we only use around half of the data as training (Table \ref{tab:ML3}). Then, we have reduced the computational cost because to perform a simulation, with half of the samples, on both single-phase and two-phase flow is less computationally demanding than performing a full simulation, with the total number of samples, on the two-phase flow. The reason behind performing a HODMD reconstruction to these data sets is to show, how this reduction in the complexity of the flow dynamics has also simplified the training of the NNs, and therefore, improving predictions.

In this sense, in order to simplify the NN training, and thus its computational time, we subtract the single-phase flow from the two-phase one, both in the original and reconstructed data sets. This is represented in Tables \ref{tab: train_cases_simple_geometry} and \ref{tab: train_cases_modified_geometry}, and Figure \ref{fig: methodology} at step $3$. These eight new data sets are the ones used to train, validate and test both NNs: RNN and CNN (Sec. \ref{sec: Neural Networks}).

\begin{table}[H]
	\centering
	\begin{tabular}{|c|c|}
		\hline
		&   Two-phase $-$ Single-phase     \\ \hline
		C1 & S2 $-$ S1 \\
		C2 & S3 $-$ S1 \\
		C1$_{ROM}$ & S2$_{ROM}$ $-$ S1$_{ROM}$ \\
		C2$_{ROM}$ & S3$_{ROM}$ $-$ S1$_{ROM}$ \\ \hline
	\end{tabular}
	\caption{Data sets obtained from subtraction of the single-phase flow from the two-phase one, for the simple geometry case (geometry without bluff body). These data sets are the ones used to train, validate and test the NNs.}
	\label{tab: train_cases_simple_geometry}
\end{table}

\begin{table}[H]
	\centering
	\begin{tabular}{|c|c|}
		\hline
		&   Two-phase $-$ Single-phase     \\ \hline
		C3 & M2 $-$ M1 \\
		C4 & M3 $-$ M1 \\
		C3$_{ROM}$ & M2$_{ROM}$ $-$ M1$_{ROM}$ \\
		C4$_{ROM}$ & M3$_{ROM}$ $-$ M1$_{ROM}$ \\ \hline
	\end{tabular}
	\caption{Data sets obtained from subtraction of the single-phase flow from the two-phase one, for the modified geometry case (geometry with bluff body). These data sets are the ones used to train, validate and test the NNs.}
	\label{tab: train_cases_modified_geometry}
\end{table}

As detailed in Section \ref{sec: hodmd} there is no difference between the spatial dimension of the original and reconstructed data sets. In this work this dimension is $(100 \times 200 \times 1)$, and after traversing the second dimension with step 2 we obtain a square shape $(100 \times 100 \times 1)$, for each snapshot. Therefore, data sets listed in Tables \ref{tab: train_cases_simple_geometry} and \ref{tab: train_cases_modified_geometry}, which are the ones used for training, validation and testing, have the same spatial dimension. In this meaning there is no need to develop a different deep learning architecture for each case (original and reconstructed). We can simply design one for both of them.

\subsection{Neural Networks} \label{sec: Neural Networks}
\begin{figure}[H]
	\centering
	\includegraphics[height=0.495\columnwidth]{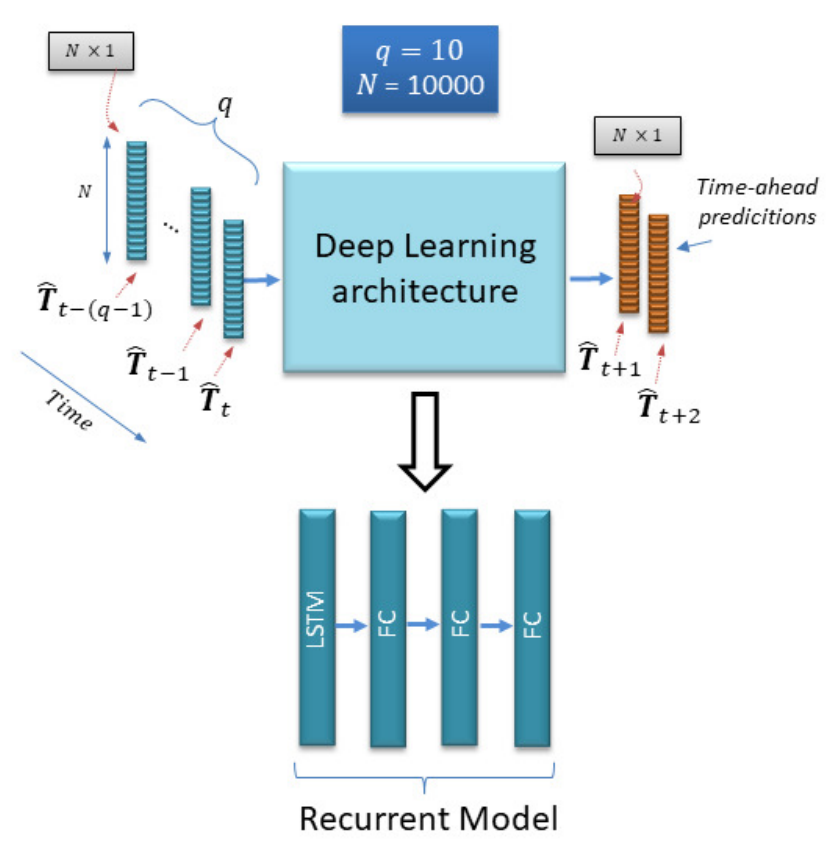}
	\includegraphics[height=0.495\columnwidth]{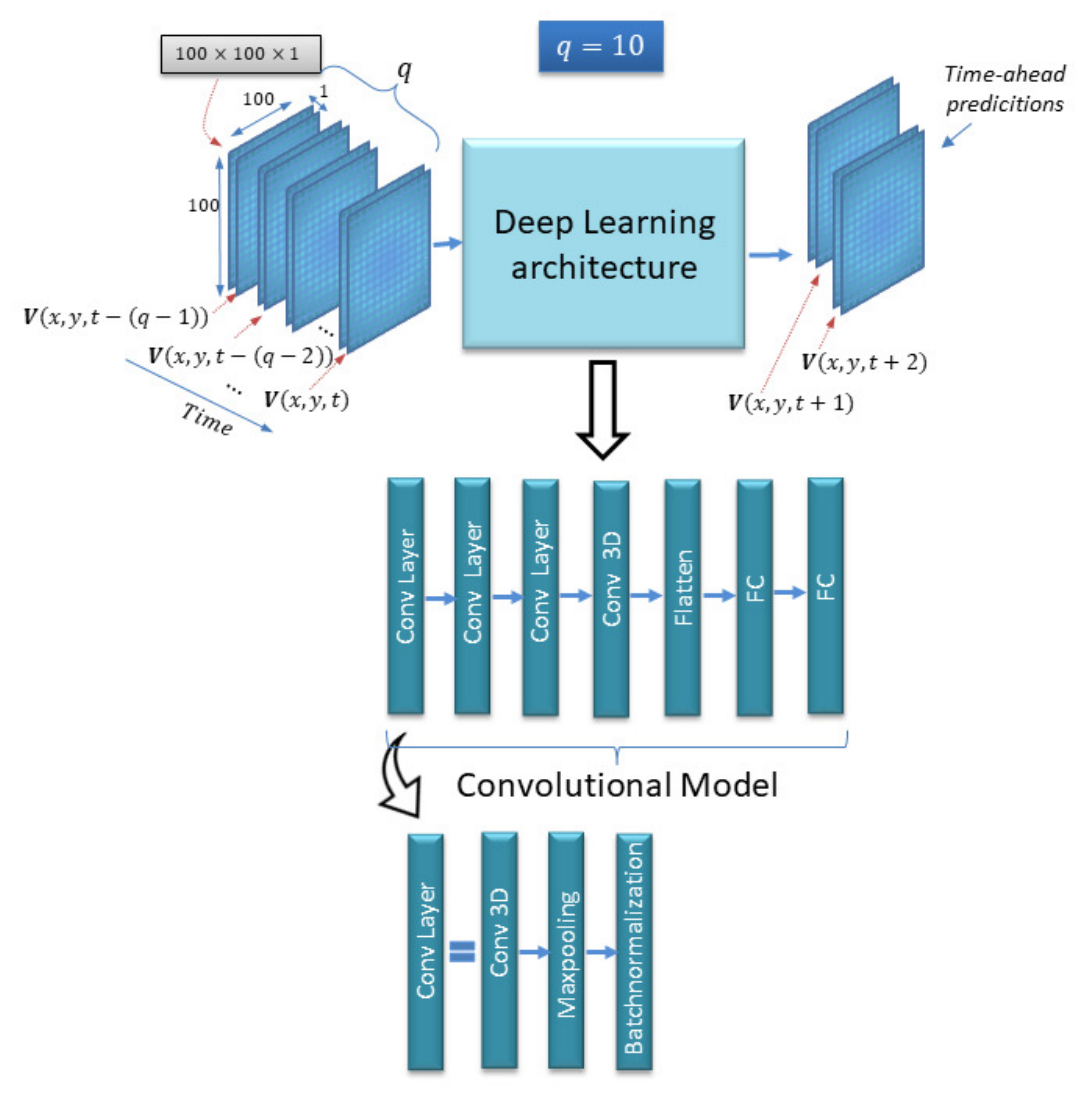}
	\caption{Graphic representation of the RNN (left) and CNN (right) models. Both of them use the previous $q=10$ samples to predict two time-ahead samples, $\bv_{t+1}$ and $\bv_{t+2}$. Since either the original and reconstructed data sets have the same spatial dimension, there is no need to develop a specific architecture for each one of them.}
	\label{fig:ML1}
\end{figure}

In this work we proposed two deep learning architectures, as shown in Figure \ref{fig:ML1} and Figure \ref{fig: methodology} at step $4$: (i) A Recurrent Neural Network (RNN) similar to the one used in Ref. \citep{Abadia22}, and (ii) a Convolutional Neural Network (CNN) which was previously used in Ref. \citep{LopezMartin21}. These architectures were designed to predict two samples in the future using $q$ samples from past, i.e., to predict $\bv_{t+1}$ and $\bv_{t+2}$ the model uses $\bv_{t}$, $\bv_{t-1}$, $\bv_{t-2}$, $\dots$, $\bv_{t-q+1}$ as training.

The idea of using these architectures, also developed by some of the authors of this article, is to test their limits of applicability to data forecasting in complex problems of fluid dynamics. Developing robust and generalizable models could suppose an advance in the field, reducing the time employed to calibrate the neural networks architectures.

The RNN is an architecture composed mainly by one long short-term memory (LSTM) layer  \citep{Yuetal19} and three fully connected (FC) layers. Where the LSTM layer was specially designed to deal with sequential data. However, even if our data is sequential, each sample is a snapshot (matrix structure), and LSTM layer only uses vector structure. Therefore, it is necessary to reshape the snapshots into vectors,
\begin{equation}
	(100 \times 100 \times 1) \Longrightarrow (10^{4}) = N.
\end{equation}
This reshaping will cause loss of information in the spatial dimension, which will negatively impact the training, as shown in Section \ref{sec: results ann}. This is the shallowest architecture proposed in this work (4 layers). A deeper architecture was studied, in which 10 additional FC layers were added at the end. However, it did not show any improvement in predictions compared to the shallower version. In addition, the reason to keep only one LSTM layer in this architecture is due to the large number of trainable parameters and the high training time required for this layer. One of the main aims of this work is to study the potential of relatively simple deep learning models when are used in complex fluid dynamics problems. Because, the more complex the architecture, the more samples are needed to train it, and in fluid dynamics problems the generation of samples may require either a high monetary (physical experiments) or computational (numerical simulations) cost and, depending on the case, the latter can be computationally prohibitive.

The other architecture, CNN, is mainly composed by four three-dimensional convolutional neural layers (Conv3D) \citep{Rawatetal2017} followed by a Maxpooling, Batchnormalization and two FC layers. However, due to the Conv3D layers, it is necessary to include a flatten function between the last Conv3D and the first FC layer, to adapt the matrix structure used by the Conv3D layers to a vector structure used by the FC layers. Tables \ref{tab:ML1} and \ref{tab:ML2} give a detailed information of the two chosen architectures: order and type of layers, number of units in each layer, activation function and its output dimension.

\begin{figure}[H]
	\centering
	\includegraphics[height=0.45\columnwidth]{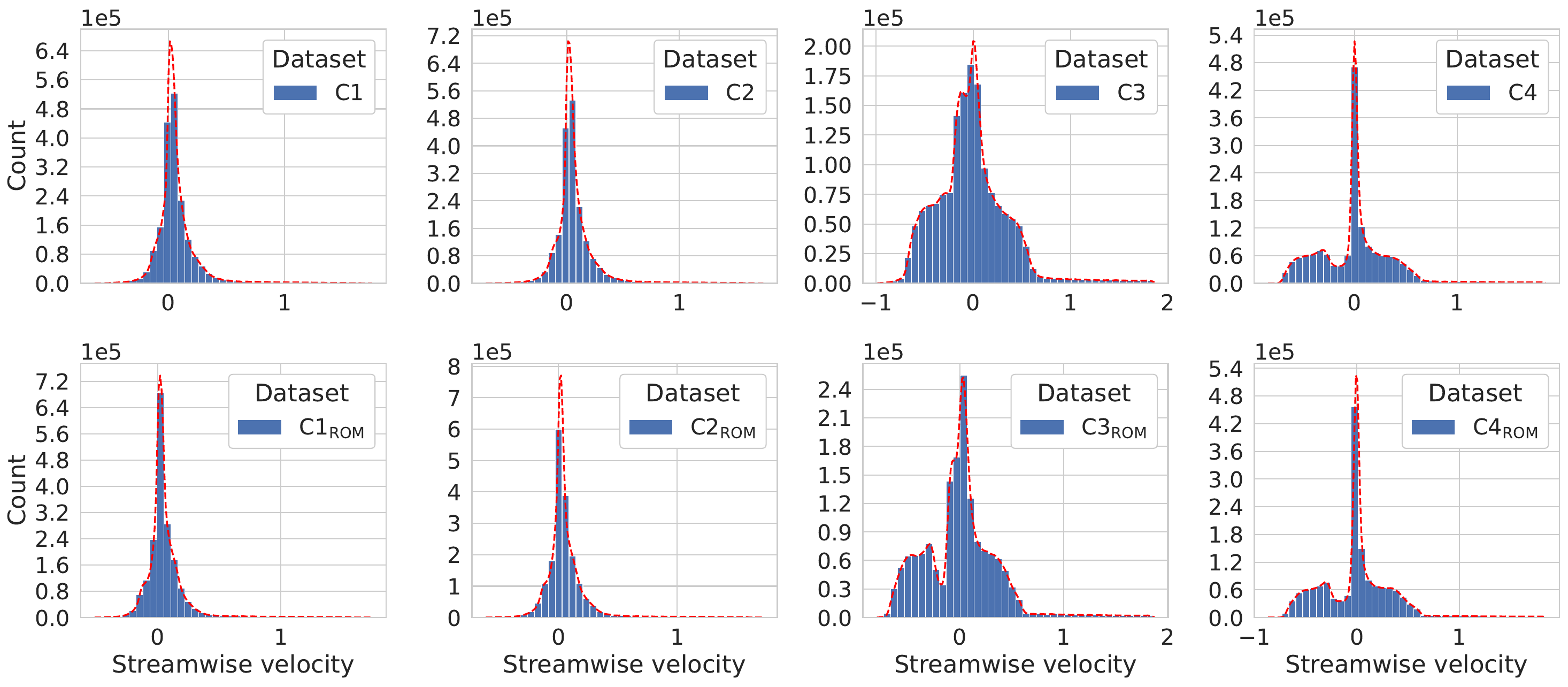}
	\caption{Histograms of all pre-normalized data sets utilized for training in this work, where the red dotted line represents the kernel density estimation (KDE). Top row shows the histograms and KDEs of data sets C1, C2, C3 and C4, respectively. Bottom row shows the same for data sets C1$_{ROM}$, C2$_{ROM}$, C3$_{ROM}$ and C4$_{ROM}$, respectively. Note that all data sets have a similar range of values, of approximately $[-0.5, 0.5]$.}
	\label{fig: histograms}
\end{figure}

As a first approach, we decided to use the same RNN model as the one proposed in Ref. \citep{Abadia22}. However, the predictions returned were not acceptable. In this meaning, we tried three different configurations: a) at preprocessing; no normalizing, between $0$ and $1$, the data set used for training, b) at the architecture; setting the last layer activation function as Linear, and finally c) combining a) and b). The cases a) and b) by itself do not yield good predictions. However, the combination of both substantially improved them. While we do not have a complete explanation to this phenomenon, we believe this could be because of the data set. It is widely acknowledged that data normalization is a beneficial practice for training deep learning models due to its positive impact on convergence speed and the prevention of vanishing/exploding gradients, among other reasons. However, it is also known that data normalization highly depends on the data set used. As an example, in this study \citep{noNormalization} the authors found out that no data normalization returns the optimal values in a LSTM architecture. Data normalization is also often applied to LSTM to improve the performance of the model by ensuring that the input data is within a similar range, however as is shown in Figure \ref{fig: histograms} all data sets used for training in this work (Tables \ref{tab: train_cases_simple_geometry} and \ref{tab: train_cases_modified_geometry}) have already a similar range, of approximately $[-0.5, 0.5]$. Just to provide validation of the results, we have given access to the source code of the RNN model and one data set, which can be used to replicate the results and check that the best predictions are obtained without normalization.

Since in this work the optimal predictions, for the RNN model, were obtained by this last configuration c), we decided not to normalize the data in this model. However, for the CNN model, the data sets used for training were normalized between $0$ and $1$, as usual, since the optimal results were obtained by normalizing the input data.

\begin{table}[H] 
	\centering
	\scalebox{0.8}{
		\begin{tabular}{|c|c|c|c|c|c|}
			\hline
			\# Layer & Layer details & Activation Function & Recurrent Activation & \# Neurons & Dimension\\\hline
			$0$ & Input & - & - & $N$ & $10\times N$\\
			$1$ & LSTM & Tanh & Sigmoid & $400$ & $400$\\
			$2$ & FC & ReLU & - & $200$ & $200$\\
			$3$ & FC & ReLU & - & $80$ & $80$\\
			$4$ & FC & Linear & - & $N$ & $N$\\\hline
		\end{tabular}
	}\caption{Architecture details in the RNN model. The layers are LSTM or FC, the number of predictors is $q=10$, $N=10^{4}$ is the 2D spatial mesh dimension reshaped to a column vector (this was done because of the LSTM layer), the activation functions are Rectified Linear Unit (ReLU) or Linear. The output tensor dimension (Dimension) and the number of Kernels/neurons (\# neurons) are indicated for each layer.}
	\label{tab:ML1}
\end{table}
\begin{table}[H] 
	\centering
	\scalebox{0.8}{
		\begin{tabular}{|c|c|c|c|c|c|c|c|}
			\hline
			\# Layer & Layer details & Kernel size & Stride& Padding& Activation& \# Neurons & Dimension\\\hline
			$0$& Input & - & - & - & - & $N$ & $10 \times N$\\
			$1$& Conv 3D & $2 \times 2 \times 2$& $1$& No & ReLU & $45$ & $9 \times 99 \times 99 \times 5$\\
			$1$& Maxpooling & $1 \times 2 \times 2$ & - & Valid & - & - & $9 \times 49 \times 49 \times 5$\\
			$1$& Batchnormalization & - & - & - & - & $20$ & $9 \times 49 \times 49 \times 5$\\
			$2$& Conv 3D & $2 \times 2 \times 2$& $1$& No & ReLU & $410$ & $8 \times 48 \times 48 \times 10$\\
			$2$& Maxpooling & $1 \times 2 \times 2$ & - & Valid & - & - & $8 \times 24 \times 24 \times 10$\\
			$2$& Batchnormalization & - & - & - & - & $40$ & $8 \times 24 \times 24 \times 10$\\
			$3$& Conv 3D & $2 \times 2 \times 2$& $1$& No & ReLU & $1620$ & $7 \times 23 \times 23 \times 20$\\
			$3$& Maxpooling & $1 \times 2 \times 2$ & - & Valid & - & - & $7 \times 11 \times 11 \times 20$\\
			$3$& Batchnormalization & - & - & - & - & $80$ & $7 \times 11 \times 11 \times 20$\\
			$4$& Conv 3D & $1 \times 1 \times 1$& $1$& No & ReLU & $42$ & $7 \times 11 \times 11 \times 2$\\
			$5$ & Flatten & - & - & - & - & - & $1694$\\
			$6$& FC & - & - & - & ReLU & $80$ & $80$ \\
			$7$& FC & - & - & - & Sigmoid & $N$ & $N$ \\\hline
	\end{tabular}}
	\caption{Same as Table \ref{tab:ML1} for the CNN model. In this architecture, $N$ is the 2D spatial mesh without reshaping, i.e., $N = (100 \times 100 \times 1)$.}
	\label{tab:ML2}
\end{table}

\begin{figure}[H]
	\centering
	\includegraphics[height=0.6\columnwidth]{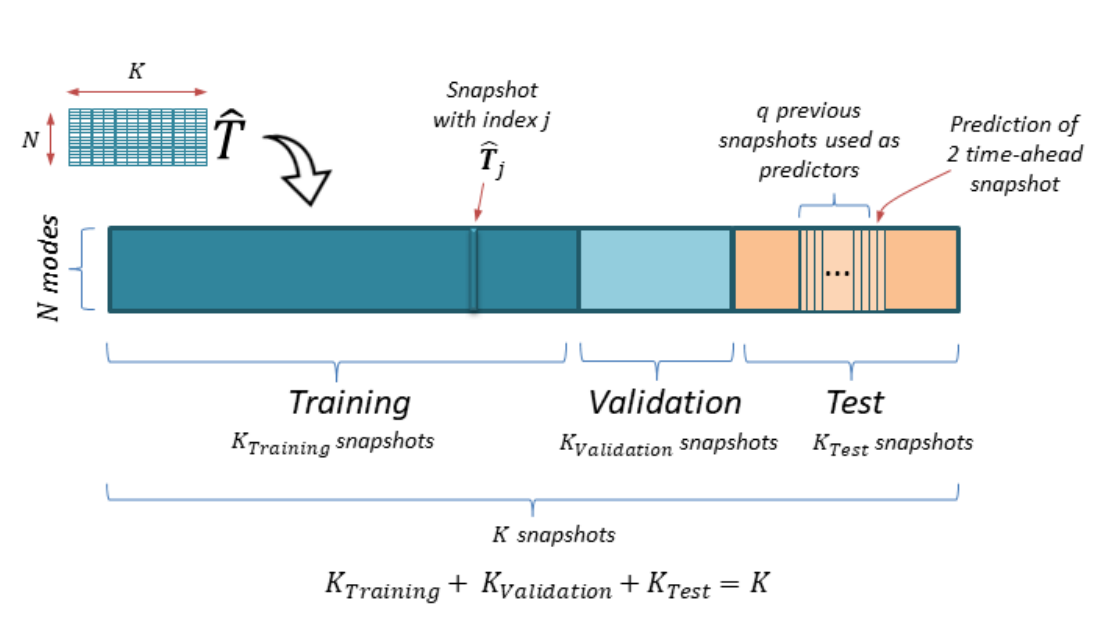}
	\caption{Data structure used for training, validation and test in both predictive models.}
	\label{fig:ML2}
\end{figure}

Our validation strategy, to guarantee generalization of results, has been to split the data set in three consecutive sets: training, validation and test, see Figure \ref{fig:ML2}. The predictions of the numerical simulations are computed over the test set, while the validation set is used for hyperparameter selection and to control the early stopping criteria, stopping training when the error on the validation set starts to rise in contrast to the error in the training set. This avoids \textit{overfitting}. In this meaning, samples of our data sets  are separated in three different subsets that are used for training, validation and testing. The number of samples in these subsets are $K_{training}$ (training), $K_{validation}$ (validation) and $K_{test}$ (test) respectively (see Table \ref{tab:ML3}), where the total number of samples is represented as $K_{training}+K_{validation}+K_{test}=K$.

\begin{figure}[H]
	\centering
	\includegraphics[height=0.32\columnwidth]{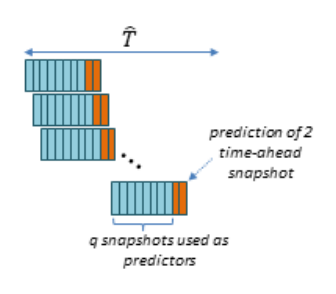}  
	\caption{Rolling window method used to extract the inputs (q) and expected outputs (two) for the predictive models.}
	\label{fig:ML3}
\end{figure}

In problems like fluid dynamics it is hard to obtain a large number of samples, most of the time it is possible to obtain just a few hundred or thousand of them. Therefore it is necesary to perform a preprocess of the data to artificially \textit{increase} the number of samples. The method most commonly used is \textit{rolling-window} as outlined in Figure \ref{fig:ML3}. This method generates data batches with $q$ inputs and two outputs as expected by the predictive model (Figure \ref{fig:ML1}). In this paper the offset considered between the successive rolling windows was chosen equal to $1$ (constant).

The loss function used for training is Mean Squared Error Loss ($MSE_{Loss}$). This loss function measures the error, during training, between the data set and the prediction performed by the NN. In order to  minimize it we used the algorithm known as batch stochastic gradient descent. $MSE_{Loss}$ is firstly calculated for each time prediction, $MSE_{Loss}(t)$ (local error), as

\begin{equation}
	MSE_{Loss}(t)=\frac{1}{M}||{\bV^{K}_{t}}^{predicted}-\bV^{K}_{t}||^2,
	\label{eq:MSEloss}
\end{equation}

where $M$ is the number of samples that composed a batch. The global loss ($MSE_{Loss}$) is computed by averaging the local loss calculated for each time prediction, over the total number of samples in the temporal matrix as
\begin{equation}
	MSE_{Loss}=\frac{1}{K_{\alpha}}\sum_{K_{\alpha}} MSE_{Loss} (t),
\end{equation}
depending on where we obtain the global loss: $K_{\alpha}$ represents the number of samples inside the validation or test set, respectively (see Figure \ref{fig:ML2}). In Table \ref{tab:ML3} is shown the number of samples for the training ($K_{training}$), validation ($K_{validation}$) and test ($K_{test}$) sets for each one of the data sets used in this work. 

\begin{table}[H] 
	\centering
	\scalebox{0.8}{
		\begin{tabular}{|c|c|c|c|c|c|}
			\hline
			Geometry & Data sets & $K_{training}$ & $K_{validation}$ & $K_{test}$ & $K$\\\hline
			Simple & C1, C2, C1$_{ROM}$, C2$_{ROM}$ & $184$ & $45$ & $122$ & $351$ \\\hline
			Modified & C3, C4, C3$_{ROM}$, C4$_{ROM}$ & $105$ & $39$ & $157$ & $301$ \\
			\hline
	\end{tabular}}
	\caption{Number of samples for the training, validation and test sets for each one of the cases, with simple and modified geometry.\label{tab:ML3}}
\end{table}

The loss ($MSE_{Loss}(t)$) over the validation set is calculated at the end of each epoch (an epoch is a complete pass of all training samples). It is not unusual that, at some point, the loss over the validation set starts to grow over the loss on the training set. This effect could lead to overfitting. In this meaning, we use {\it early stopping} over the validation set to stop training when this effect is detected and also when the loss over this set is not reduced after a certain number of epochs ({\it patience period}). Both early stopping and patience period were only used when training was performed with data sets from Table \ref{tab: train_cases_simple_geometry} (simple geometry case).

The training parameters for the NNs used in this work are the following: 
\begin{itemize}
	\item[a)] Mini-batch gradient descent with a batch size of $5$ and a training length of $70$ epochs for the CNN model and $140$ epochs for the RNN model. Given the number of samples available for training is no more than $184$, \textit{early stopping} with a patience period of $10$ epochs is used as an additional regularization method to avoid overfitting.
	\item[b)] Adam method was set as optimization \citep{Kingmaetal} using the default values for the parameters ($\alpha=0.001$ for the learning rate, $\epsilon=10^{-8}$, $\beta_1=0.9$ and $\beta_2=0.999$ see details in Ref. \citep{Kingmaetal}).
	\item[c)] $10$ samples used to predict the two next time-ahead samples (i.e., $q=10$).
\end{itemize}
As shown in Section \ref{sec: results ann}, the CNN model can achieve fairly good predictions even with such a low number of epochs used for training. However, the RNN model was not able to improve its predictions, even though increasing the number of epochs. Therefore, we believe that the poor performance of this model may be due to its architecture and the format of the input data. On our previous work \citep{Abadia22} however, the RNN model achieves the best results because the input data was not obtained by just flattening the original snapshots, but by using Singular Value Decomposition.

Finally, the Relative Root Mean Square Error (RRMSE) is also computed for each sample belonging to the test set, to measure the error on the predictions carried out by the NNs as
\begin{equation}
	RRMSE(t)=\frac{||{\bV^{K}_t}^{predicted}-\bV^{K}_t||}{||\bV^{K}_t||}.\label{eq:RMSEtimeNN}
\end{equation}
Where $\bV^{K}_t$ are the original/reconstructed data sets, ${\bV^{K}_t}^{predicted}$ are the predictions performed by the NN and $t$ represents the $t$-th sample inside the test set.

\subsubsection{Hardware specifications}
To train both models, RNN and CNN, we used \textit{Google Colaboratory} or \textit{Google Colab} with GPU accelerator. Specifically the GPU used in this work was an Nvidia Tesla T4 with 16Gb (gigabytes) of RAM.

\section{Flow physics}\label{sec: simulation results}
In this section, the results obtained in the numerical simulations are discussed to understand the main differences and the expected complexity of the different cases modelled using neural networks.

As explained in section \ref{Initial conditions and boundary conditions}, the problem starts with the computational volume filled with phase 2 at rest and suddenly both phases are injected through the inlet tubes. Thus, there will be an initial transient regime in which the static state vanishes and the movement of the fluid begins. This initial regime is characterized by the absence of clearly defined patterns in the fluid due to their constant evolution. The approximate duration of this regime depends on each case, but in general it lasts around the first 100 time units in all the cases studied. The studies  will be carried out without considering the initial transient stage.

\begin{figure}[H]
	\centering
	\begin{subfigure}[b]{0.95\textwidth}
		\centering
		\includegraphics[trim=0 50 0 0, clip,width=\textwidth]{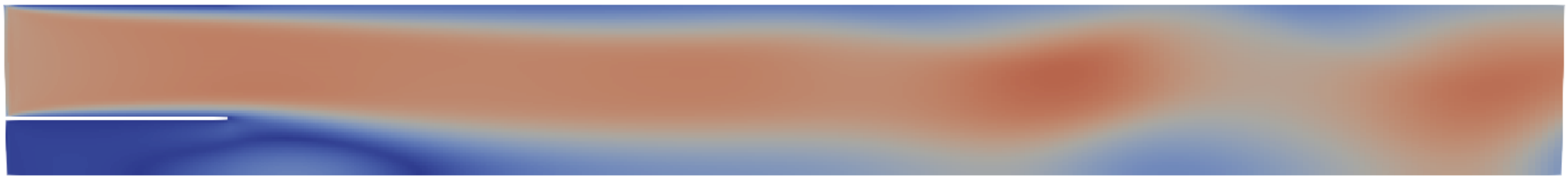}
		\subcaption{S1}
	\end{subfigure}
	\vfill
	\begin{subfigure}[b]{0.95\textwidth}
		\centering
		\includegraphics[trim=0 50 0 0, clip,width=\textwidth]{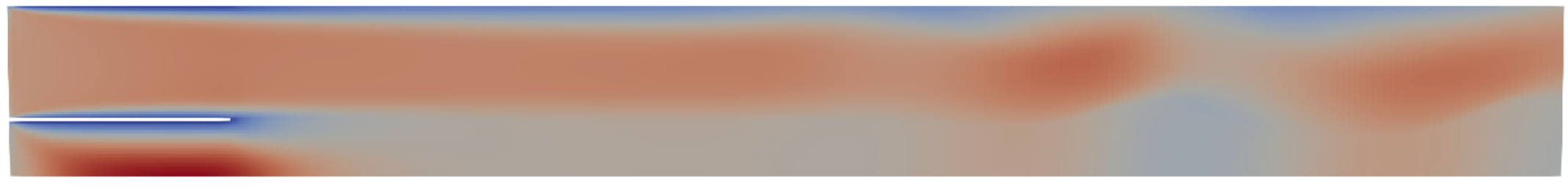}
		\subcaption{S3}
	\end{subfigure}
	\vfill
	\begin{subfigure}[b]{0.95\textwidth}
		\centering
		\includegraphics[trim=0 50 0 0, clip,width=\textwidth]{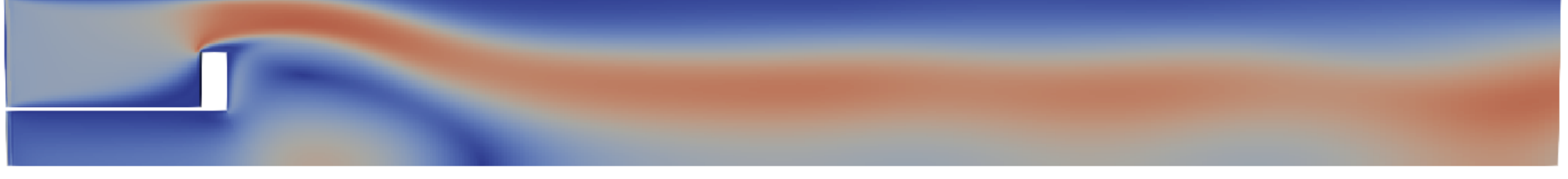}
		\subcaption{M2}
	\end{subfigure}
	\vfill
	\begin{subfigure}[b]{0.95\textwidth}
		\centering
		\includegraphics[trim=0 50 0 0, clip,width=\textwidth]{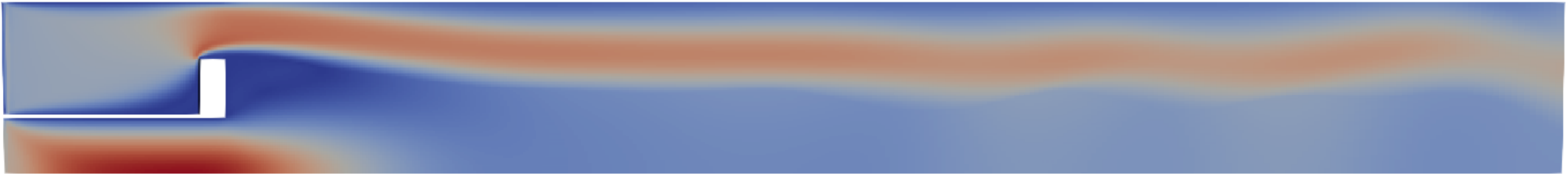}
		\subcaption{M3}
	\end{subfigure}
	\vfill
	\begin{subfigure}[b]{0.95\textwidth}
		\centering
		\includegraphics[width=0.6\textwidth]{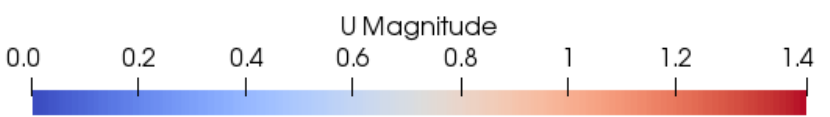}
	\end{subfigure}
	\caption{Velocity magnitude at a representative instant once the permanent stage has been reached for four different multiphase cases.}
	\label{fig:VelMag}
\end{figure}

The saturated regime  is characterized by the existence of patterns and by the constant process of interaction between both jets due to a shear layer instability. Figure  \ref{fig:VelMag} shows a representative snapshot of the velocity magnitude for some of the cases studied. 

As seen, during the permanent state, the jet interaction process gives rise to two distinct zones: (i) Primary zone, where continuity is maintained in the jets as the fluid advances to the right. At the beginning, the value of velocity is quasi-stationary but as the value of x grows, mechanisms appear on a microscopic scale that make the jets oscillate increasingly as they are amplified with the advance of the fluid. At some point these oscillations reach an amplitude large enough to cause the jet to break. (ii) Breaking zone, where the jet becomes unstable and breaks up, giving rise to smaller structures that detach over time. In this zone, the greatest degree of interaction and atomization between both phases occurs. The functioning of this process is due to the fact that the lower jet develops lobes that tend to ascend as they advance and divide the upper jet into different packages.

\begin{figure}[H]
	\centering
	\includegraphics[width=0.7\textwidth]{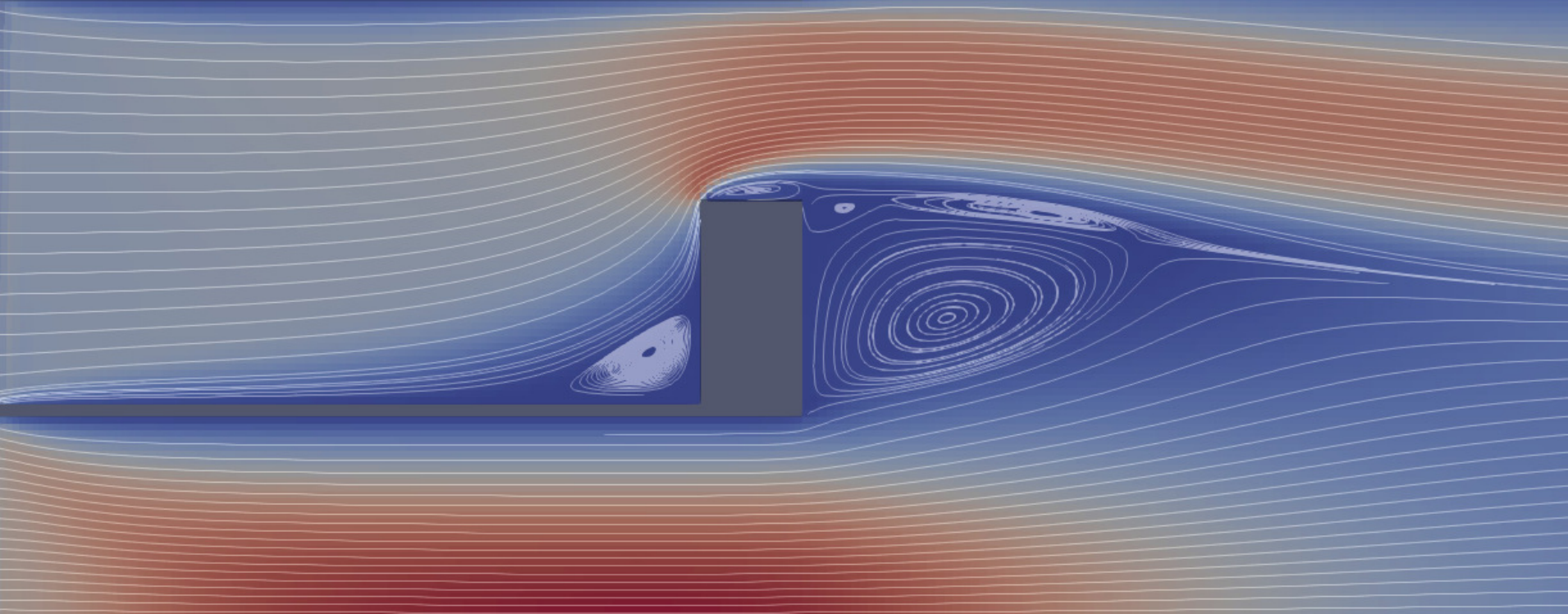}
	\caption{Streamlines around the bluff body. Colors represent velocity magnitude, where red implies large velocity magnitude and blue refers to small velocity.}
	\label{fig: Bluff body zoom}
\end{figure}

Regarding the physics surrounding the bluff body, a recirculation zone exists at the rear face of the element as can be seen in Figure \ref{fig: Bluff body zoom}, giving rise to a region of low pressure. Such a pressure drop is responsible for causing the upper jet, instead of continuing straight in the main chamber, to be oriented downward and interacting with the lower jet. A shear layer exists between the recirculation bubble and the upper jet, giving rise to a continuous detachment of eddies. Hence, as presented, the bluff body has the ability to enhance mixing between the two phases.

As mentioned at the beginning, the cases have been calculated both with and without surface tension. Now, the effect of this physical property on the solution is studied. 

\begin{figure}[H]
	\centering
	\begin{subfigure}[b]{0.95\textwidth}
		\centering
		\includegraphics[trim=0 48 0 0, clip,width=\textwidth]{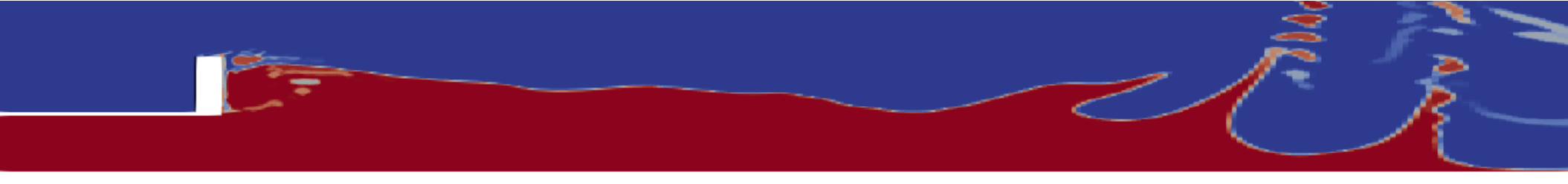}
	\end{subfigure}
	\vfill
	\begin{subfigure}[b]{0.95\textwidth}
		\centering
		\includegraphics[trim=0 48 0 0, clip,width=\textwidth]{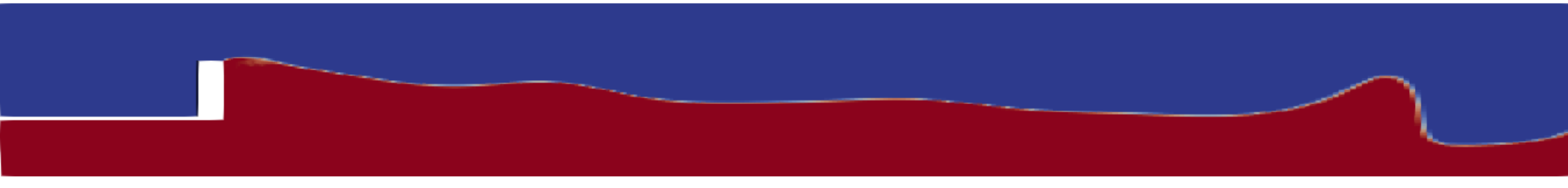}
	\end{subfigure}
	\begin{subfigure}[b]{0.95\textwidth}
		\centering
		\includegraphics[width=0.6\textwidth]{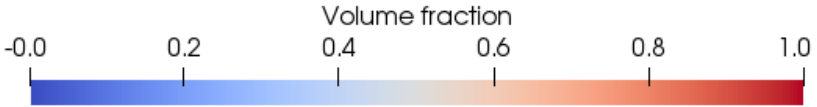}
	\end{subfigure}
	\caption{Volume fraction $\alpha$ shows the difference in atomization and interaction when not including surface tension (top) or when including it (bottom).}
	\label{fig:Bluffbodyperformance}
\end{figure}

Surface tension is defined as the amount of energy necessary to increase the surface of a fluid by unit area. An alternative definition is that surface tension is the tangential force by unit length parallel to the surface of the liquid. As a consequence, some phenomena arise due to the presence of surface tension: decrease in the degree of atomization due to the higher amount of energy that is present in the fluid in the form of surface energy (recall that the system will try to minimize its energy and therefore will tend to reduce the area of contact between both surfaces) or the resistance exerted by the fluid to be penetrated through its surface due to the higher amount of work that must be done. Surface tension also has an influence at the boundaries of the domain, resulting in a contact angle between the surface and the fluid, as occurs with water droplets on a glass surface. In this simulation, as discussed in Section \ref{Initial conditions and boundary conditions}, zero volume fraction gradient has been imposed at the boundaries, so that the contact angle between the fluid and the surface is equal to 90º.
Figure \ref{fig:Bluffbodyperformance} shows the difference in the volume fraction distribution resulting from including surface tension in the model.

As we would expect, the case which does not include surface tension presents a greater degree of atomization and interaction of both jets. It can also be verified that the area of the interface is smaller when surface tension is included, while in the other case the interface covers a much larger area because, as we explained before, the larger the surface tension, the higher the amount of energy that will be associated to the interface. In cases in which injection velocity is greater, the system will have more kinetic energy, which can be converted into surface energy, and will result in a higher degree of atomization. Finally, another notable effect of surface tension is the tendency of the fluids to stick to the walls of the bluff body, making the recirculation zone only composed of a single phase, influencing to some extent the intensity of the subsequent mixing.

Therefore, as a summary, it can be concluded that, in order to enhance interaction between the two phases, it is of interest that surface tension remains low and that the system has enough energy to overcome its effect, thus causing a higher amount of atomization. Another important aspect is related to the complexity of the flow, since surface tension makes the flow topology simpler, which is something to take into account when analyzing the data and trying to make predictions of a certain magnitude of the flow.
\subsection{Computation time}
As was explained at the beginning, the main goal of this study is to reduce the computation time of multiphase flow numerical simulations replacing them by a single-phase case. There are some aspects to take into account when looking at the computation time. First of all, the more equations the system has, the larger it will take to compute, so obviously solving for the VOF equations is more demanding than the single-phase equations. Another aspect is related to the mesh since, the larger the number of elements, the longer it will take to compute. In this case the same mesh has been used in both the multiphase and the single-phase simulations. A last important aspect to consider is the time step, which has been chosen to be as large as possible without breaking the numerical stability barrier in order to reduce computation time.

As can be seen in Table \ref{tab: Times}, simulating a single-phase case is much less computationally demanding compared to the multiphase case, which requires longer computation times. In fact, the multiphase case requires more than twice the time it takes to simulate a single-phase case. On the other hand, it is observed that the case with surface tension requires slightly less time compared to the case without surface tension, which may be due to the fact that the solution fields are more regular because of the lower mixing. 
%
\begin{table}[H]   
	\centering
	\begin{tabular}{|c|c|c|} 
		\hline
		Case &  Computation time &  Speed-up\\
		\hline
		S1 & 2.15 &  -\\
		S2 & 5.74 & 2.67 \\   
		S3 & 5.54 &  2.58\\ 
		\hline
		M1 & 1.66 &  -\\ 
		M2 & 5.24 & 3.16\\ 
		M3 & 4.38 & 2.64\\ 
		\hline
	\end{tabular}
	\caption{Time comparisons, in hours, among the different simulations. All simulations have been calculated over 500 seconds using 12 processors. The speed-up compares computation time of a multiphase case compared to the corresponding single-phase simulation.}
	\label{tab: Times}
\end{table}

\section{Patterns identification (using HODMD)}\label{sec: results hodmd}
The way to proceed in this analysis is, first, to determine the parameter ranges in which the HODMD works correctly (parameters are $d$, the window size; $\epsilon$ and $\epsilon_1$, both tolerances). This means finding the parameter ranges for which enough physical modes are retained (physical modes usually appear repeated along many parameter choices in which the model works well) so that the reconstruction error is as small as possible and, on the other hand, the amount of noise modes remains low (these modes are different in every parameter choice). The next step will be to jointly perform the HODMD analysis of the cases that provide good results to find out which are the physical frequencies. From there, the parameters are set to a particular case (the one that works best), finally obtaining the modes that govern the physics of the problem. The last step is to develop a Reduced Order Model (ROM), which only considers the most important physical modes, and to compare it with the simulation data to examine its performance. By doing this, one can build a simple but reliable version of the complex dynamical system consisting of a multiphase flow just by retaining a few physical modes. For a more extensive explanation of HODMD calibration, see Ref. \citep{LeClaincheetalJFM20}.


%
%

\begin{figure}[H]
	\centering
	\includegraphics[width=\textwidth]{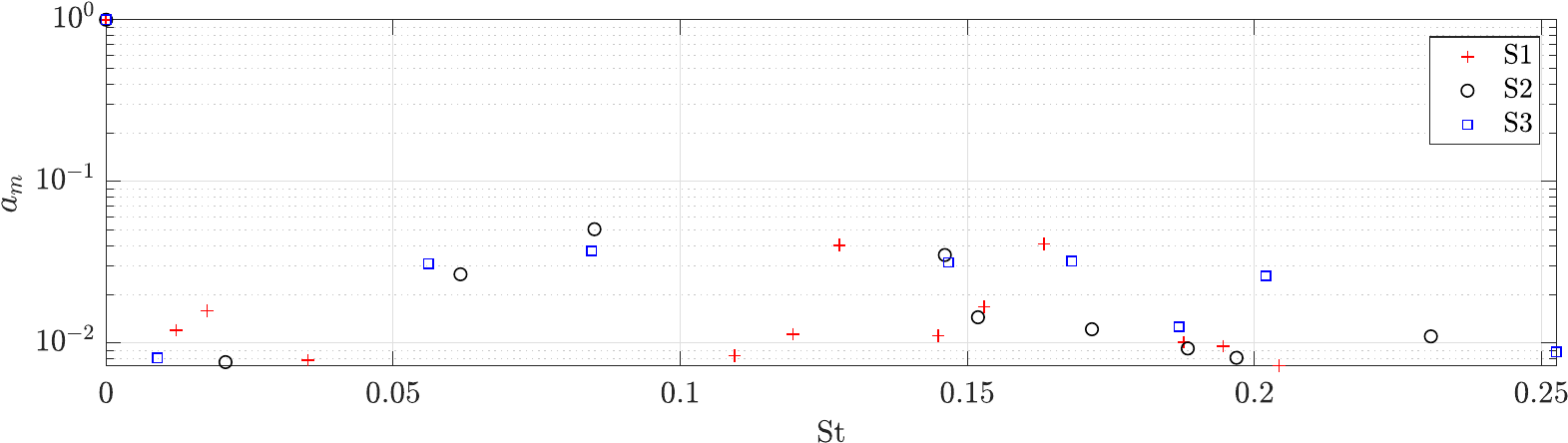}
	\caption{Normalized aplitude vs St comparison between three different cases for normal geometry. A window size of $d=100$ has been used for the single-phase case and $d=60$ for both multiphase cases. The tolerances have been set to $\epsilon=\epsilon_1=7\cdot 10^{-3}$ for the three cases.}
	\label{Simple comparacion 3 casos}
\end{figure}

\begin{figure}[H]
	\centering
	\includegraphics[width=\textwidth]{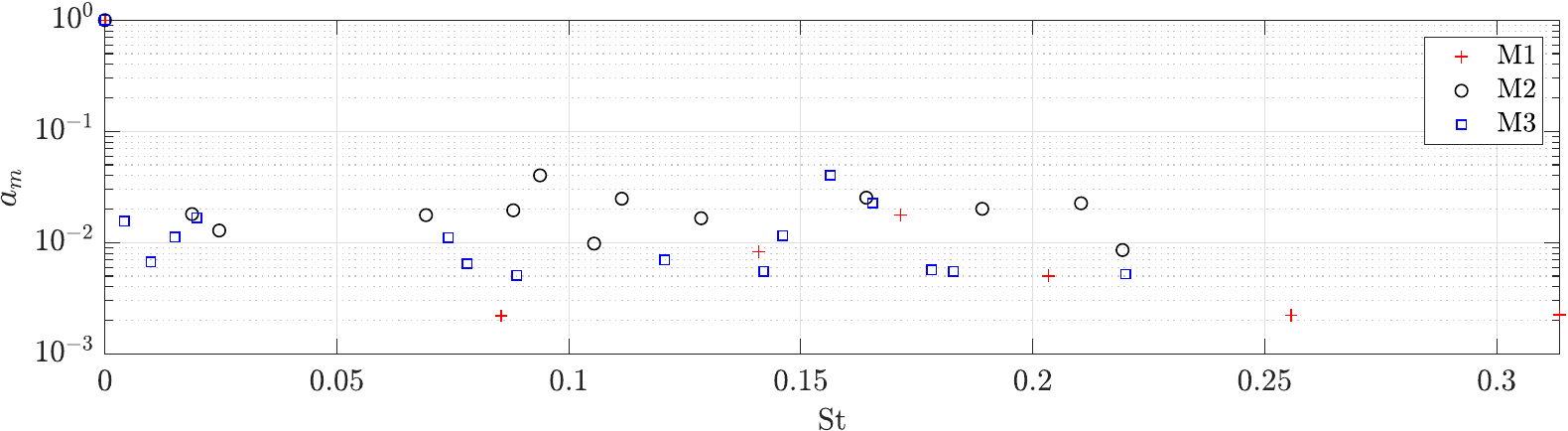}
	\caption{Normalized amplitude vs St comparison between three different cases for modified geometry. For the single-phase case d=60 and $\epsilon=\epsilon_1=2\cdot 10^{-3}$, for the multiphase without surface tension d=100 and $\epsilon=\epsilon_1=8\cdot 10^{-3}$, for the multiphase with surface tension d=100 and $\epsilon=\epsilon_1=5\cdot 10^{-3}$.}
	\label{Modified comparacion 3 casos}
\end{figure}

Figure \ref{Simple comparacion 3 casos} shows the HODMD modes selected in the single-phase and both multiphase (with and without surface tension) cases in the simple geometry (without bluff body). The amplitudes are divided by that of the dominant mode and the frequencies are expressed in dimensionless form by means of the Strouhal number ($St=\omega h/({2\pi}U)$). Regarding both multiphase cases, the modes appear as pairs in specific frequencies, which is due to the reduced effect of surface tension in this geometry. Since both jets interact in a limited way, the effect of surface tension is low and the modes from both cases are similar. On the other hand, the single-phase modes show a larger variety when compared to the multiphase case. Some frequencies are very close to the ones obtained before such as $St=0.15$ or $St=0.2$ as well as the low amplitude modes. However, other frequencies do not appear in the single-phase case such as those between $St=0.05$ and $St=0.1$. This suggests that the dynamics in the single-phase case is simpler than in the two-phase flow at the conditions studied. In Figure \ref{Modified comparacion 3 casos} the same information is plotted but corresponding to the modified geometry (with bluff body). In this case, the results are not as clear as before since the dynamics are more complex due to the increase of interaction between both jets and in this case surface tension has a higher effect in the results, resulting in larger differences between the two multiphase cases. The rise in flow complexity is also reflected in the higher amplitude associated to the modes, showing that a larger number of DMD modes is required to properly reconstruct the flow and to develop a ROM. 

\begin{figure}[H]
	
	\centering 
	\hskip 1cm Simulation \hskip 4cm Reconstruction\\
	\vskip 0.1cm
	
	\includegraphics[width=0.49\textwidth, angle =0]{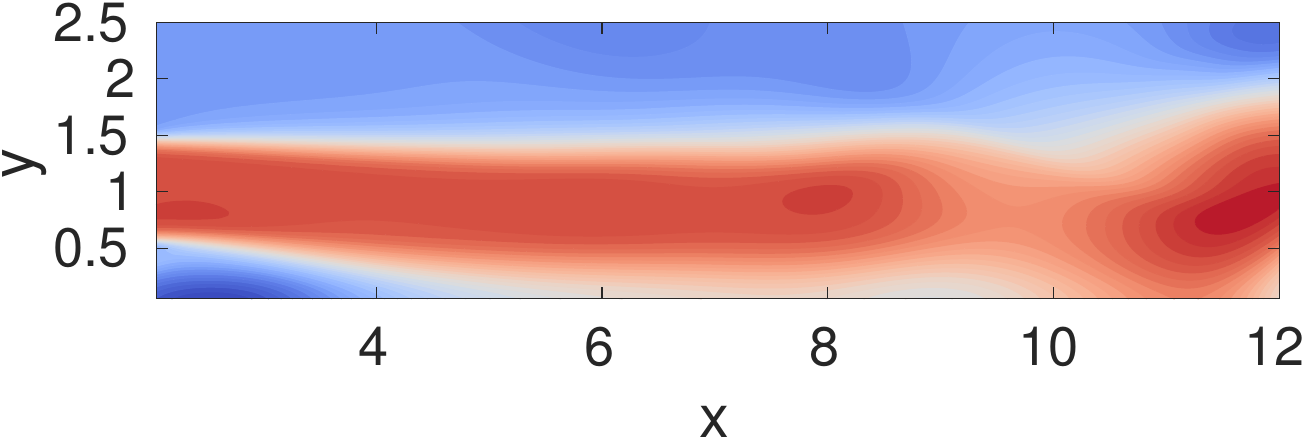}
	\includegraphics[width=0.49\textwidth, angle =0]{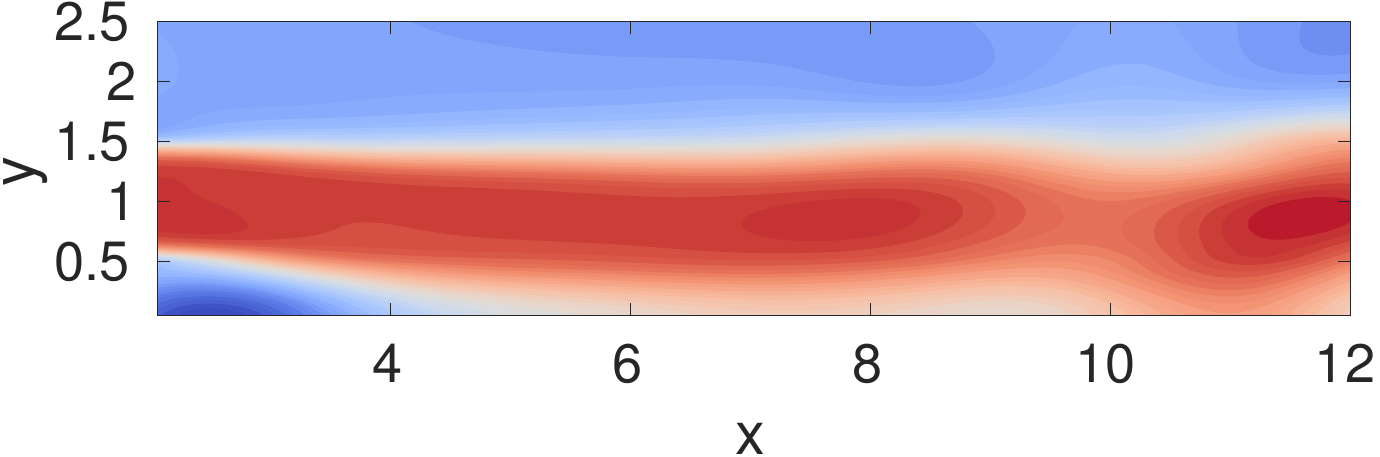}\\
	\vskip-0.2cm
	\textbf{(a)} Case S1\\
	\vskip0.2cm
	
	\includegraphics[width=0.49\textwidth, angle =0]{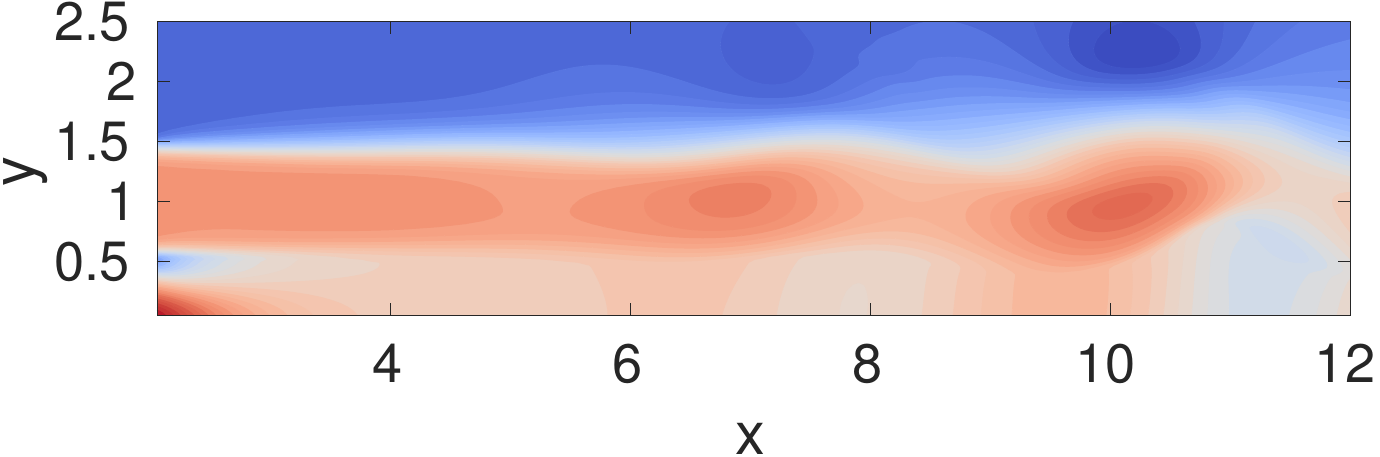}
	\includegraphics[width=0.49\textwidth, angle =0]{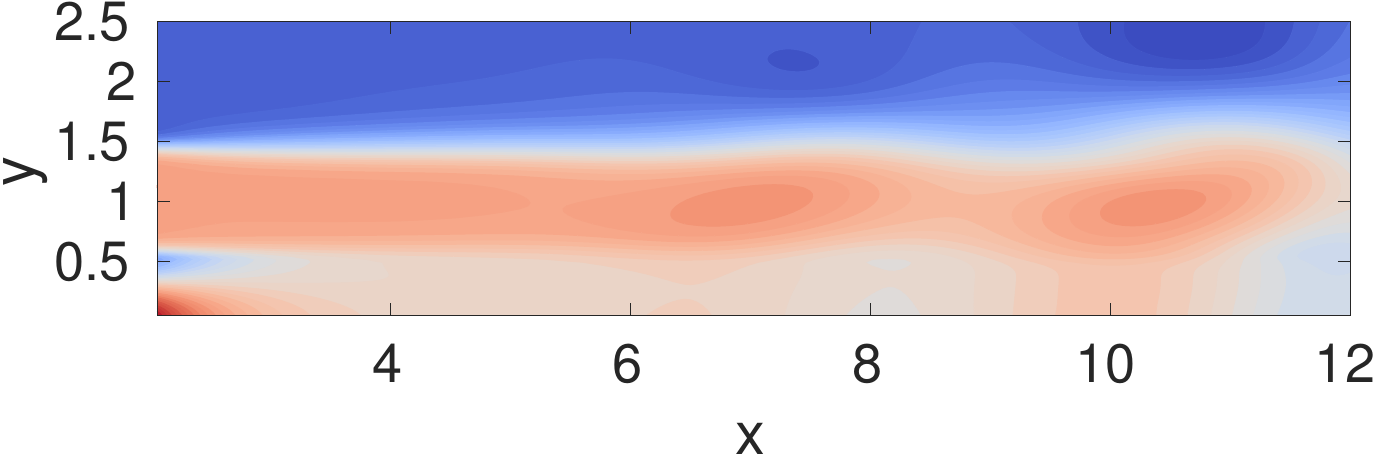}\\
	\vskip-0.2cm
	\textbf{(b)} Case S2\\
	\vskip0.2cm
	
	\includegraphics[width=0.49\textwidth, angle =0]{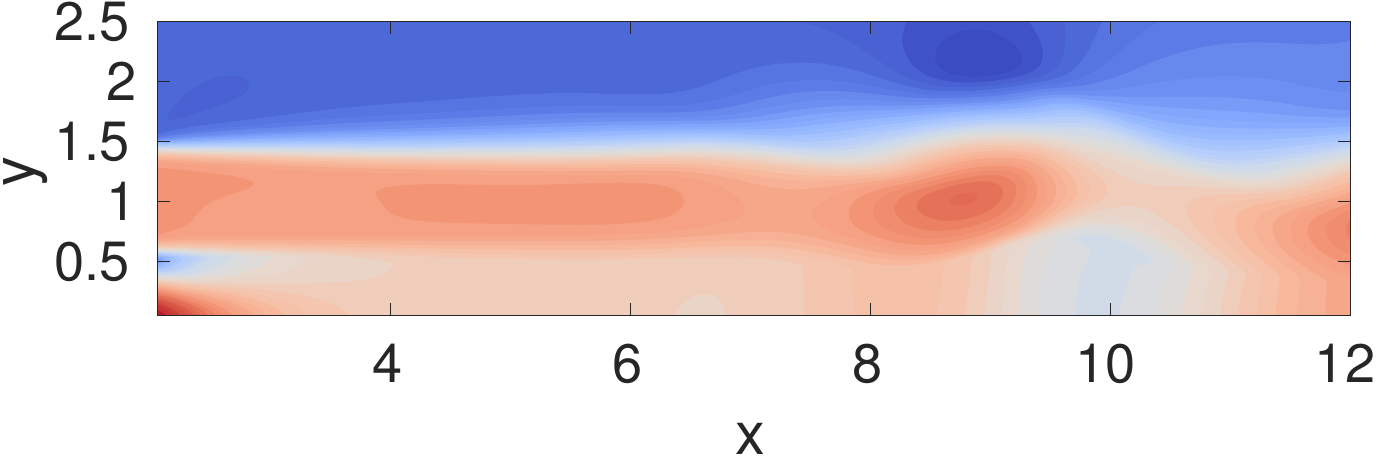}
	\includegraphics[width=0.49\textwidth, angle =0]{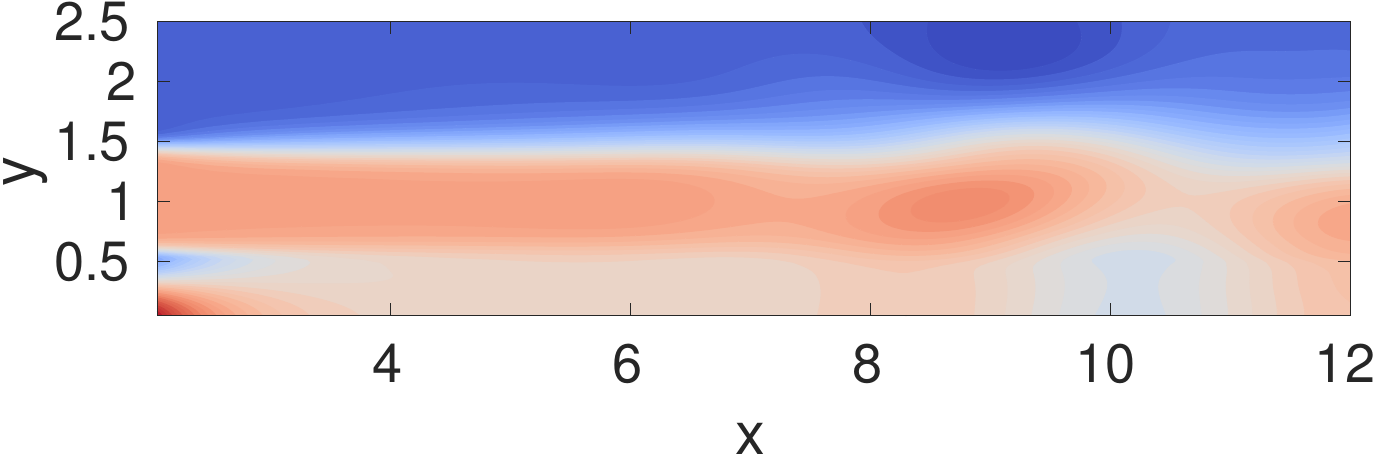}\\
	\vskip-0.2cm
	\textbf{(c)} Case S3\\
	\vskip0.2cm
	
	\caption{Original snapshot (left) and reconstruction using a reduced order model (right) for a specific time instant ($t=207$). Single phase (top), multiphase without surface tension (middle) and multiphase with surface tension (bottom).}
	\label{Reconstruccion normal}
\end{figure}

\begin{figure}[H]
	
	\centering 
	\hskip 1cm Simulation \hskip 4cm Reconstruction\\
	\vskip 0.1cm
	
	\includegraphics[width=0.49\textwidth, angle =0]{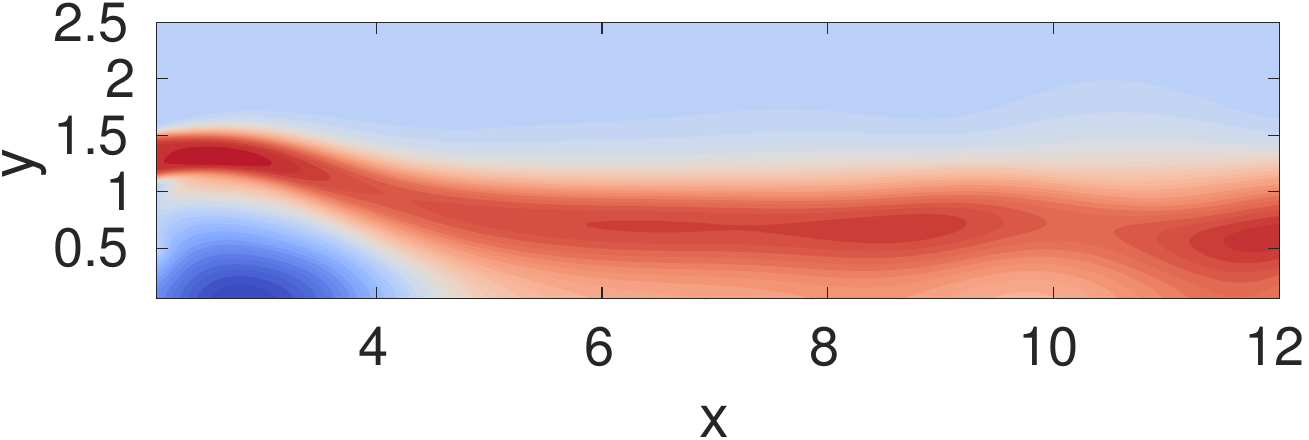}
	\includegraphics[width=0.49\textwidth, angle =0]{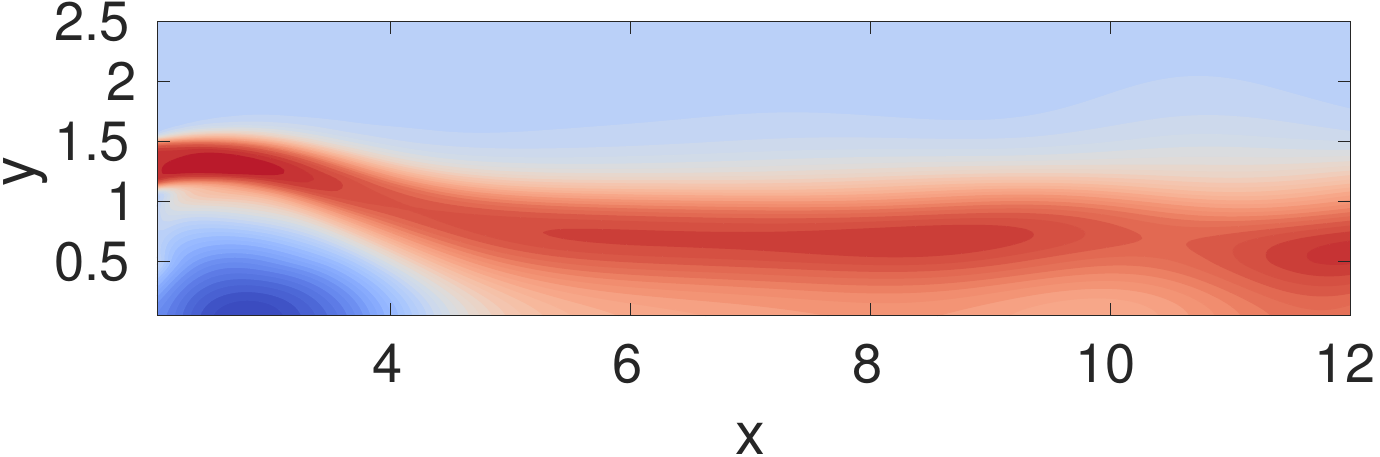}\\
	\vskip-0.2cm
	\textbf{(a)} Case M1\\
	\vskip0.2cm
	
	\includegraphics[width=0.49\textwidth, angle =0]{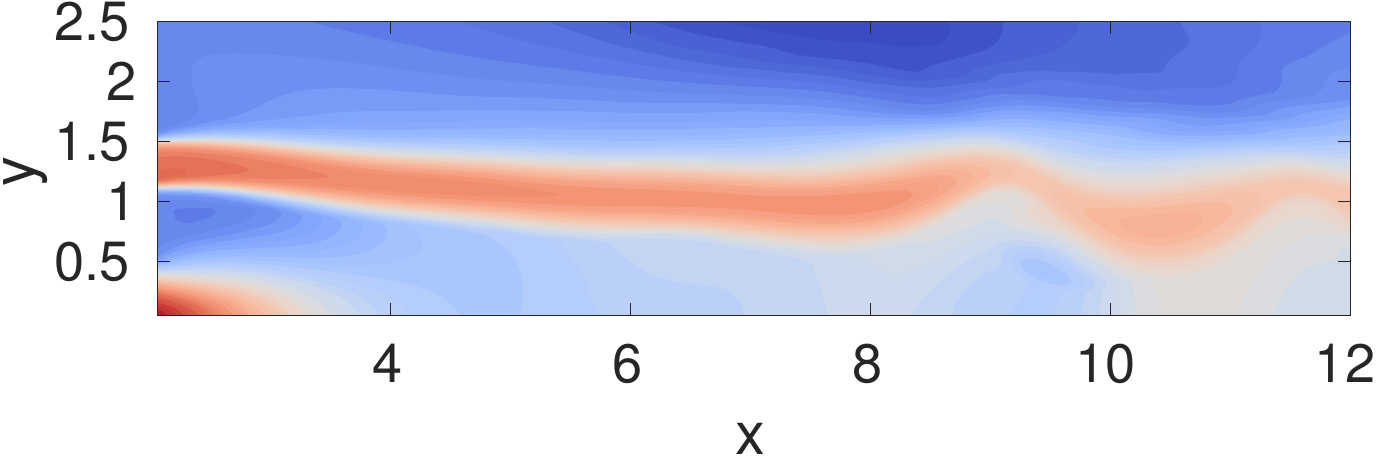}
	\includegraphics[width=0.49\textwidth, angle =0]{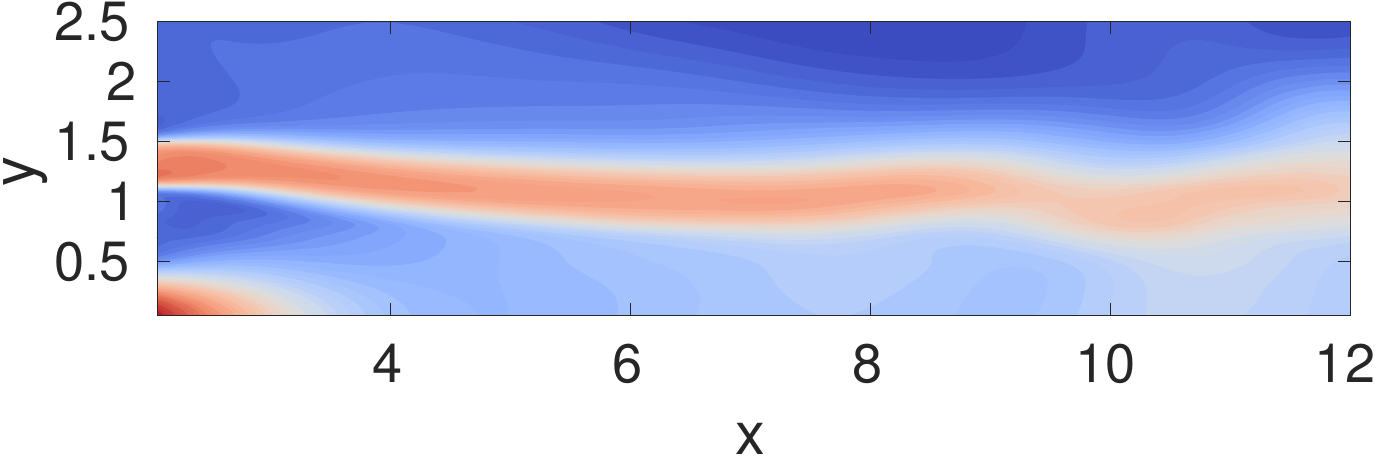}\\
	\vskip-0.2cm
	\textbf{(b)} Case M2\\
	\vskip0.2cm
	
	\includegraphics[width=0.49\textwidth, angle =0]{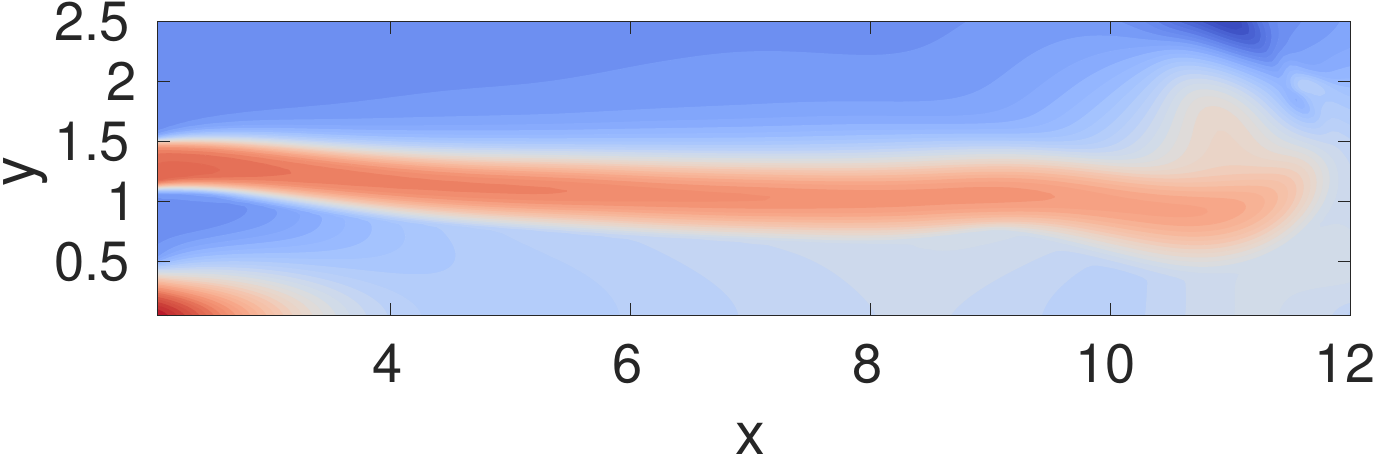}
	\includegraphics[width=0.49\textwidth, angle =0]{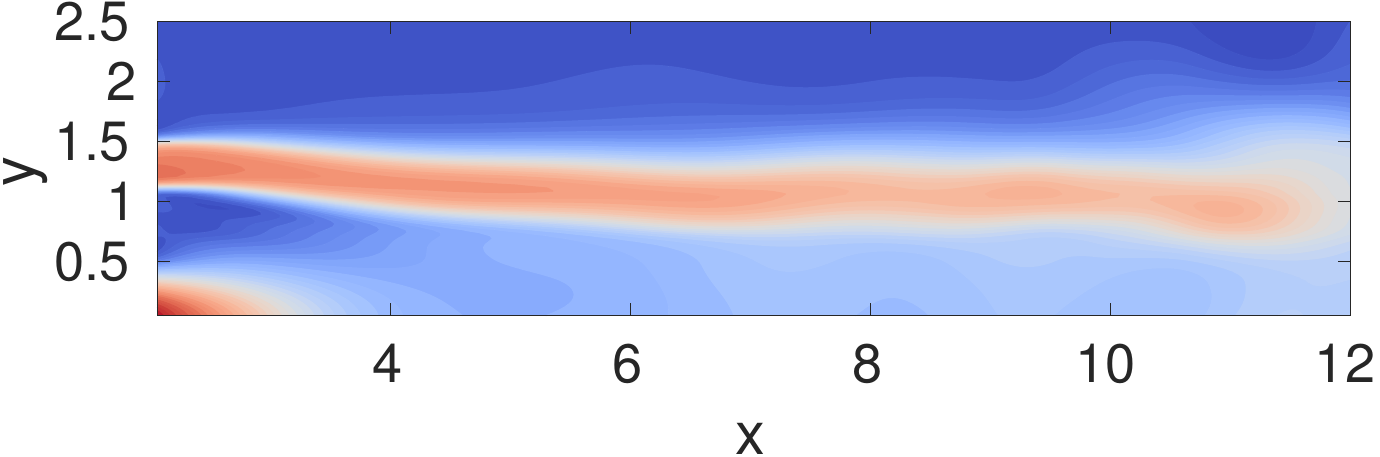}\\
	\vskip-0.2cm
	\textbf{(c)} Case M3\\
	\vskip0.2cm
	
	\caption{Original snapshot (left) and reconstruction using a reduced order model (right) for a specific time instant ($t=300$). Single phase (top), multiphase without surface tension (middle) and multiphase with surface tension (bottom).}
	\label{Reconstruccion bluff}
\end{figure}

Once the modes that will constitute the ROM have been selected, the solution is reconstructed and compared to the original data by calculating the RRMSE in the same way as in eq. (\ref{eq:RMSEtimeNN}) which is shown in Table \ref{RRMSE reconstruccion}. 
\begin{table}[H]
		\centering
		\begin{tabular}{|c|c|c|}
			\hline
			Case & Reconstruction error & Number of modes\\
			\hline 
			S1$_{ROM}$ & 0.069 & 13 \\
			S2$_{ROM}$ & 0.069 & 10\\
			S3$_{ROM}$ & 0.074 & 9\\
			\hline
			M1$_{ROM}$ & 0.032 & 7 \\
			M2$_{ROM}$ & 0.155 & 13 \\
			M3$_{ROM}$ & 0.091 & 16 \\
			\hline
		\end{tabular}
		\label{tab:rrmsesinsigma}
	\caption{RRMSE obtained when reconstructing the flow velocity using a ROM and number of DMD modes involved. Test case nomenclature defined in Table  \ref{tab: Nomenclatura}.}
	\label{RRMSE reconstruccion}
\end{table}
Figures \ref{Reconstruccion normal} and \ref{Reconstruccion bluff} show the reconstructed solution and the original snapshots for all the computed cases. Considering the simple geometry, all the reconstructions have a low RRMSE (about $7\%$) due to the small interaction between both jets, which is very acceptable for a ROM. On the other hand, the modified geometry has been more difficult to reconstruct due to the higher interaction of the jets, which gives place to more complex dynamics and a larger amount of error. In general, the best reconstructions are obtained for the single-phase case. As a summary, HODMD has provided reliable reconstructions in complex dynamics, which is useful to reduce the complexity of dynamics and the amount of noise, which is crucial when using the neural networks, as will be explained in the next sections.

\section{Predictions via artificial neural networks}\label{sec: results ann}
In this section we show the predictions obtained from the NNs described in Section \ref{sec: Neural Networks}. As was stated in Section \ref{sec: preprocessing_data}, the training data set is obtained by subtracting the single-phase flow from the two-phase one (Tables \ref{tab: train_cases_simple_geometry} and \ref{tab: train_cases_modified_geometry}). Once the NN is trained with these eight data sets, predictions are generated. Lastly, we add the respective single-phase flow to each prediction in order to undo the previous subtraction. To generate predictions we input ten samples to the NNs $\{v_{t}, v_{t-1}, \dots, v_{t-9}\}$ and obtained the forecasting of the following two $\{v_{t+1}, v_{t+2}\}$.

\subsection{Simple Geometry: geometry without bluff body (S2, S3, S2$_{ROM}$ and S3$_{ROM}$).}
This section compares the predictions obtained from NNs when they are trained with data sets from Table \ref{tab: train_cases_simple_geometry} (i.e., data sets corresponding to geometry without bluff body). In this meaning, predictions will be compared with cases S2, S3, S2$_{ROM}$ and S3$_{ROM}$ (Table \ref{tab: orig_hodmd_nomenclatura}).

\begin{figure}[H]
	\centering
	\begin{tabular}{cccc}
		Sample & Simulation & RNN & CNN\\
		$t = 244$&
		\includegraphics[width=0.27\textwidth, angle=0]{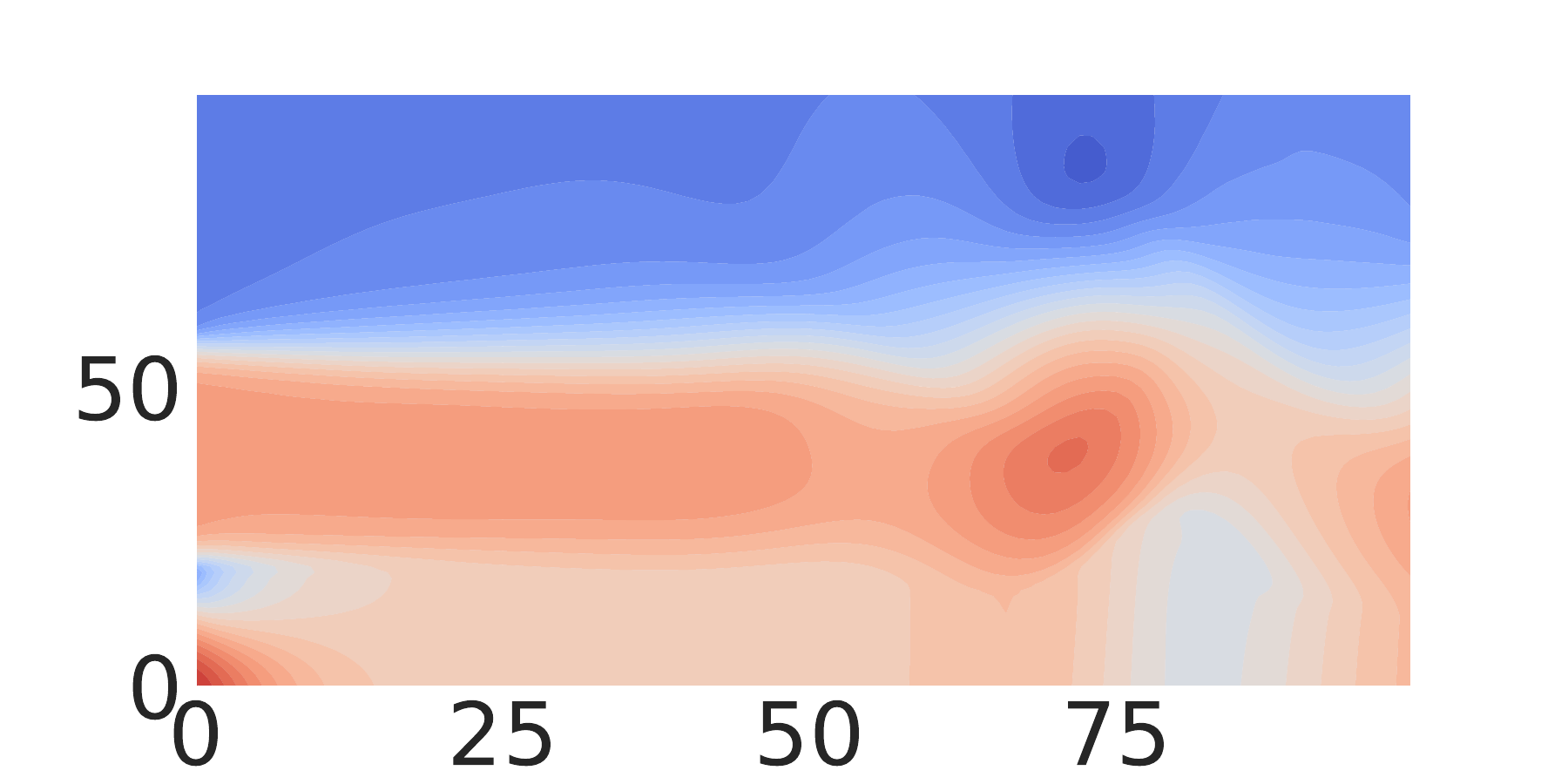}&
		\includegraphics[width=0.27\textwidth, angle=0]{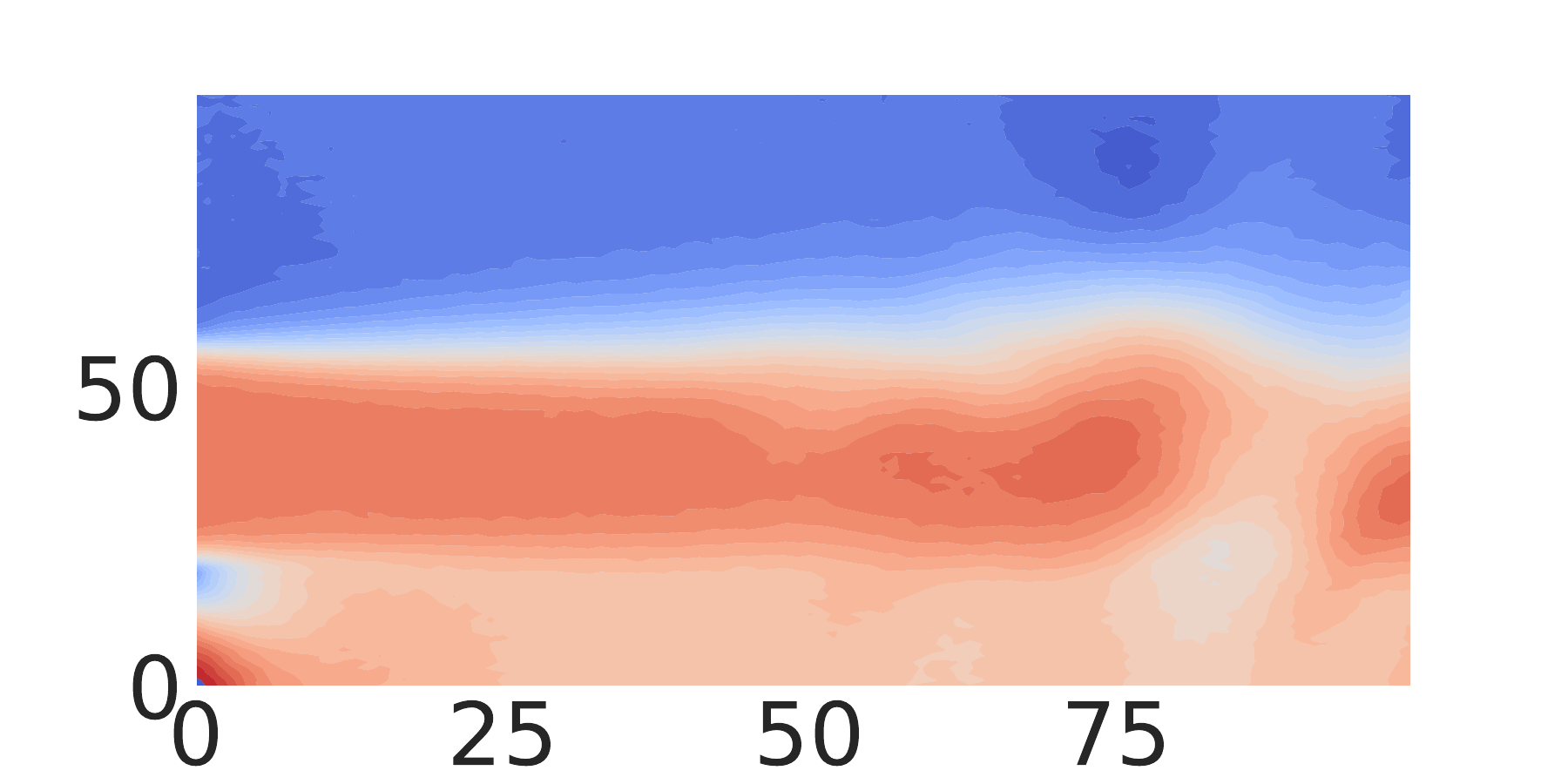}&
		\includegraphics[width=0.27\textwidth, angle=0]{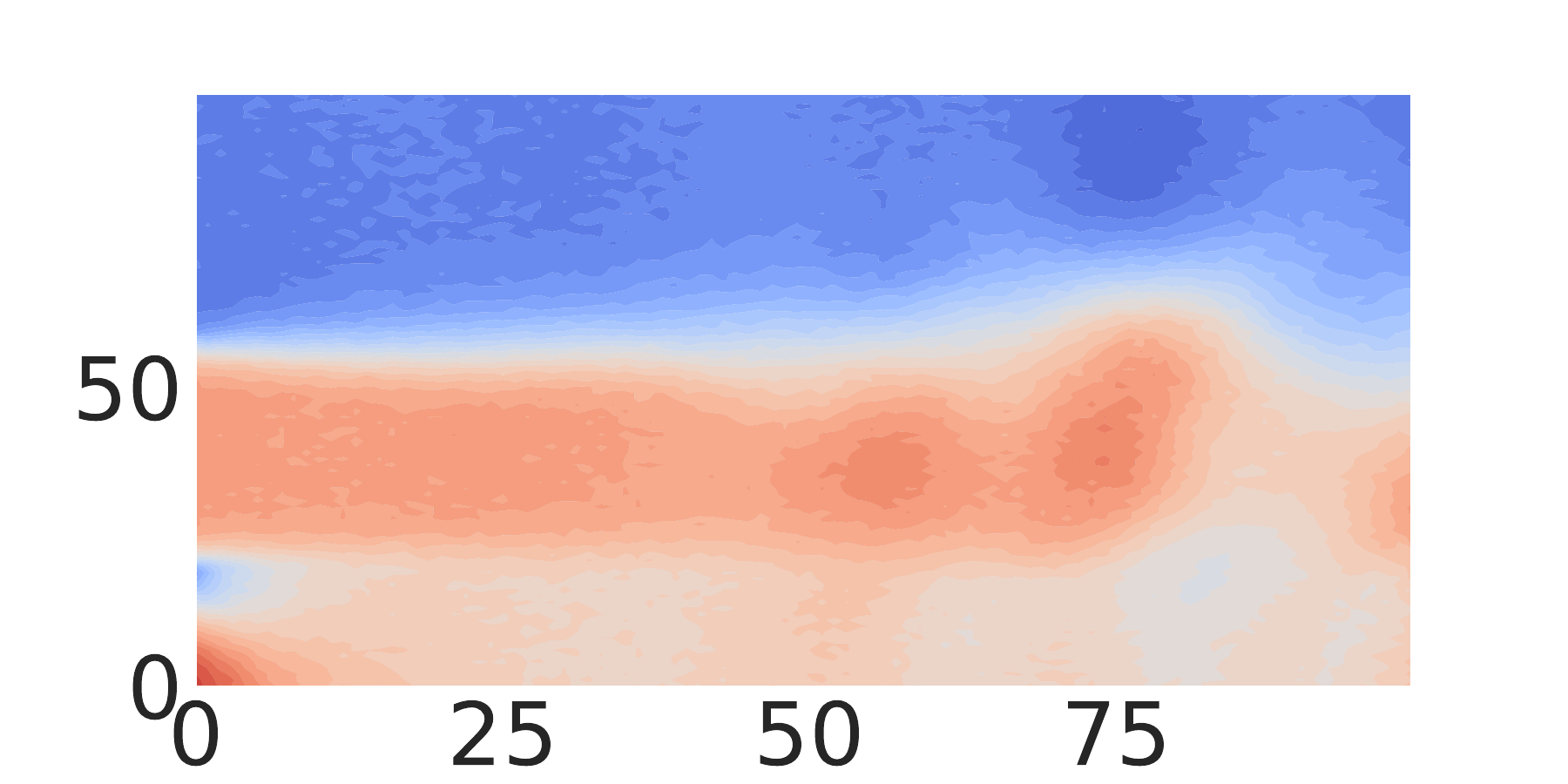}\\
		$t=245$&
		\includegraphics[width=0.27\textwidth, angle=0]{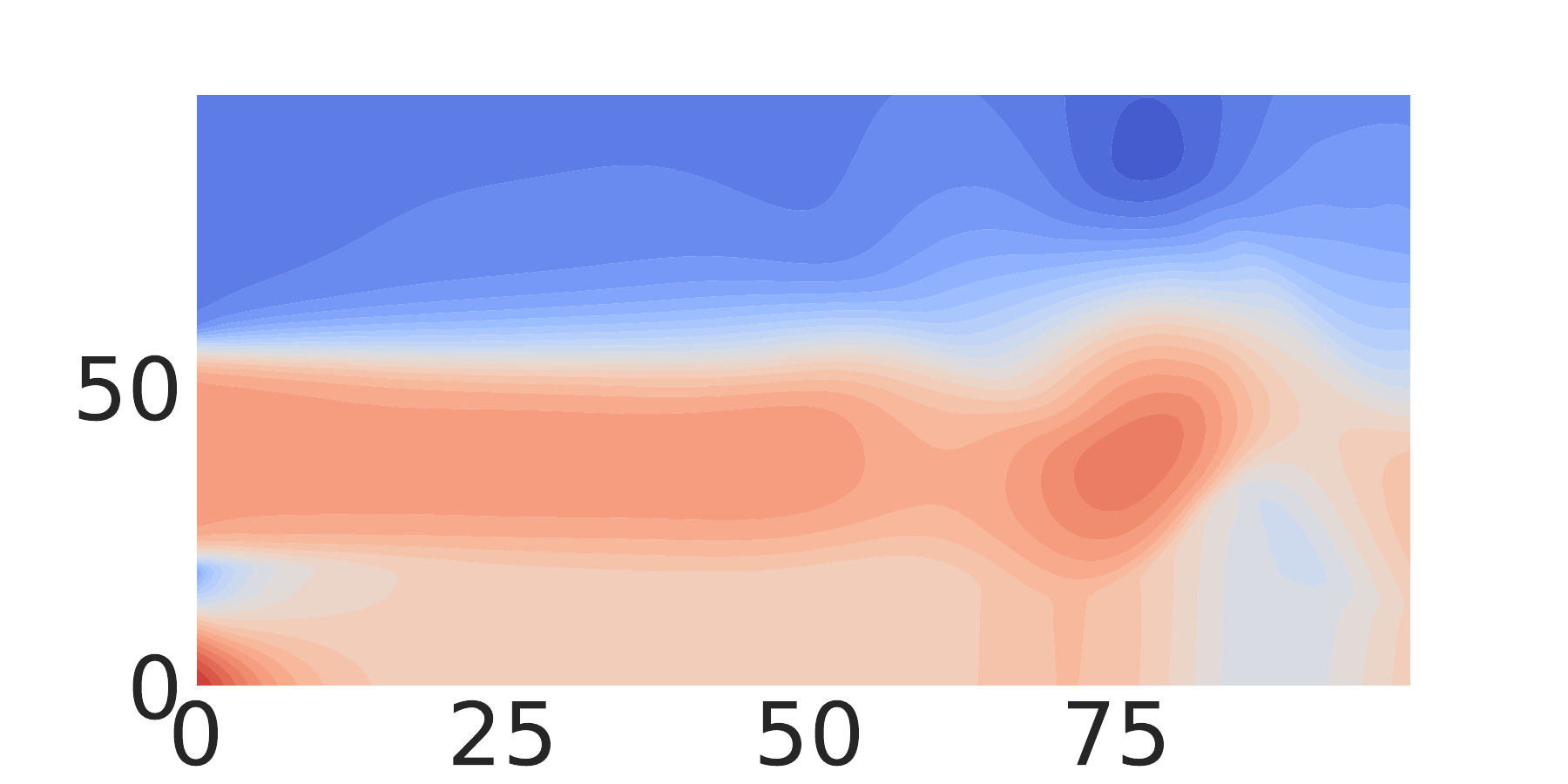}&
		\includegraphics[width=0.27\textwidth, angle=0]{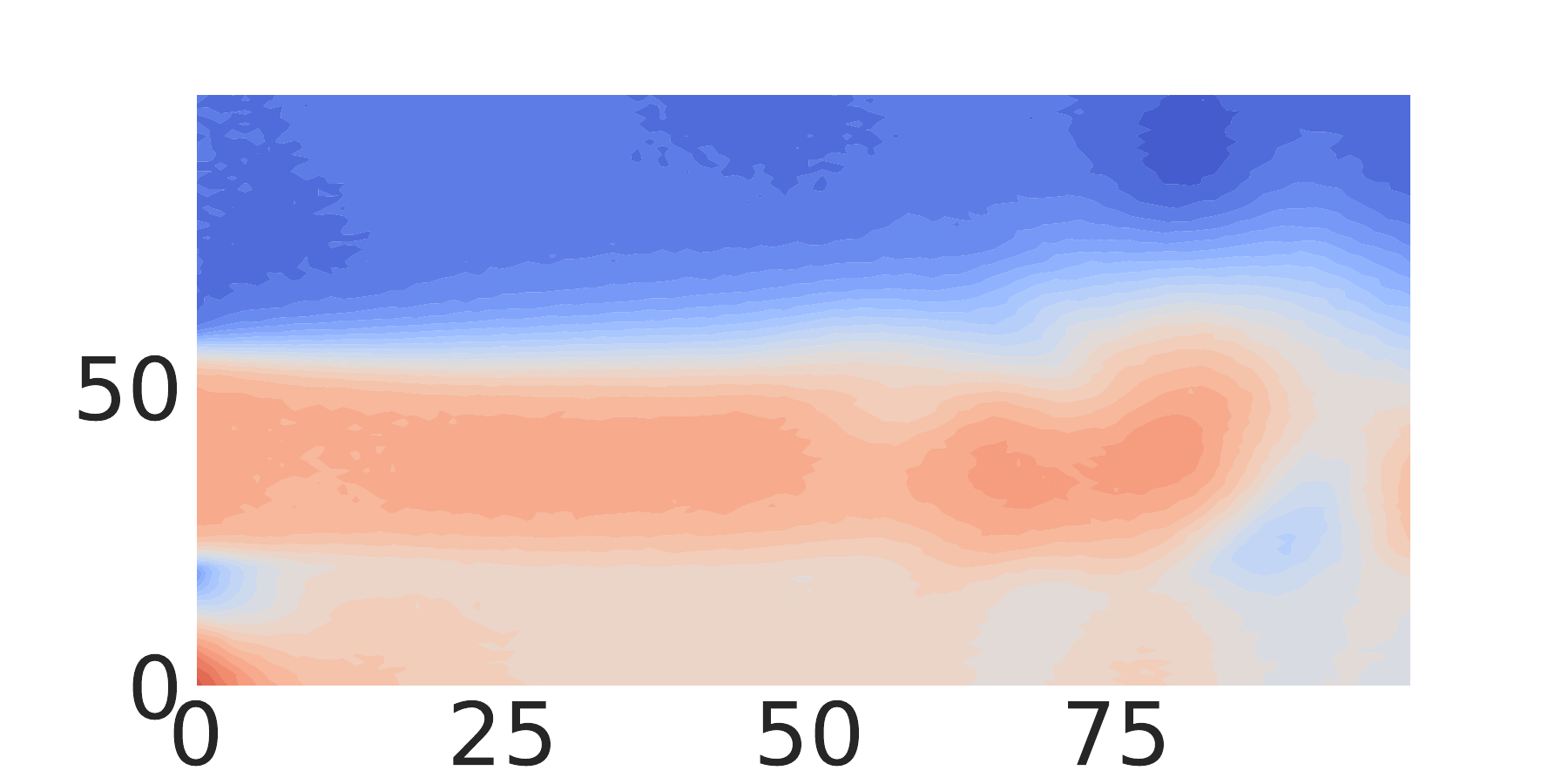}&
		\includegraphics[width=0.27\textwidth, angle=0]{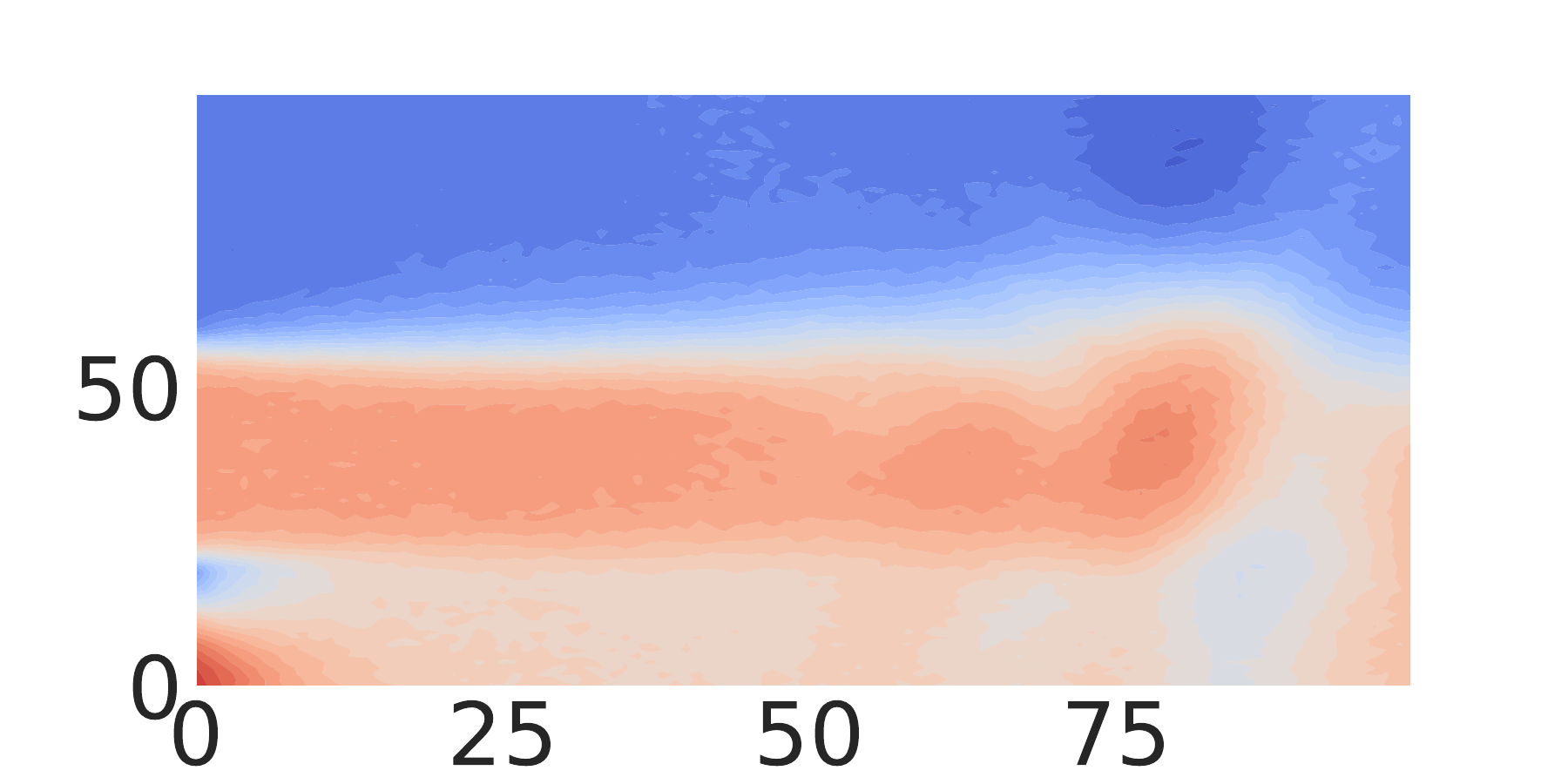}\\
		$t = 259$&
		\includegraphics[width=0.27\textwidth, angle=0]{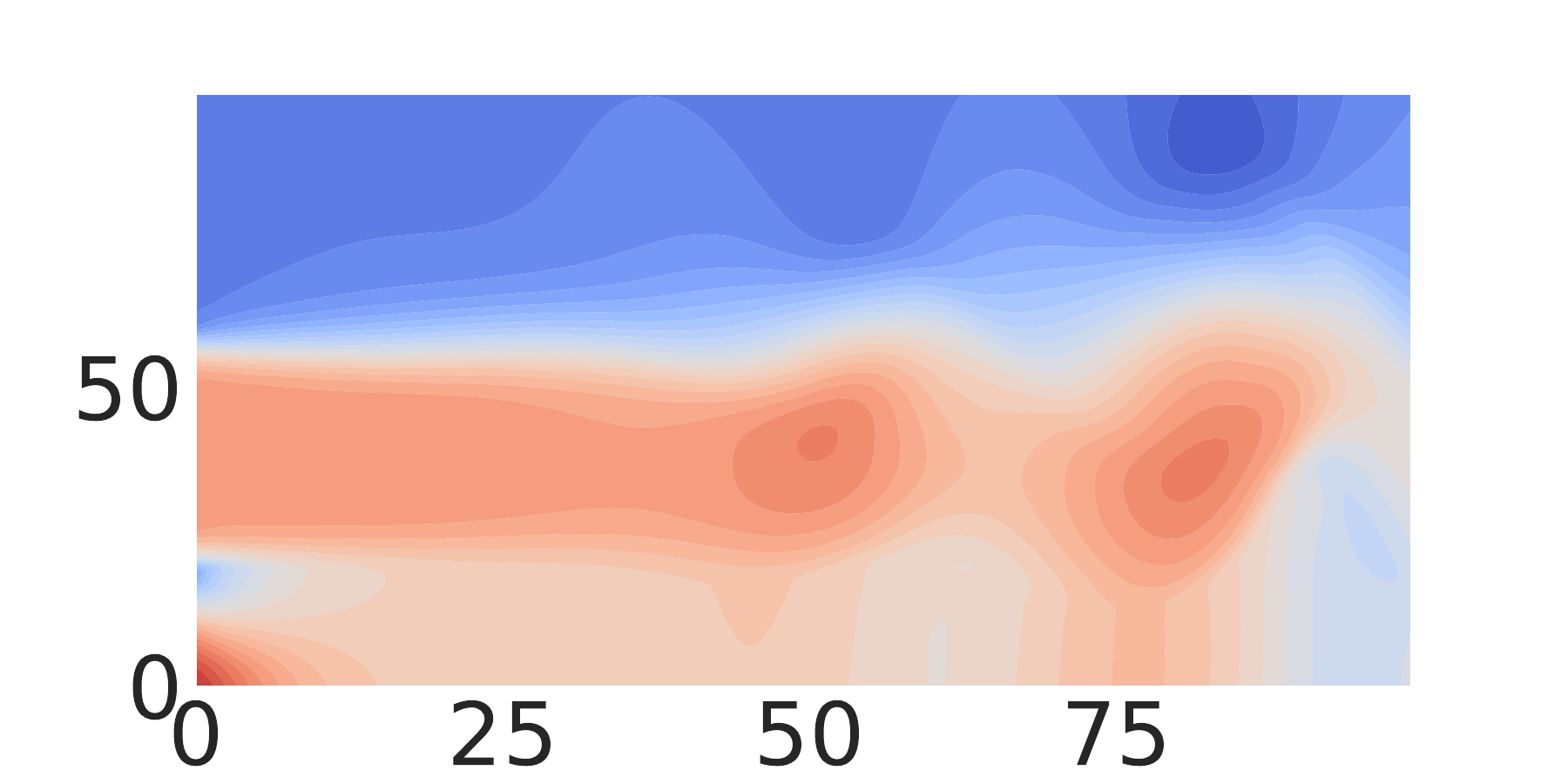}&
		\includegraphics[width=0.27\textwidth, angle=0]{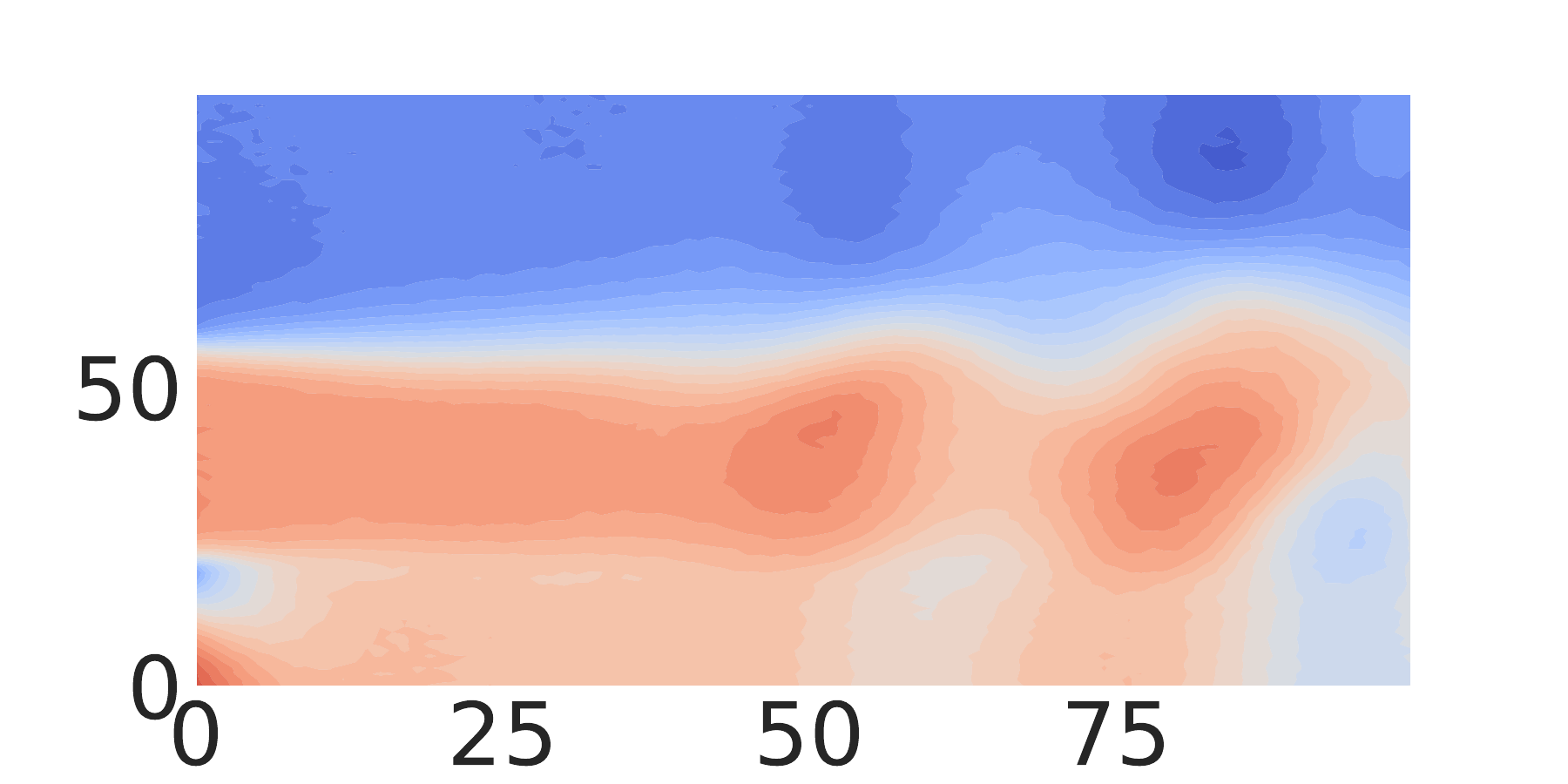}&
		\includegraphics[width=0.27\textwidth, angle=0]{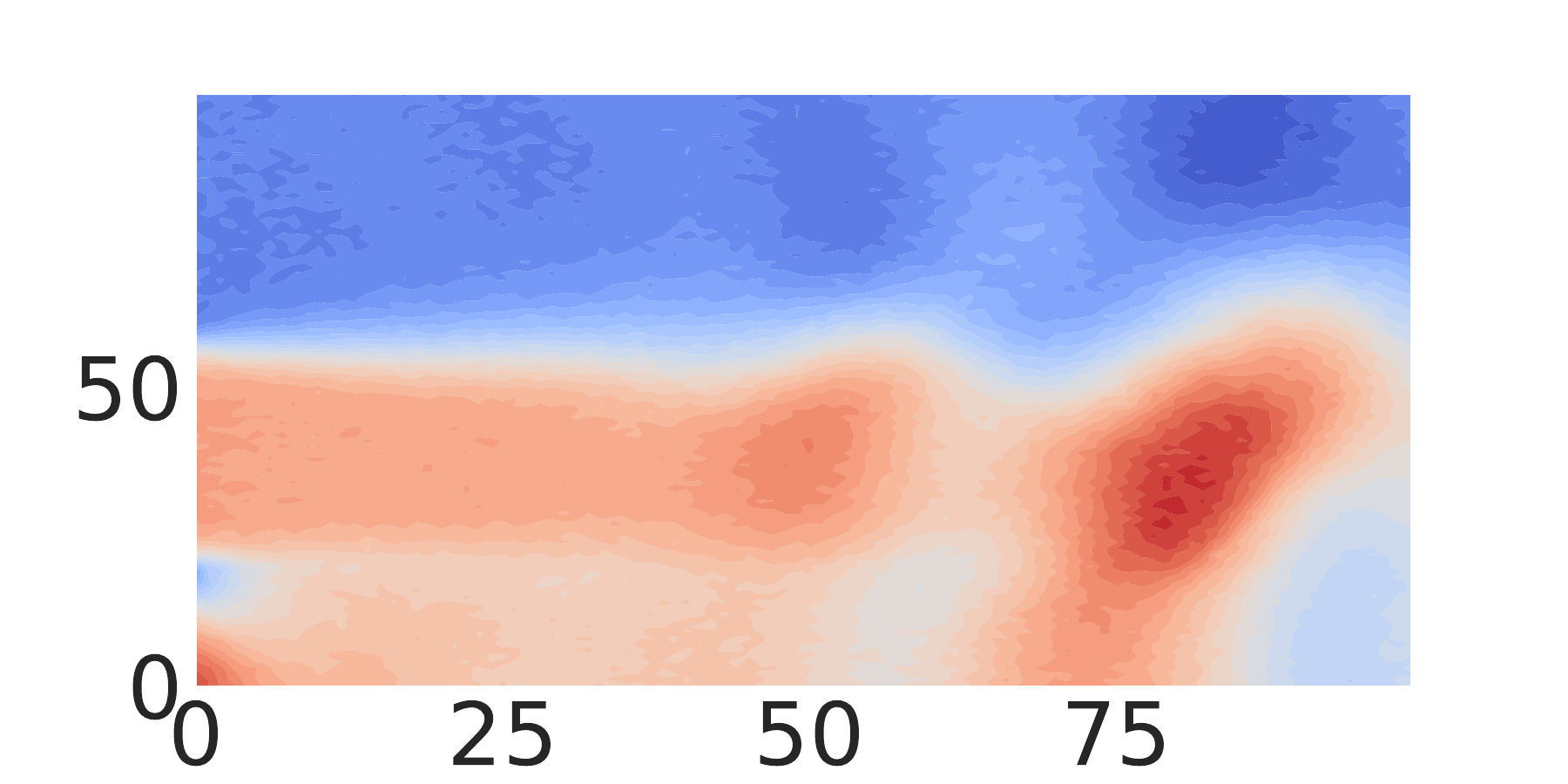}\\
		$t=260$&
		\includegraphics[width=0.27\textwidth, angle=0]{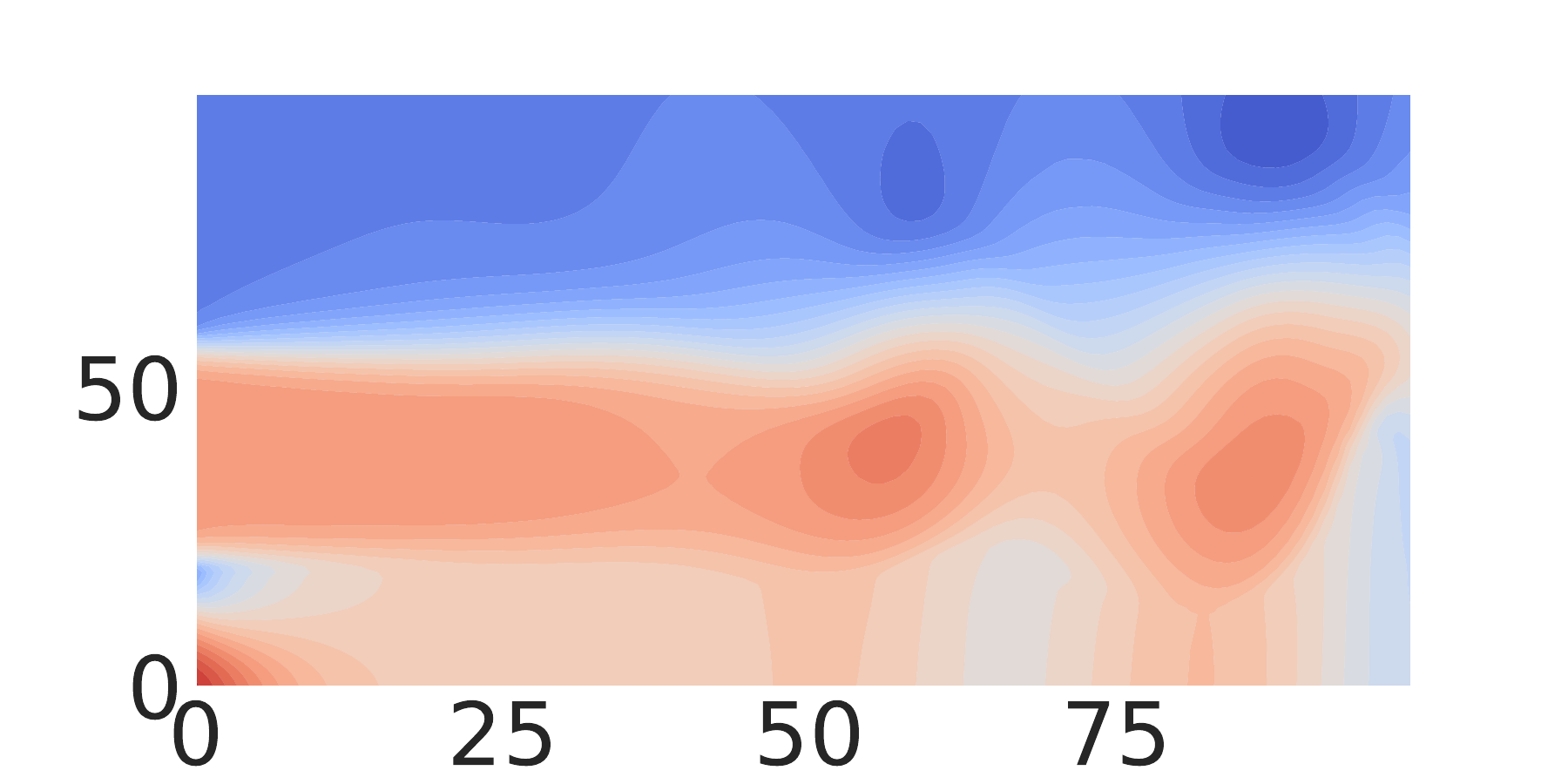}&
		\includegraphics[width=0.27\textwidth, angle=0]{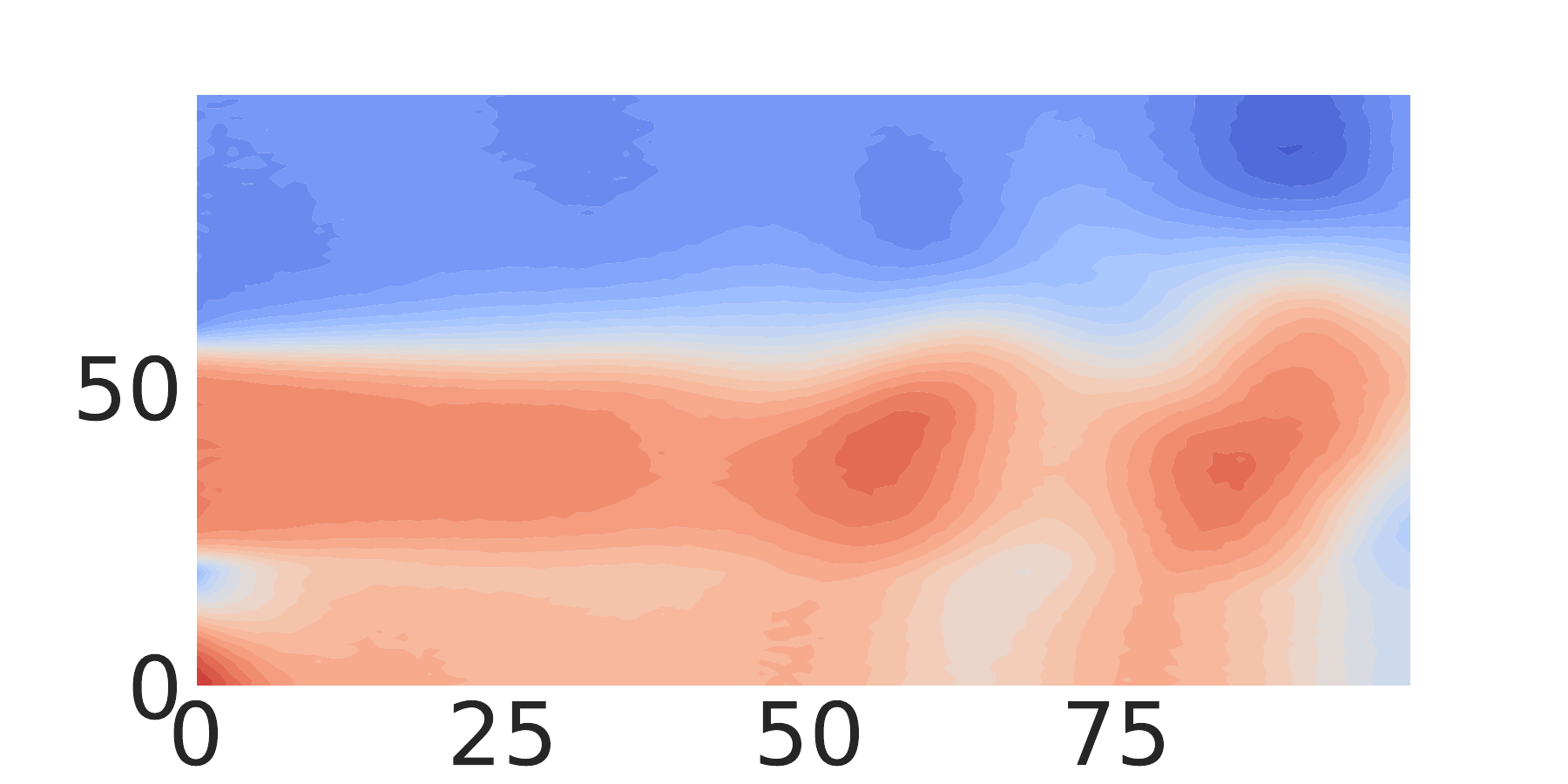}&
		\includegraphics[width=0.27\textwidth, angle=0]{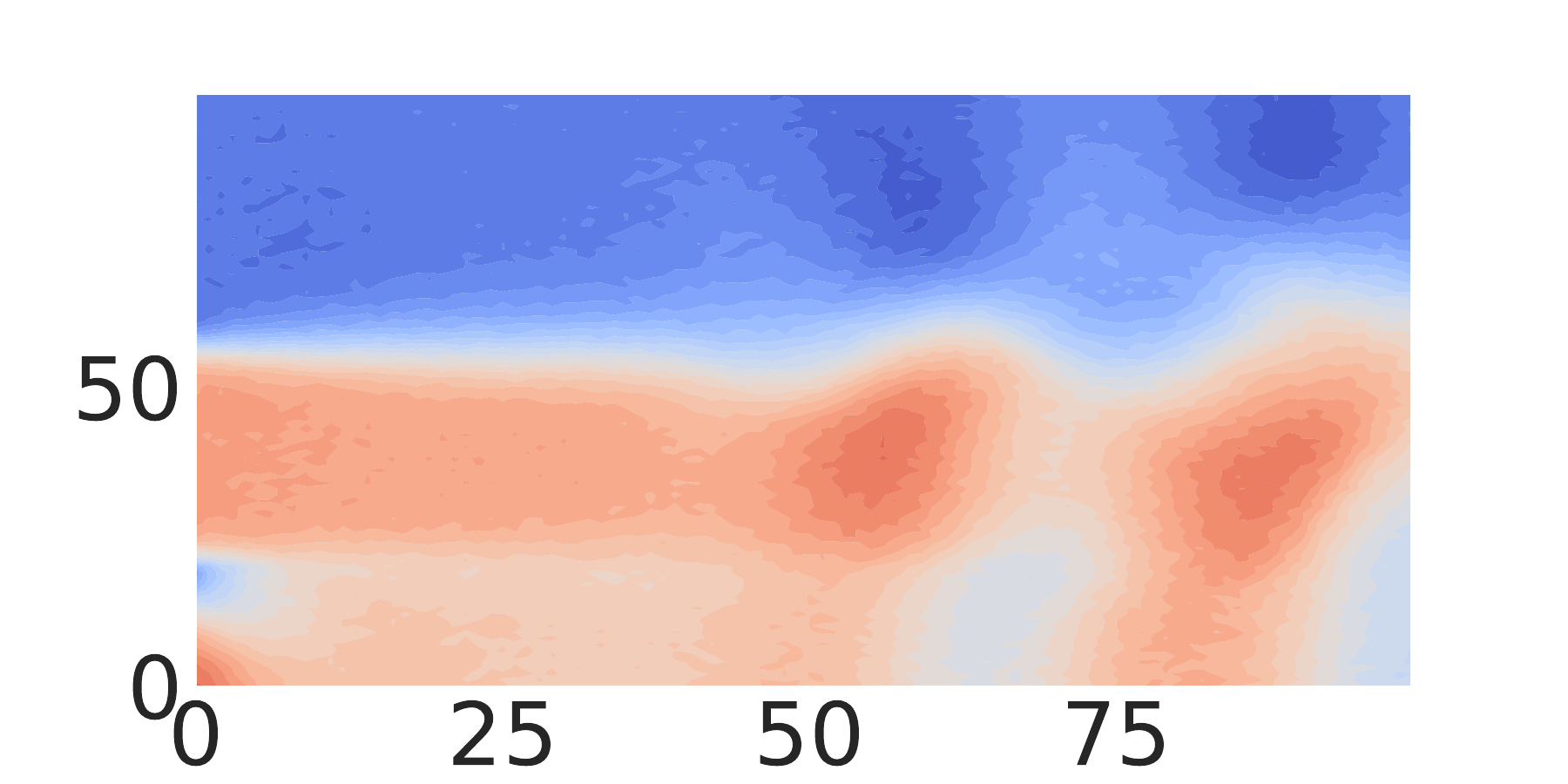}
	\end{tabular}
	\caption{From left to right: Snapshots from original data set, prediction of RNN model and prediction of CNN model, respectively, corresponding to case S2. To generate these predictions the following samples were sent to both NNs; $\{v_{243}, v_{242}, \dots, v_{234}\}$ and $\{v_{258}, v_{257}, \dots, v_{249}\}$. Note that all these samples belong to the test set, Table \ref{tab:ML3}.}
	\label{fig: ann_normal_sts_sim}
\end{figure}

\begin{figure}[H]
	\centering
	\begin{tabular}{lccc}
		Sample & Simulation & RNN & CNN\\
		$t = 244$&
		\includegraphics[width=0.27\textwidth, angle=0]{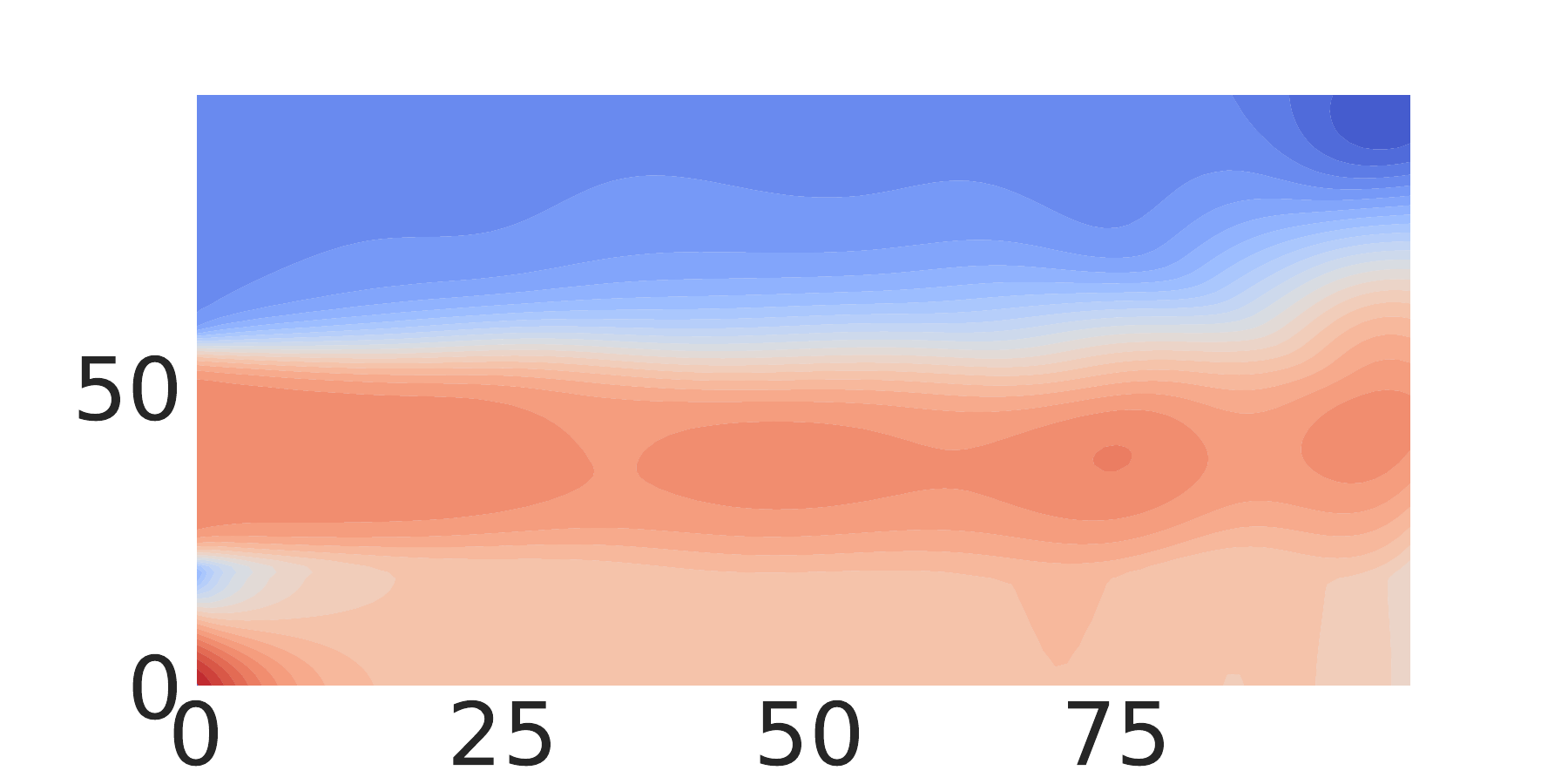}&
		\includegraphics[width=0.27\textwidth, angle=0]{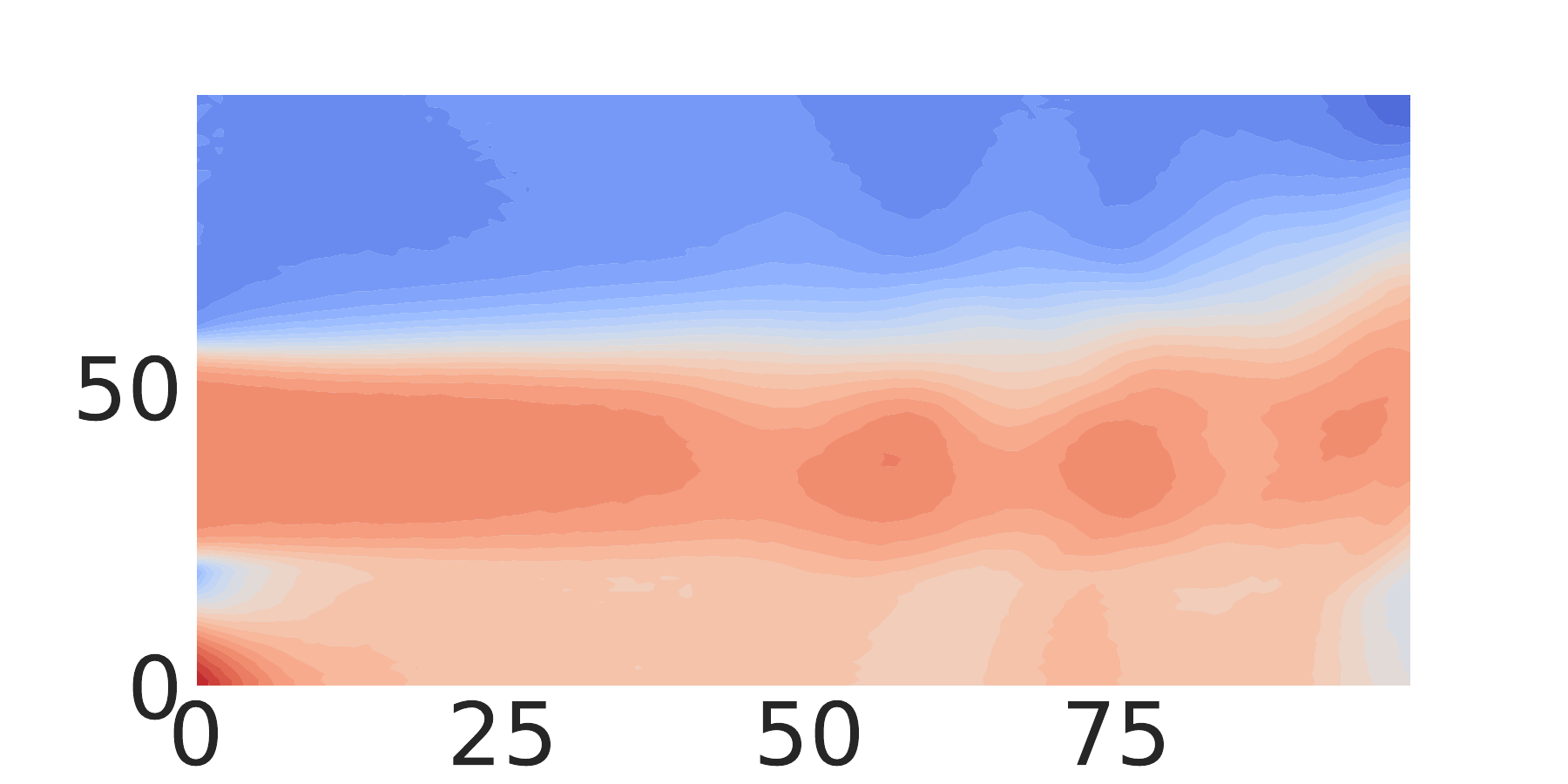}&
		\includegraphics[width=0.27\textwidth, angle=0]{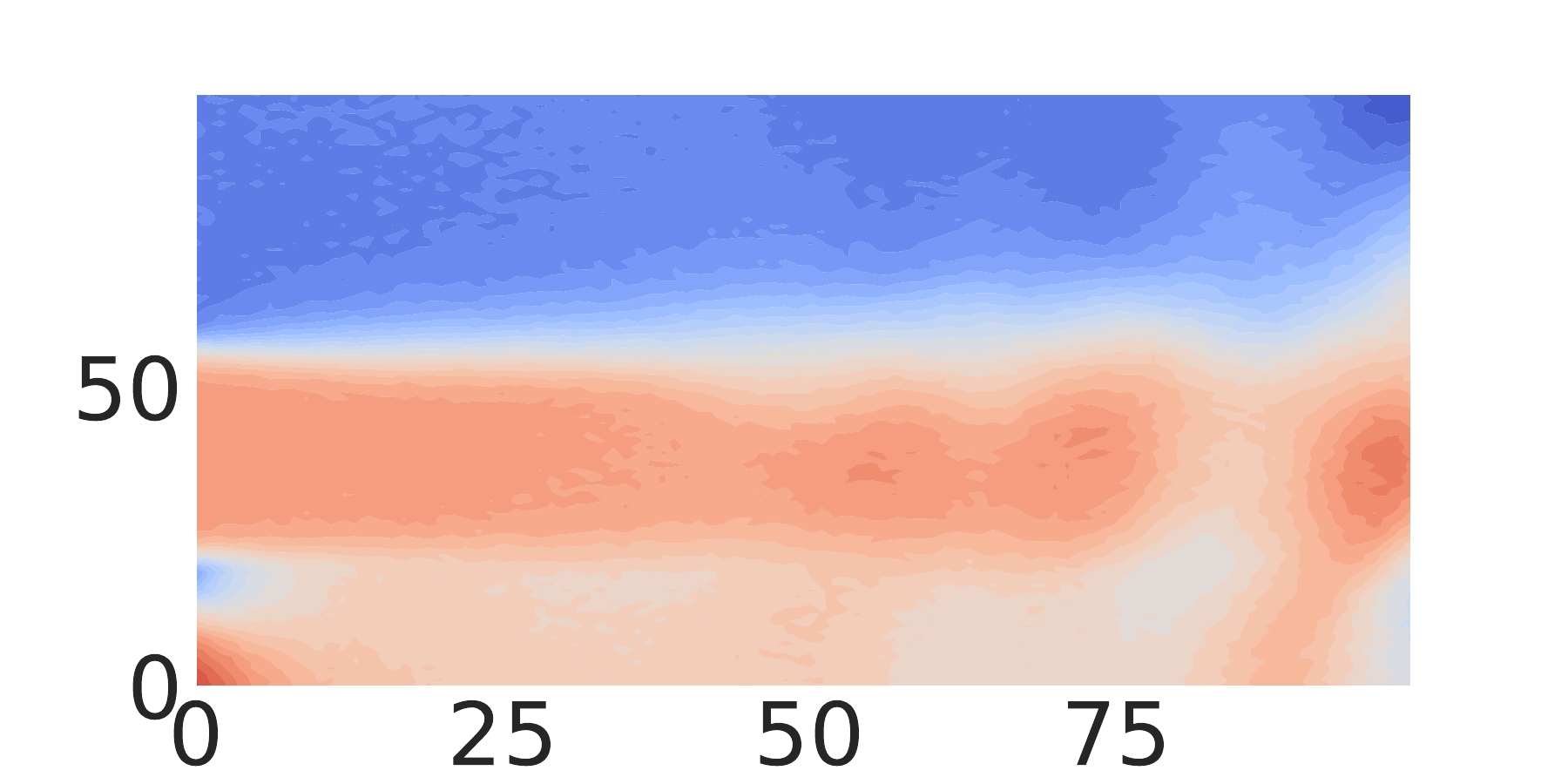}\\
		$t=245$&
		\includegraphics[width=0.27\textwidth, angle=0]{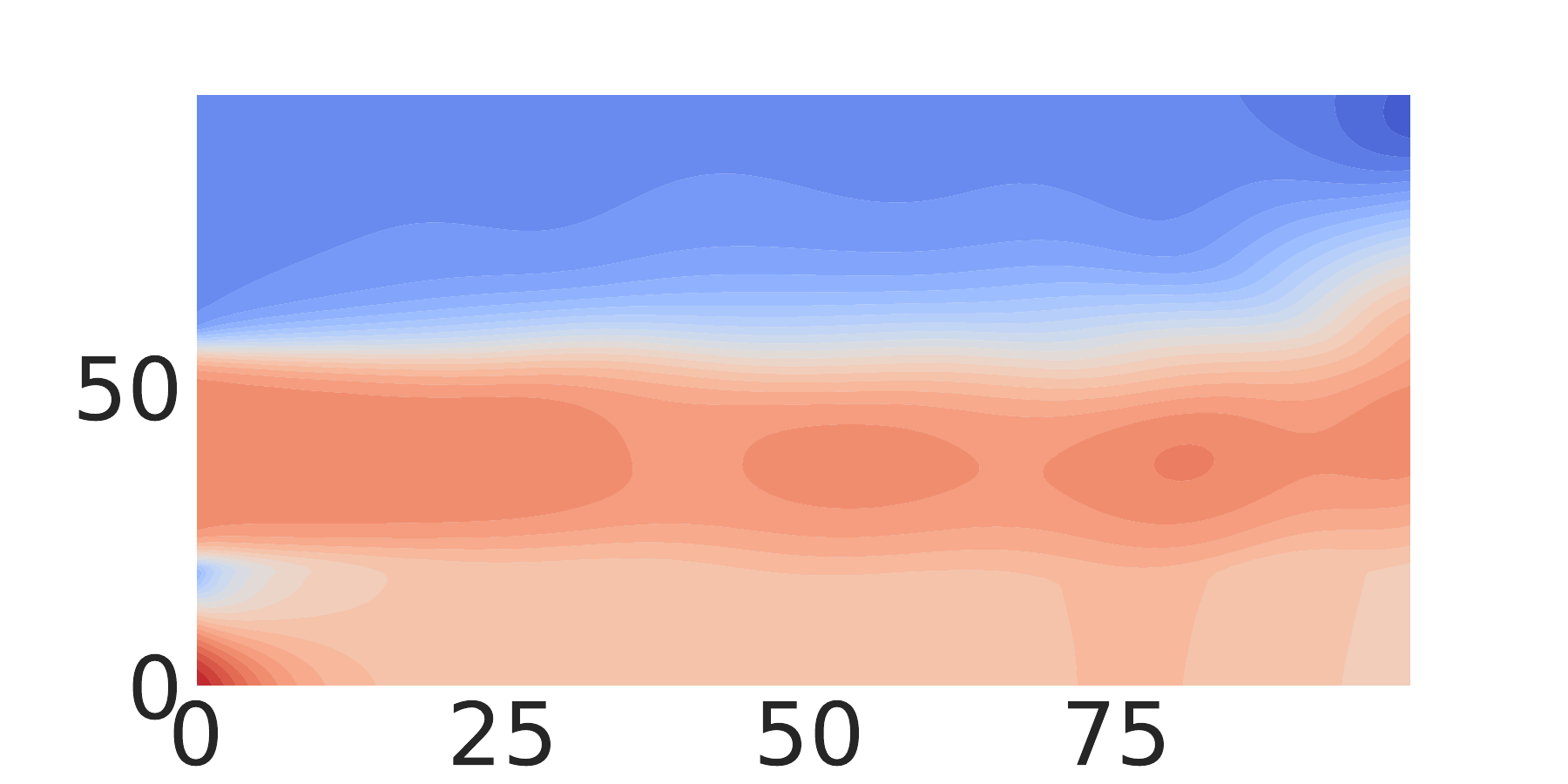}&
		\includegraphics[width=0.27\textwidth, angle=0]{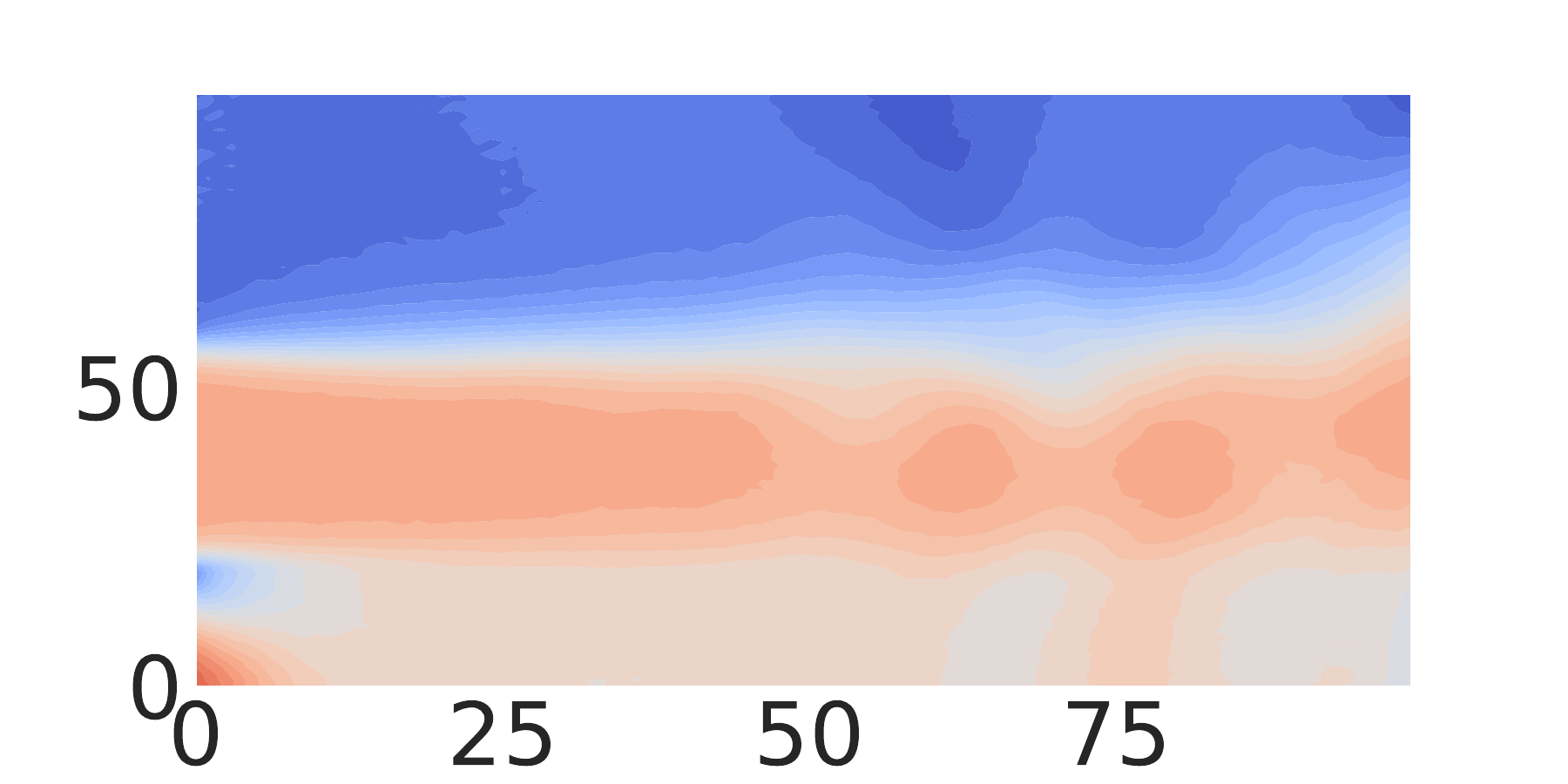}&
		\includegraphics[width=0.27\textwidth, angle=0]{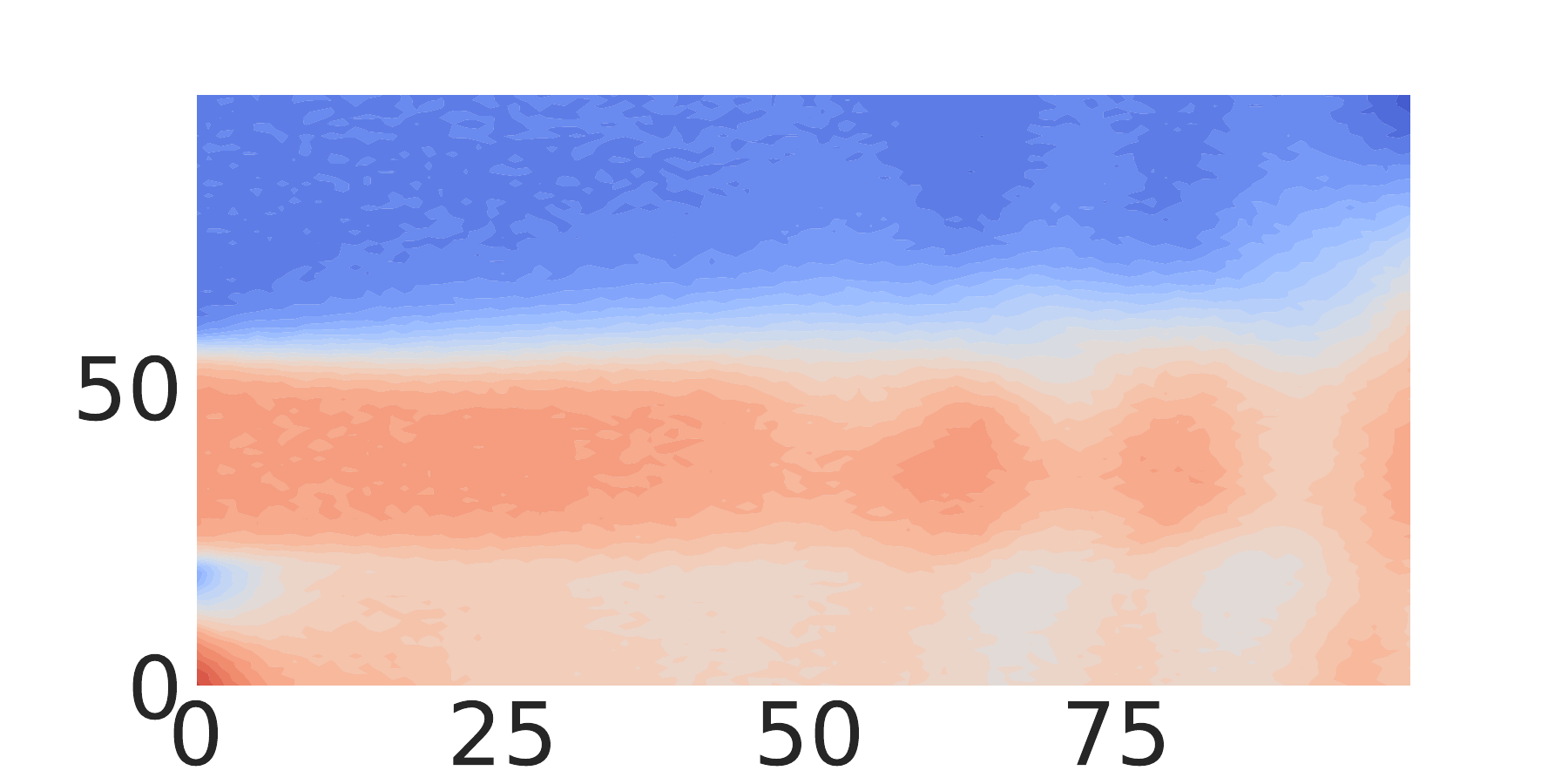}\\
		$t = 259$&
		\includegraphics[width=0.27\textwidth, angle=0]{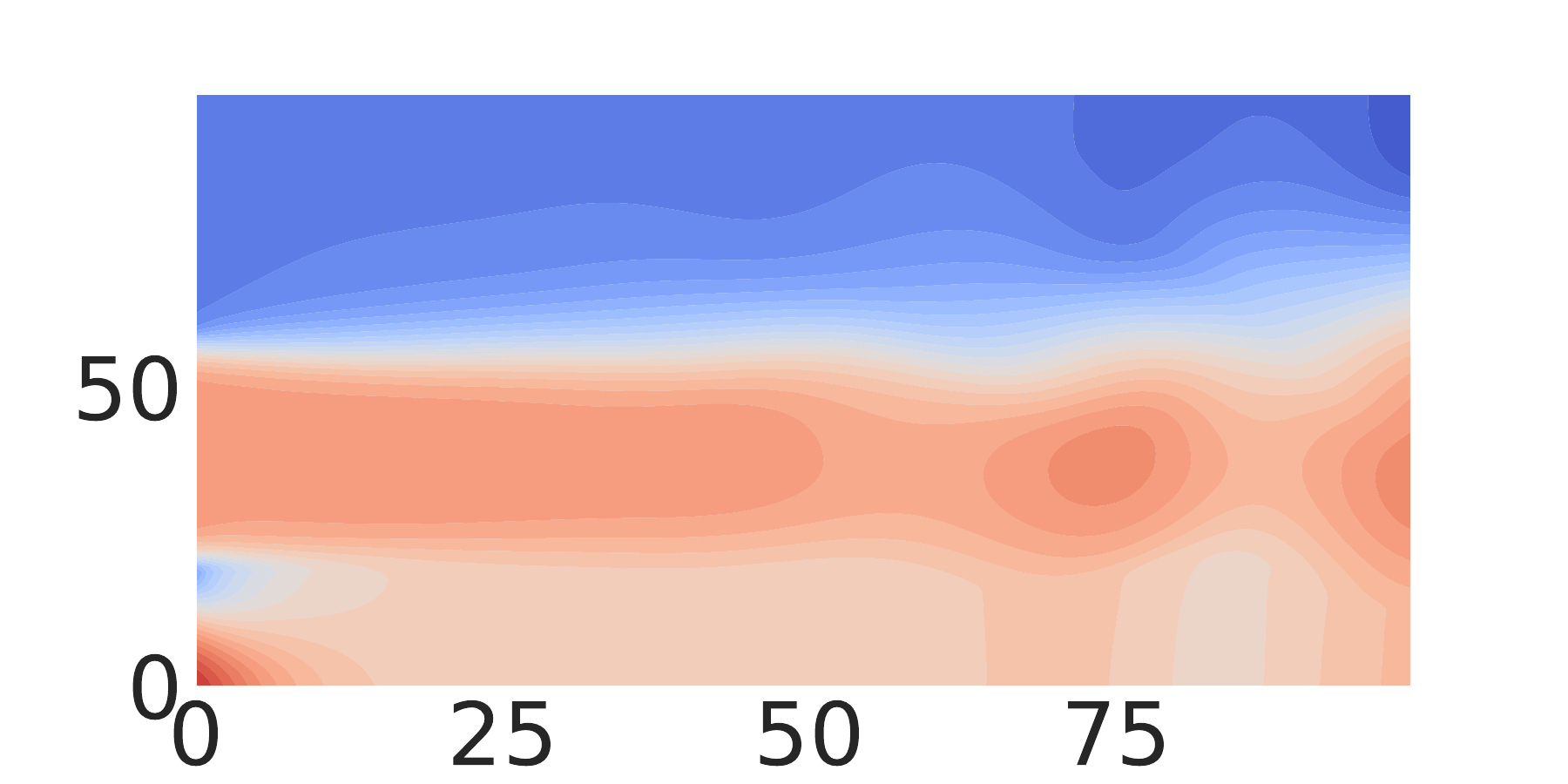}&
		\includegraphics[width=0.27\textwidth, angle=0]{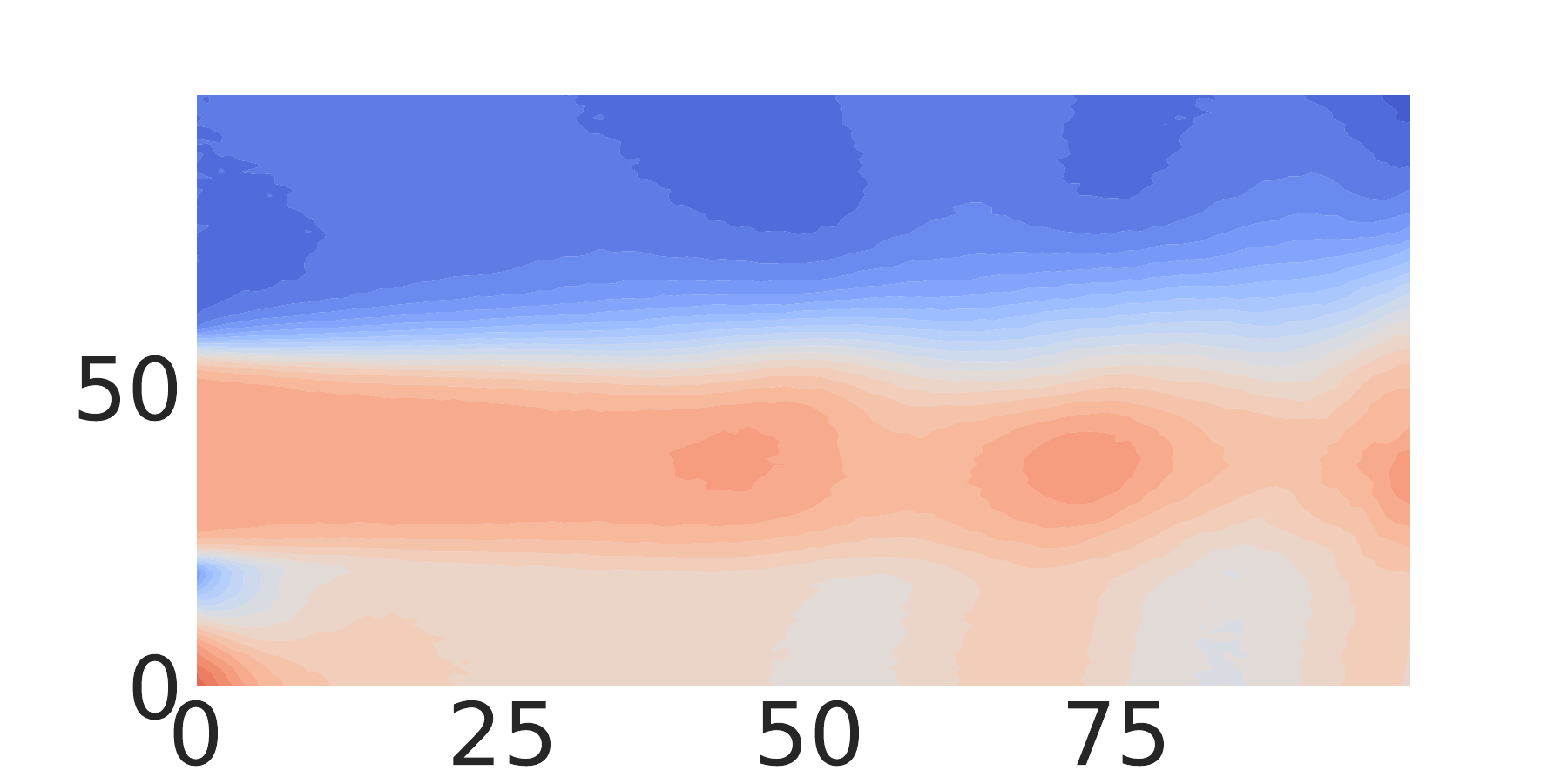}&
		\includegraphics[width=0.27\textwidth, angle=0]{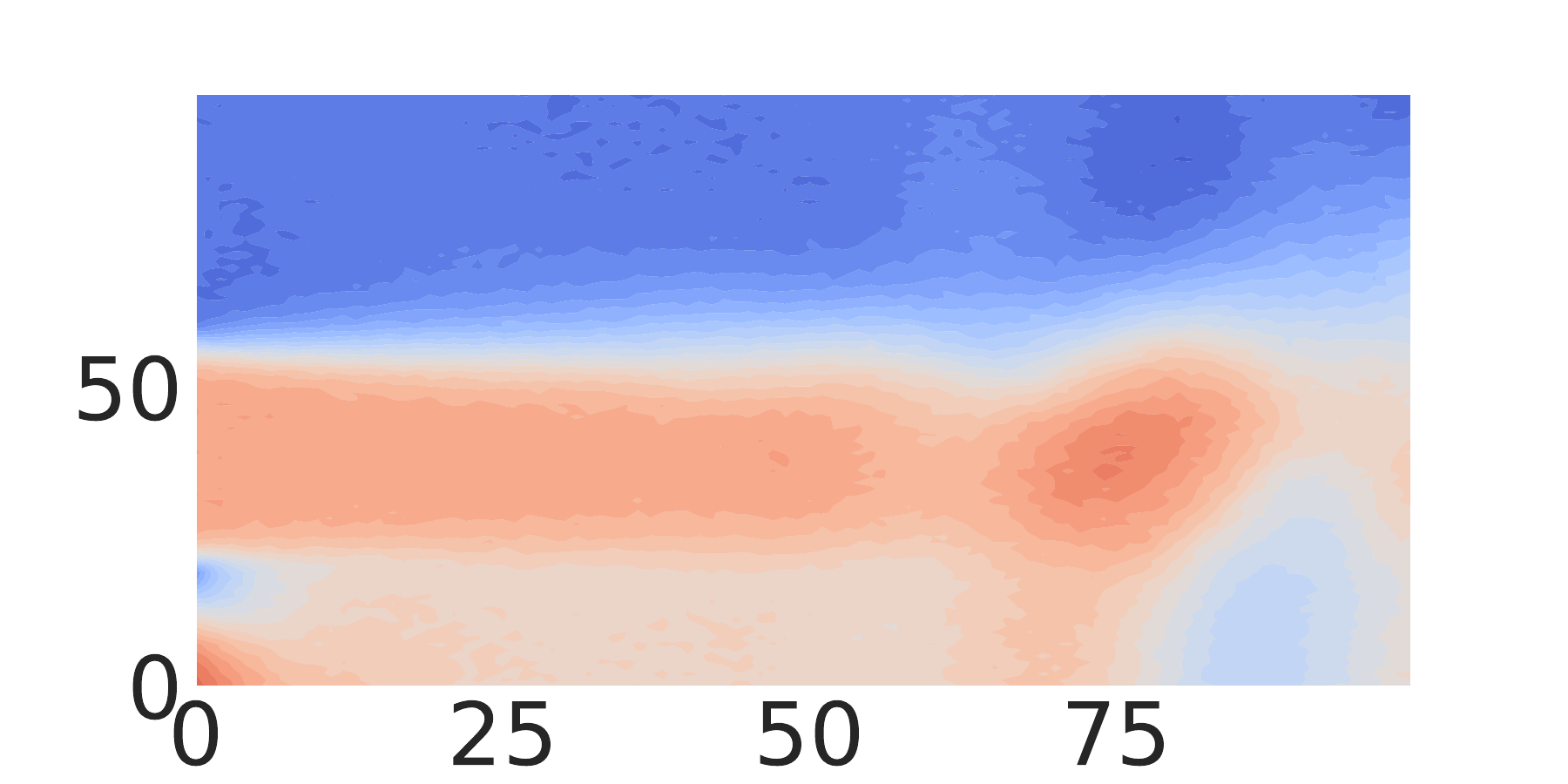}\\
		$t=260$&
		\includegraphics[width=0.27\textwidth, angle=0]{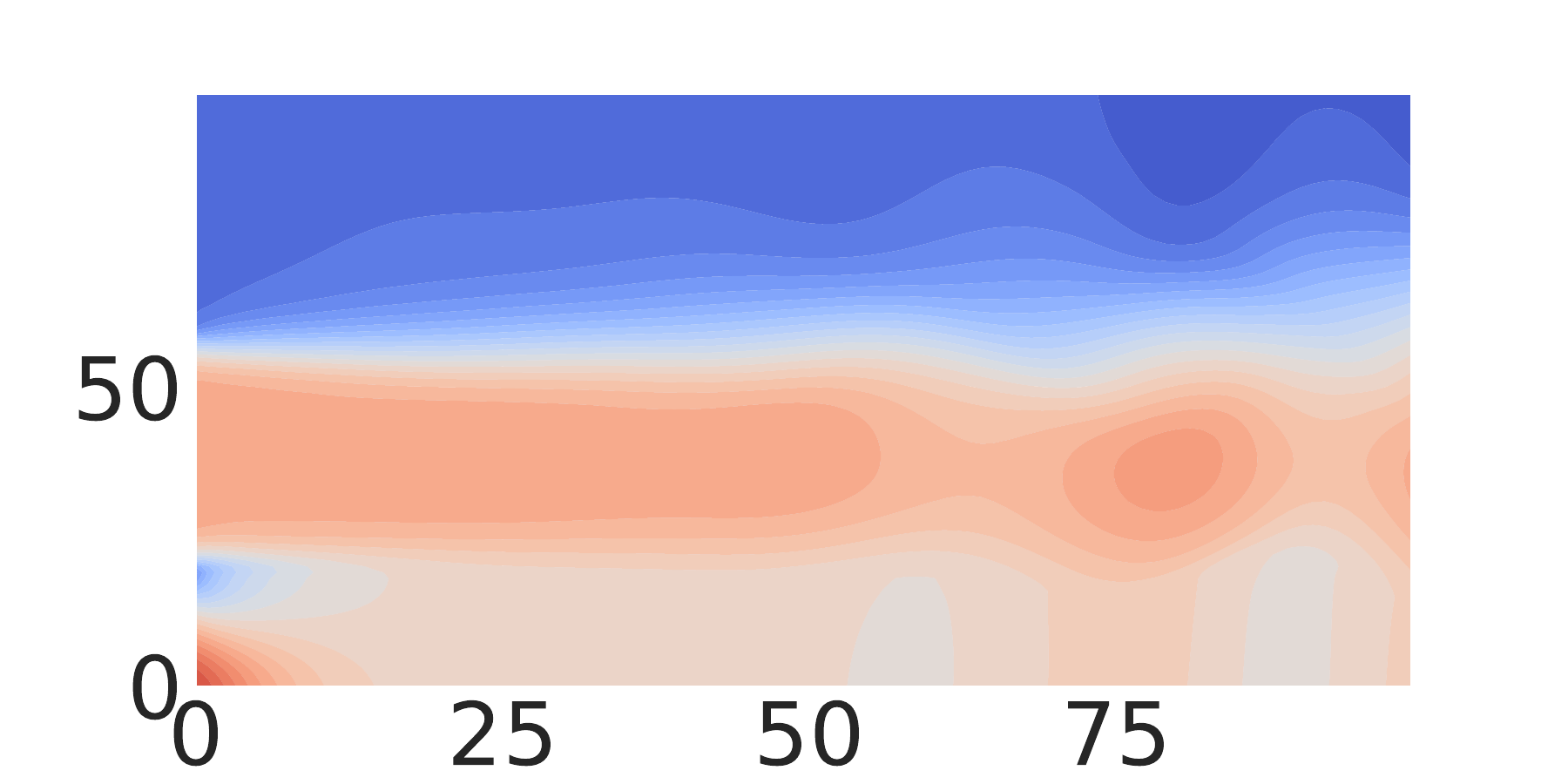}&
		\includegraphics[width=0.27\textwidth, angle=0]{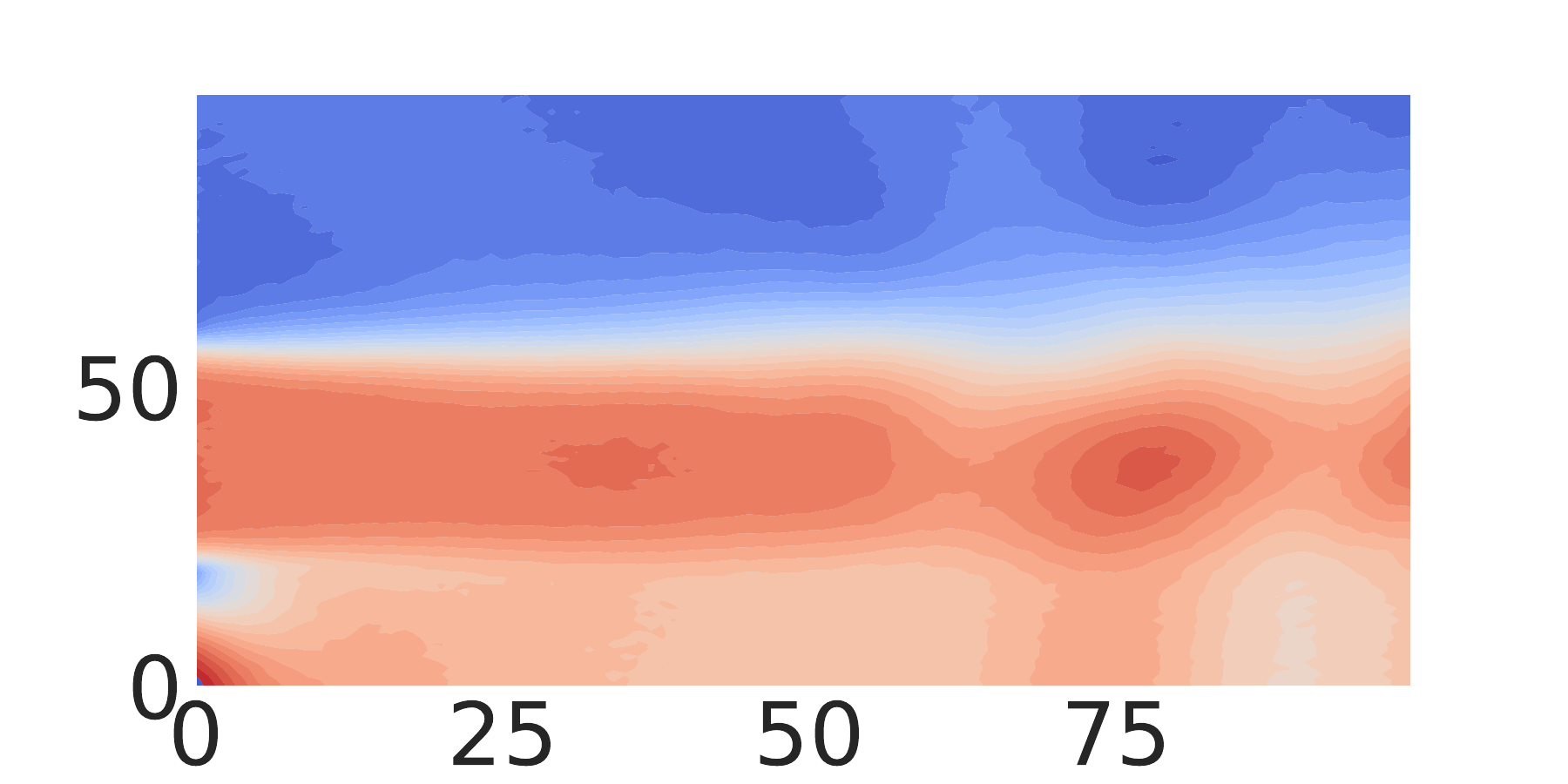}&
		\includegraphics[width=0.27\textwidth, angle=0]{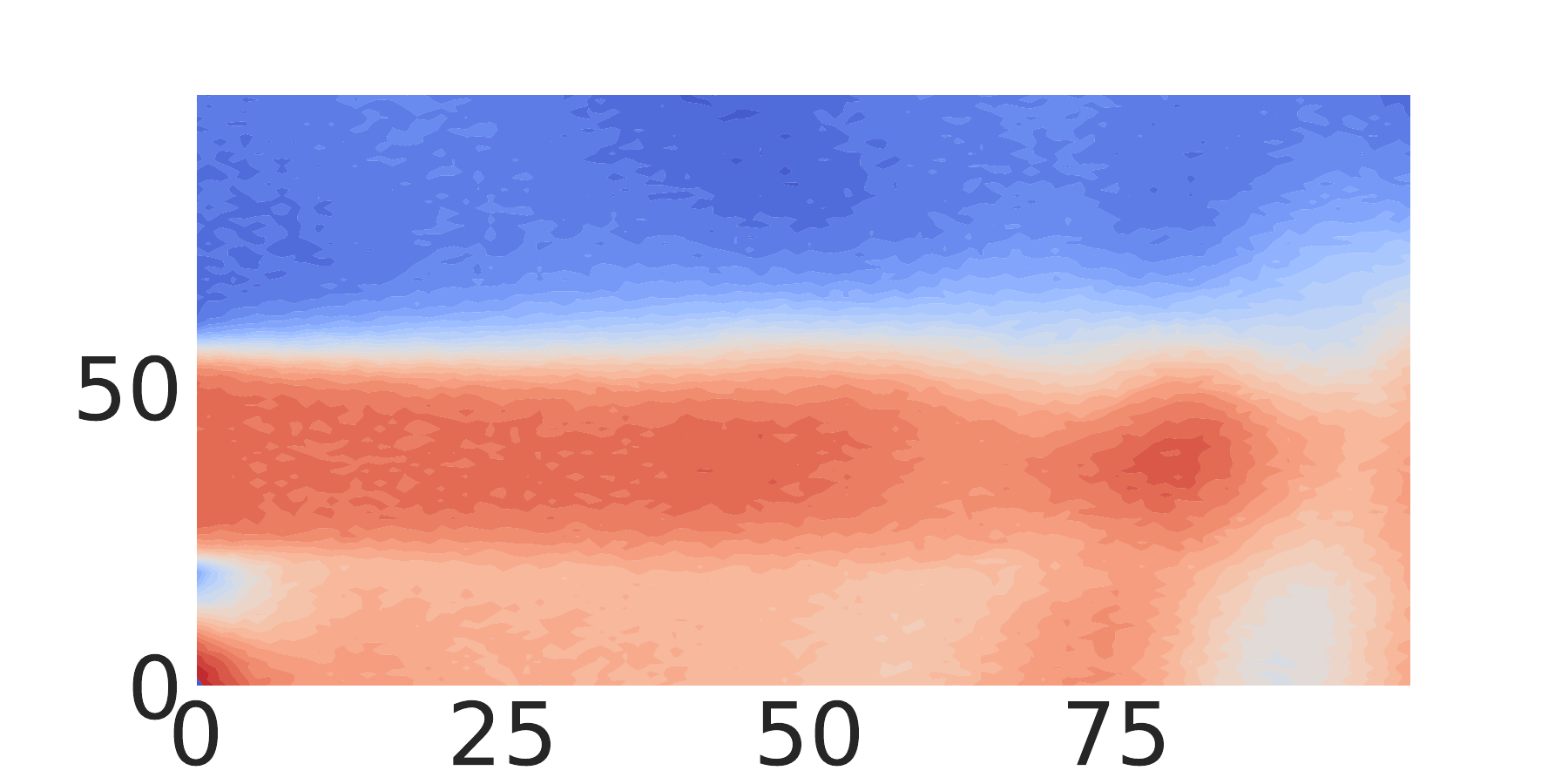}\\
	\end{tabular}
	\caption{Same as Figure \ref{fig: ann_normal_sts_sim} for case S3.}
	\label{fig: ann_normal_cts_sim}
\end{figure}

\begin{figure}[H]
	\centering
	\begin{tabular}{lccc}
		Sample & Simulation & RNN & CNN\\
		$t = 244$&
		\includegraphics[width=0.27\textwidth, angle=0]{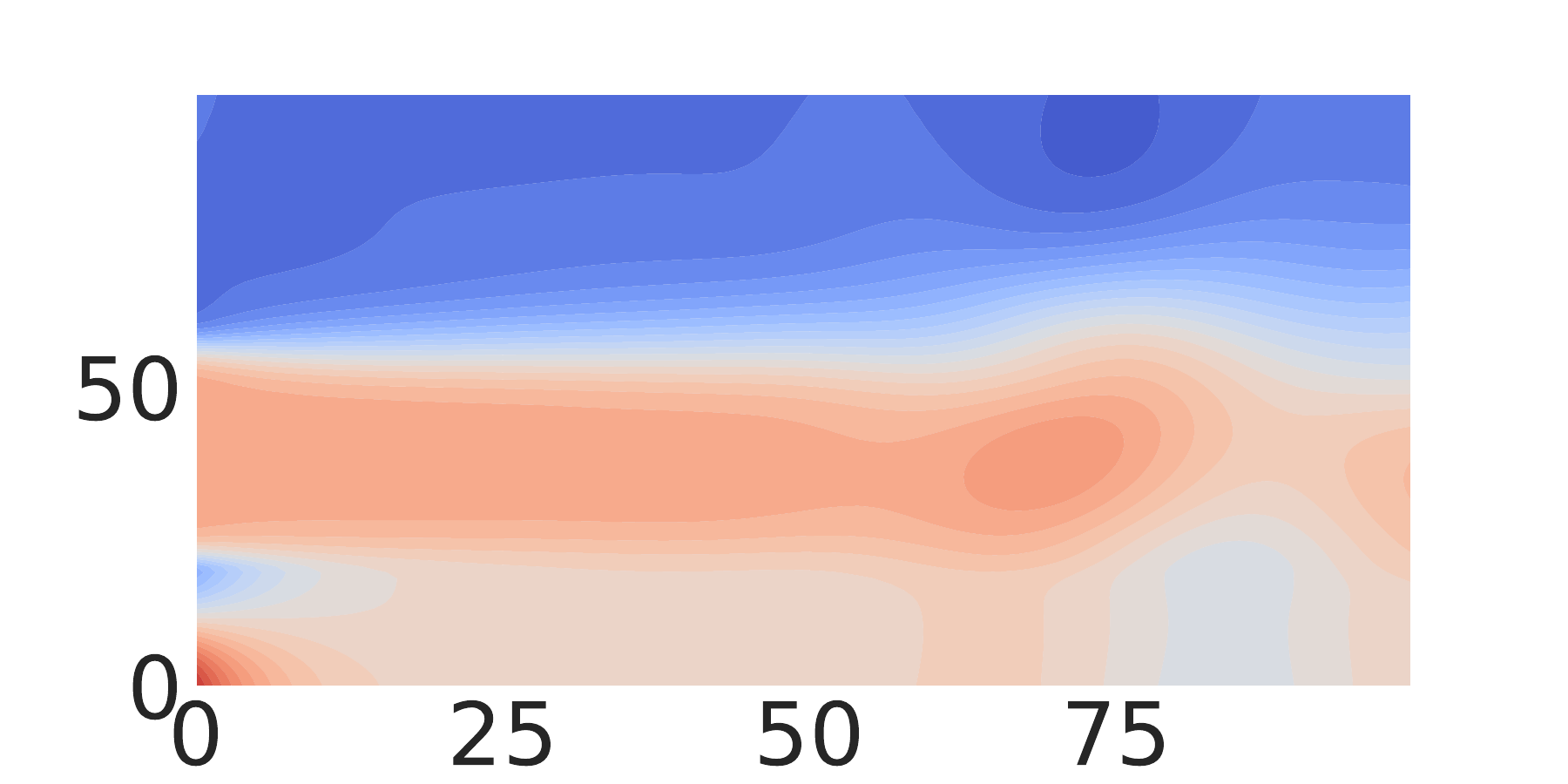}&
		\includegraphics[width=0.27\textwidth, angle=0]{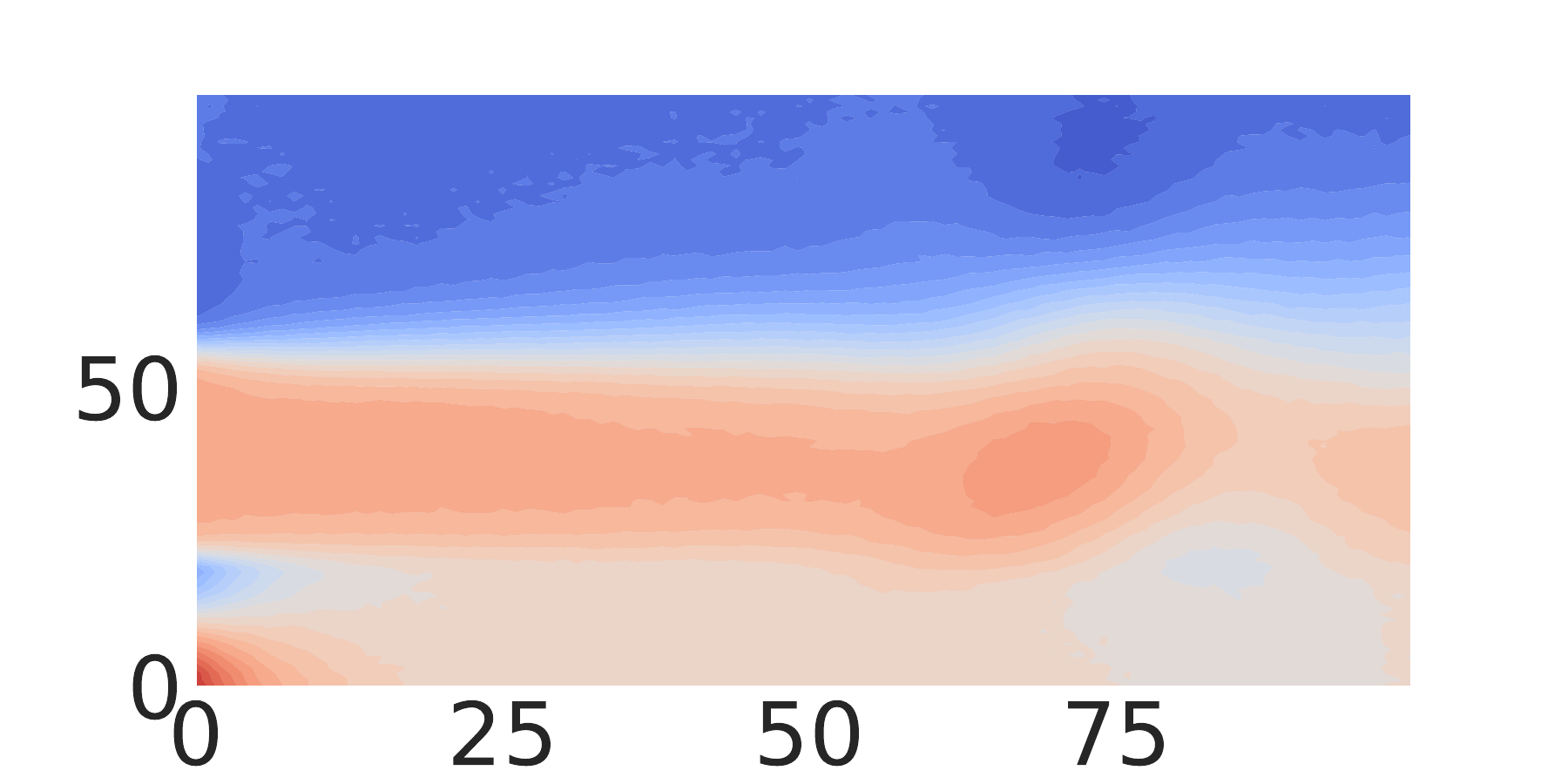}&
		\includegraphics[width=0.27\textwidth, angle=0]{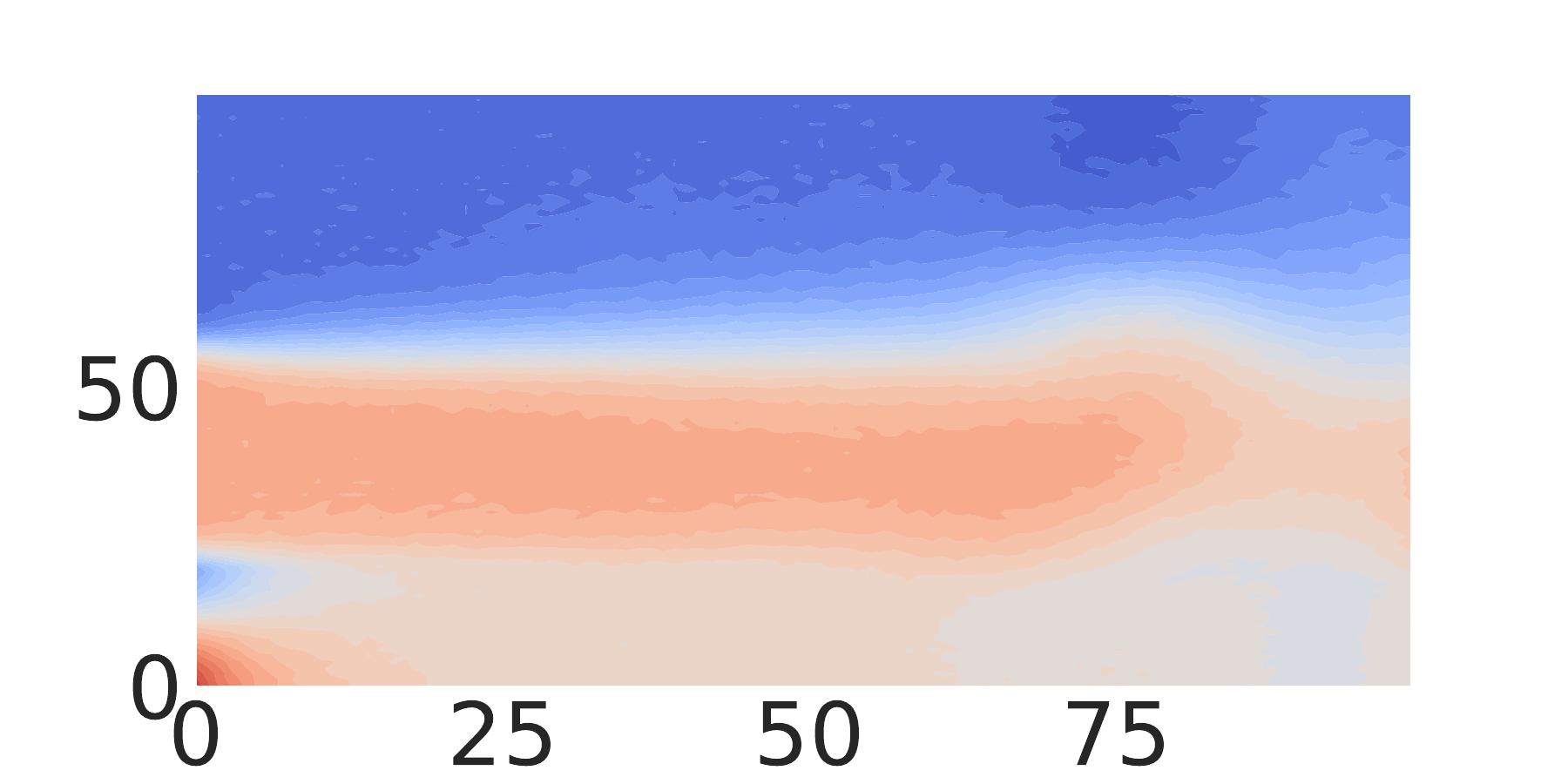}\\
		$t=245$&
		\includegraphics[width=0.27\textwidth, angle=0]{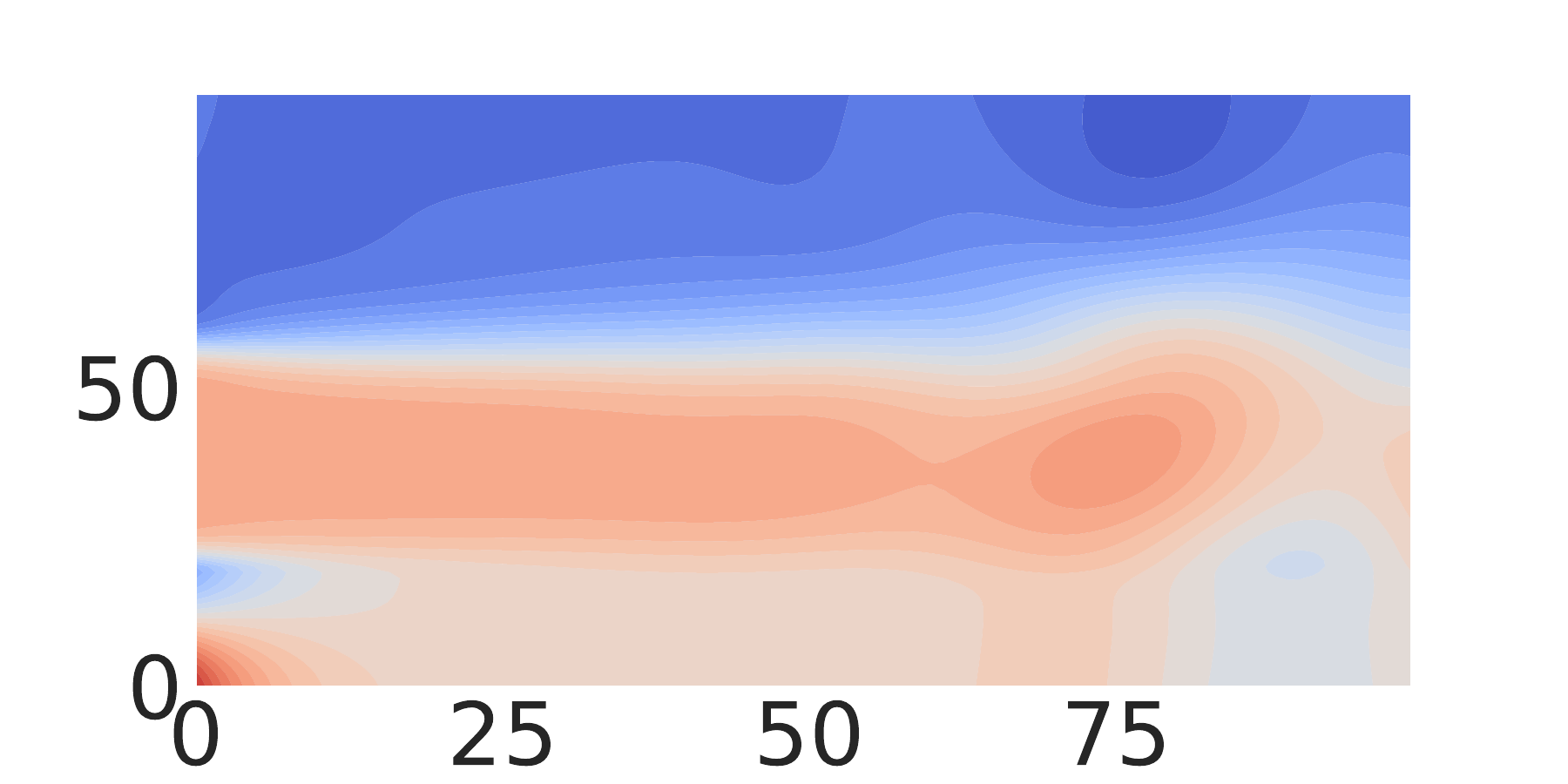}&
		\includegraphics[width=0.27\textwidth, angle=0]{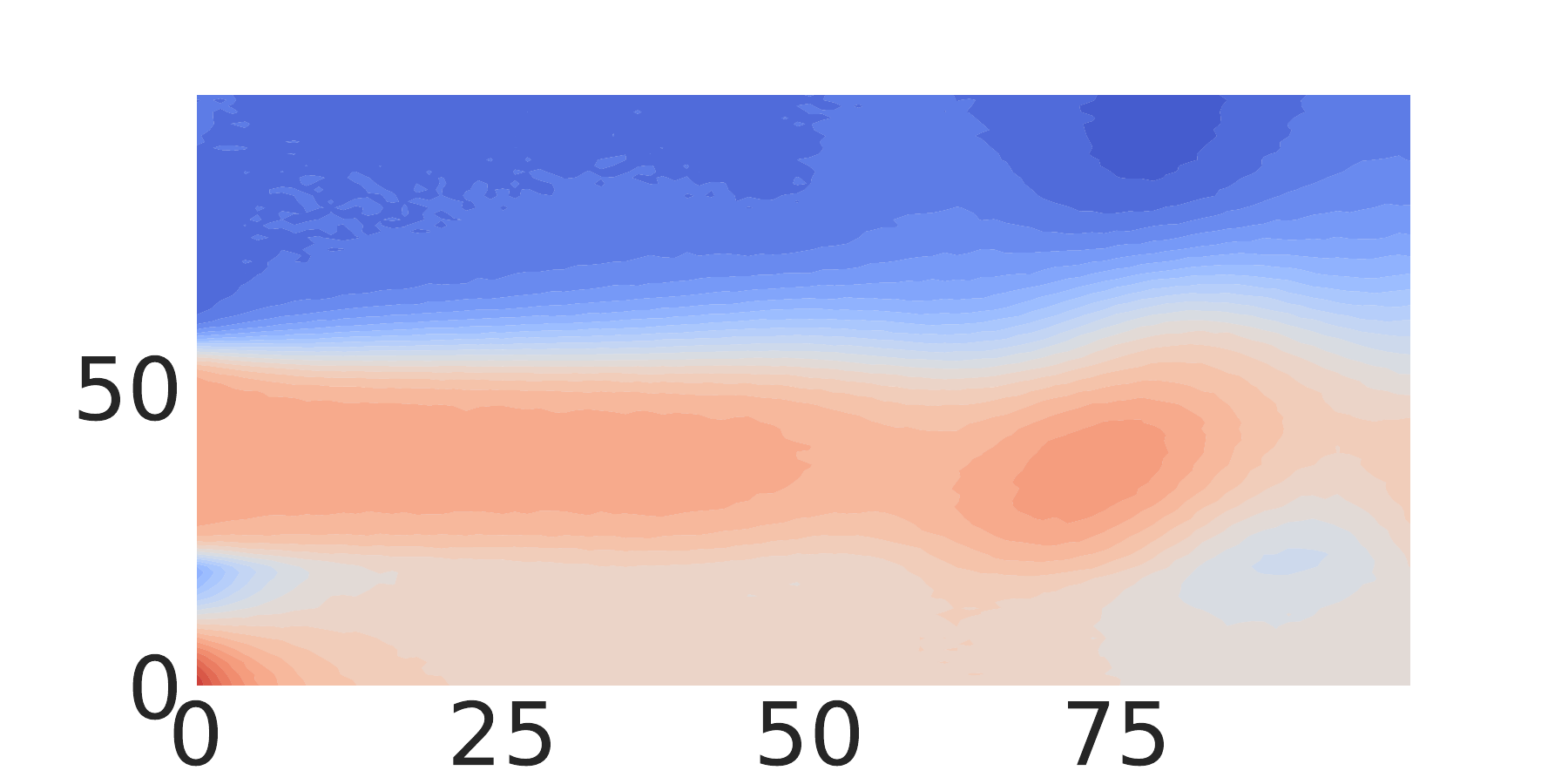}&
		\includegraphics[width=0.27\textwidth, angle=0]{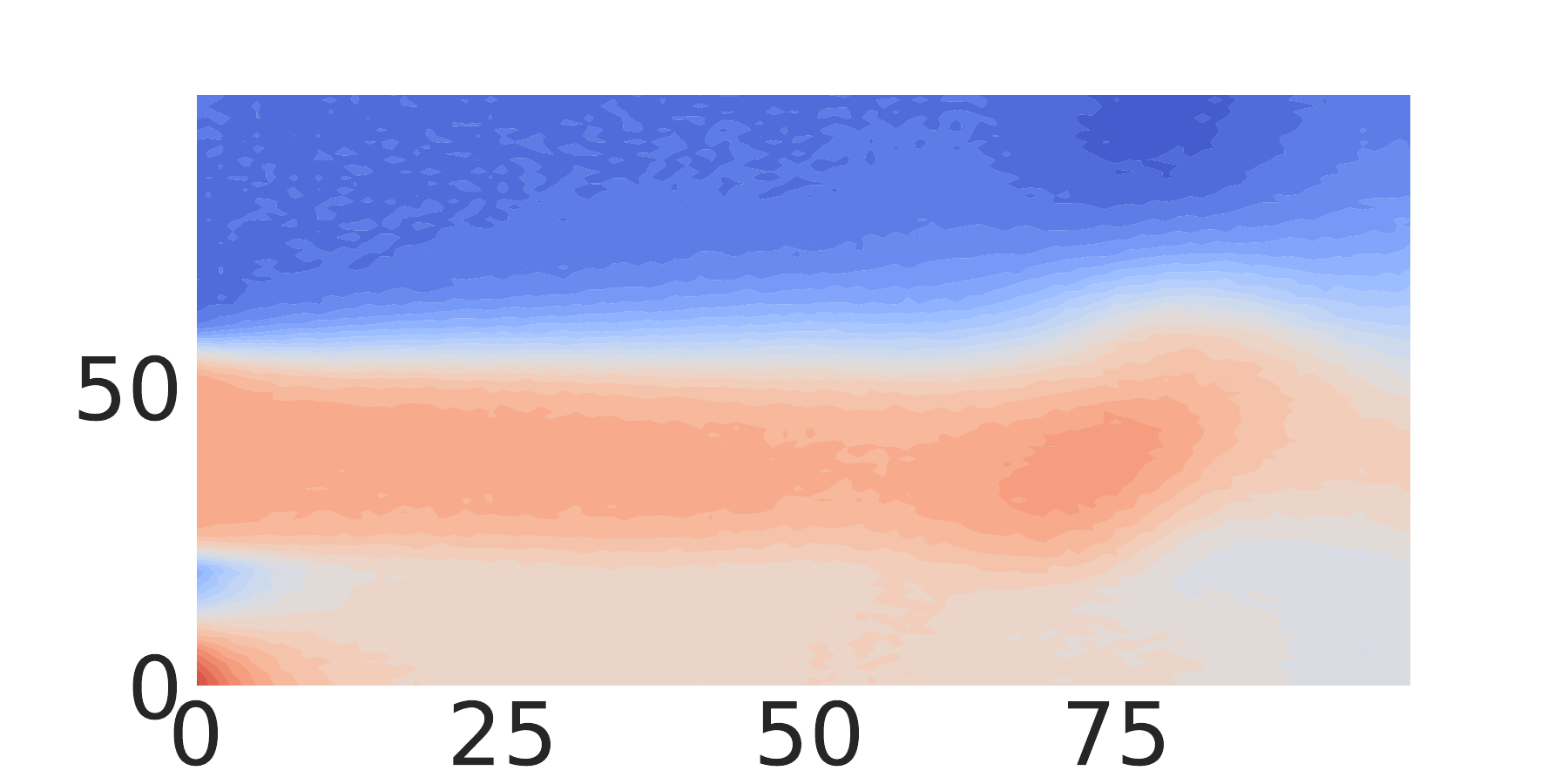}\\
		$t = 259$&
		\includegraphics[width=0.27\textwidth, angle=0]{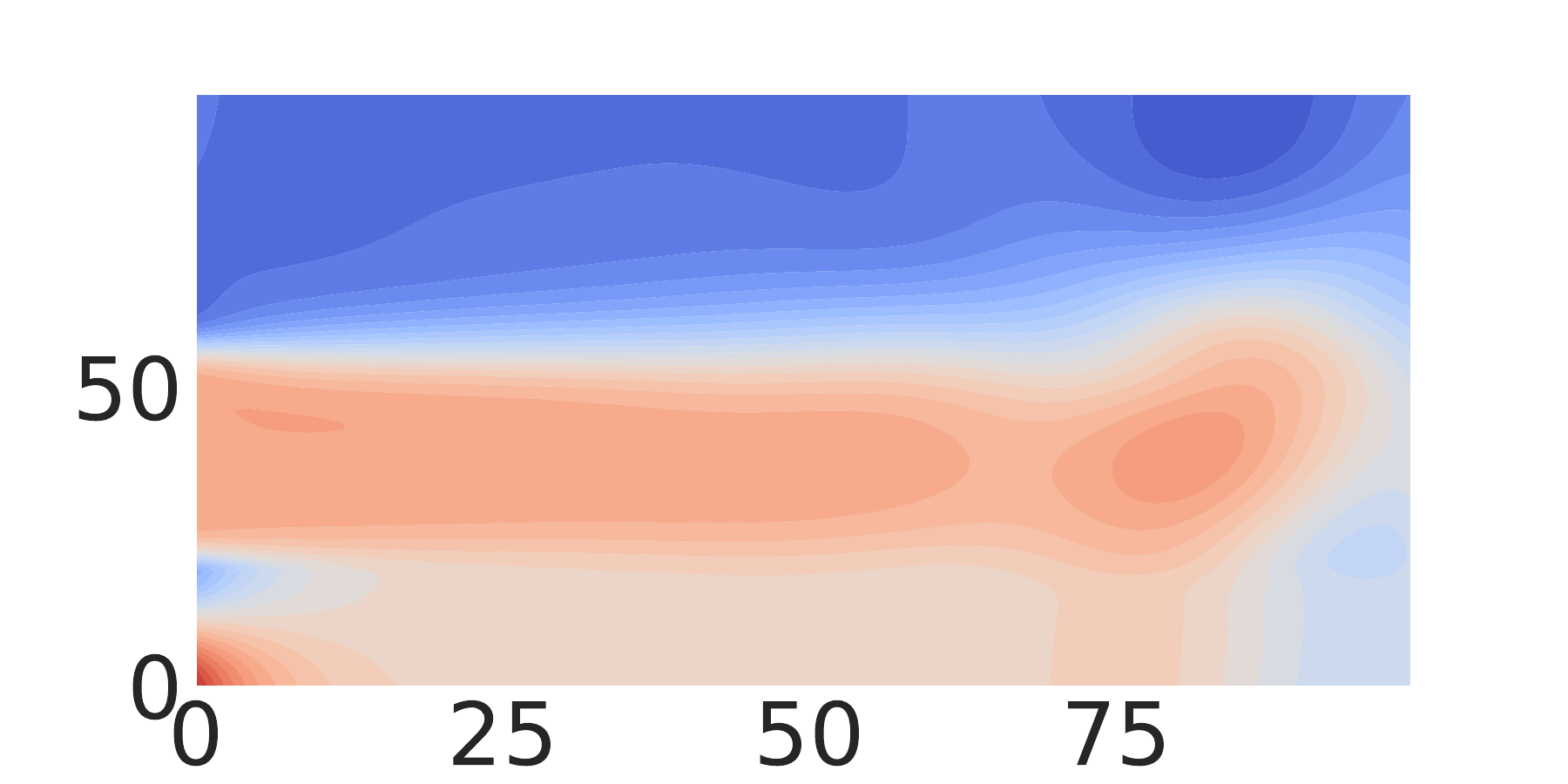}&
		\includegraphics[width=0.27\textwidth, angle=0]{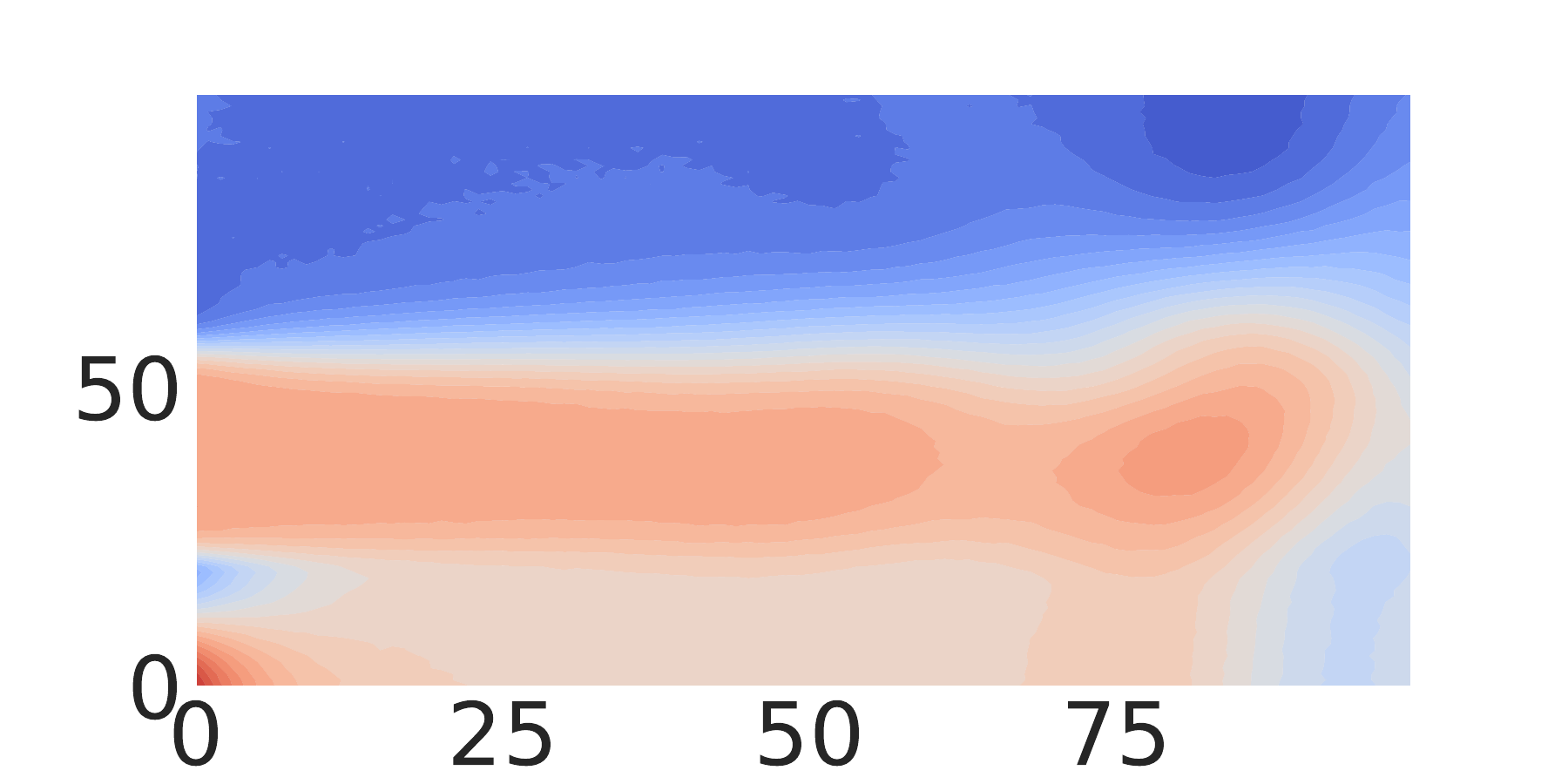}&
		\includegraphics[width=0.27\textwidth, angle=0]{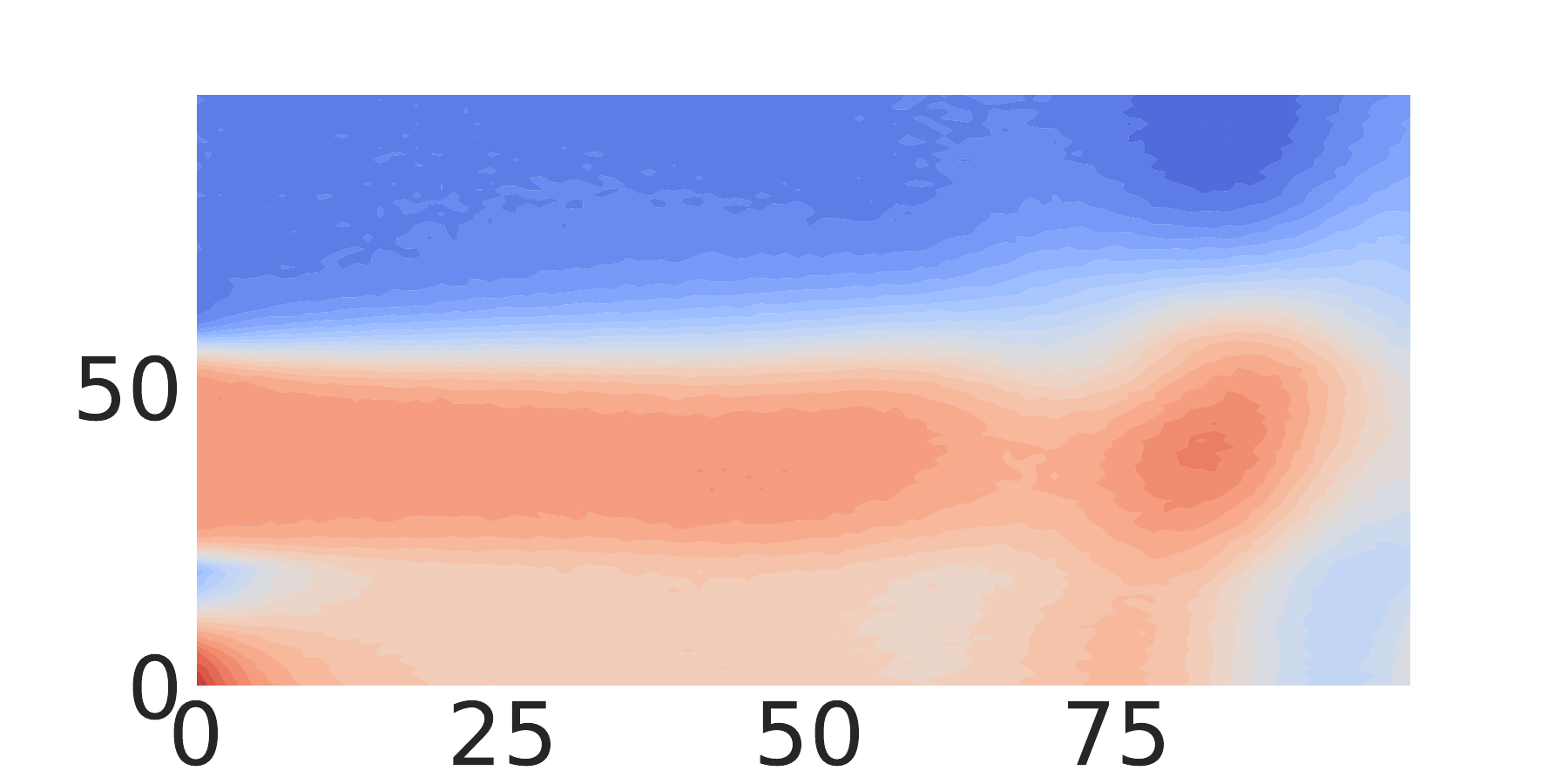}\\
		$t=260$&
		\includegraphics[width=0.27\textwidth, angle=0]{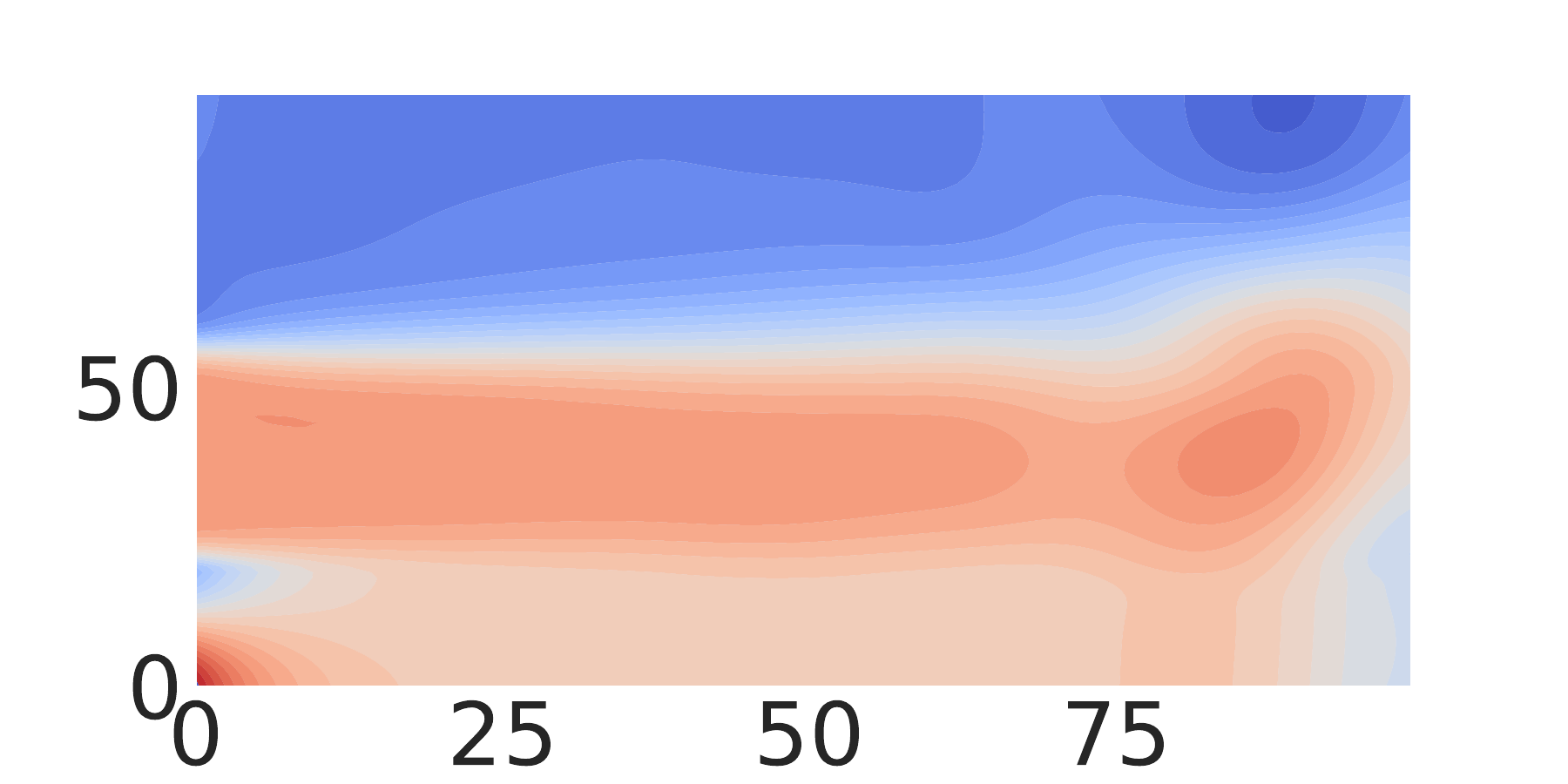}&
		\includegraphics[width=0.27\textwidth, angle=0]{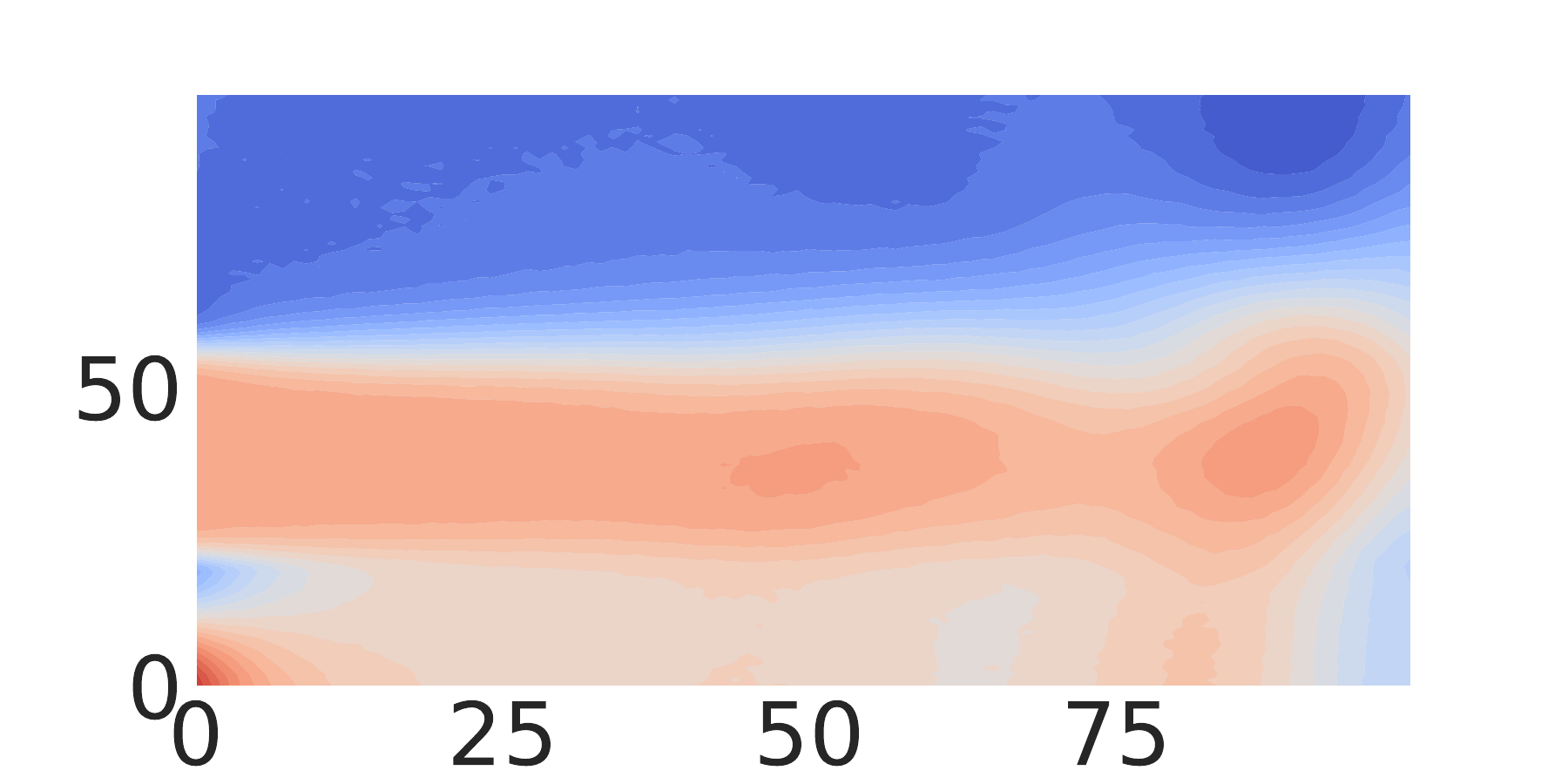}&
		\includegraphics[width=0.27\textwidth, angle=0]{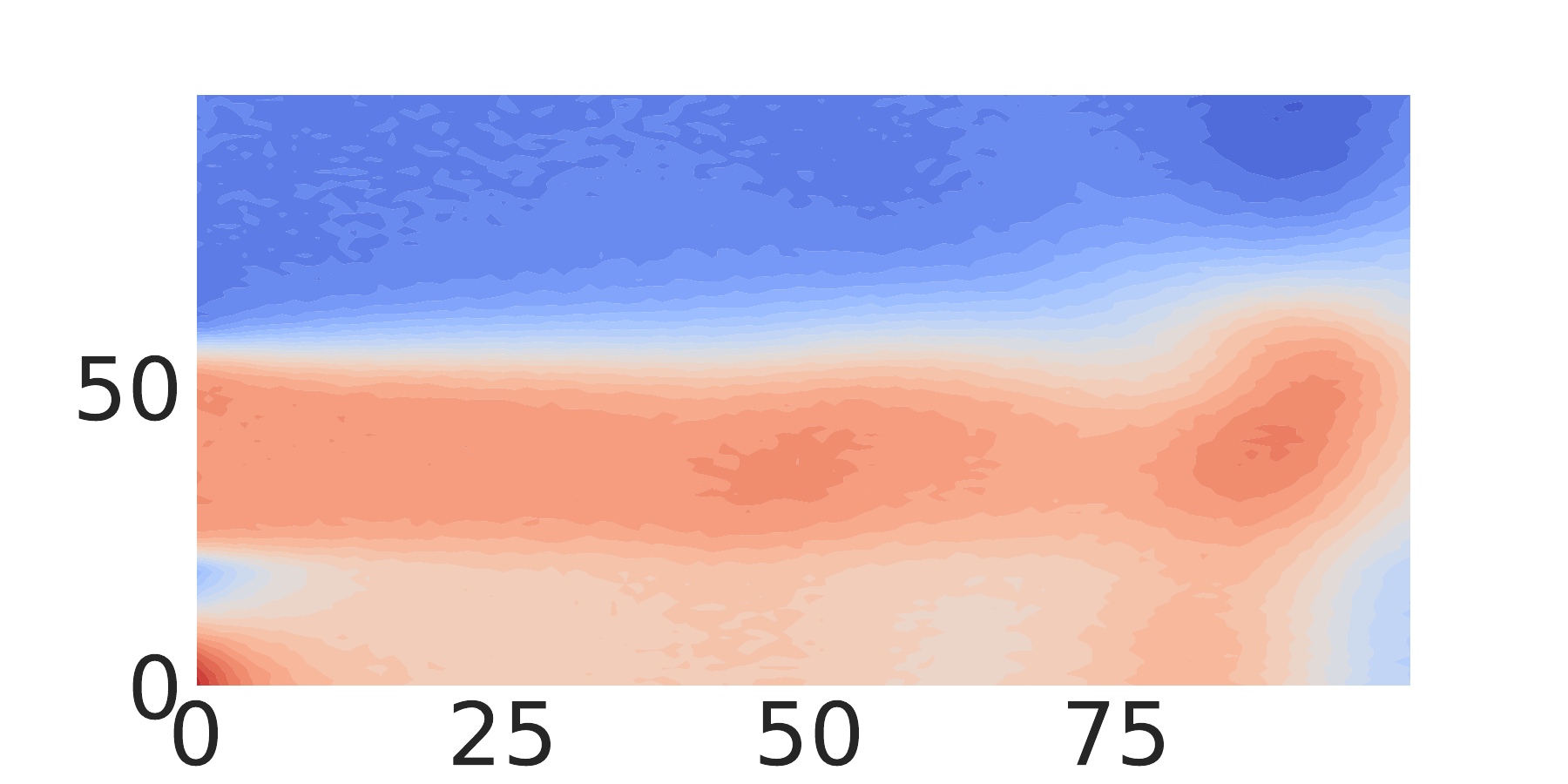}\\
	\end{tabular}
	\caption{From left to right: Snapshots from reconstructed data set (through HODMD), prediction of RNN model and prediction of CNN model, respectively, corresponding to case S2$_{ROM}$. To generate these predictions the following samples were sent to both NNs; $\{v_{243}, v_{242}, \dots, v_{234}\}$ and $\{v_{258}, v_{257}, \dots, v_{249}\}$. Note that all these samples belong to the test set, Table \ref{tab:ML3}.}
	\label{fig: ann_normal_sts_rom}
\end{figure}

\begin{figure}[H]
	\centering
	\begin{tabular}{lccc}
		Sample & Simulation & RNN & CNN\\
		$t = 244$&
		\includegraphics[width=0.27\textwidth, angle=0]{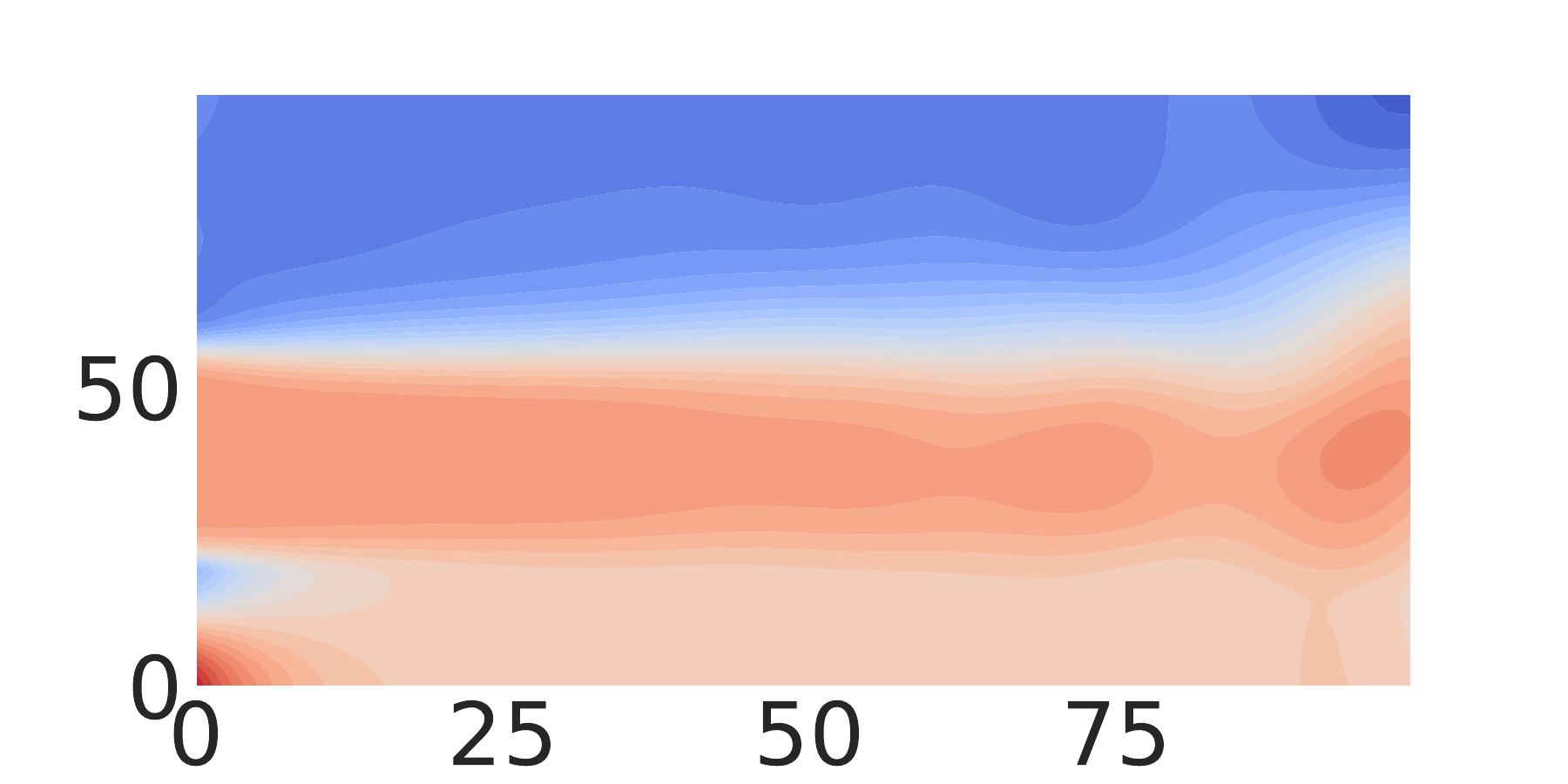}&
		\includegraphics[width=0.27\textwidth, angle=0]{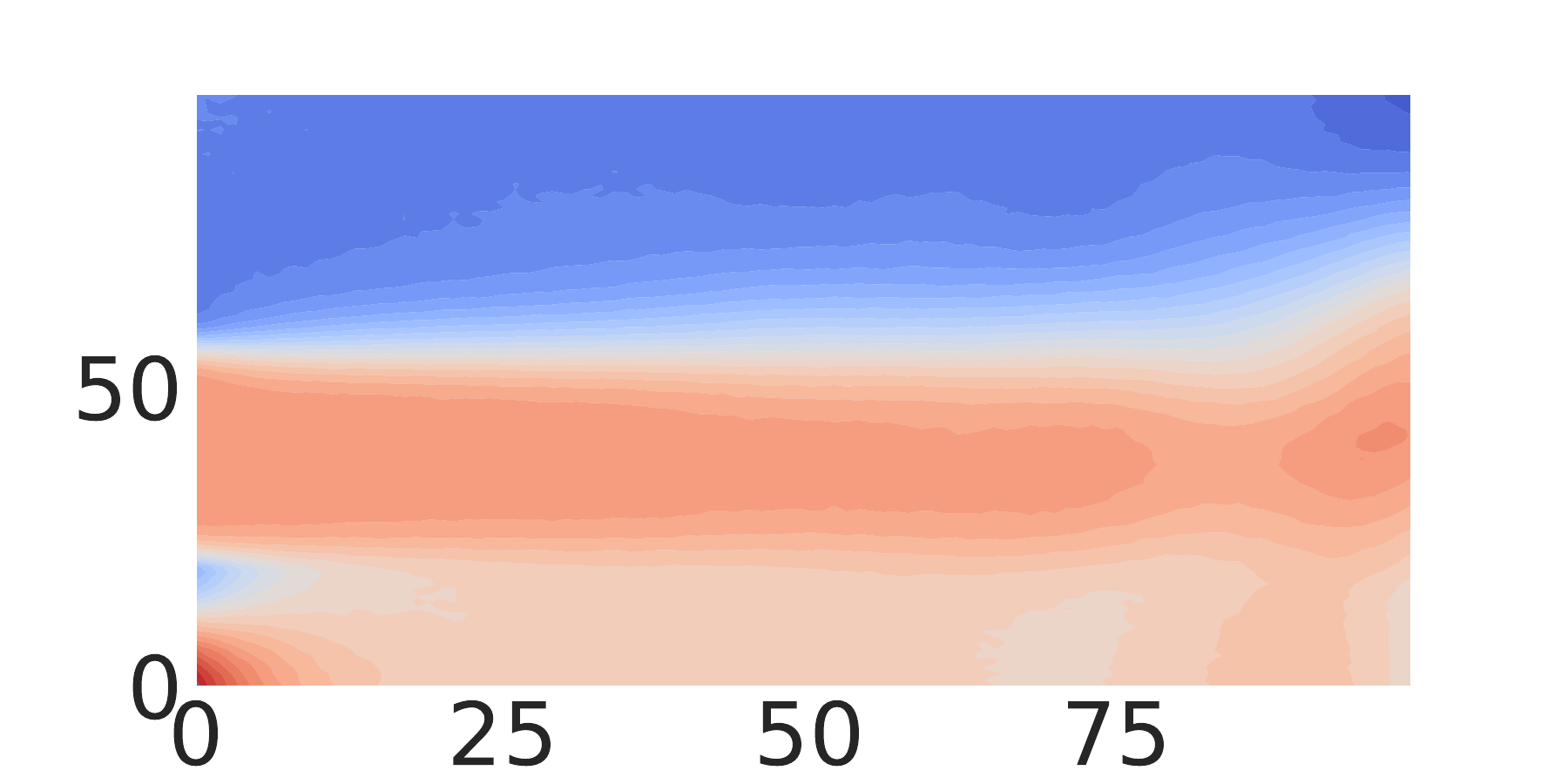}&
		\includegraphics[width=0.27\textwidth, angle=0]{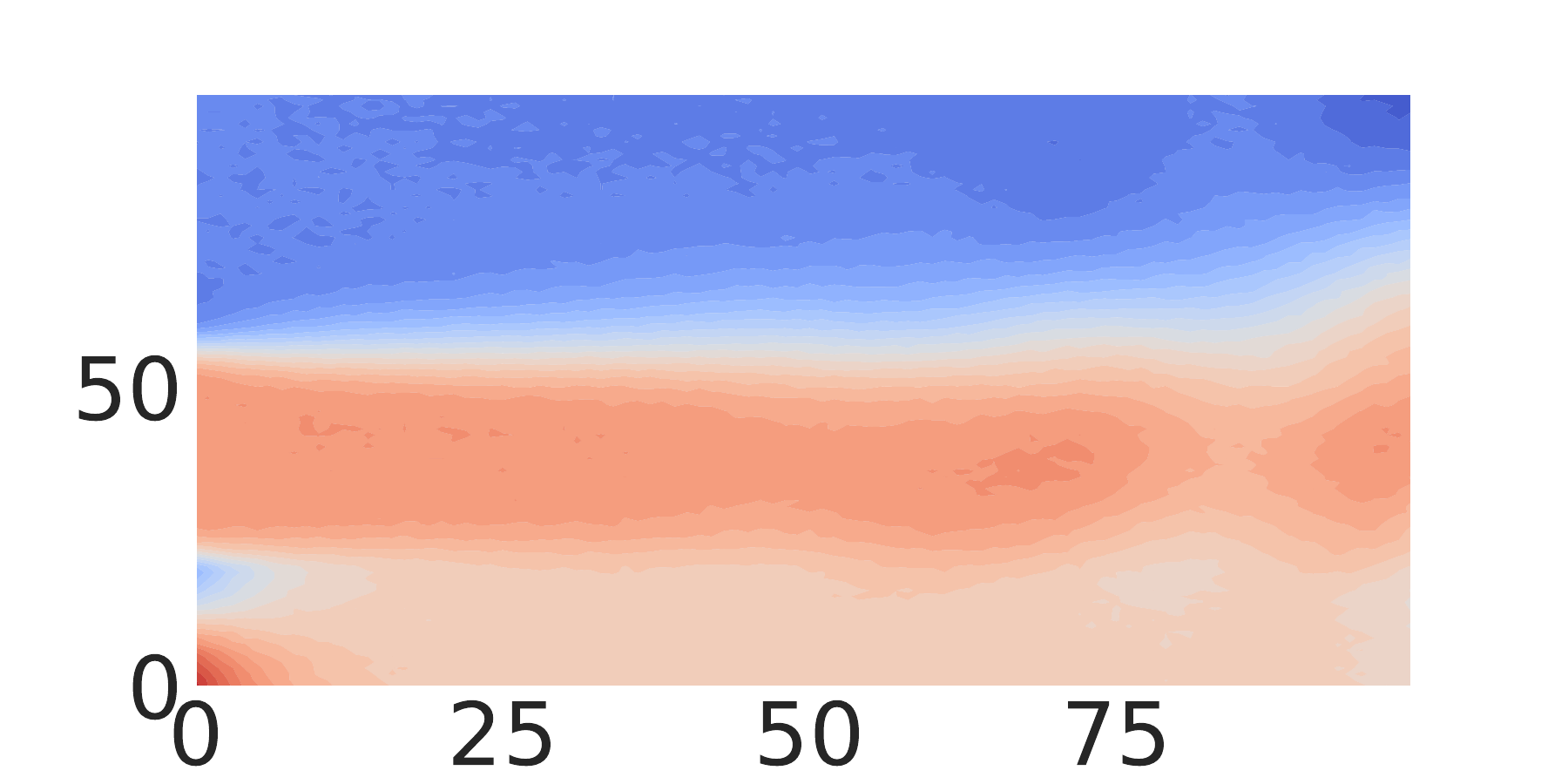}\\
		$t=245$&
		\includegraphics[width=0.27\textwidth, angle=0]{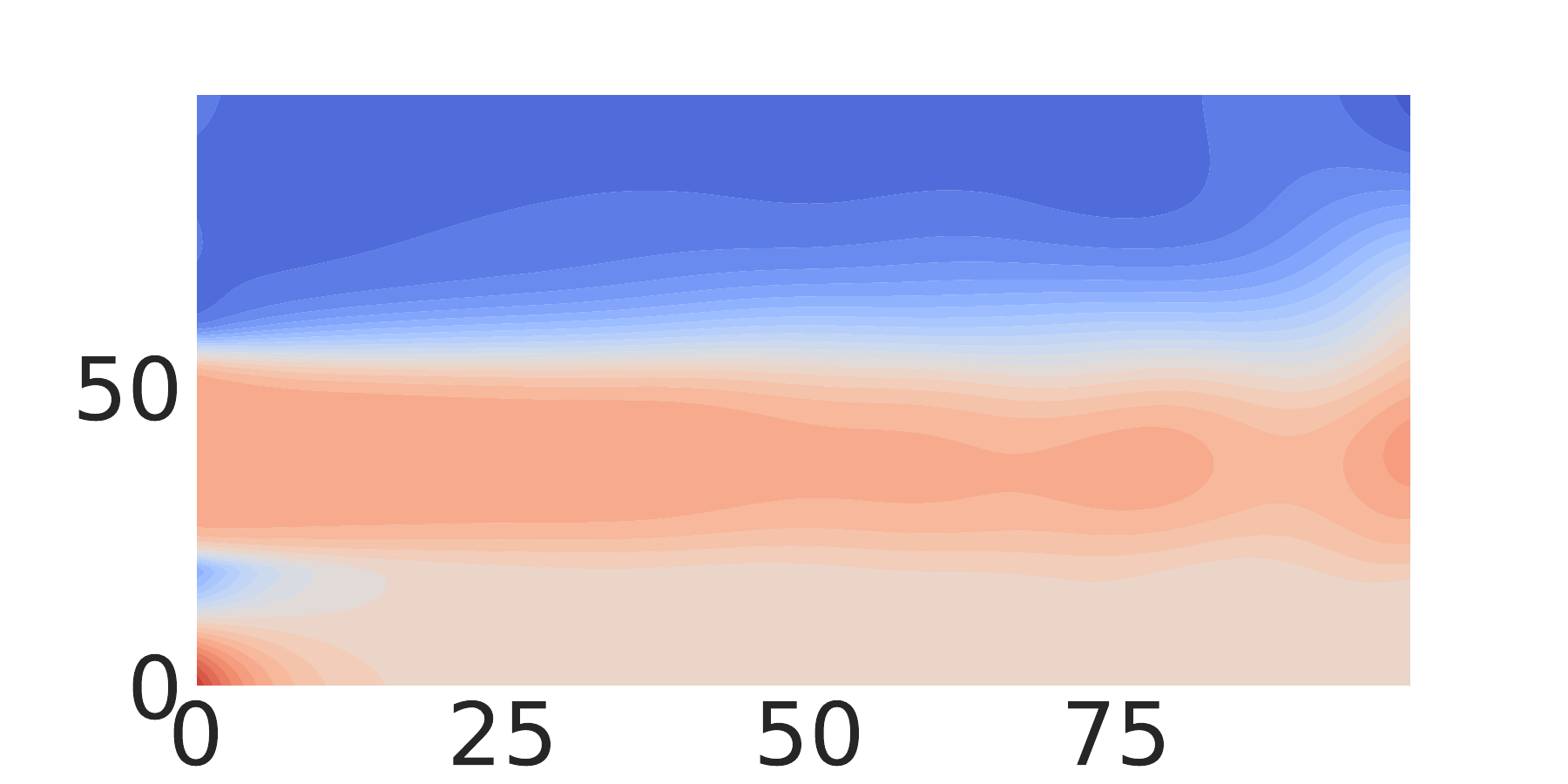}&
		\includegraphics[width=0.27\textwidth, angle=0]{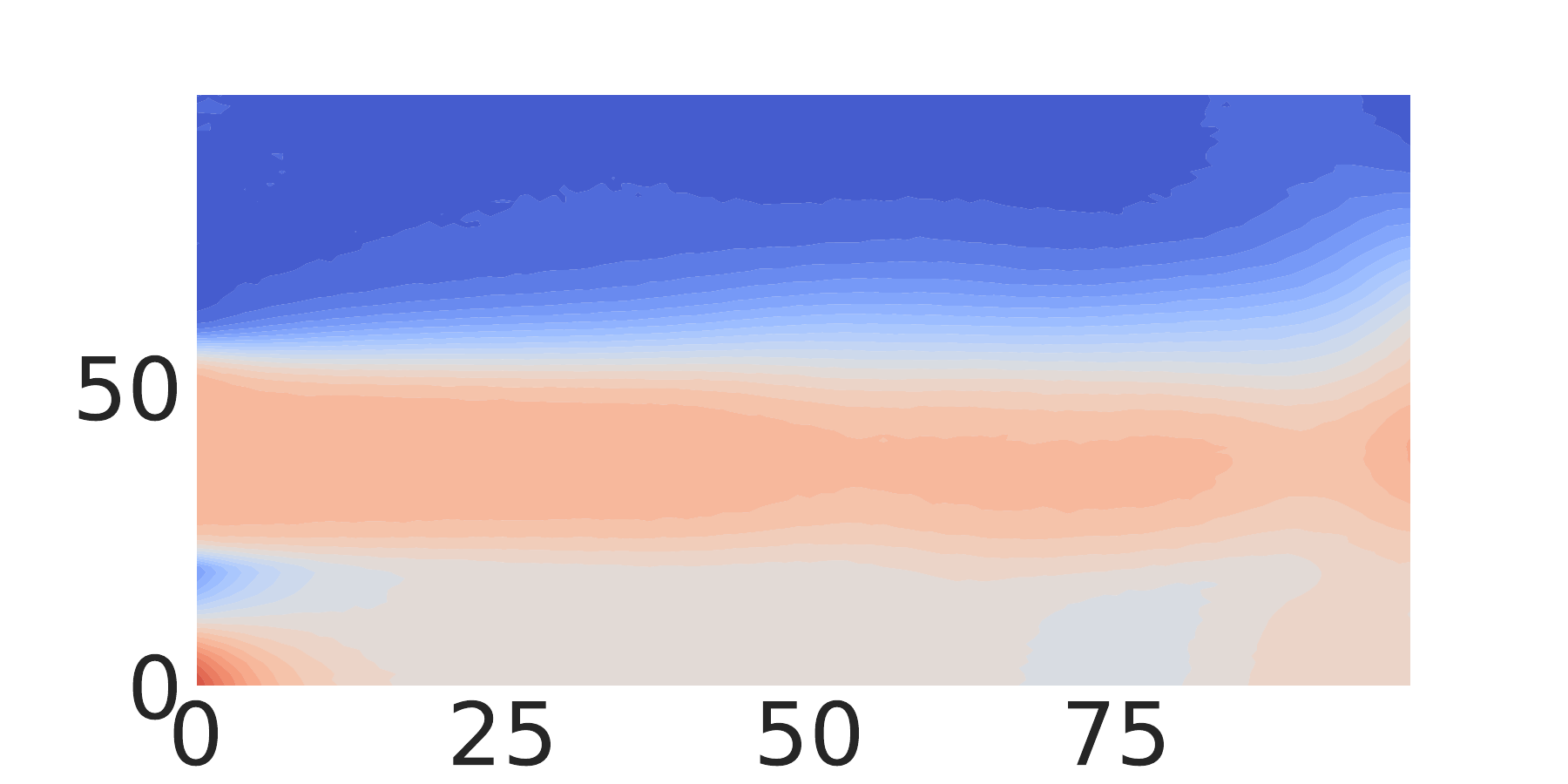}&
		\includegraphics[width=0.27\textwidth, angle=0]{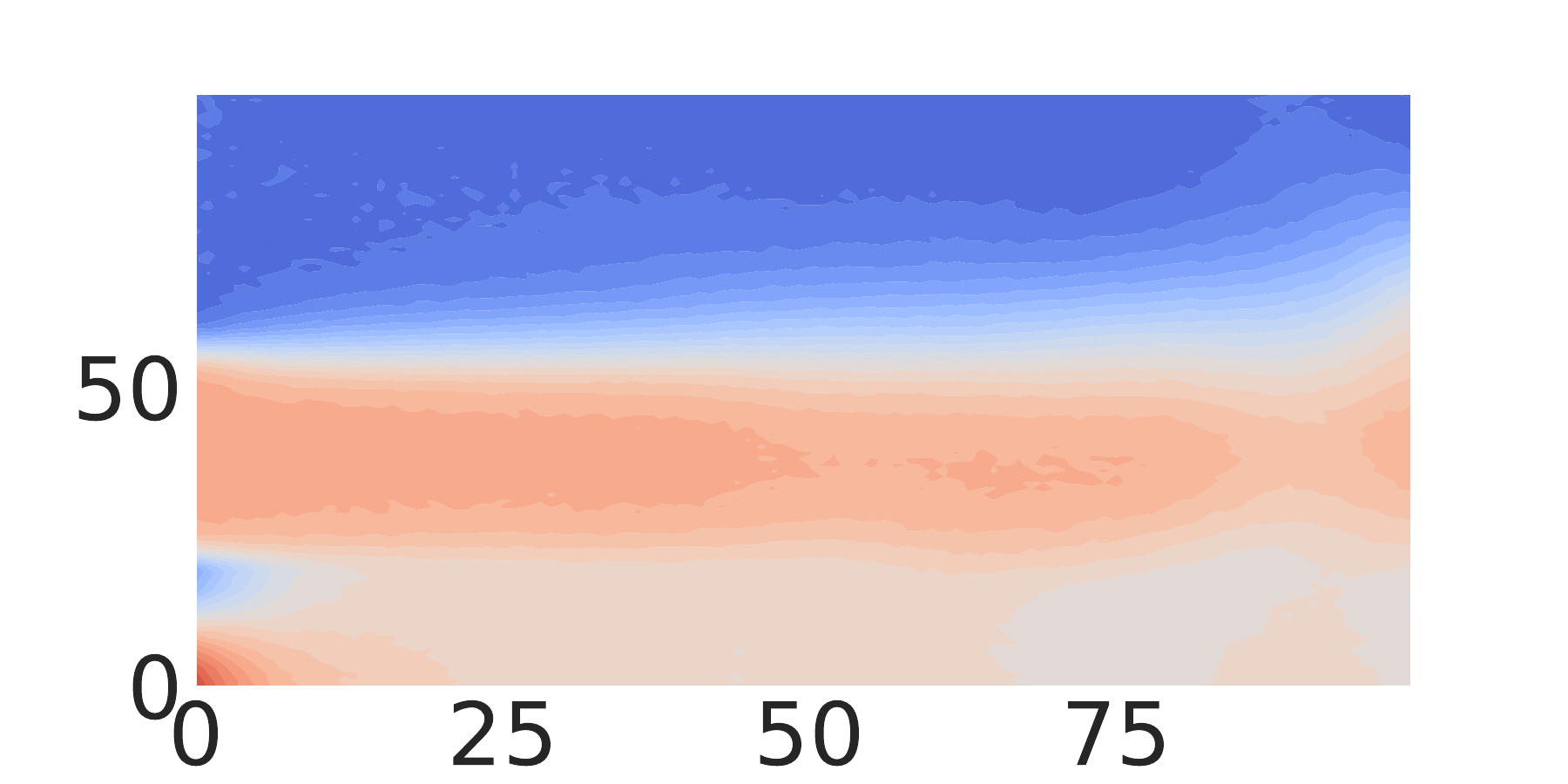}\\
		$t = 259$&
		\includegraphics[width=0.27\textwidth, angle=0]{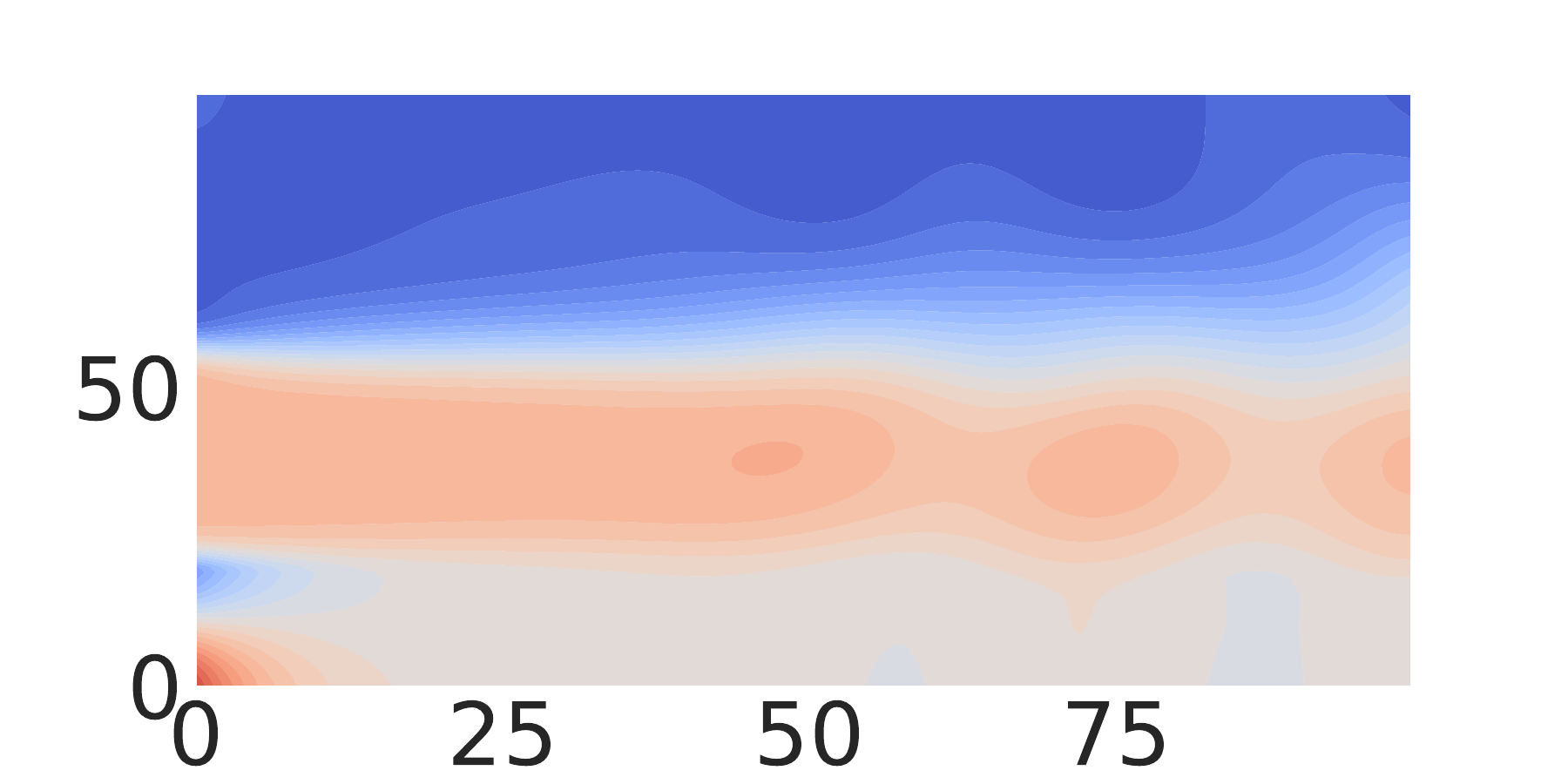}&
		\includegraphics[width=0.27\textwidth, angle=0]{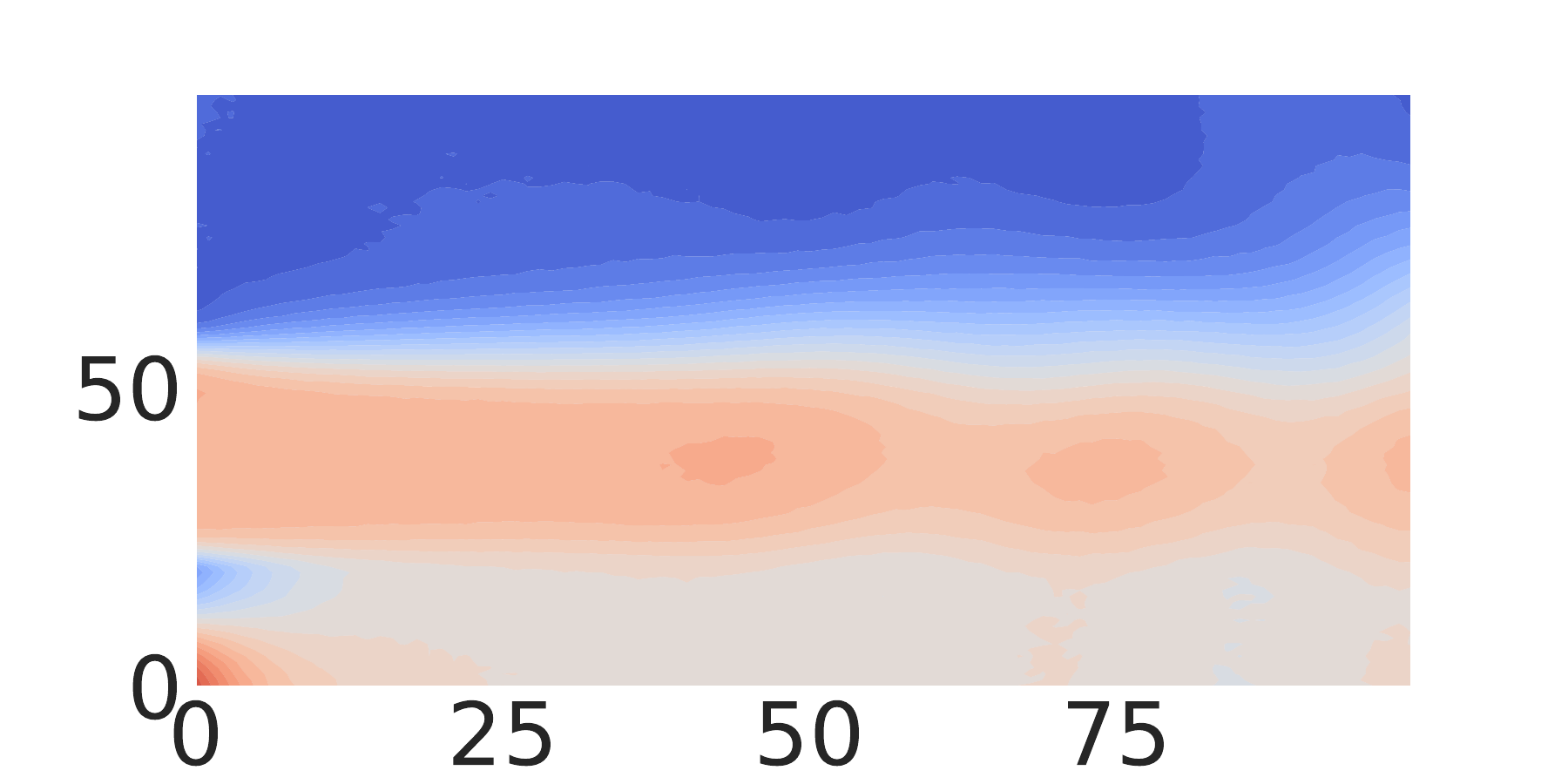}&
		\includegraphics[width=0.27\textwidth, angle=0]{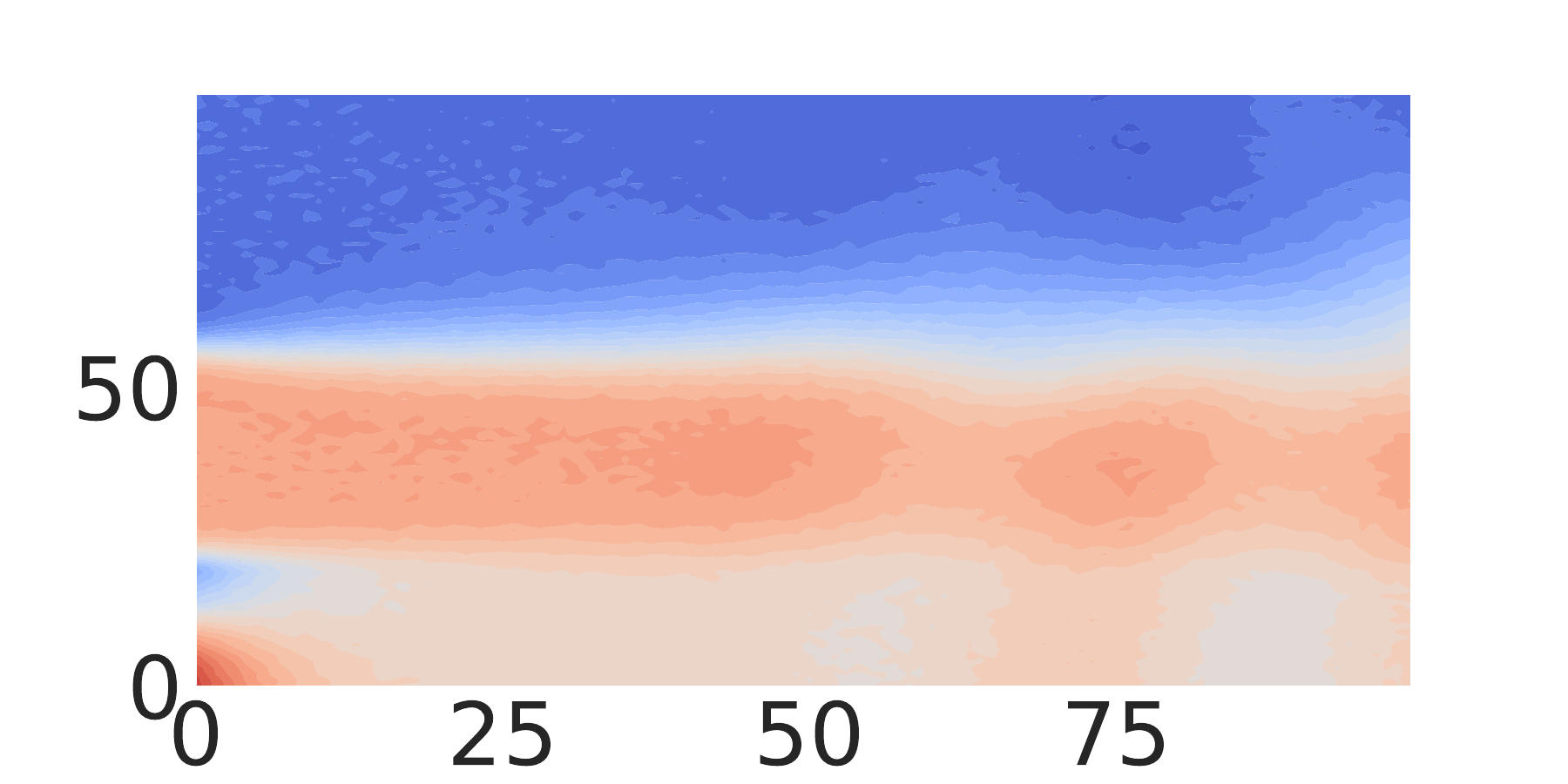}\\
		$t=260$&
		\includegraphics[width=0.27\textwidth, angle=0]{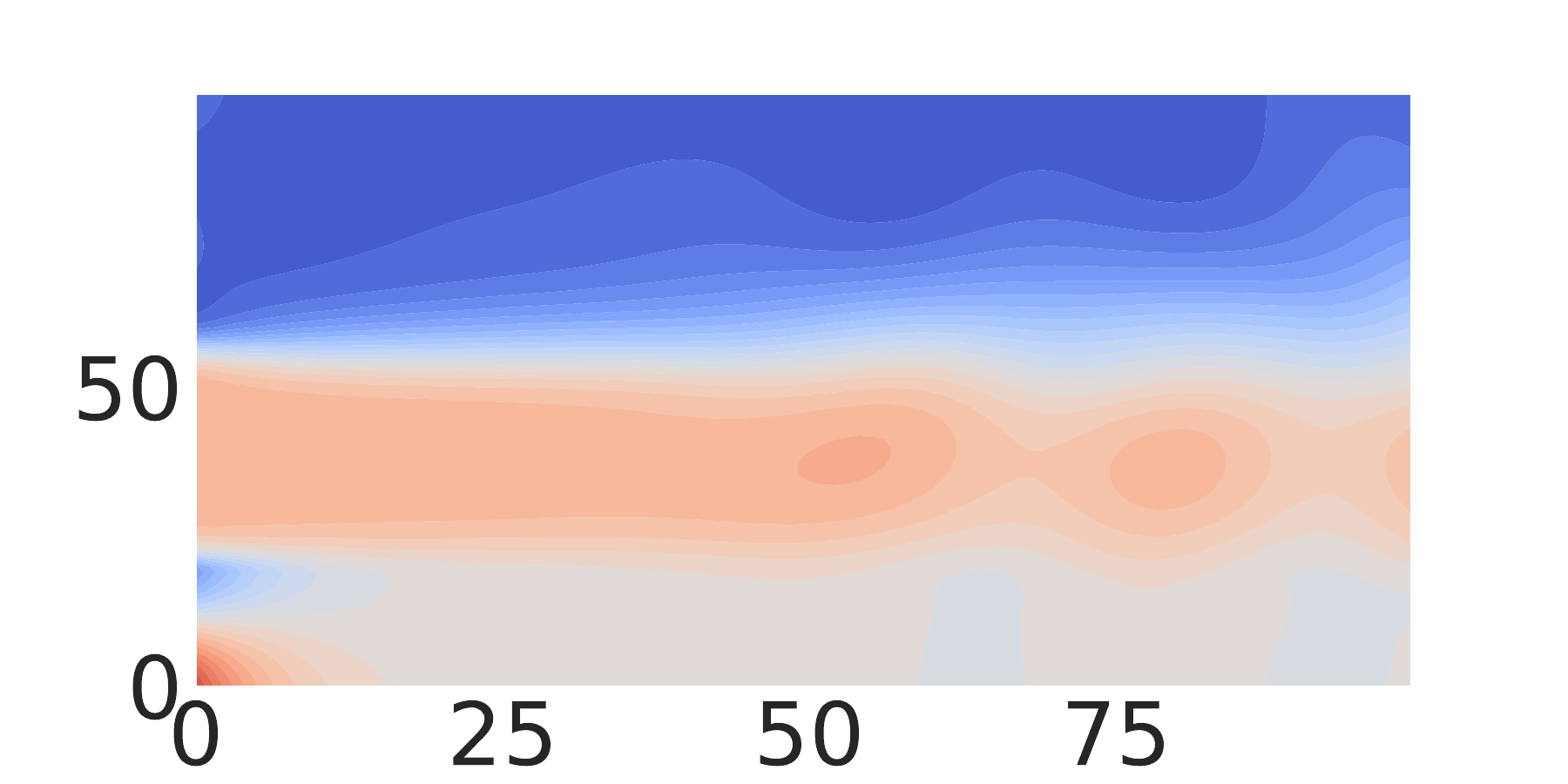}&
		\includegraphics[width=0.27\textwidth, angle=0]{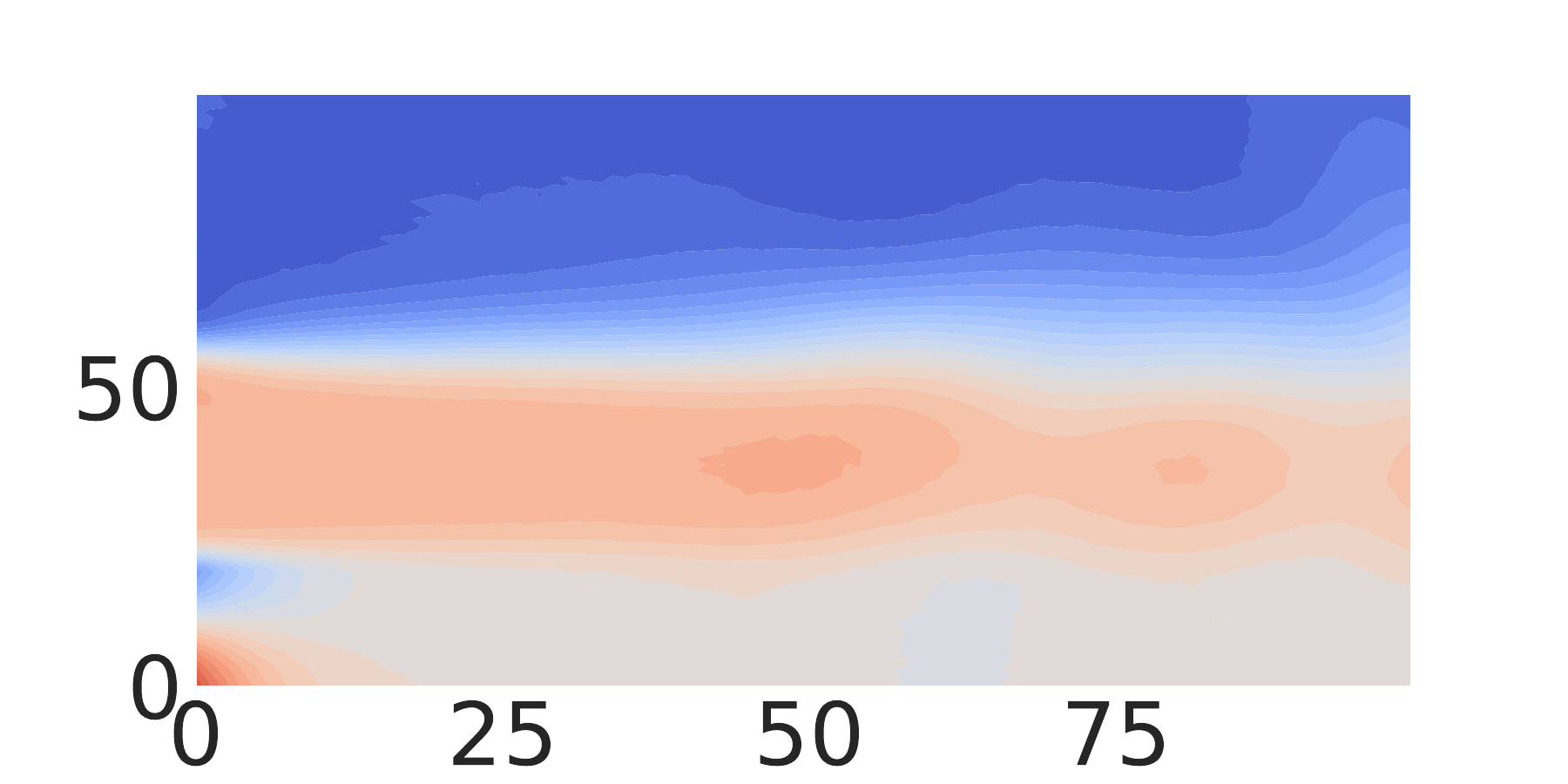}&
		\includegraphics[width=0.27\textwidth, angle=0]{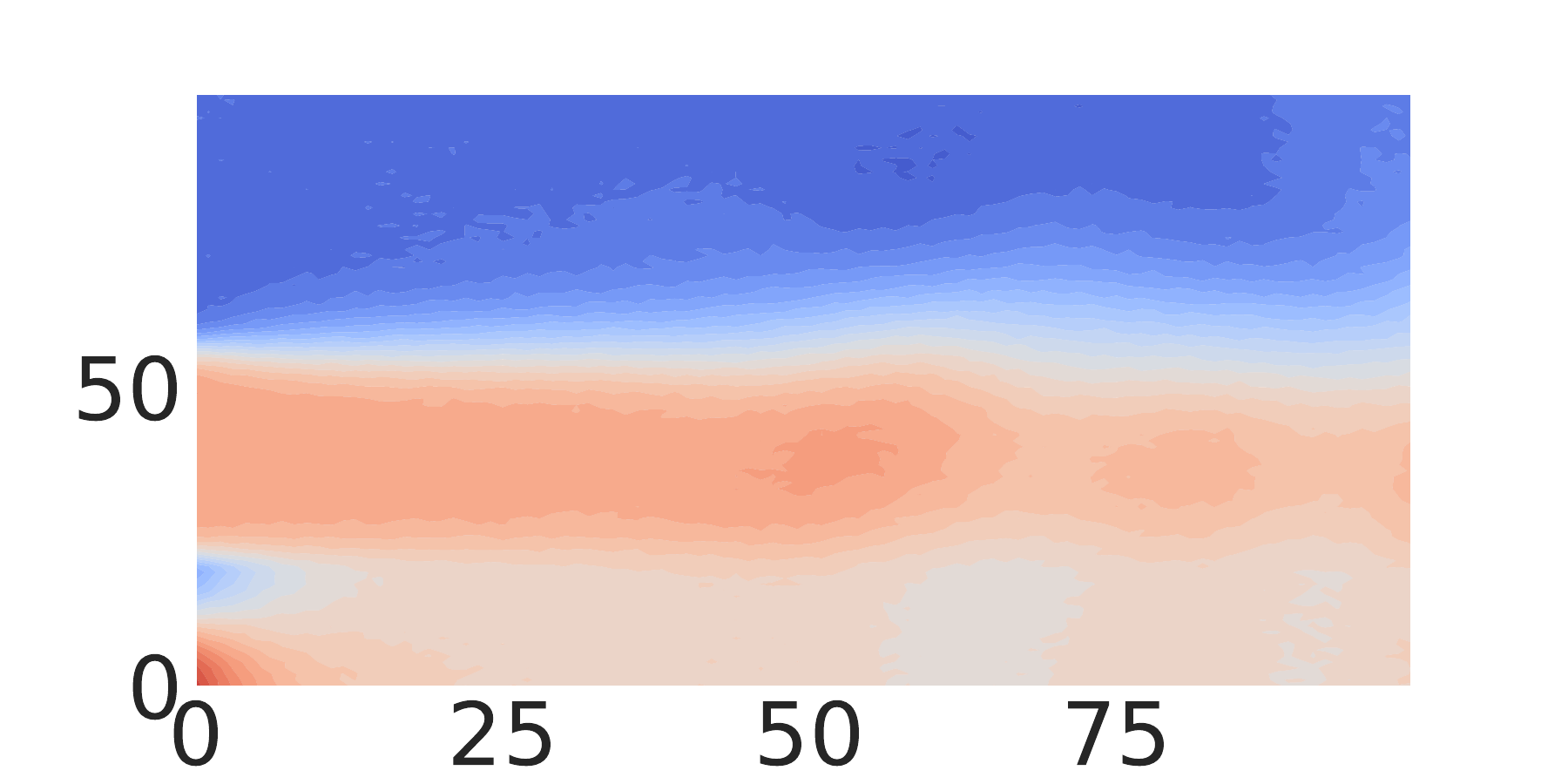}\\
	\end{tabular}
	\caption{Same as Figure \ref{fig: ann_normal_sts_rom} for case S3$_{ROM}$.}
	\vspace{0.5cm}
	\label{fig: ann_normal_cts_rom}
\end{figure}

In this paper we use Relative Root Mean Square Error (RRMSE), which is defined in (\ref{eq:RMSEtimeNN}), to calculate the error performed by NN predictions. The corresponding prediction errors for cases S2, S3, S2$_{ROM}$ and S3$_{ROM}$ are listed in Table \ref{tab: rrmse_s2_s3}. Note that RRMSE value is expressed in a scale from 0 to 1, not as a percentage.

\begin{table}[H]
	\centering
	\begin{tabular}{|l|c|c|c|c|}
		\hline
		& S2 & S2$_{ROM}$ & S3 & S3$_{ROM}$ \\ \hline
		RNN & $0.161$ & $0.087 \;\, \& \;\, 0.069 $ & $0.189$ & $0.079 \;\, \& \;\, 0.074$ \\ \hline
		CNN & $0.064$ & $0.041 \;\, \& \;\, 0.069 $ & $0.068$ & $0.045 \;\, \& \;\, 0.074$ \\ \hline
	\end{tabular}
	\caption{RRMSE from the predictions obtained using the RNN and CNN models, for cases S2, S2$_{ROM}$, S3 and S3$_{ROM}$. The sub-index $ROM$ represents reconstructed data sets through HODMD. Note that in both cases S2$_{ROM}$ and S3$_{ROM}$, we indicate the actual prediction error as well as the reconstruction error, listed in Table \ref{RRMSE reconstruccion}, also measured with RRMSE, where $0.069$ is the reconstruction error for case S2$_{ROM}$ and $0.074$ for case S3$_{ROM}$.}
	\label{tab: rrmse_s2_s3}
\end{table}

\subsection{Modified Geometry: geometry with bluff body (M2, M3, M2$_{ROM}$ and M3$_{ROM}$).}
This section compares the predictions obtained from NNs when they are trained with data sets from Table \ref{tab: train_cases_modified_geometry} (i.e., data sets corresponding to geometry with bluff body). In this meaning, the predictions will be compared with cases M2, M3, M2$_{ROM}$ and M3$_{ROM}$ (Table \ref{tab: orig_hodmd_nomenclatura}).

\begin{figure}[H]
	\centering
	\begin{tabular}{lccc}
		Sample & Simulation & RNN & CNN\\
		$t = 159$&
		\includegraphics[width=0.27\textwidth, angle=0]{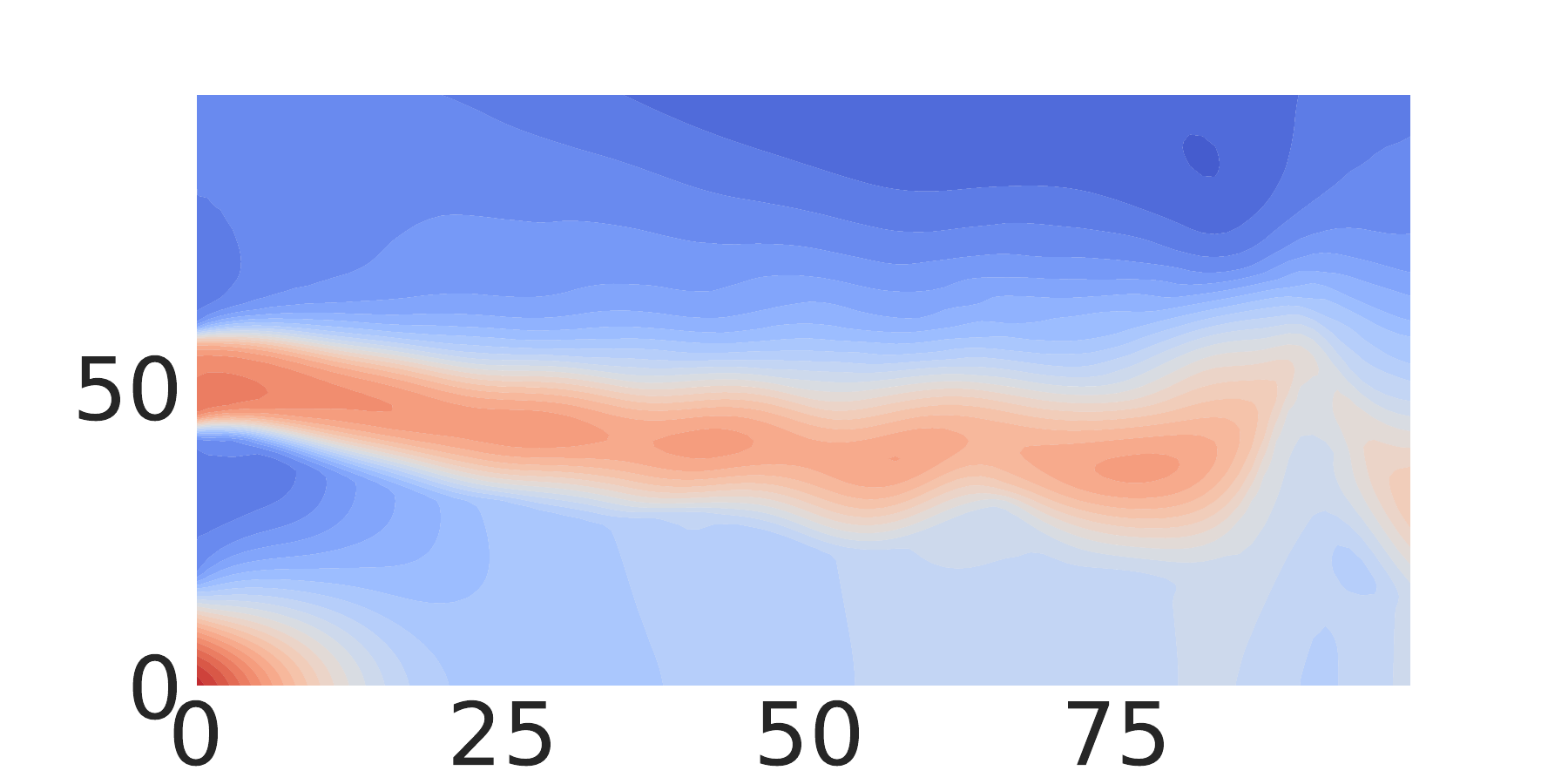}&
		\includegraphics[width=0.27\textwidth, angle=0]{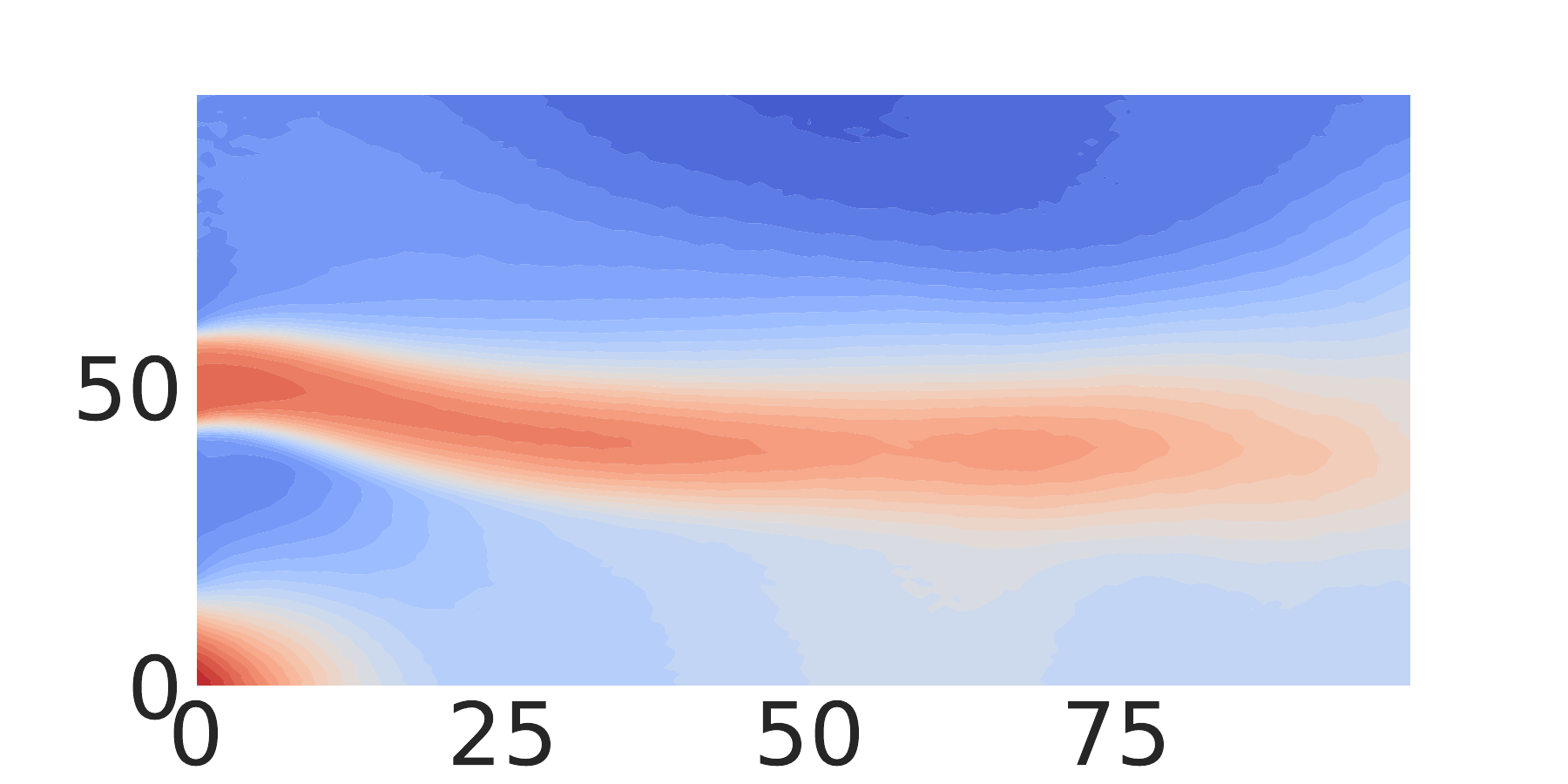}&
		\includegraphics[width=0.27\textwidth, angle=0]{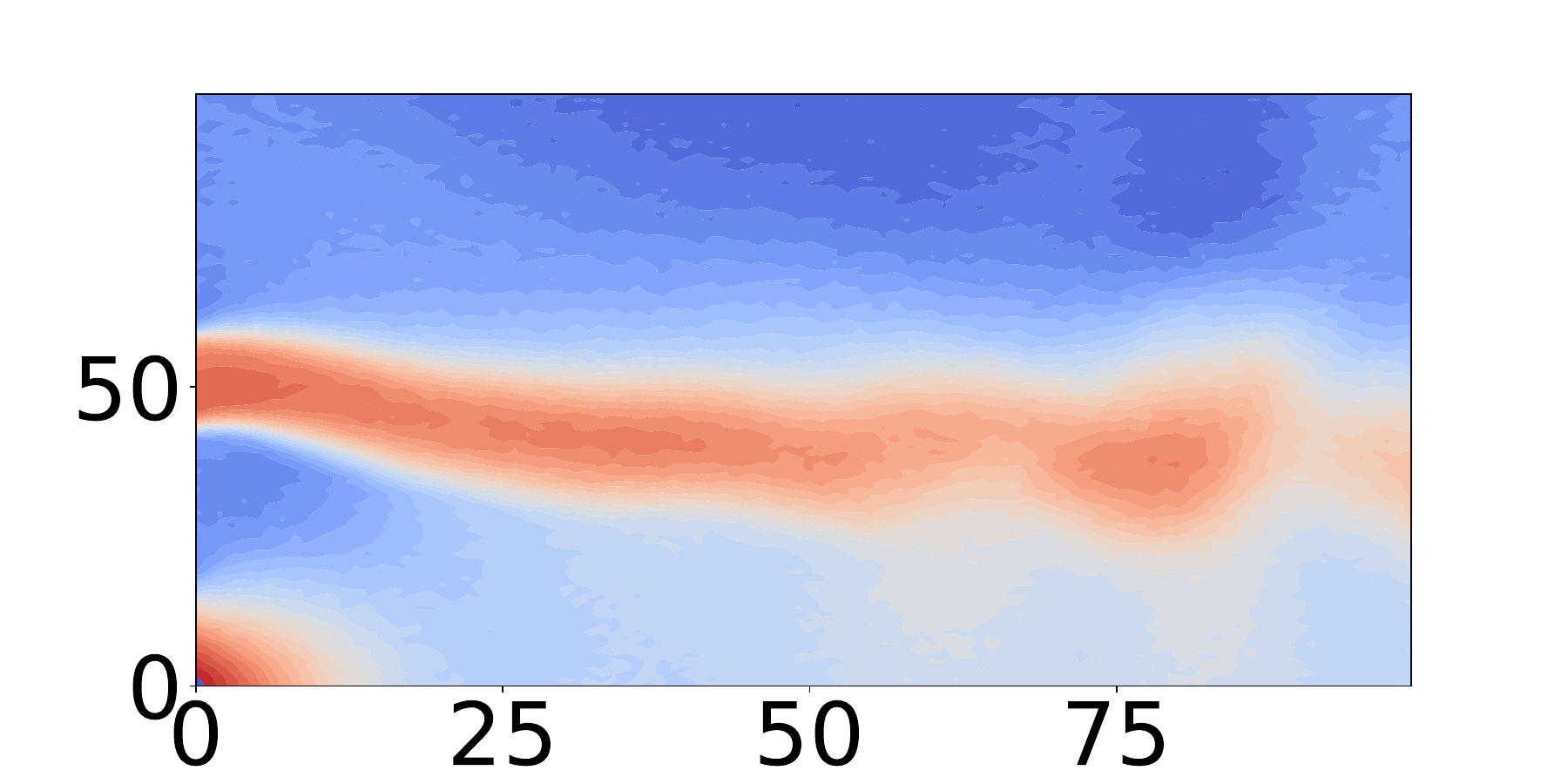}\\
		$t=160$&
		\includegraphics[width=0.27\textwidth, angle=0]{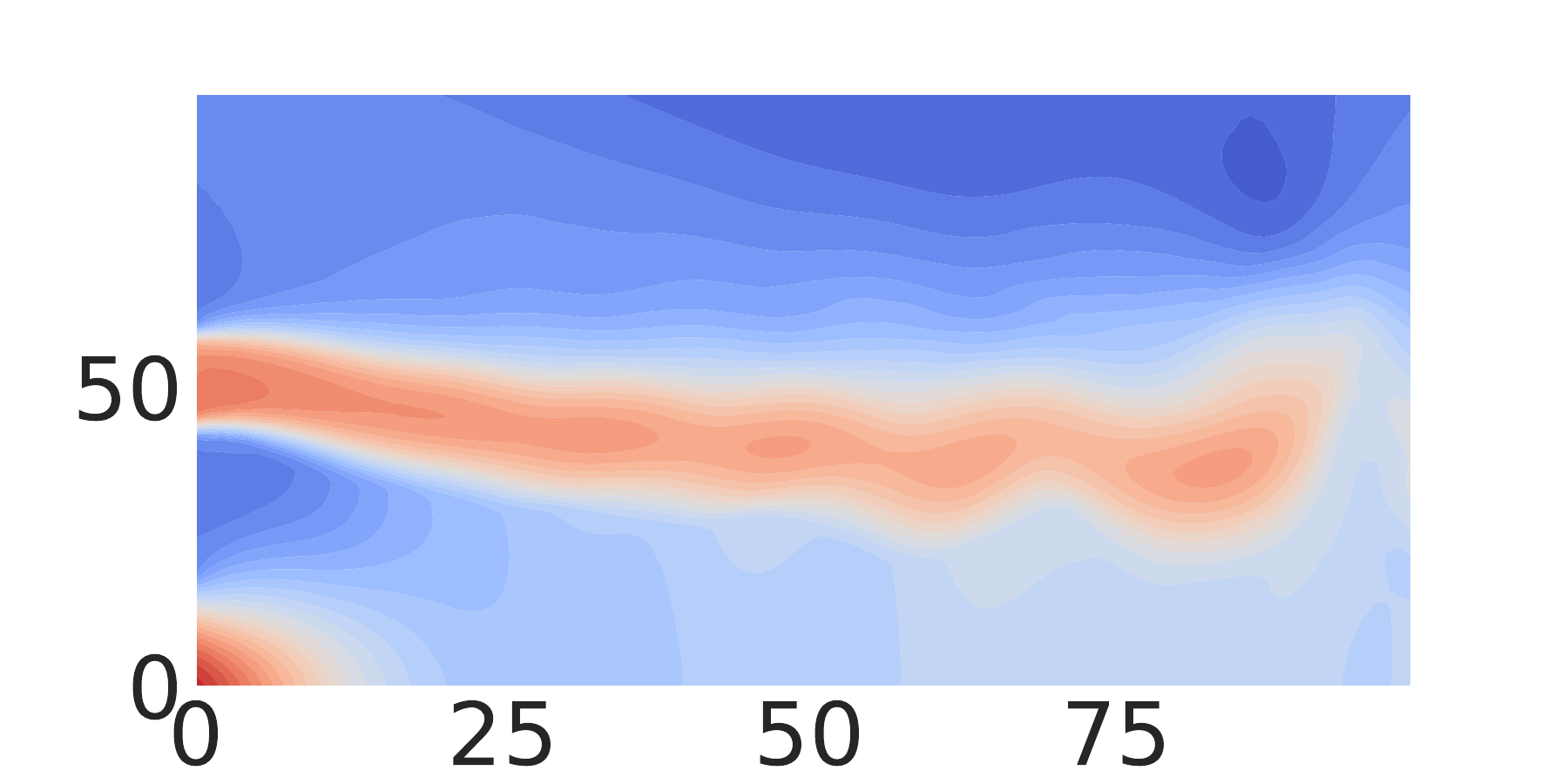}&
		\includegraphics[width=0.27\textwidth, angle=0]{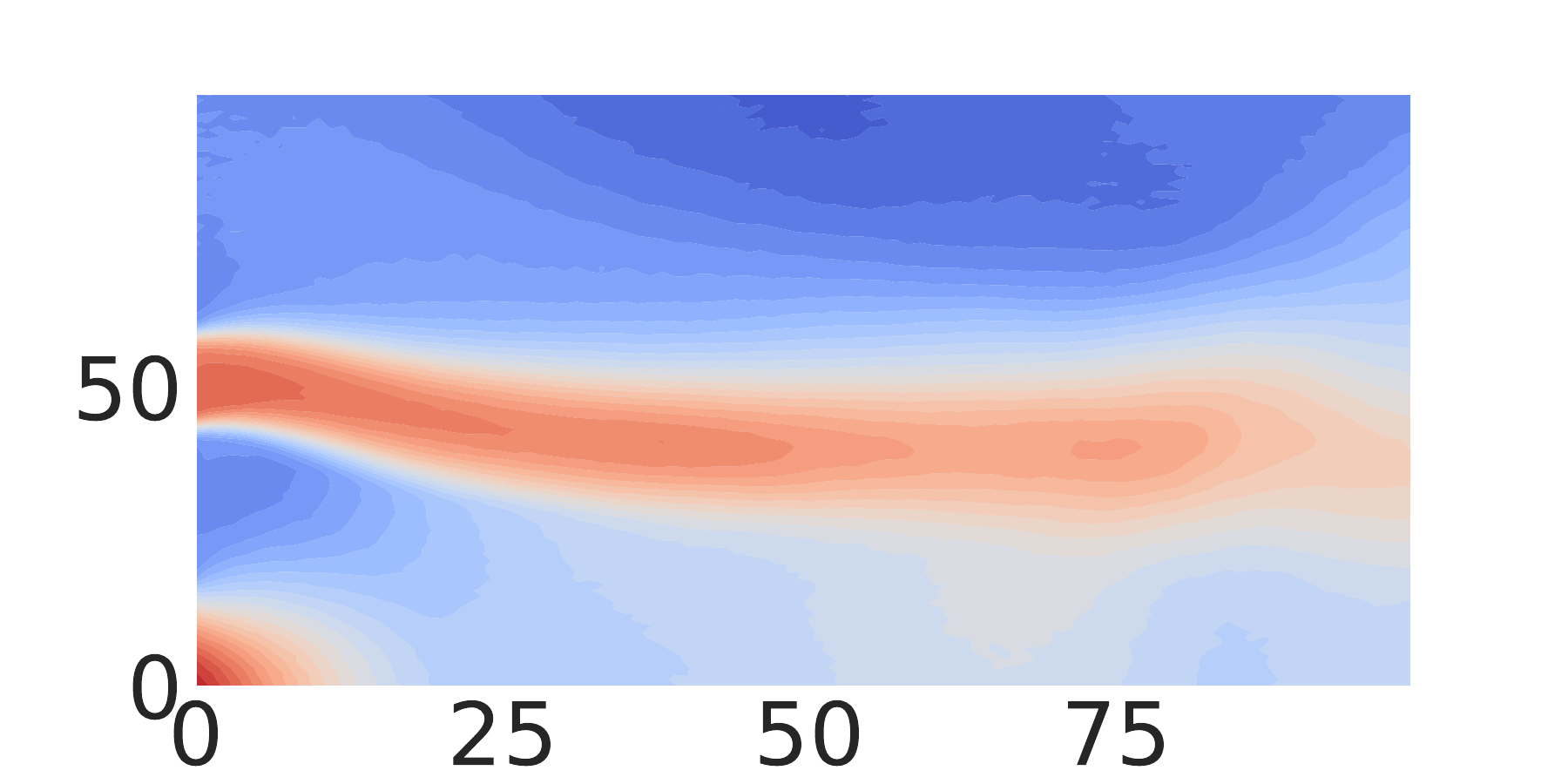}&
		\includegraphics[width=0.27\textwidth, angle=0]{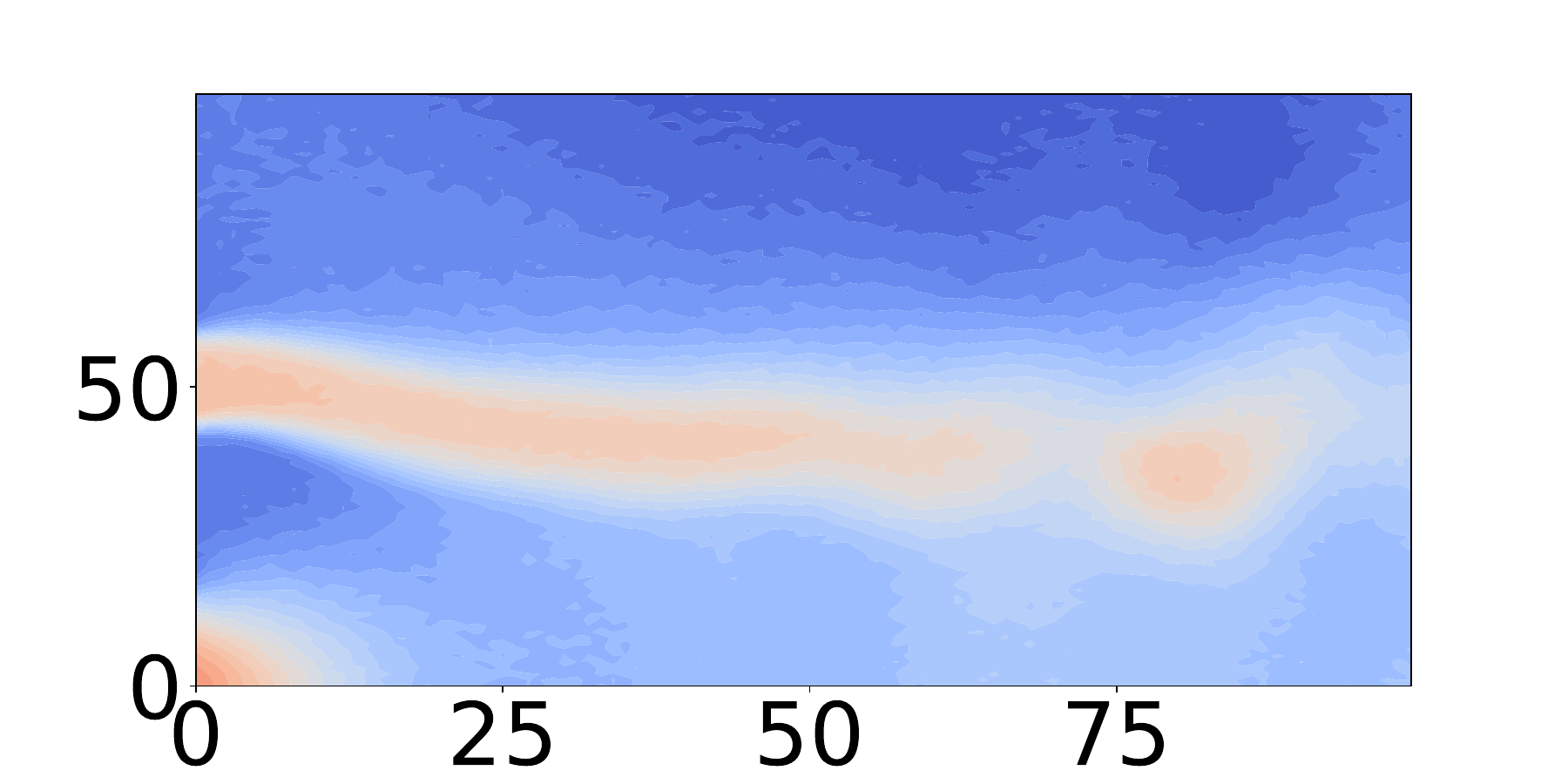}\\
		$t = 174$&
		\includegraphics[width=0.27\textwidth, angle=0]{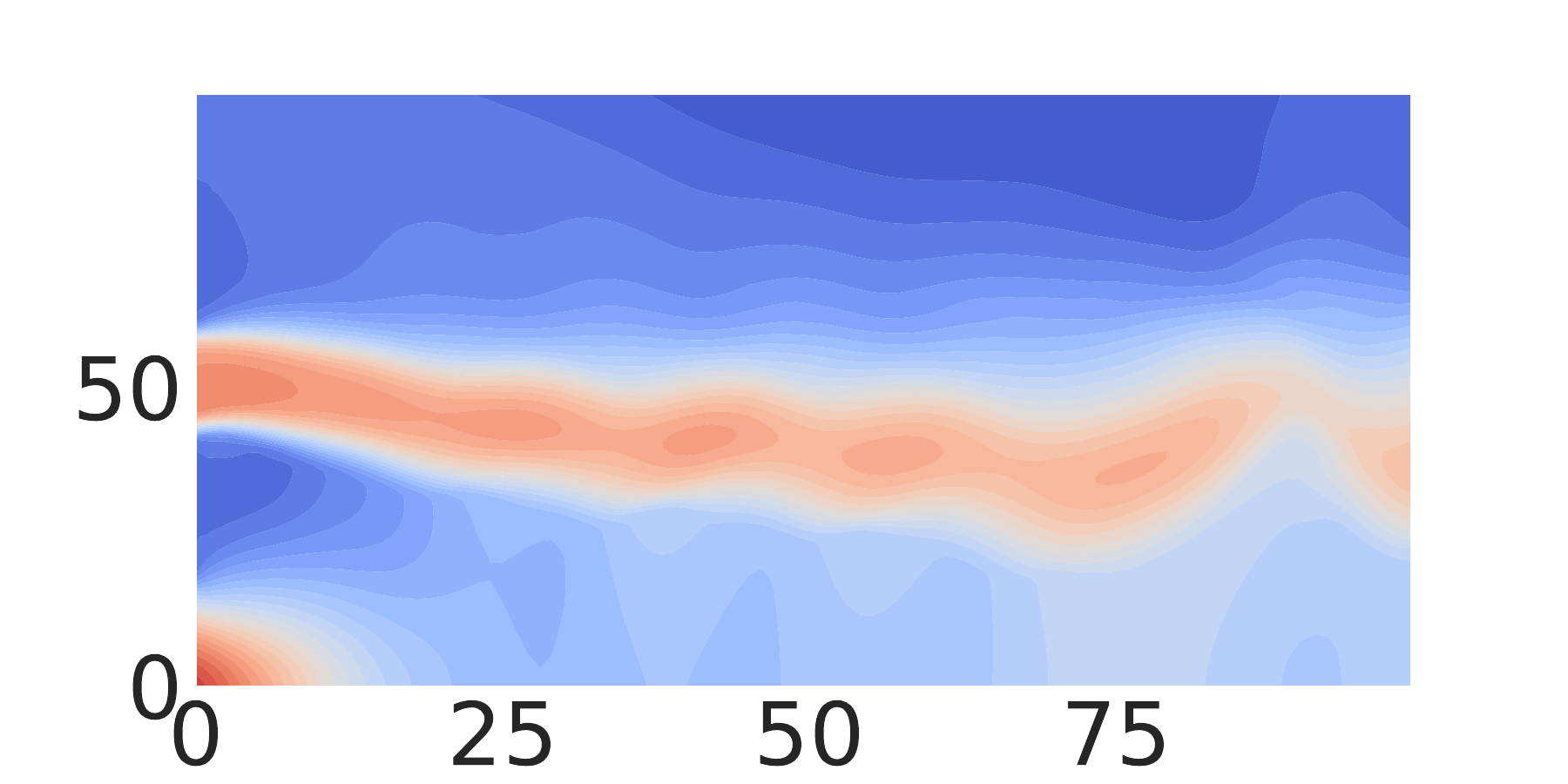}&
		\includegraphics[width=0.27\textwidth, angle=0]{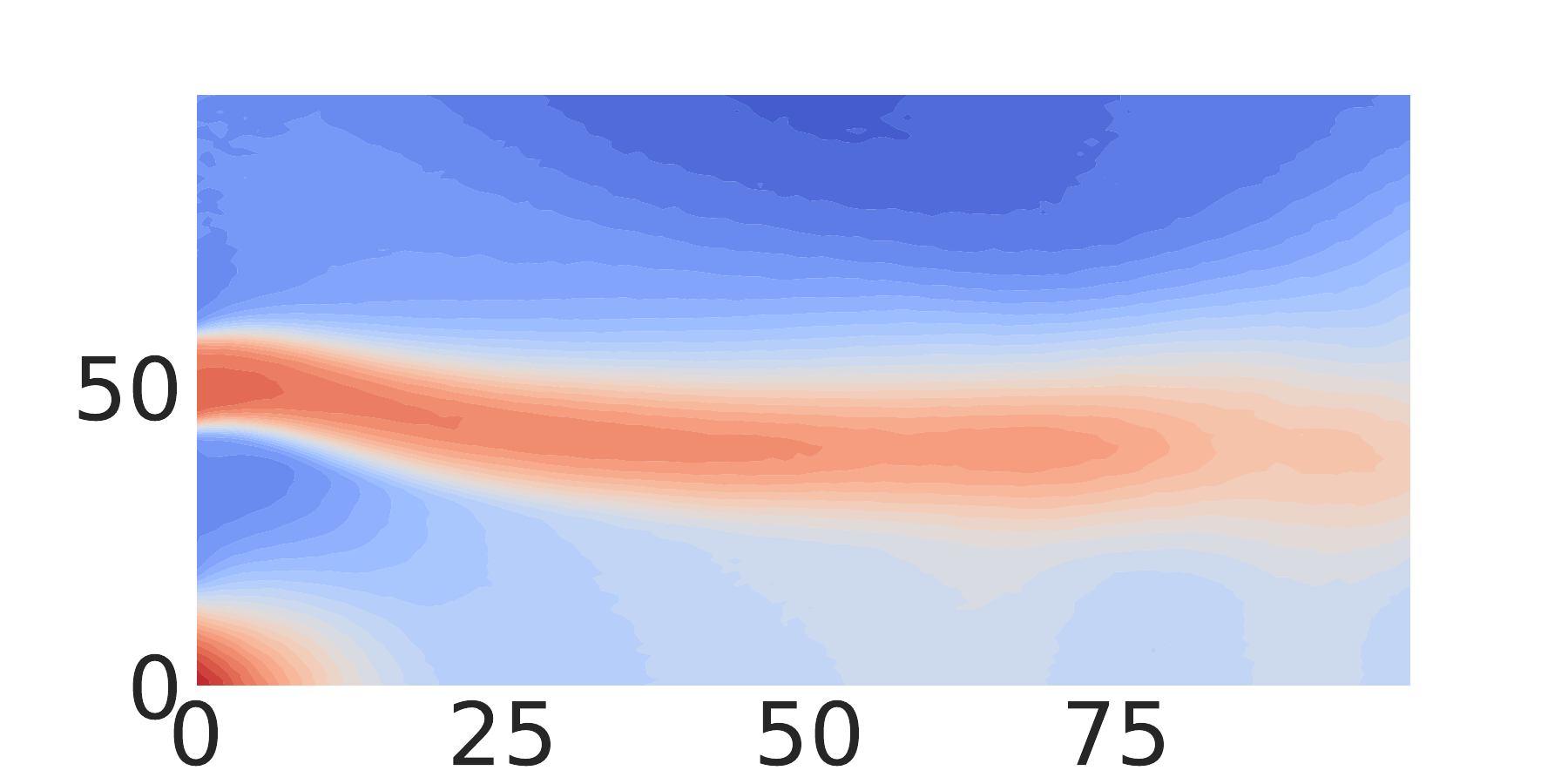}&
		\includegraphics[width=0.27\textwidth, angle=0]{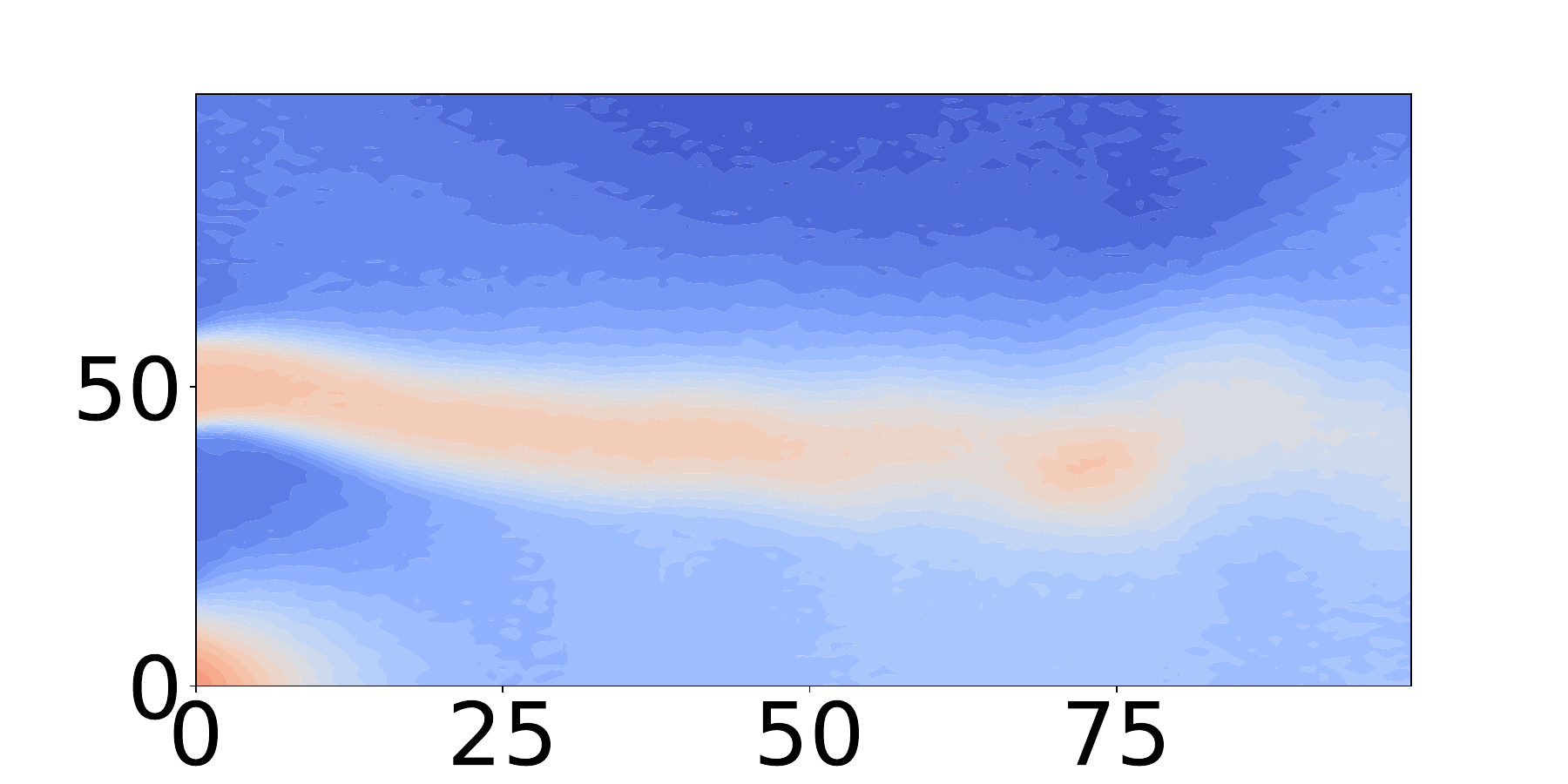}\\
		$t=175$&
		\includegraphics[width=0.27\textwidth, angle=0]{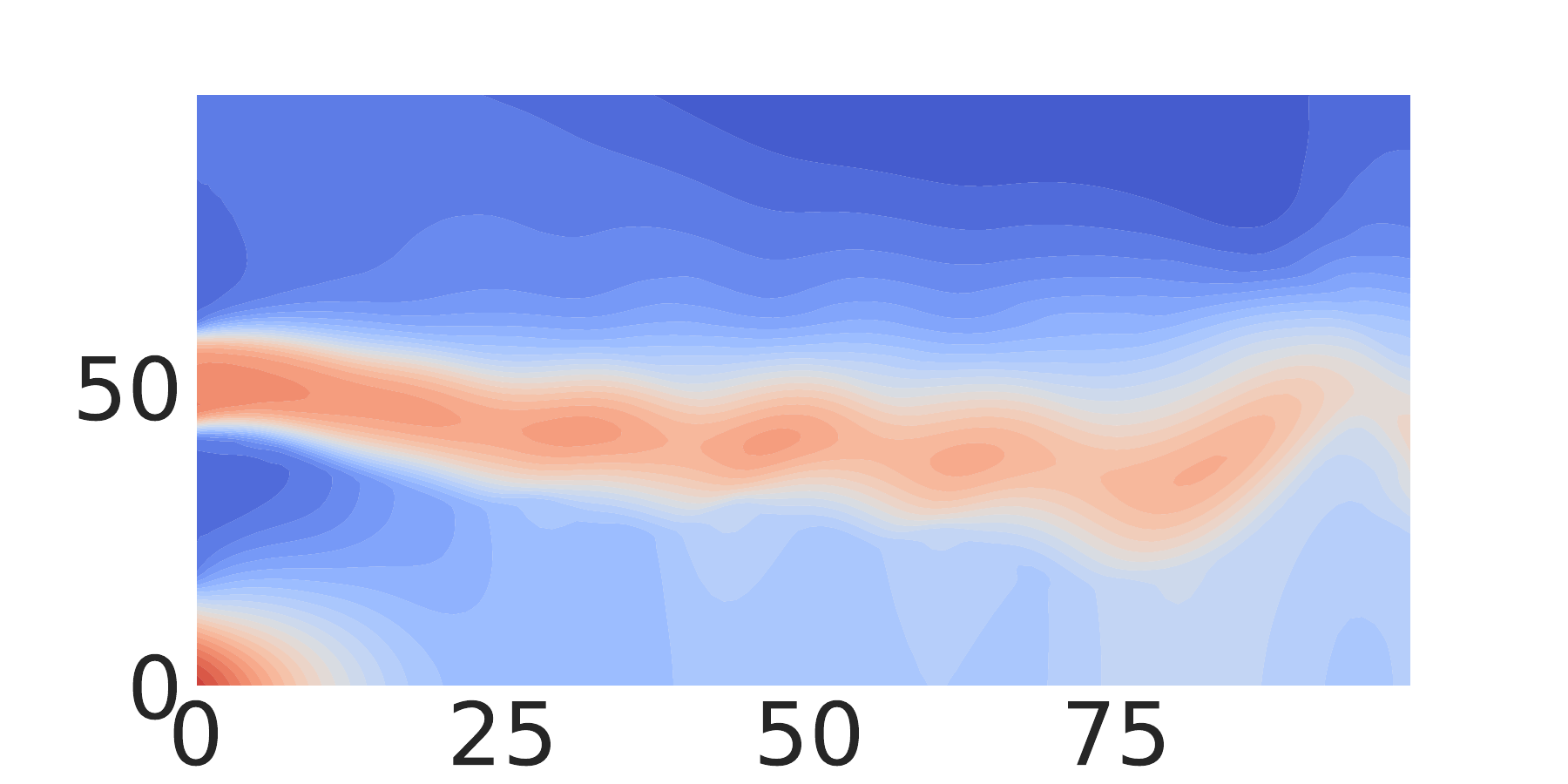}&
		\includegraphics[width=0.27\textwidth, angle=0]{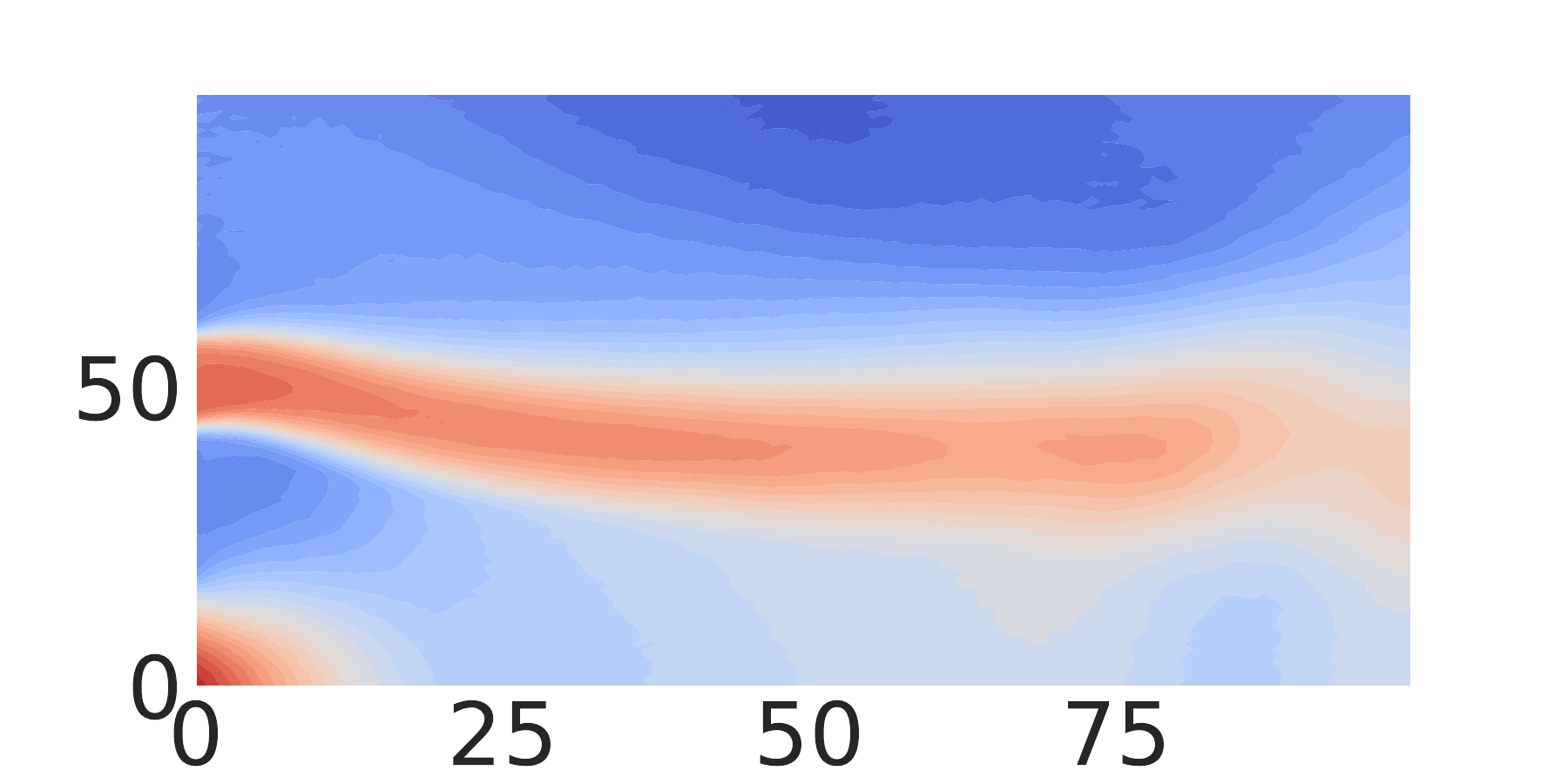}&
		\includegraphics[width=0.27\textwidth, angle=0]{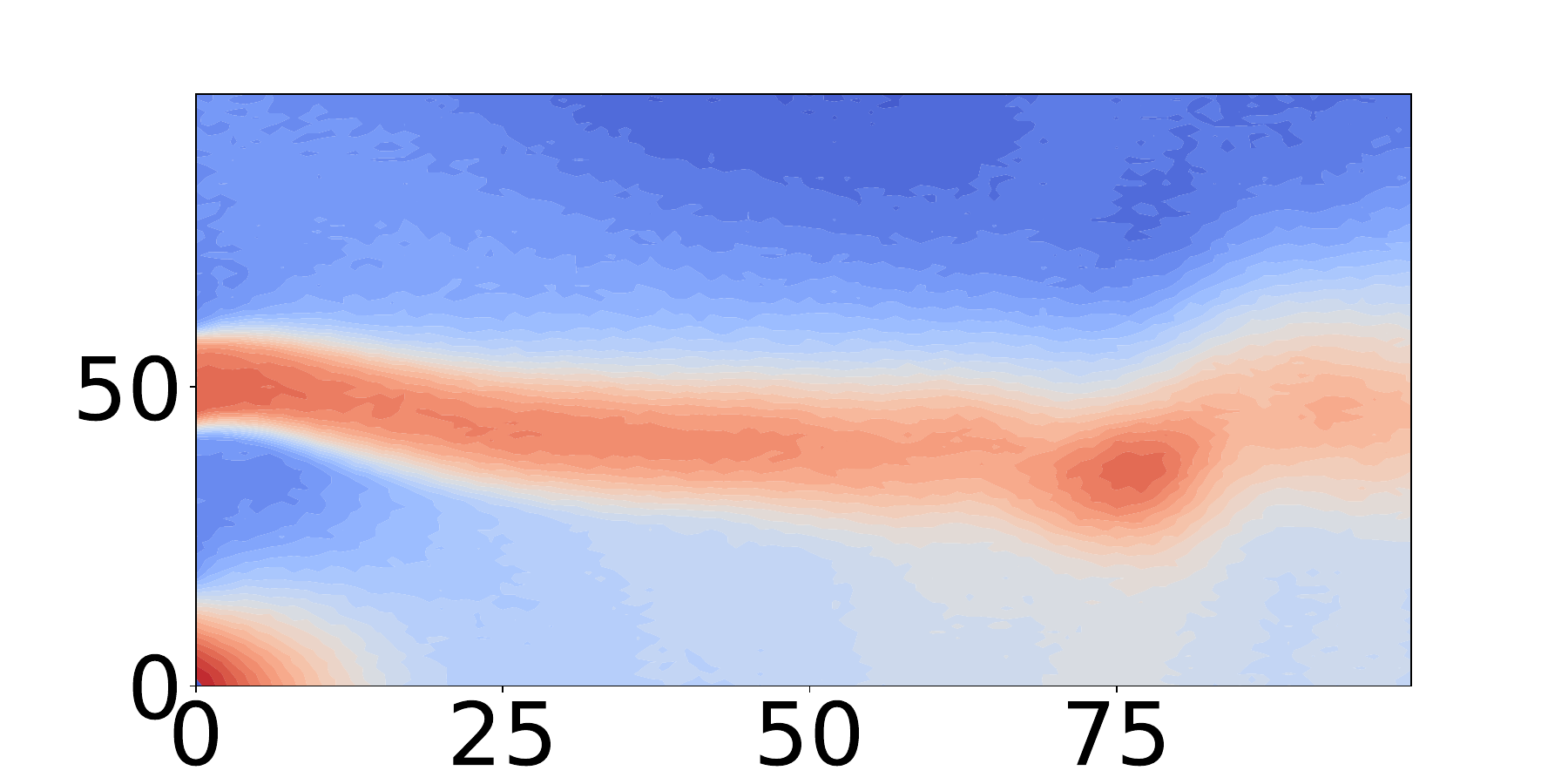}\\
	\end{tabular}
	\caption{From left to right: Snapshots from original data set, prediction of RNN model and prediction of CNN model, respectively, corresponding to case M2. To generate these predictions the following samples were sent to both NNs; $\{v_{158}, v_{157}, \dots, v_{149}\}$ and $\{v_{173}, v_{172}, \dots, v_{164}\}$. Note that all these samples belong to the test set, Table \ref{tab:ML3}.}
	\label{fig: ann_bluff_sts_sim}
\end{figure}

\begin{figure}[H]
	\centering
	\begin{tabular}{lccc}
		Sample & Simulation & RNN & CNN\\
		$t = 159$&
		\includegraphics[width=0.27\textwidth, angle=0]{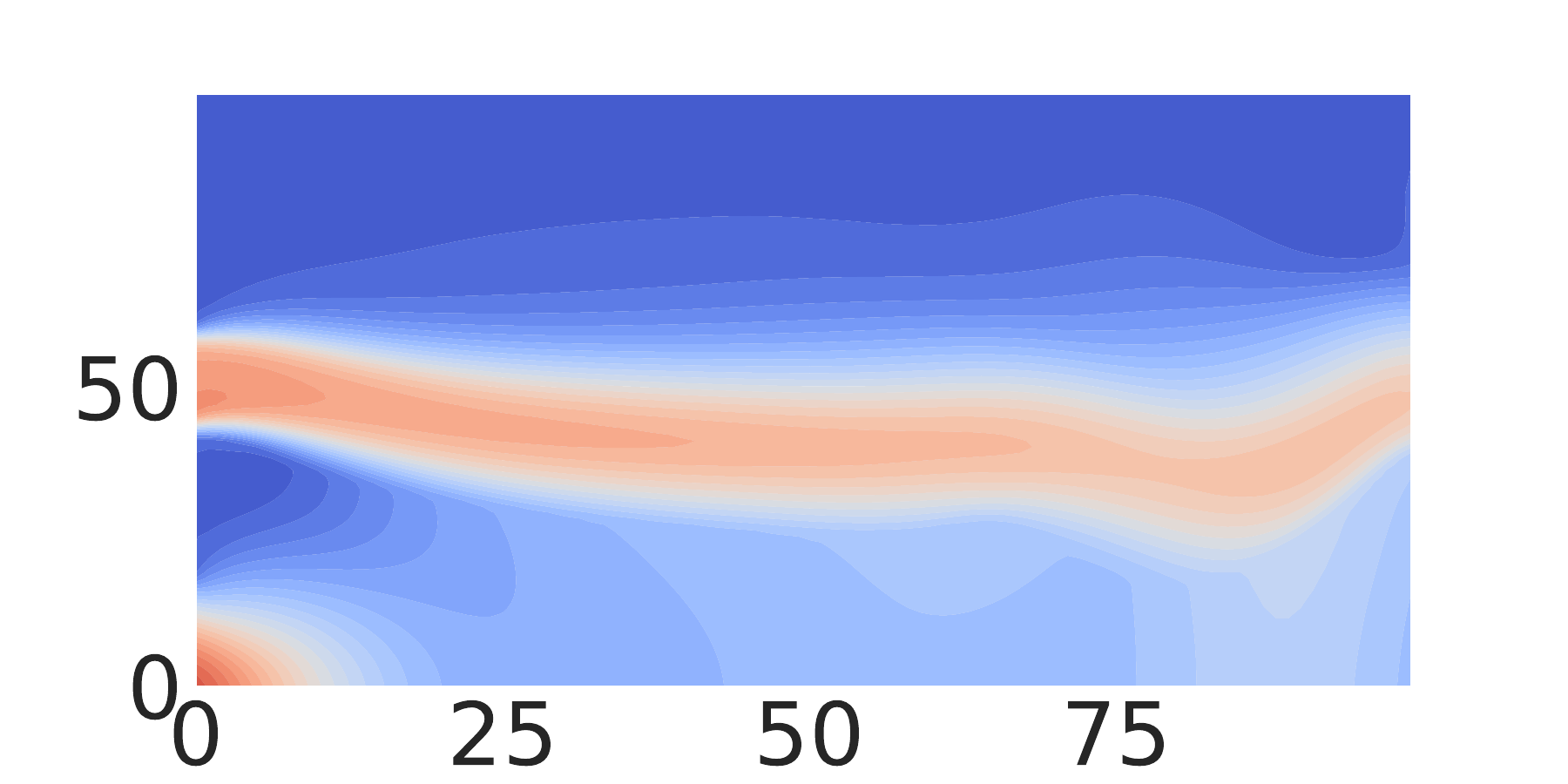}&
		\includegraphics[width=0.27\textwidth, angle=0]{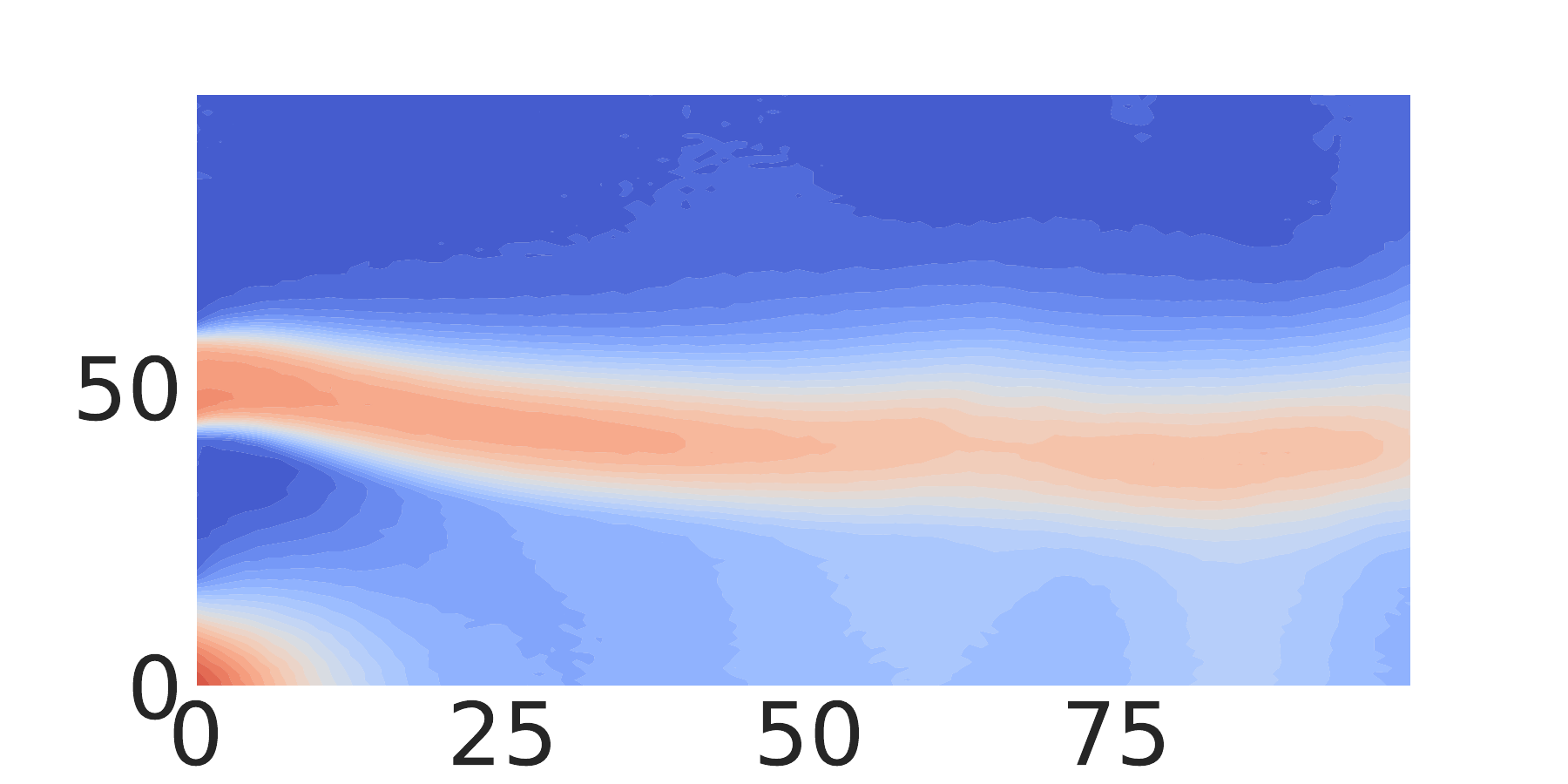}&
		\includegraphics[width=0.27\textwidth, angle=0]{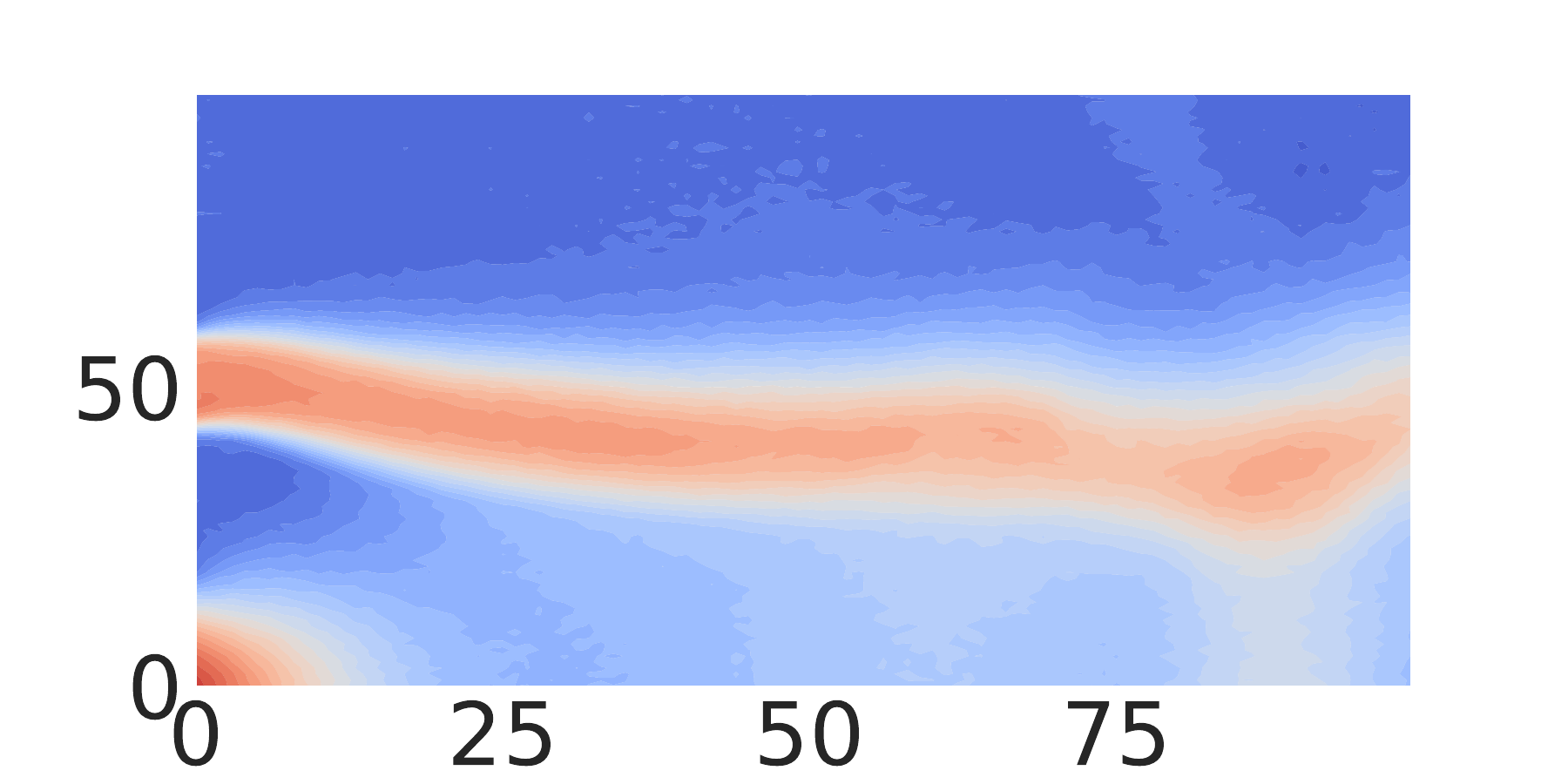}\\
		$t=160$&
		\includegraphics[width=0.27\textwidth, angle=0]{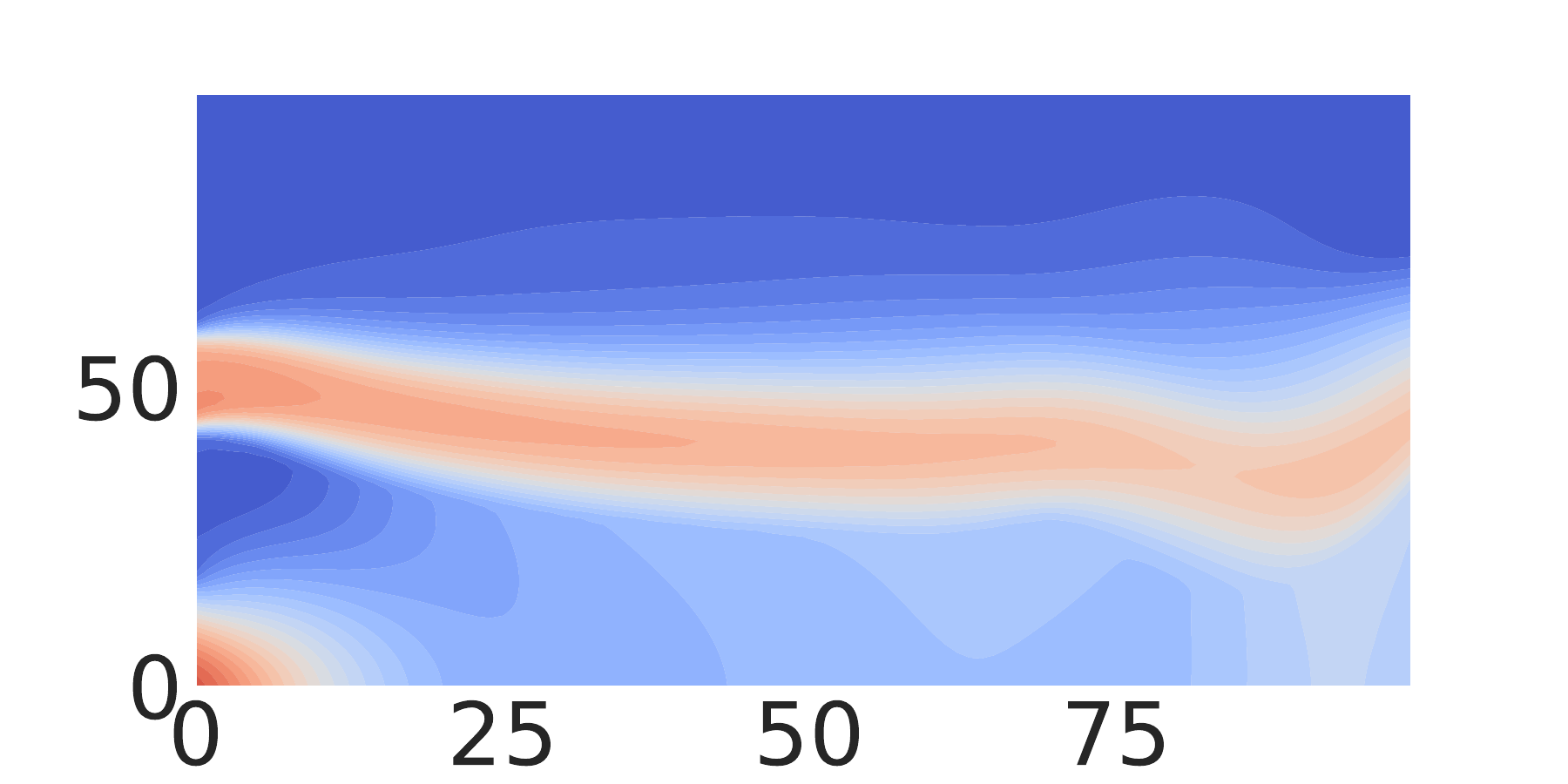}&
		\includegraphics[width=0.27\textwidth, angle=0]{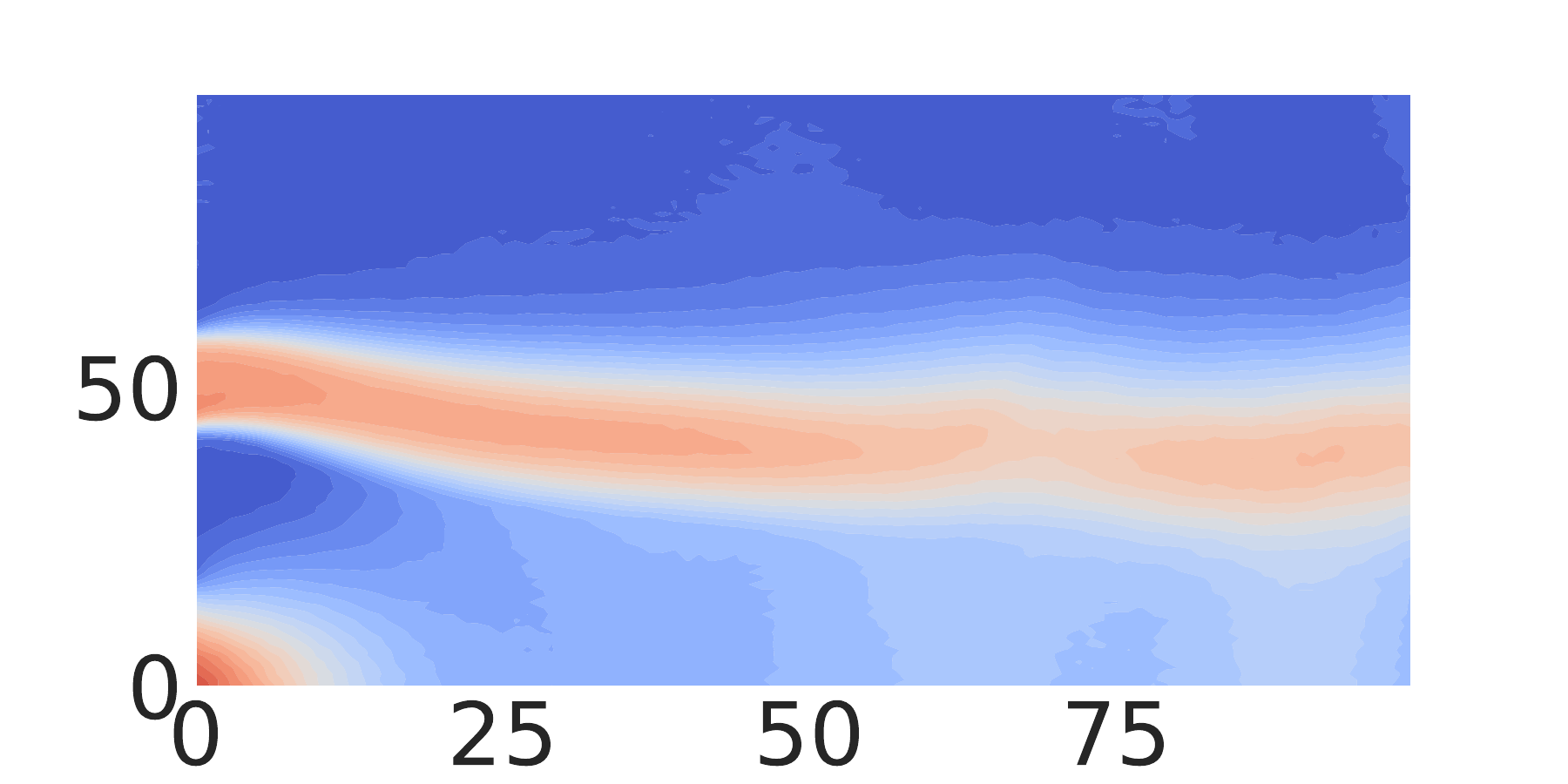}&
		\includegraphics[width=0.27\textwidth, angle=0]{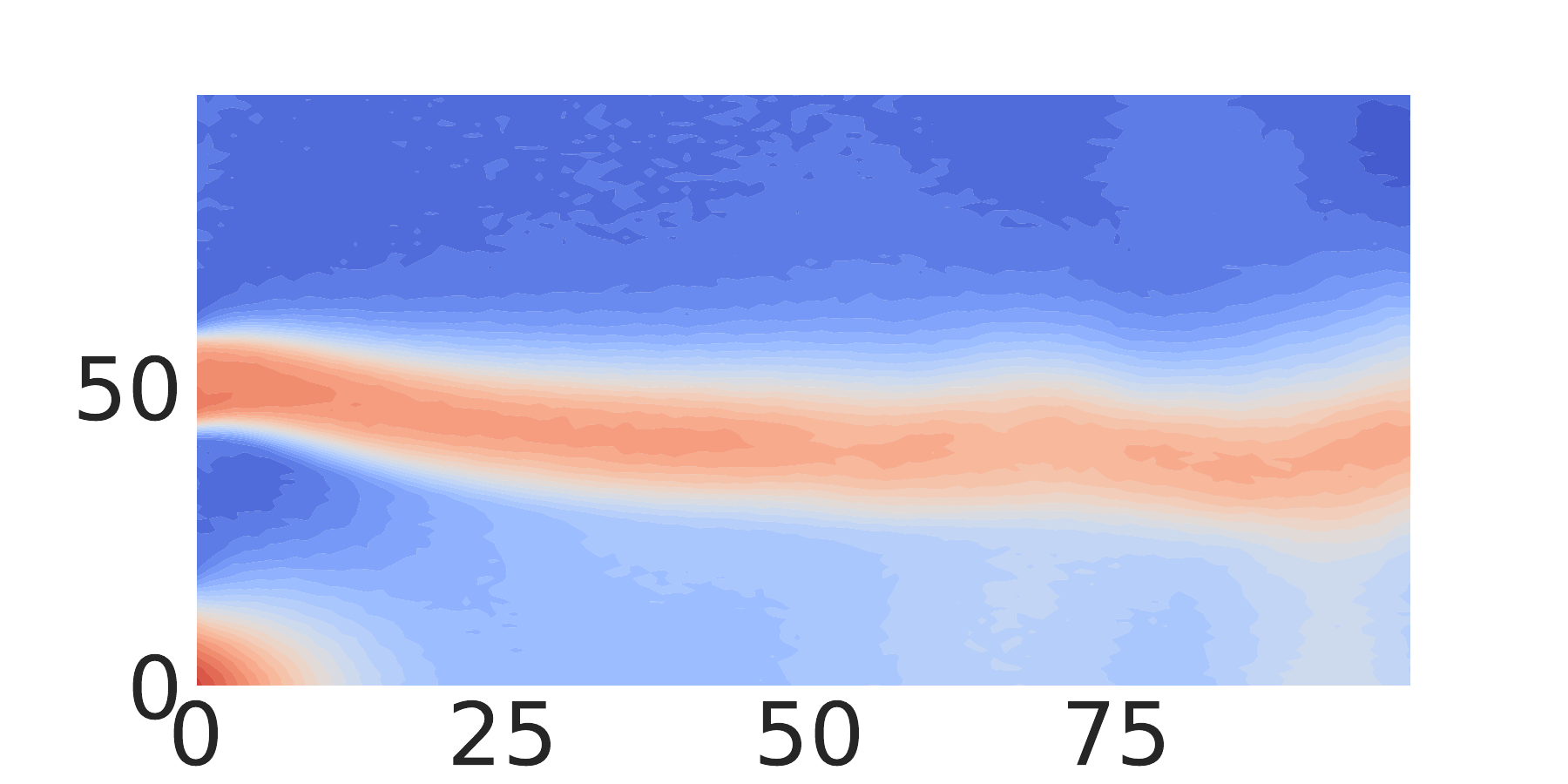}\\
		$t = 174$&
		\includegraphics[width=0.27\textwidth, angle=0]{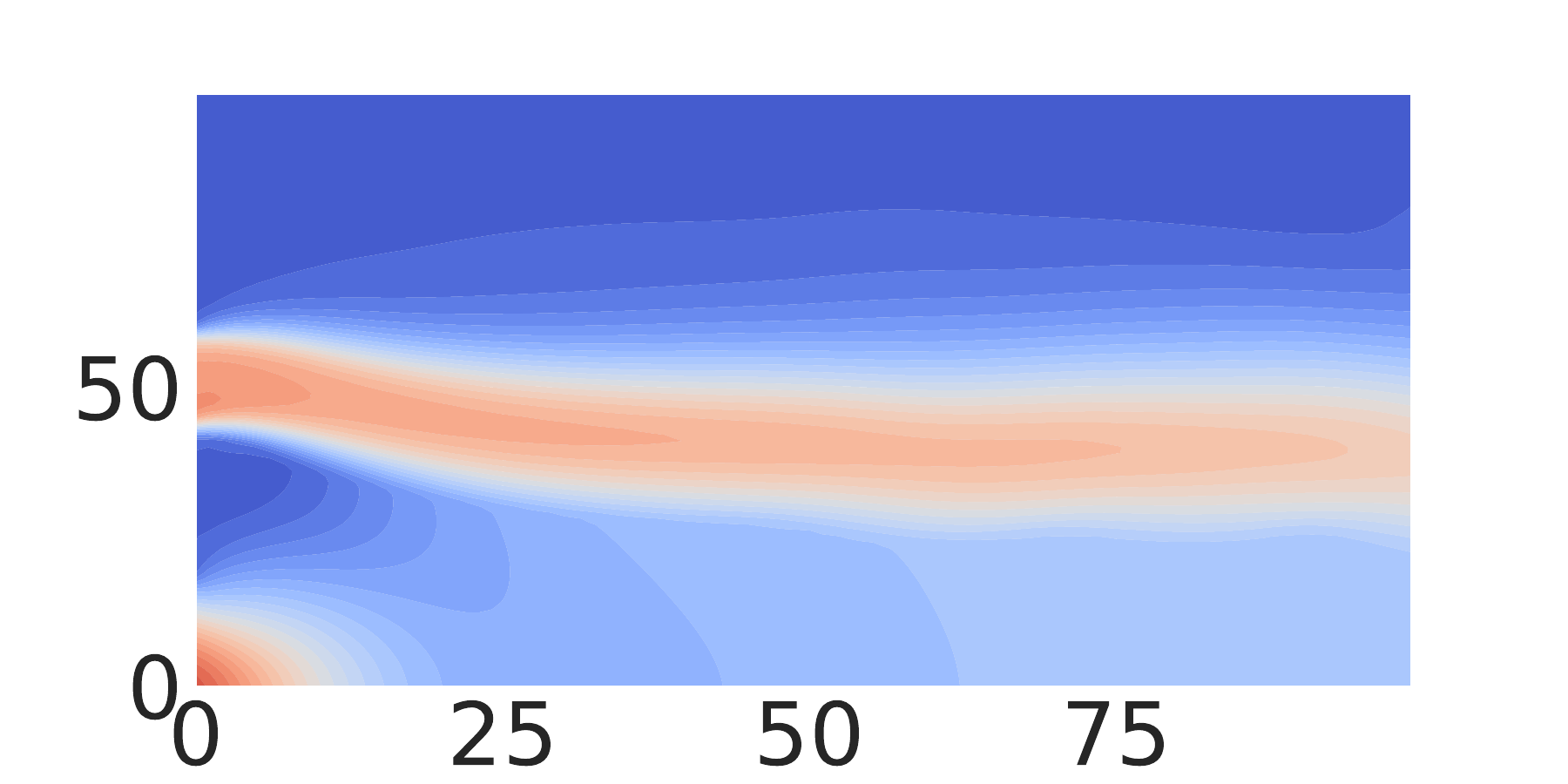}&
		\includegraphics[width=0.27\textwidth, angle=0]{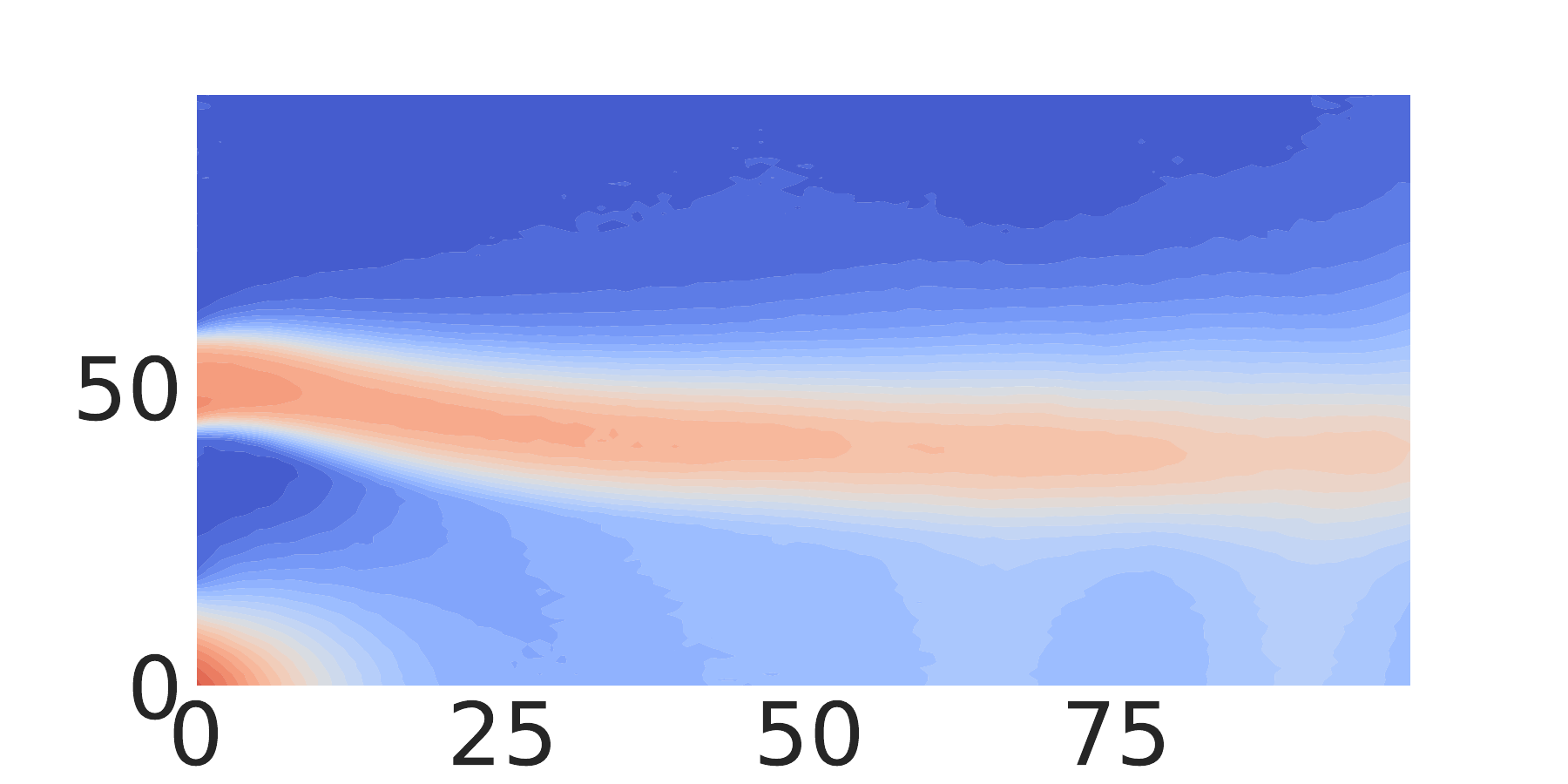}&
		\includegraphics[width=0.27\textwidth, angle=0]{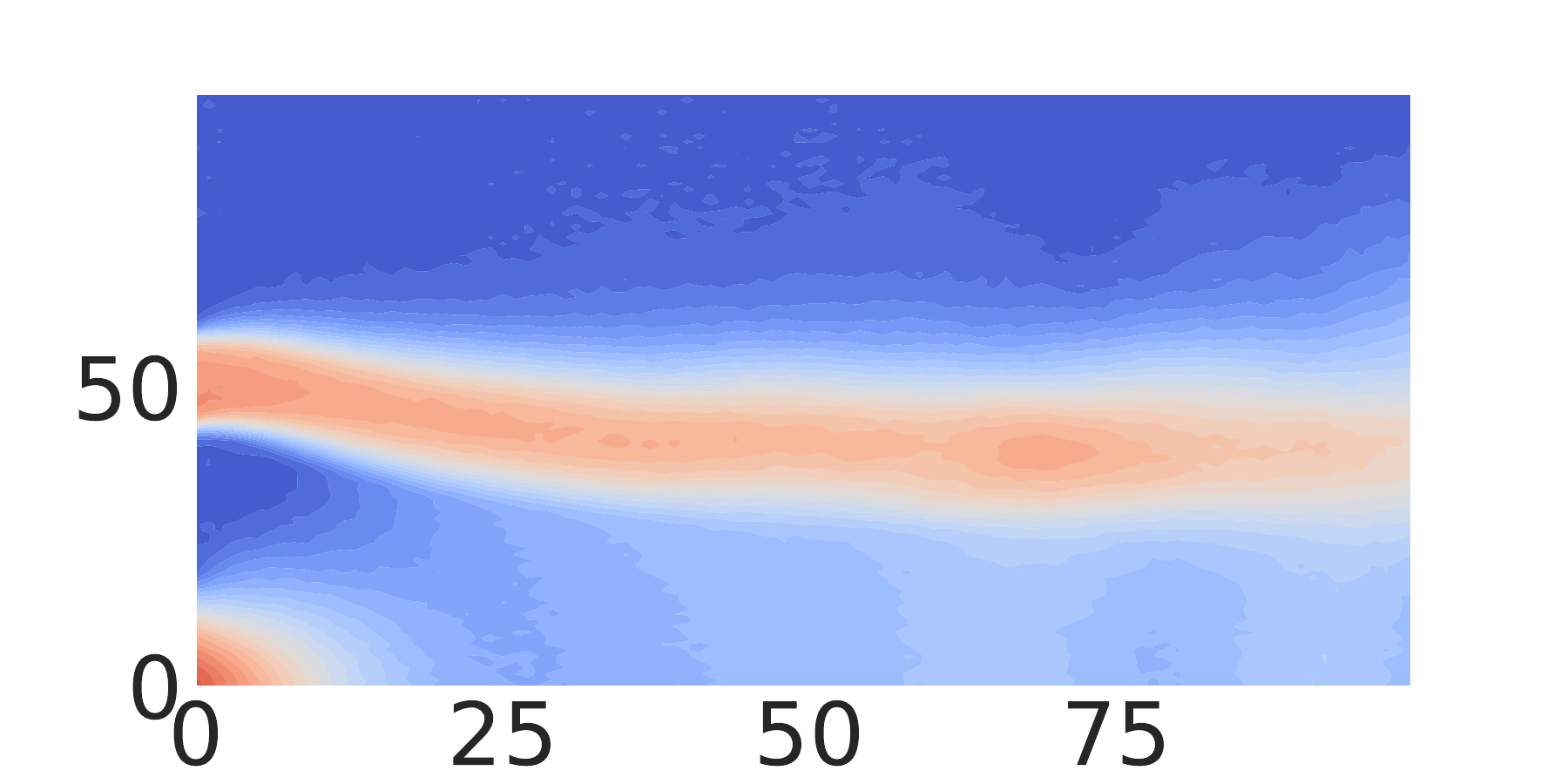}\\
		$t=175$&
		\includegraphics[width=0.27\textwidth, angle=0]{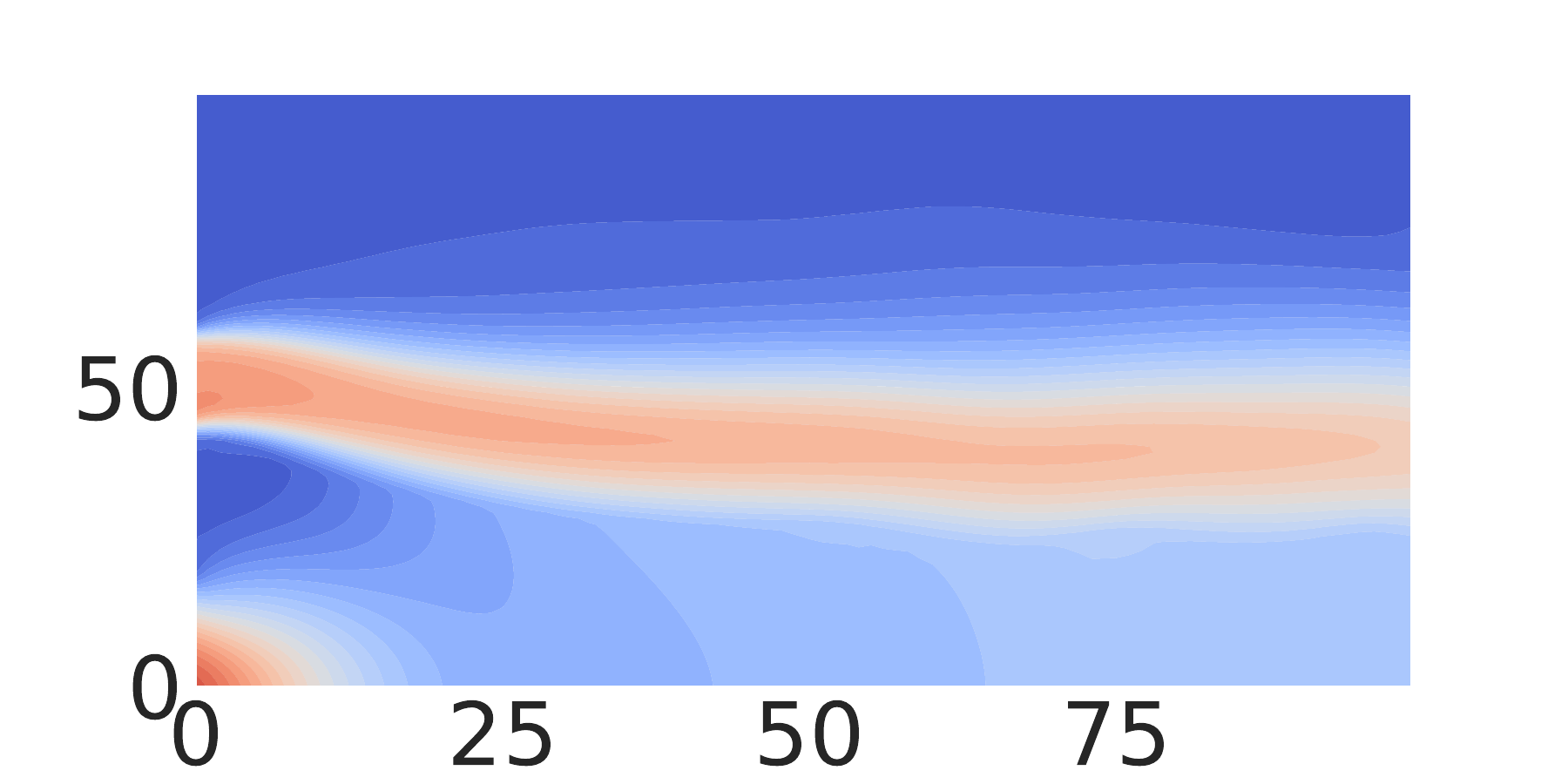}&
		\includegraphics[width=0.27\textwidth, angle=0]{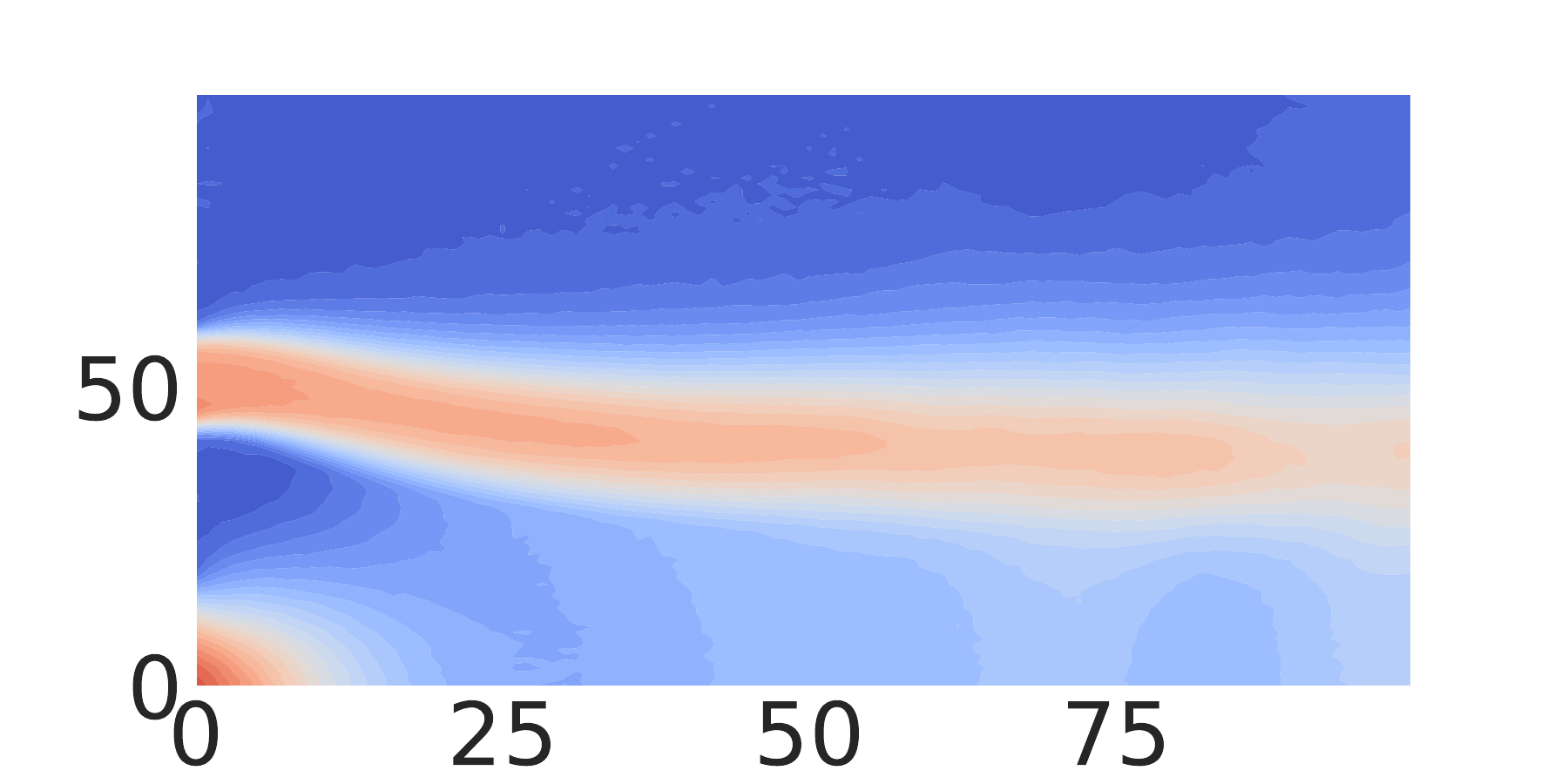}&
		\includegraphics[width=0.27\textwidth, angle=0]{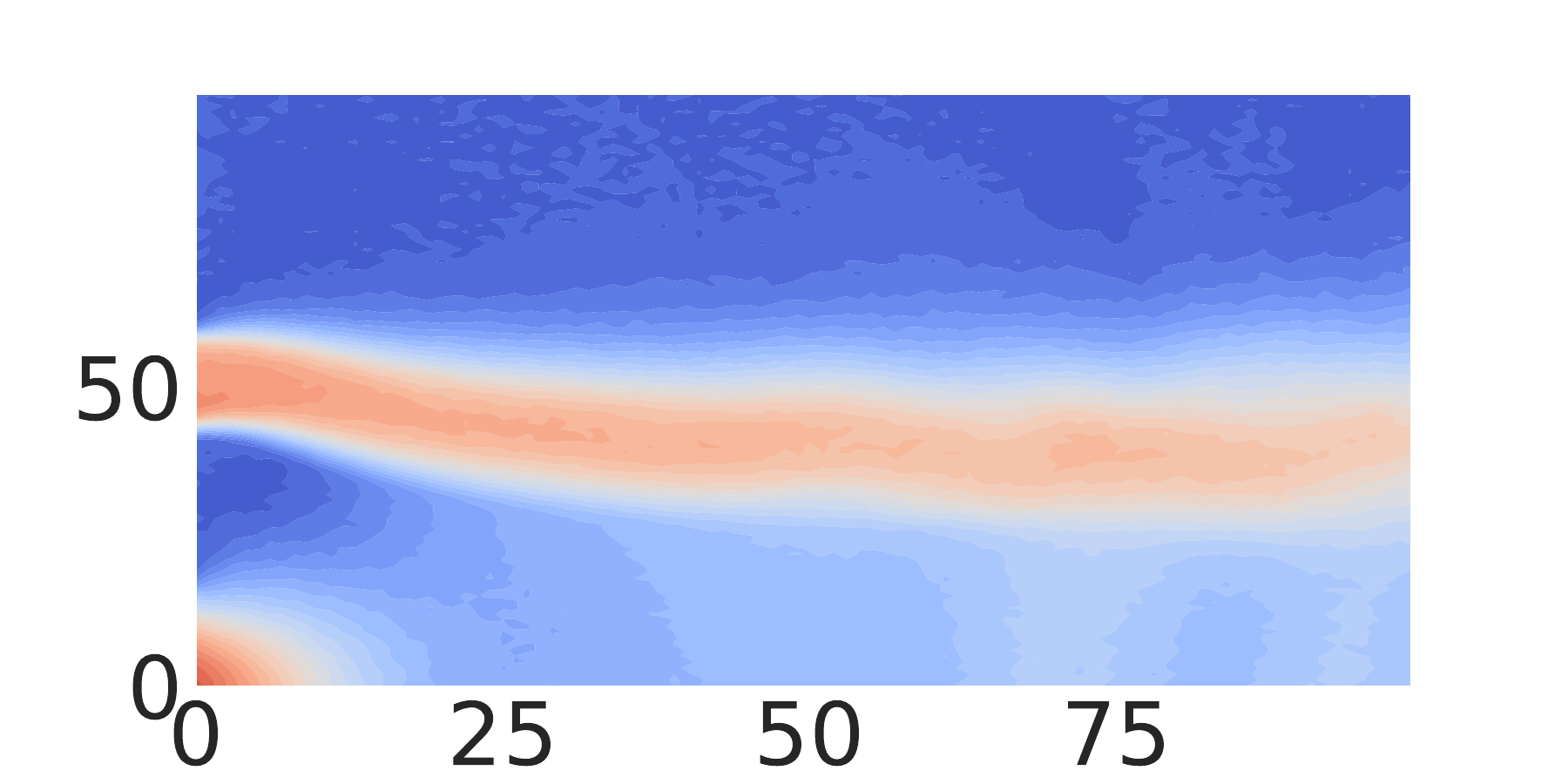}\\
	\end{tabular}
	\caption{Same as Figure \ref{fig: ann_bluff_sts_sim} for case M3.}
	\label{fig: ann_bluff_cts_sim}
\end{figure}

\begin{figure}[H]
	\centering
	\begin{tabular}{lccc}
		Sample & Simulation & RNN & CNN\\
		$t = 159$&
		\includegraphics[width=0.27\textwidth, angle=0]{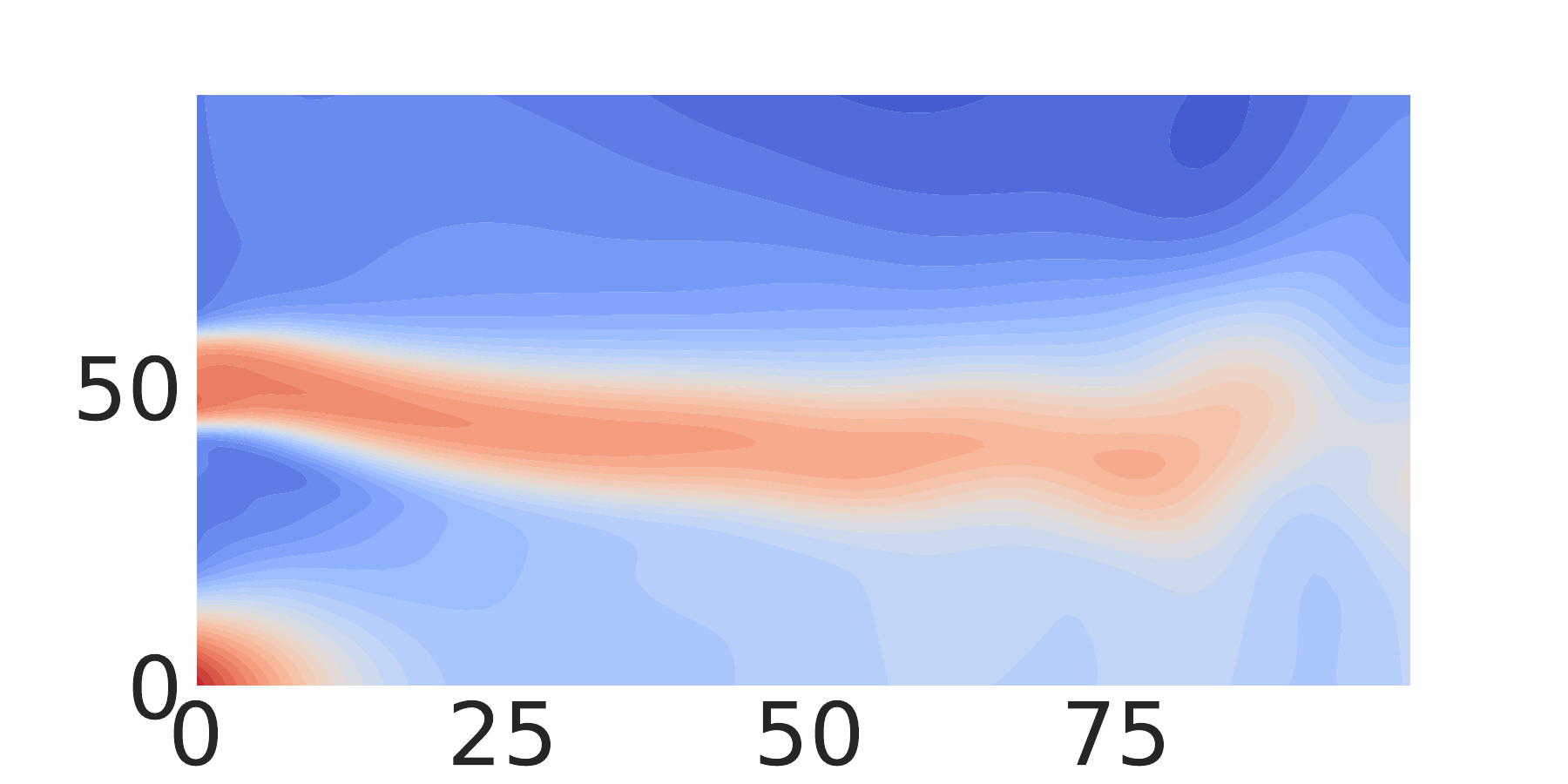}&
		\includegraphics[width=0.27\textwidth, angle=0]{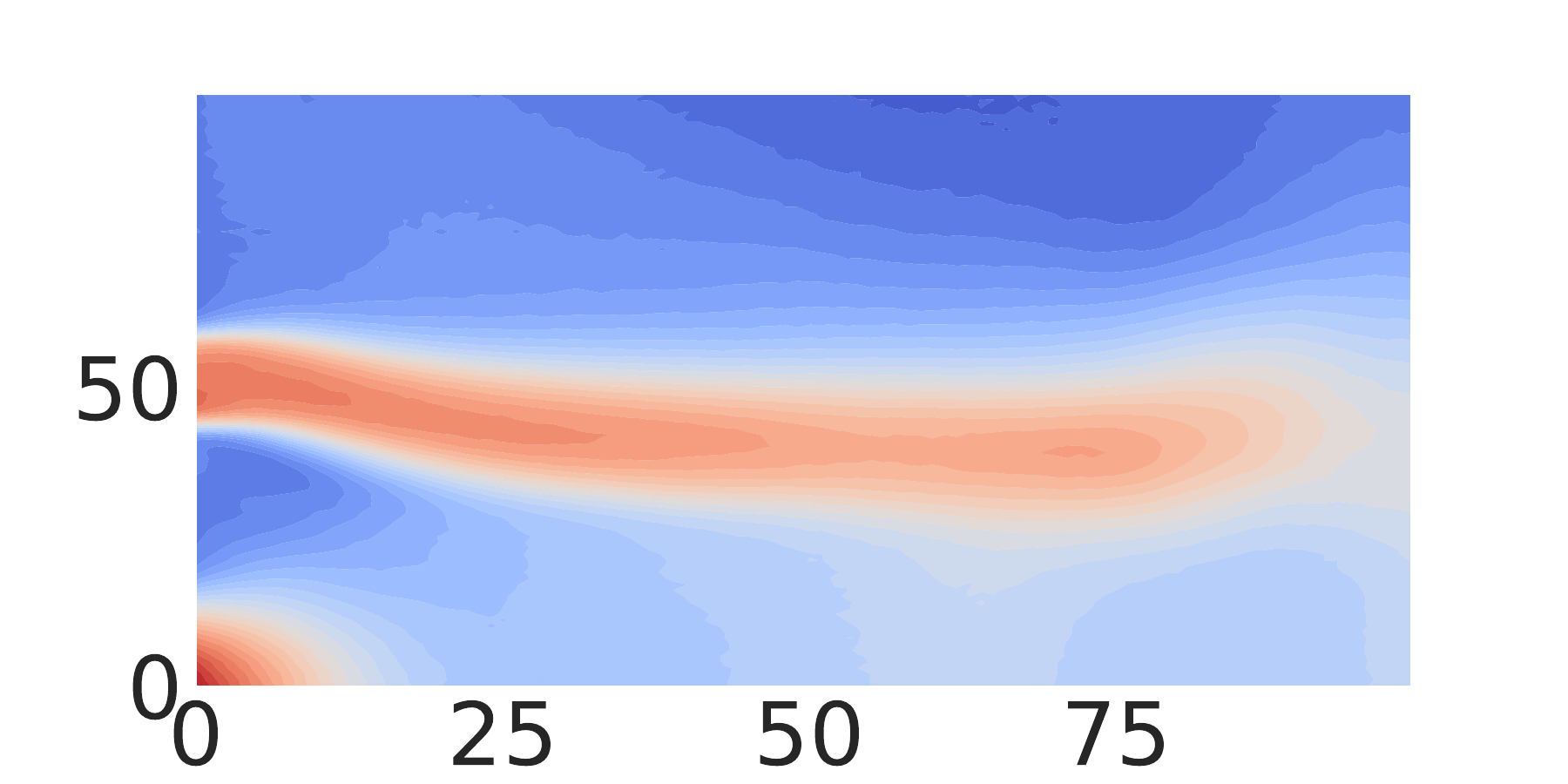}&
		\includegraphics[width=0.27\textwidth, angle=0]{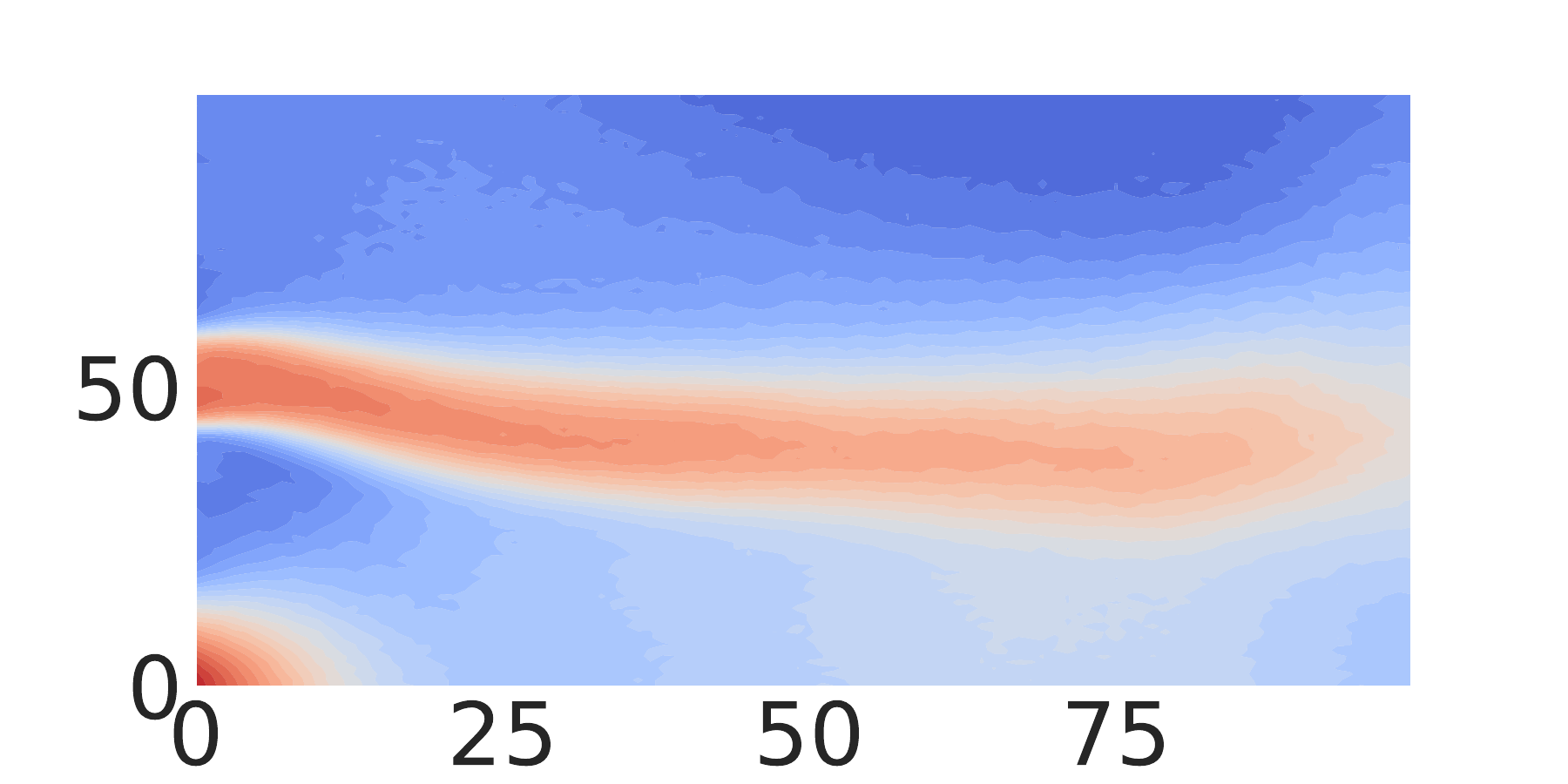}\\
		$t=160$&
		\includegraphics[width=0.27\textwidth, angle=0]{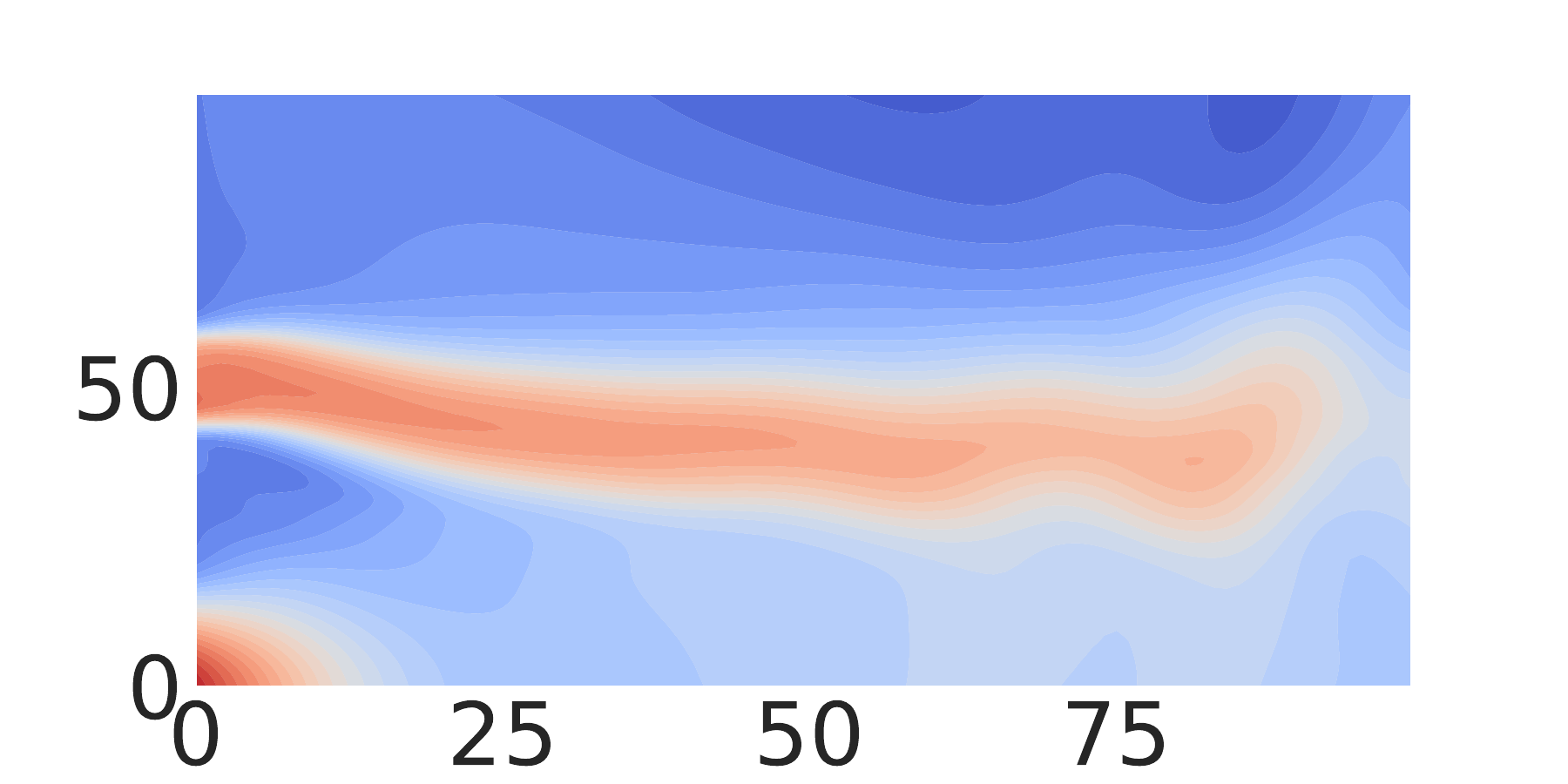}&
		\includegraphics[width=0.27\textwidth, angle=0]{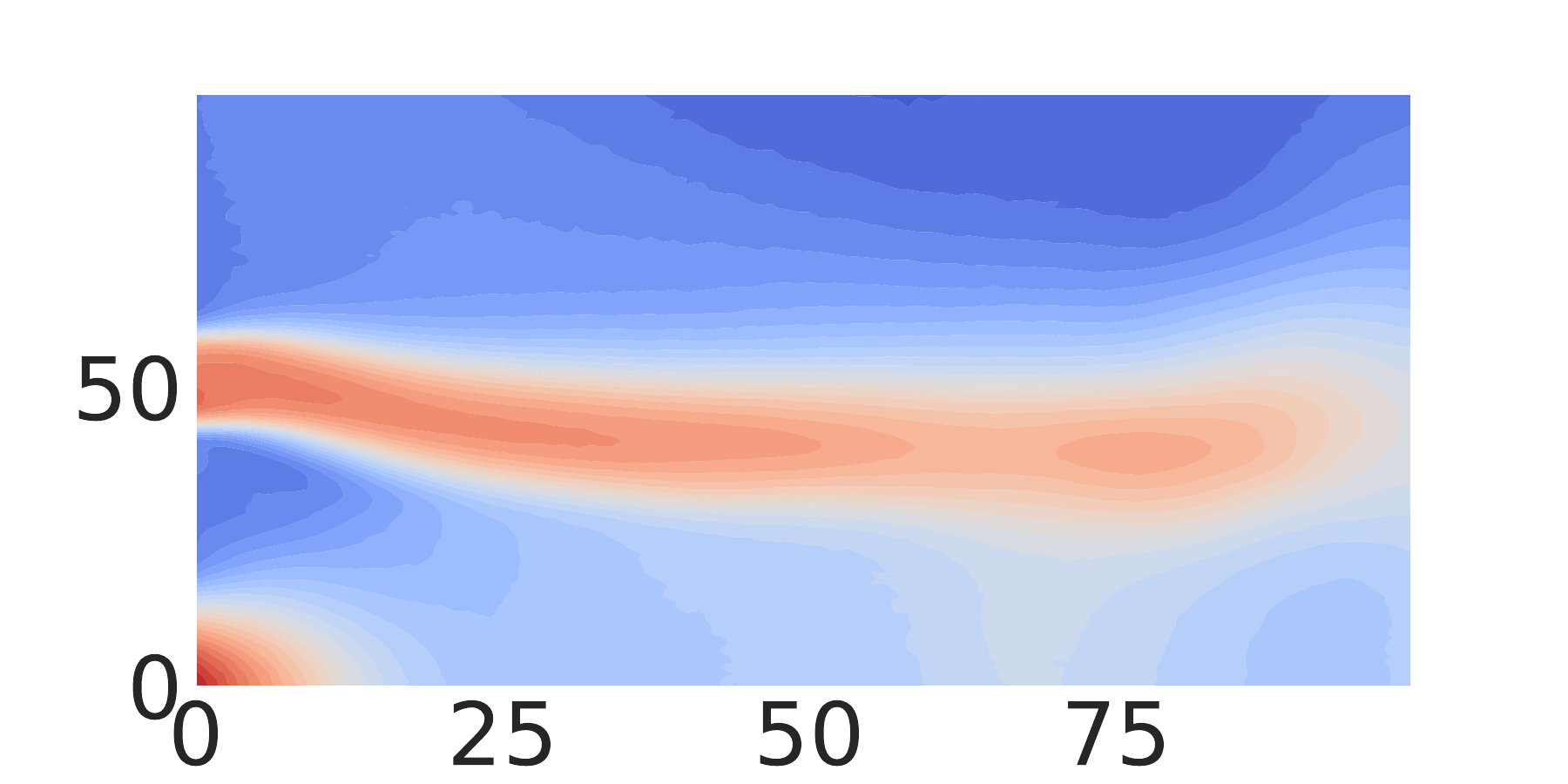}&
		\includegraphics[width=0.27\textwidth, angle=0]{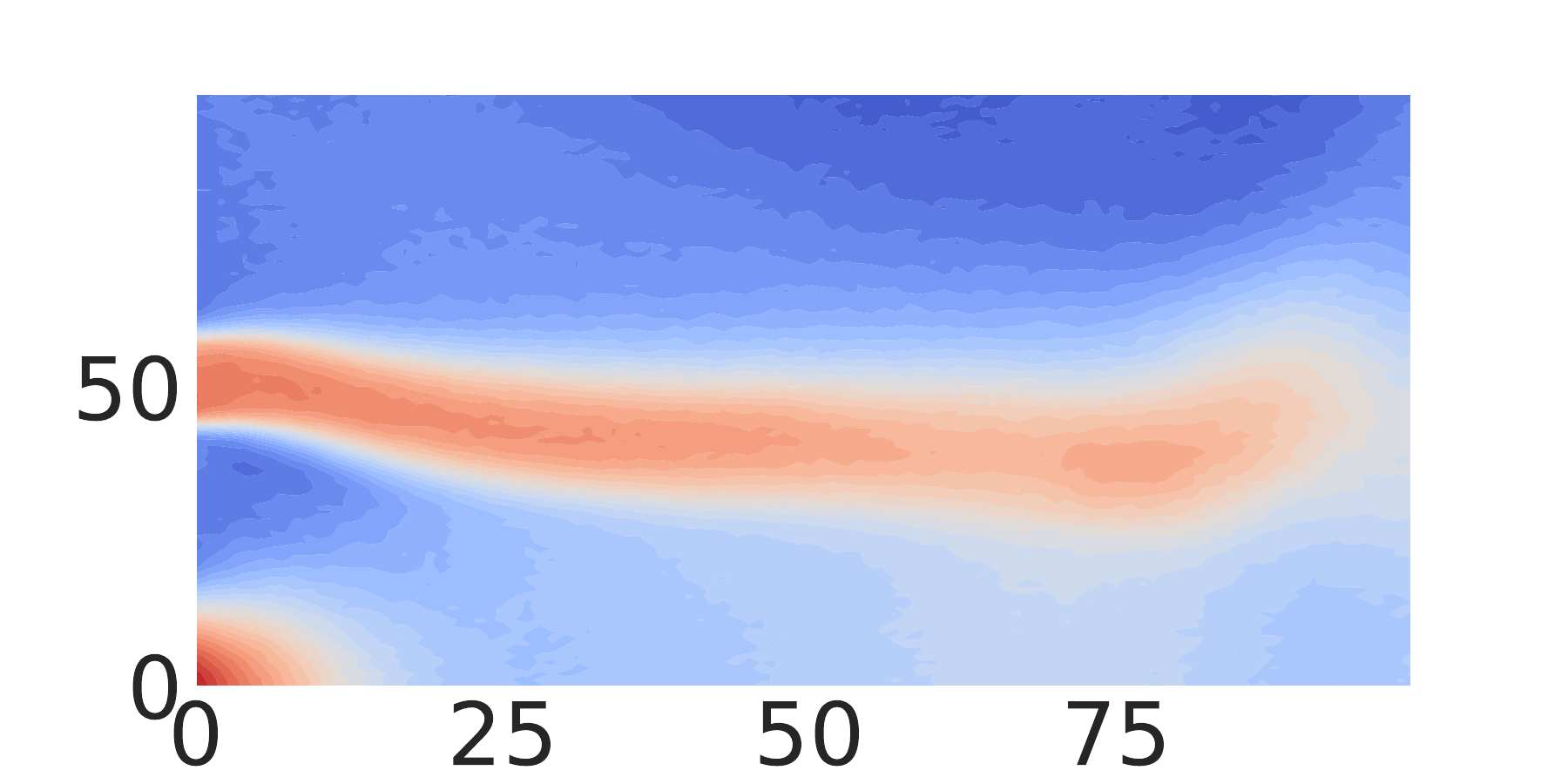}\\
		$t = 174$&
		\includegraphics[width=0.27\textwidth, angle=0]{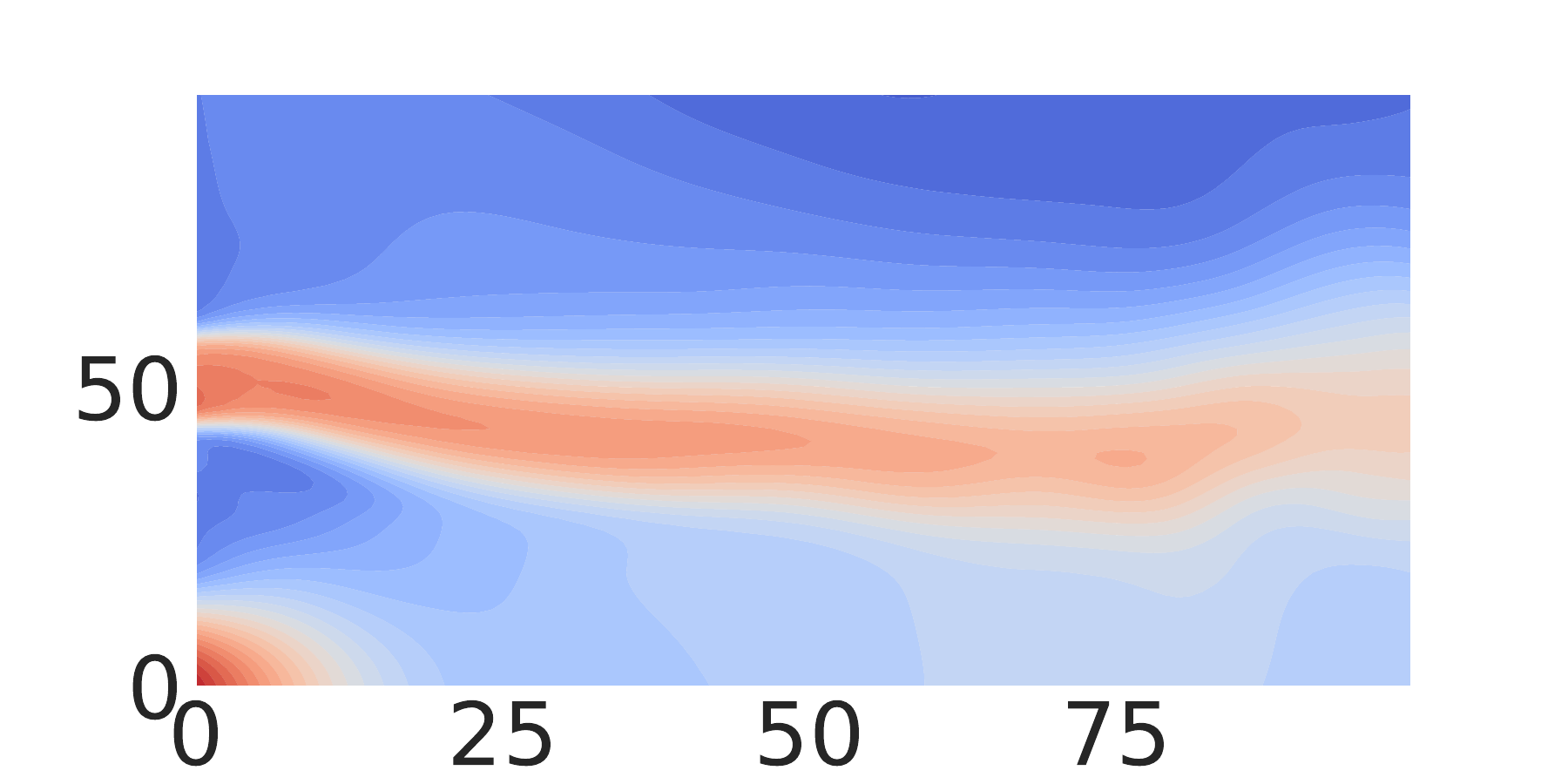}&
		\includegraphics[width=0.27\textwidth, angle=0]{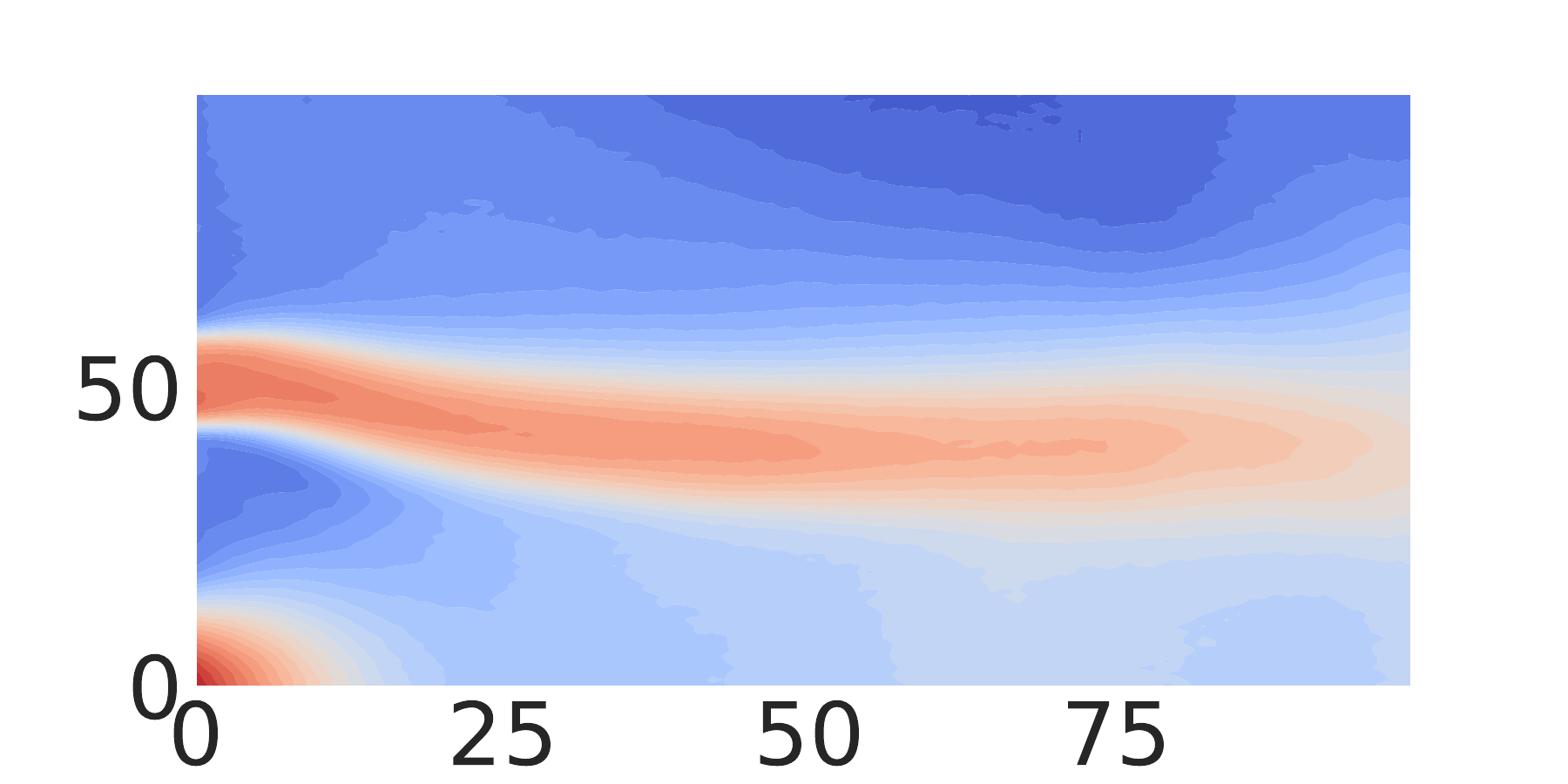}&
		\includegraphics[width=0.27\textwidth, angle=0]{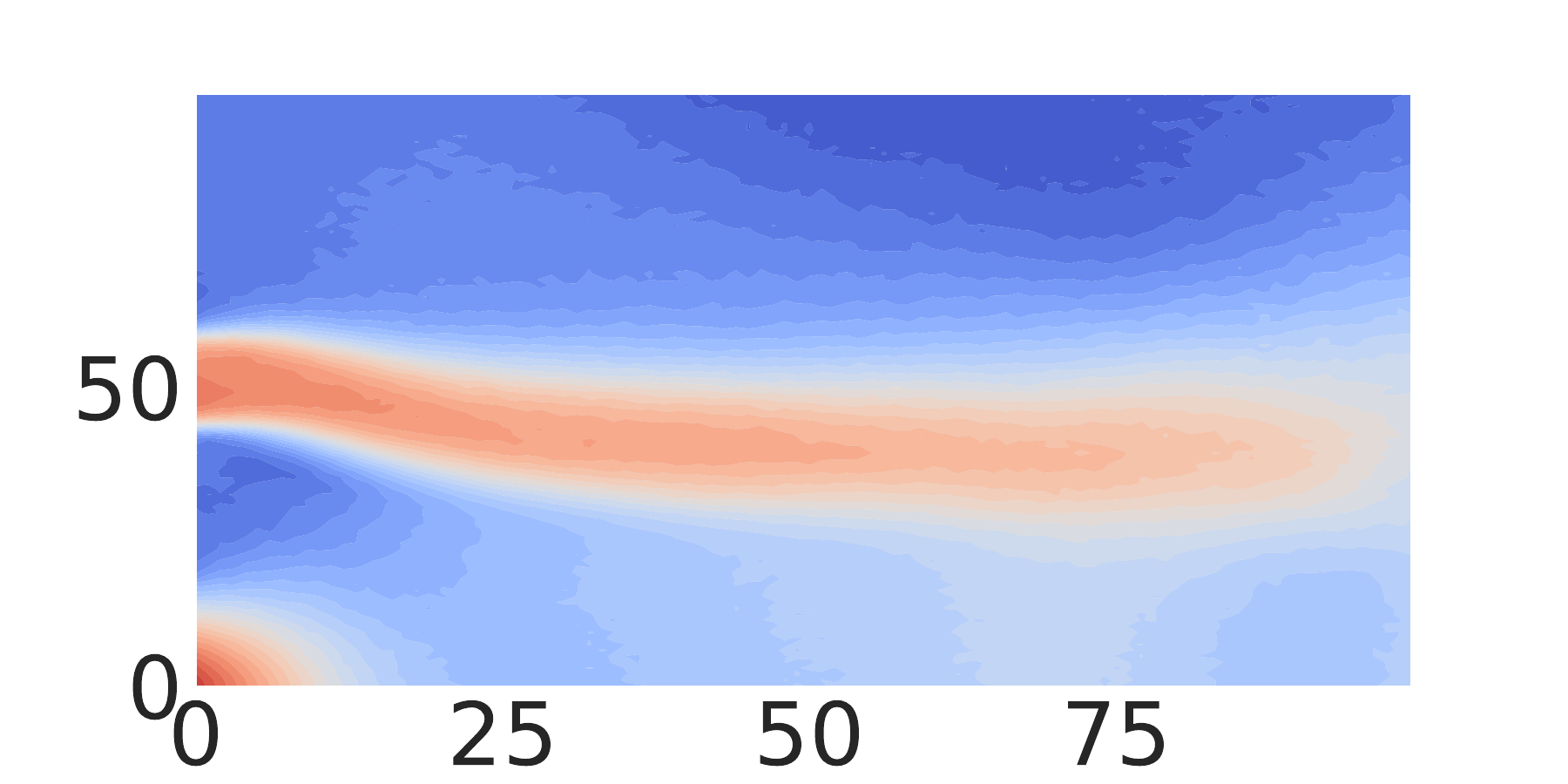}\\
		$t=175$&
		\includegraphics[width=0.27\textwidth, angle=0]{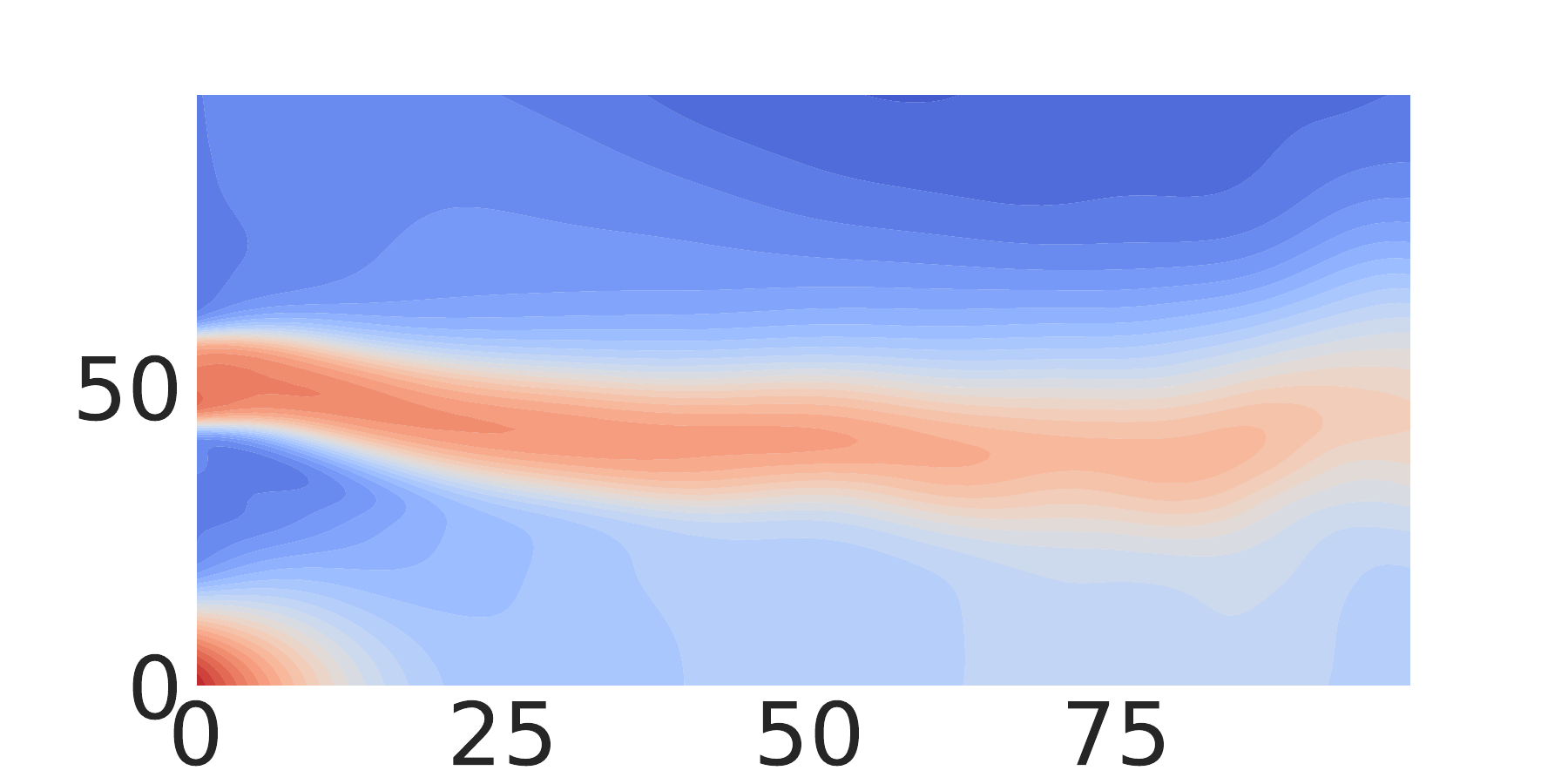}&
		\includegraphics[width=0.27\textwidth, angle=0]{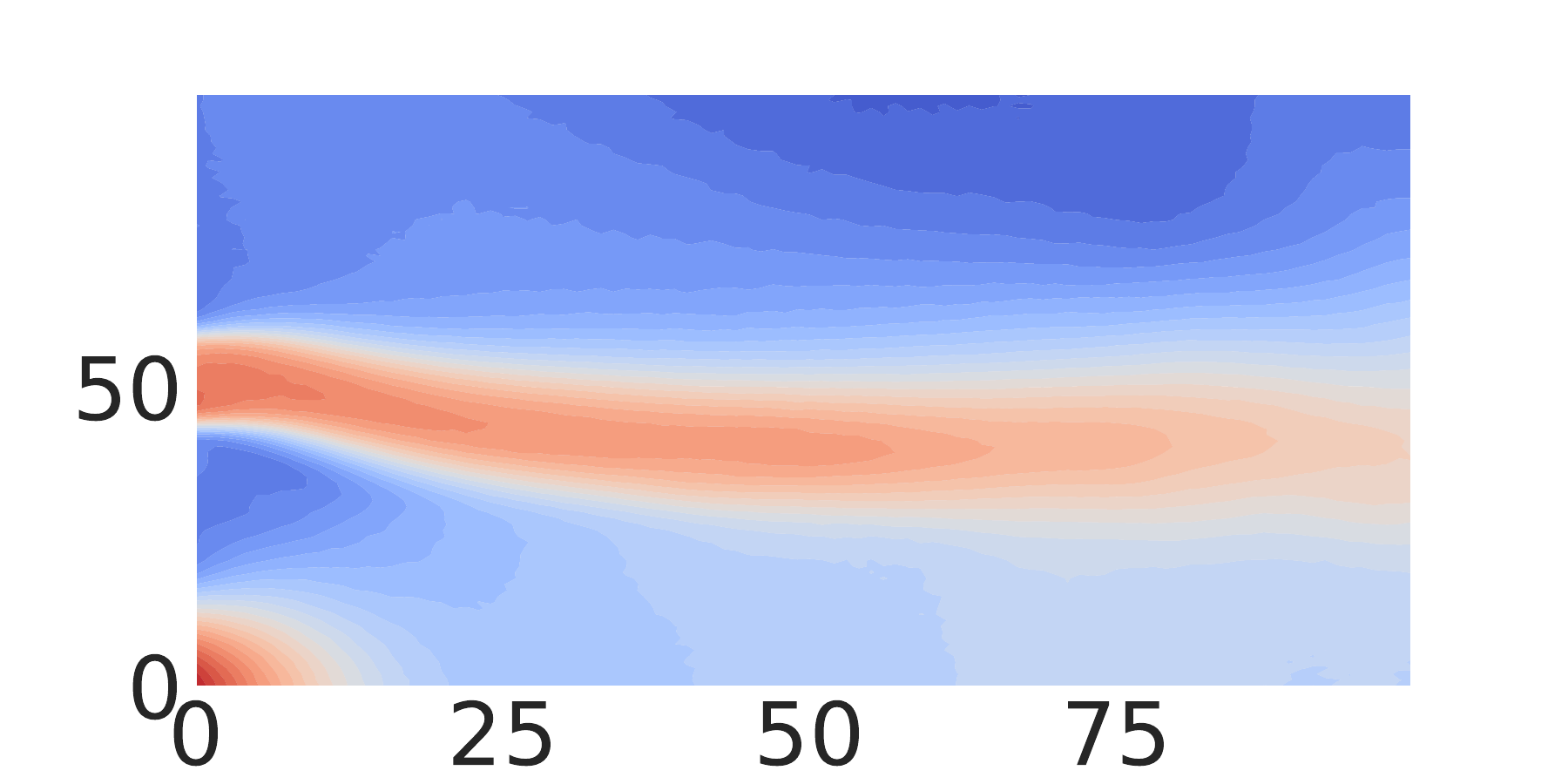}&
		\includegraphics[width=0.27\textwidth, angle=0]{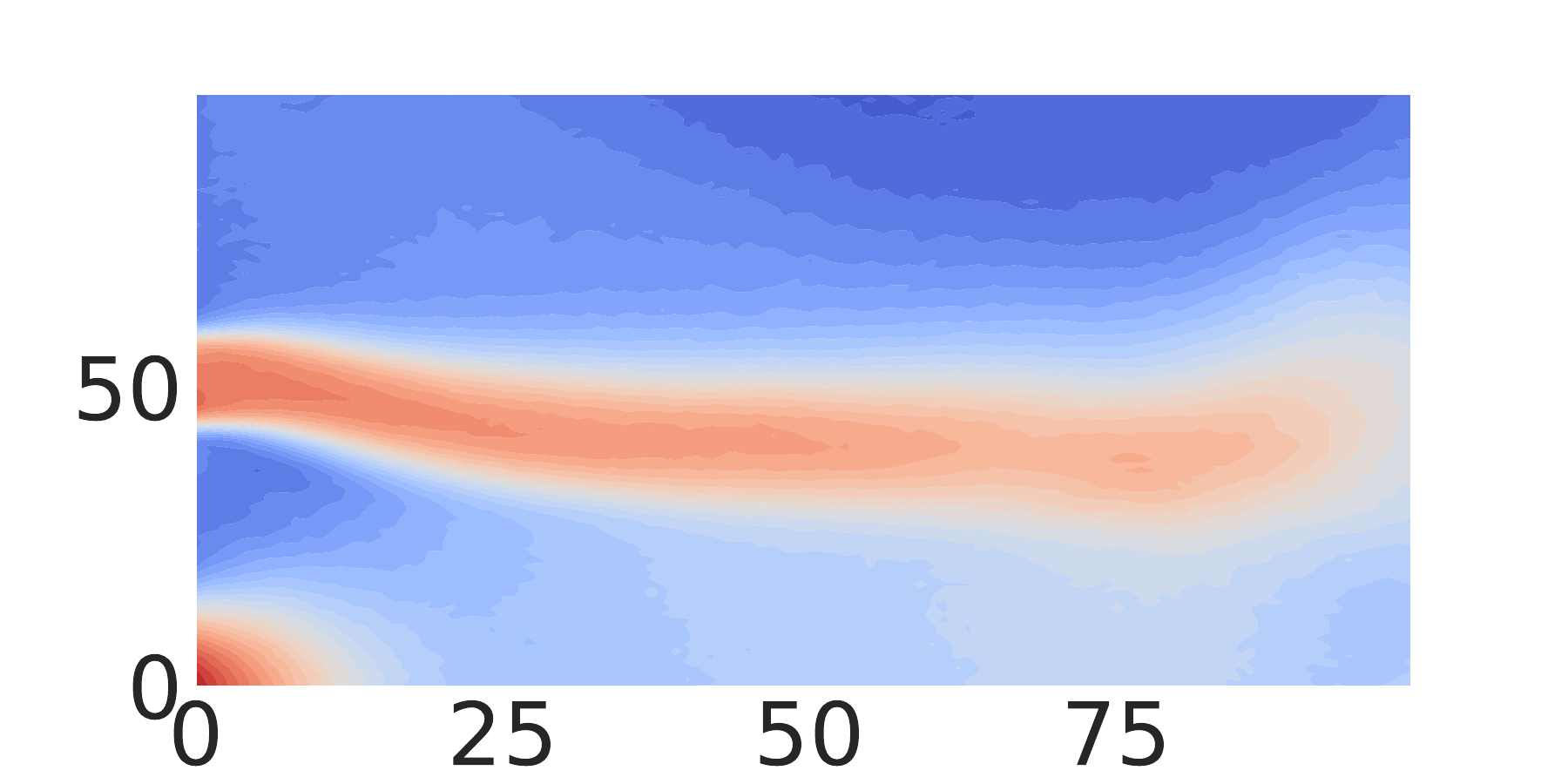}\\
	\end{tabular}
	\caption{From left to right: Snapshots from reconstructed data set (through HODMD), prediction of RNN model and prediction of CNN model, respectively, corresponding to case M2$_{ROM}$. To generate these predictions the following samples were sent to both NNs; $\{v_{158}, v_{157}, \dots, v_{149}\}$ and $\{v_{173}, v_{172}, \dots, v_{164}\}$. Note that all these samples belong to the test set, Table \ref{tab:ML3}.}
	\label{fig: ann_bluff_sts_rom}
\end{figure}

\begin{figure}[H]
	\centering
	\begin{tabular}{lccc}
		Sample & Simulation & RNN & CNN\\
		$t = 159$&
		\includegraphics[width=0.27\textwidth, angle=0]{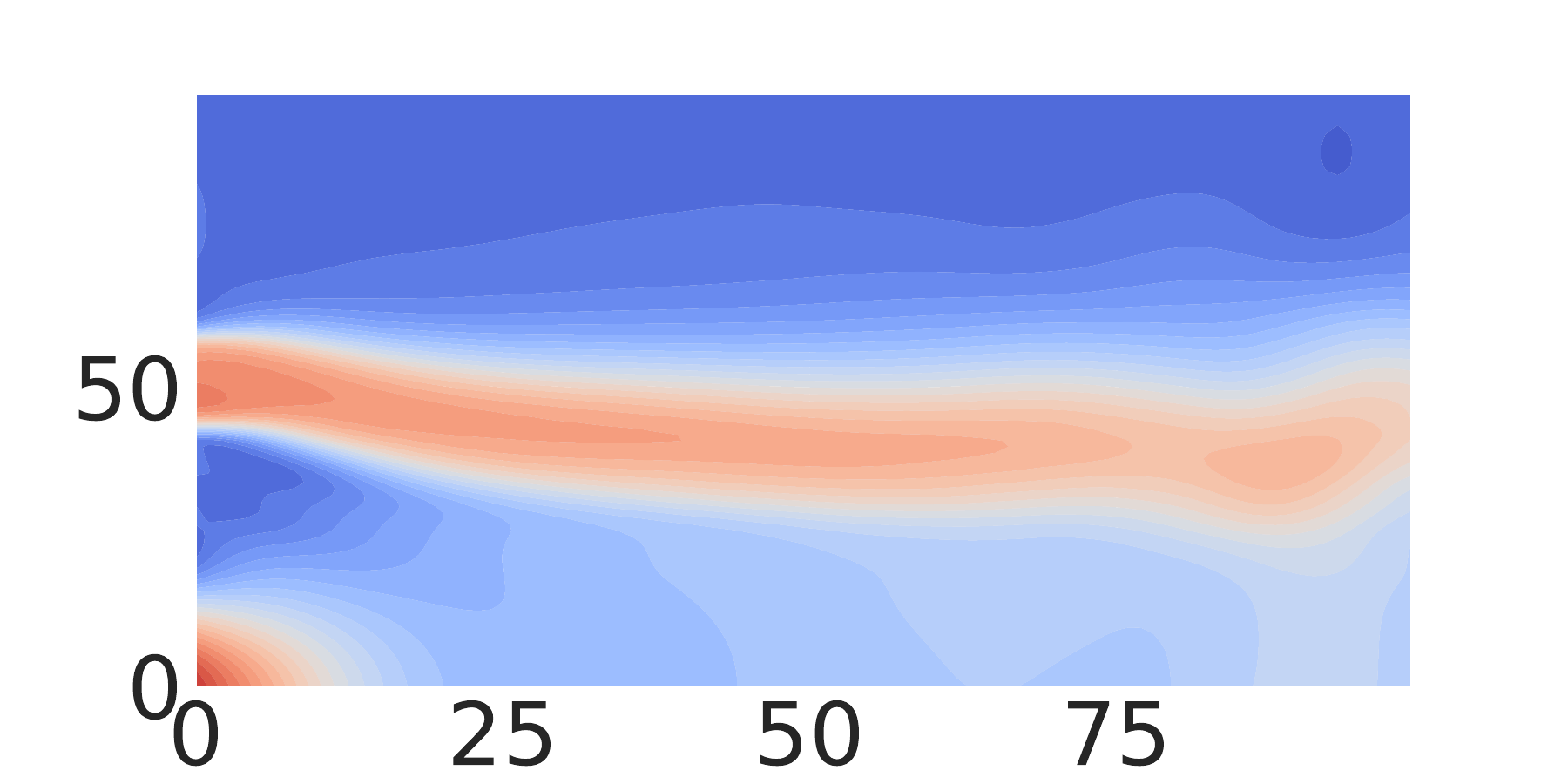}&
		\includegraphics[width=0.27\textwidth, angle=0]{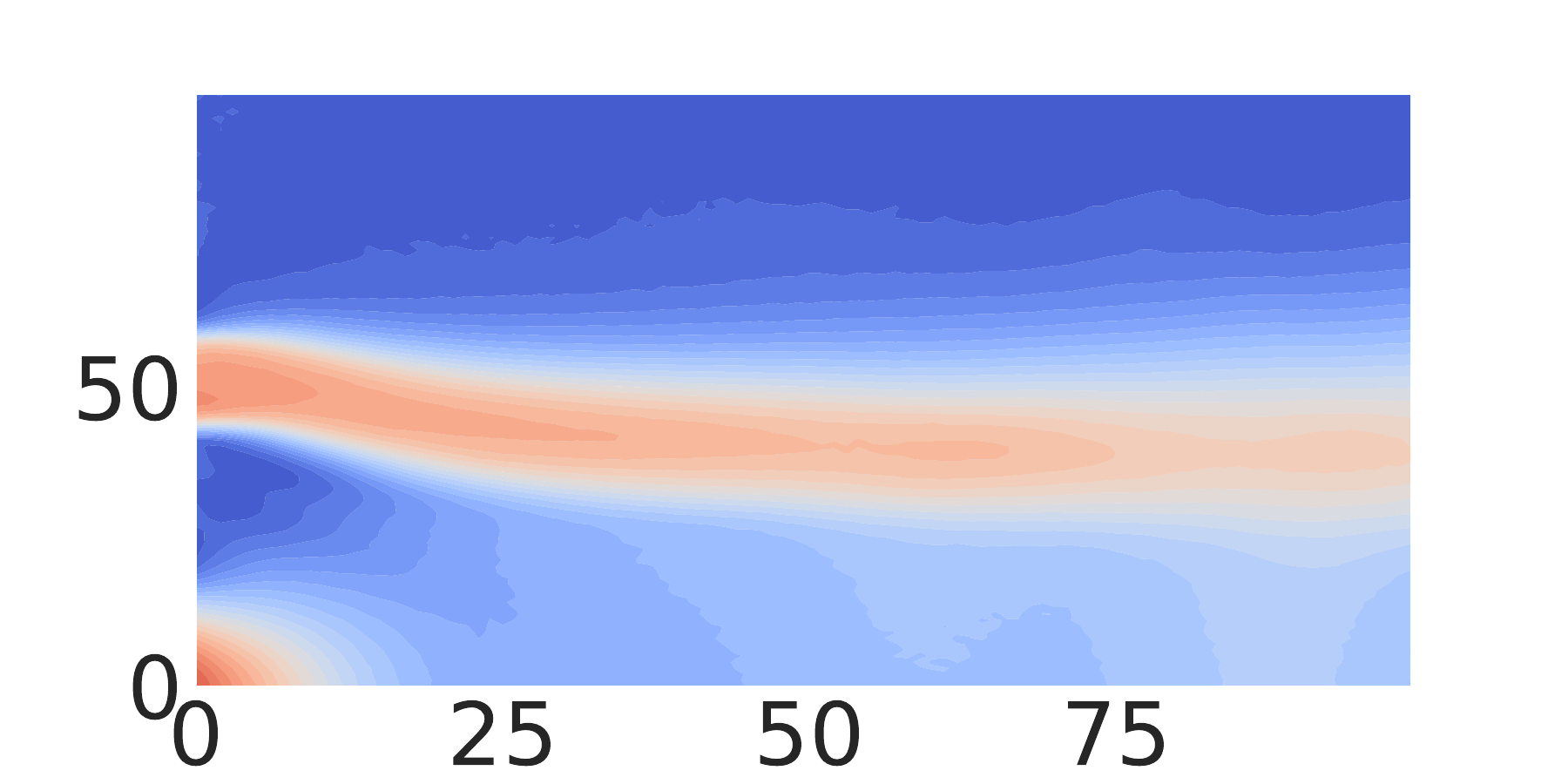}&
		\includegraphics[width=0.27\textwidth, angle=0]{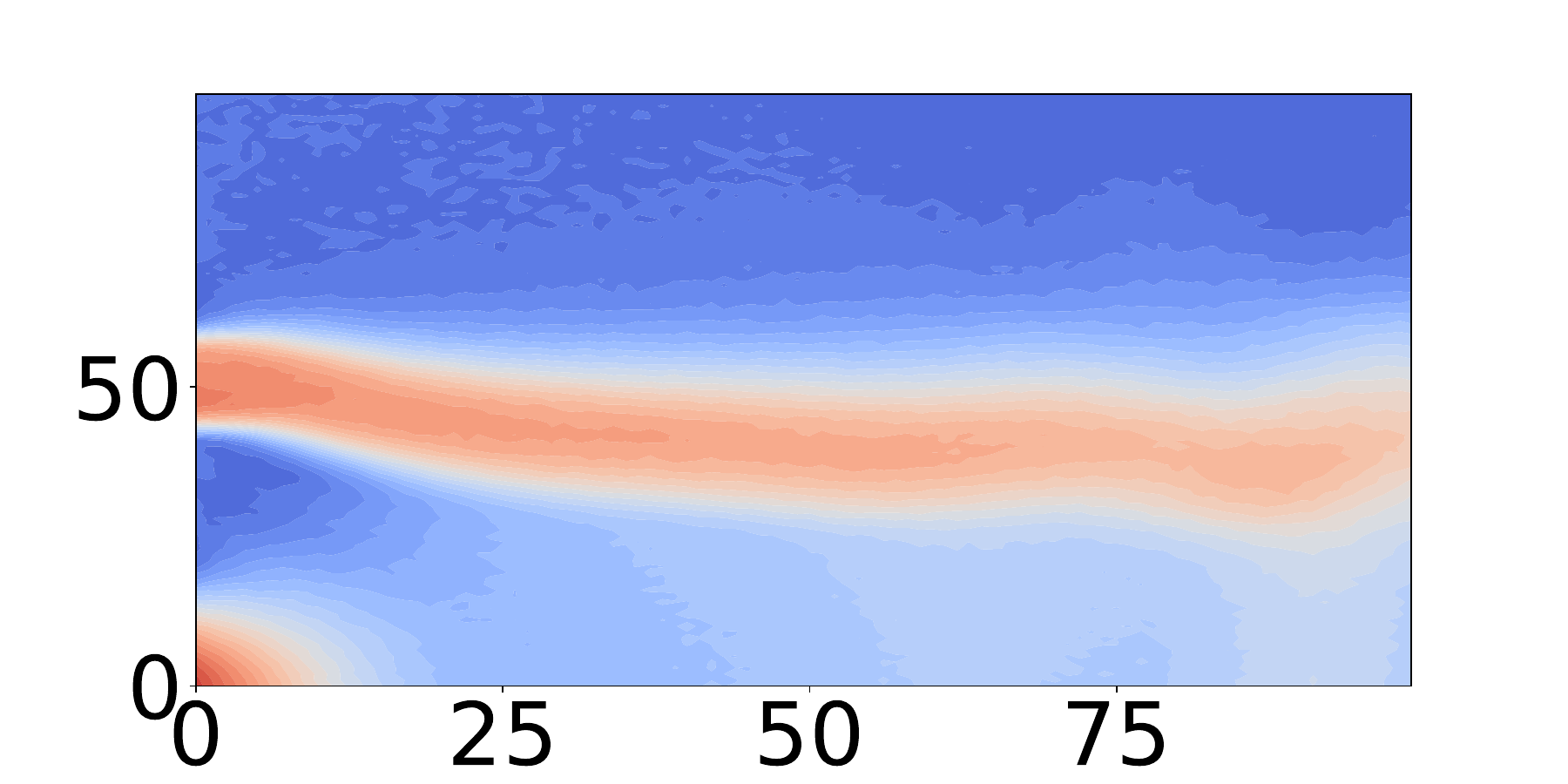}\\
		$t=160$&
		\includegraphics[width=0.27\textwidth, angle=0]{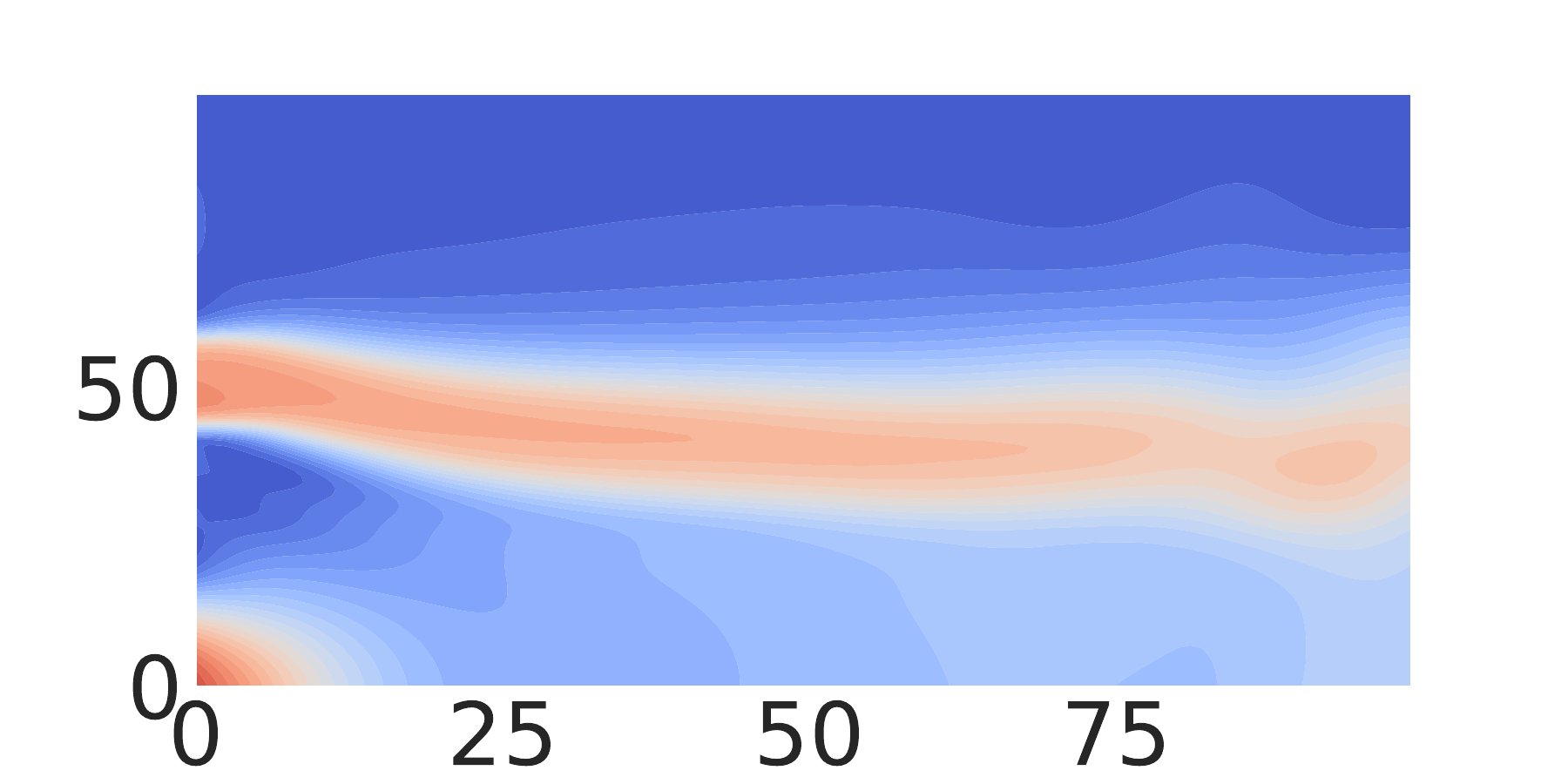}&
		\includegraphics[width=0.27\textwidth, angle=0]{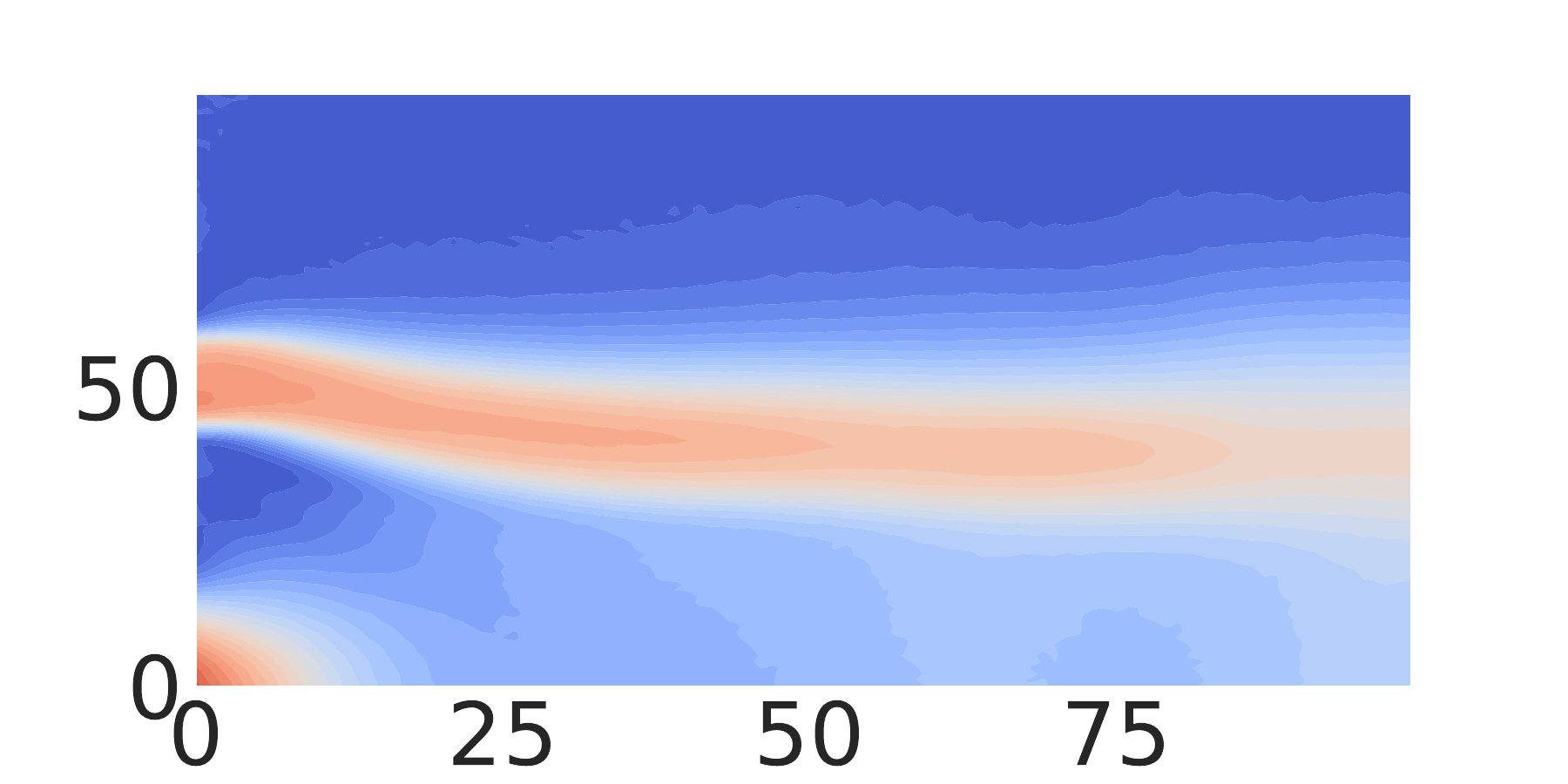}&
		\includegraphics[width=0.27\textwidth, angle=0]{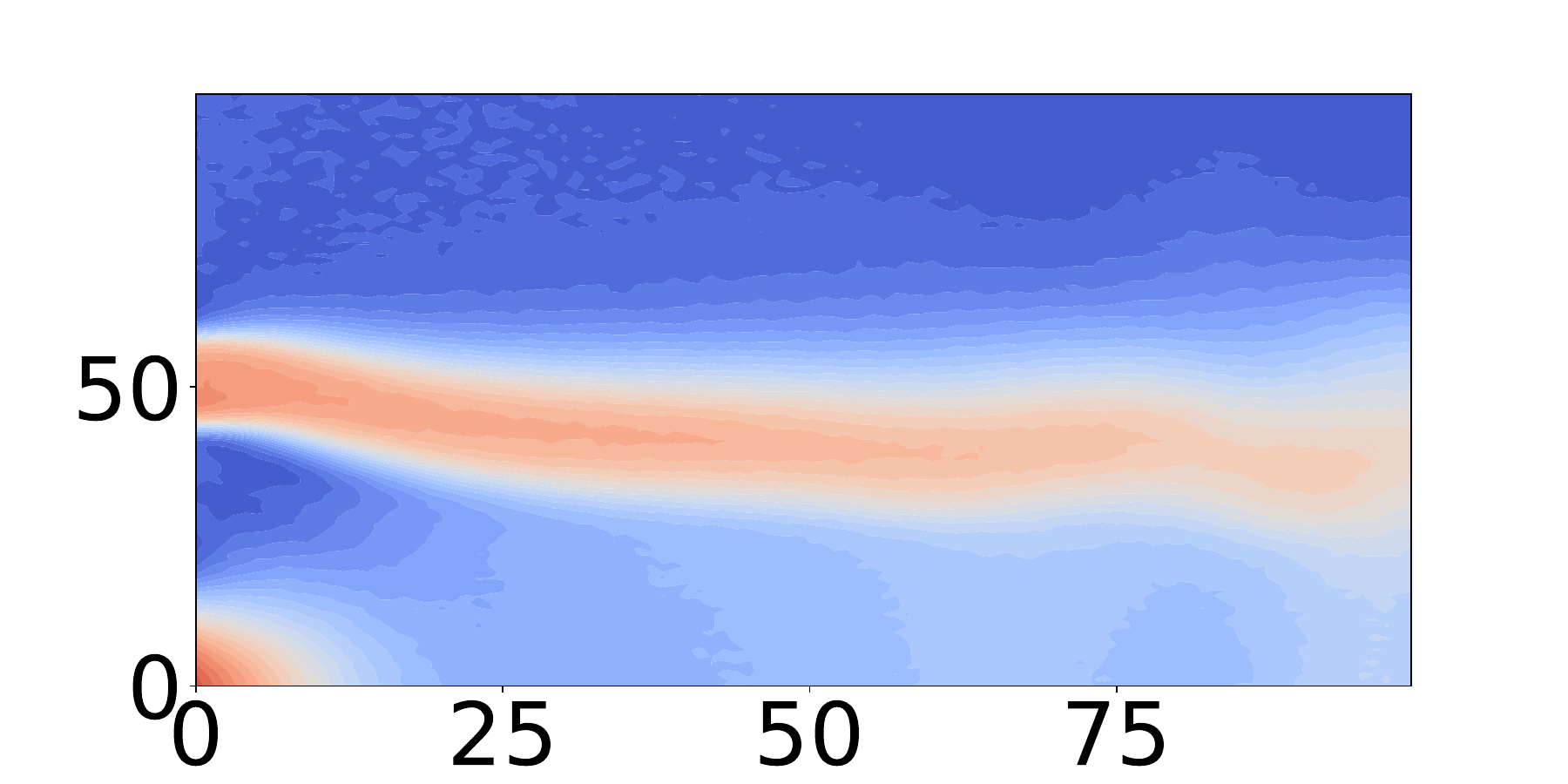}\\
		$t = 174$&
		\includegraphics[width=0.27\textwidth, angle=0]{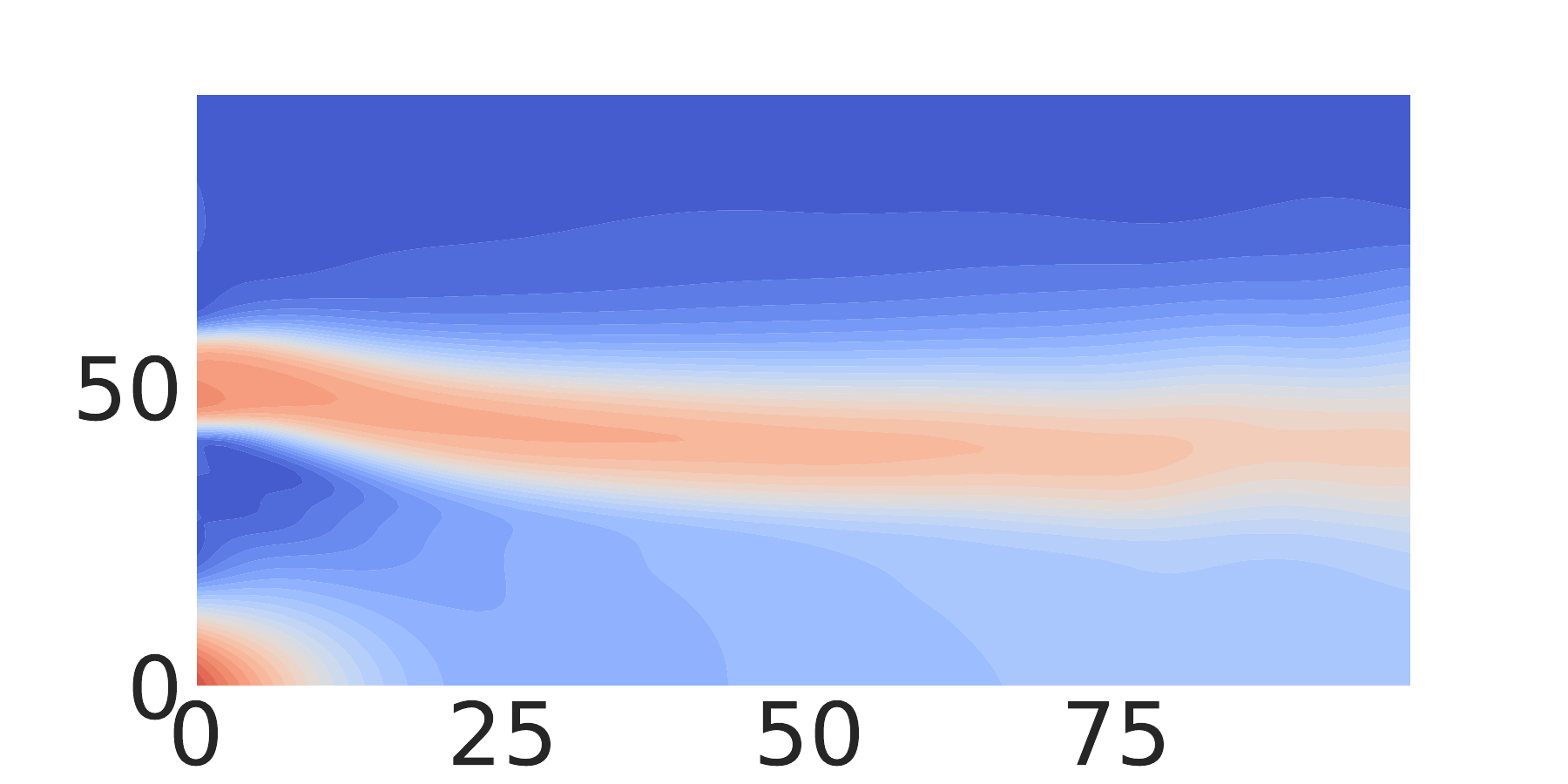}&
		\includegraphics[width=0.27\textwidth, angle=0]{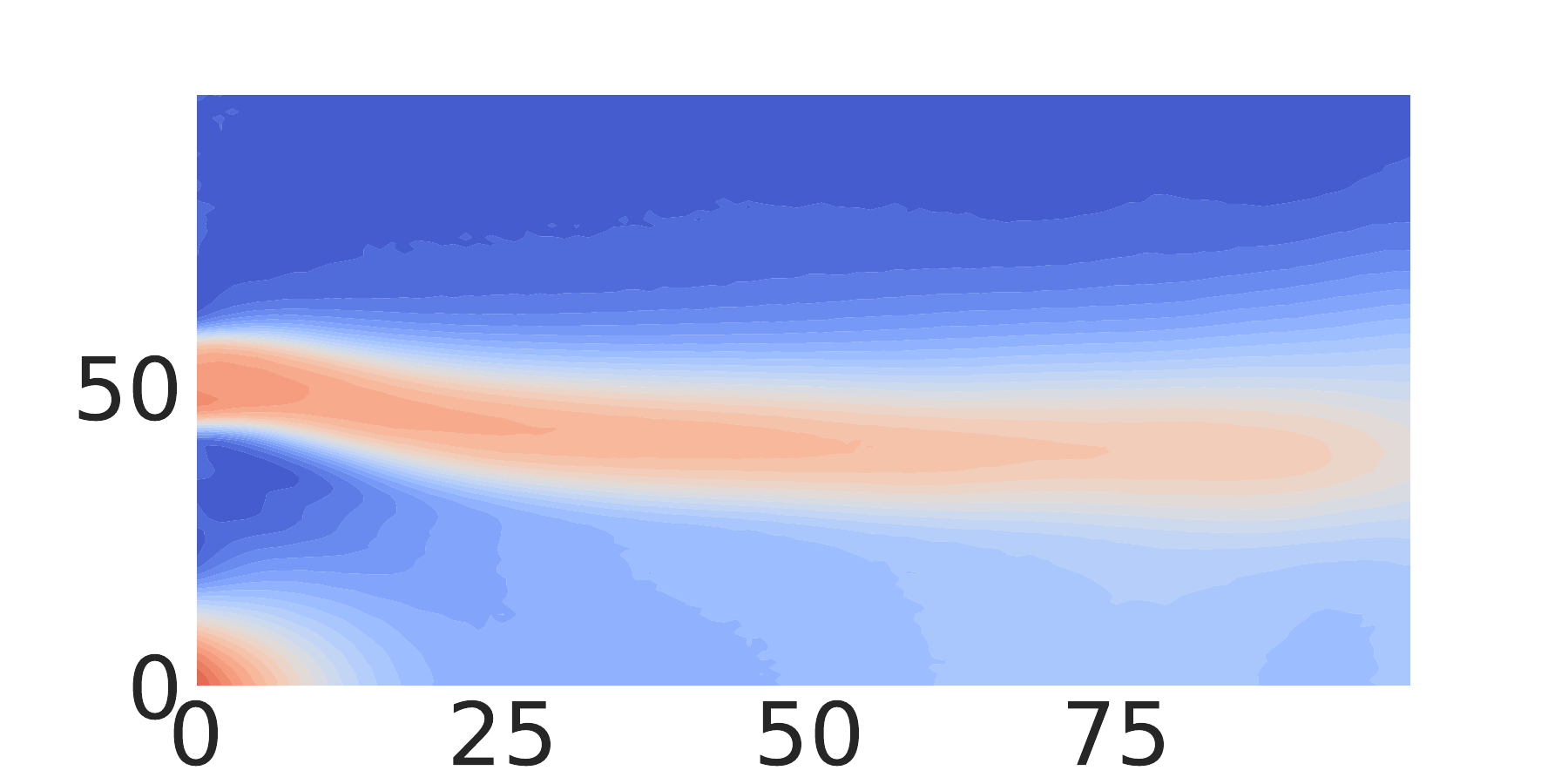}&
		\includegraphics[width=0.27\textwidth, angle=0]{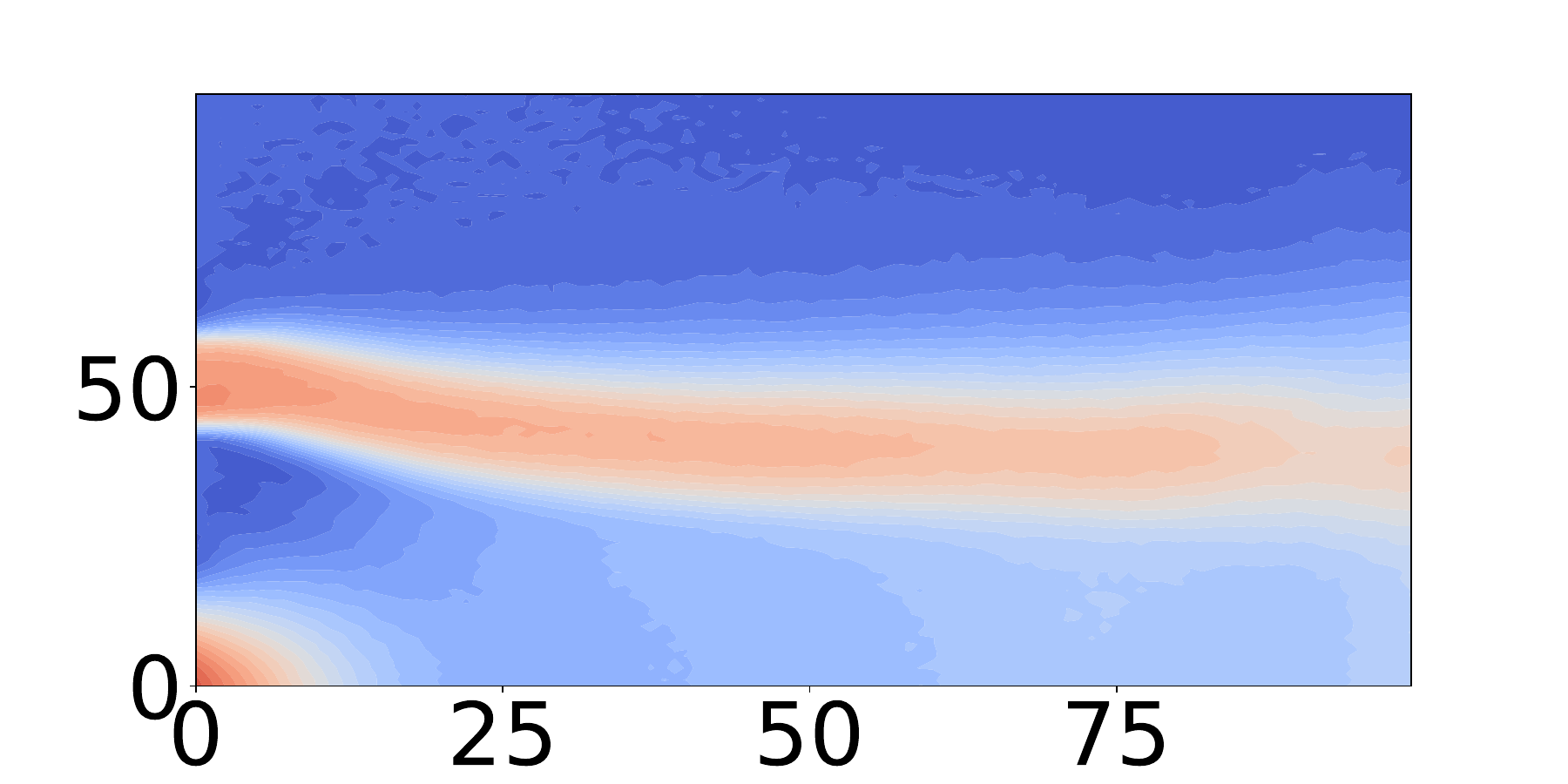}\\
		$t=175$&
		\includegraphics[width=0.27\textwidth, angle=0]{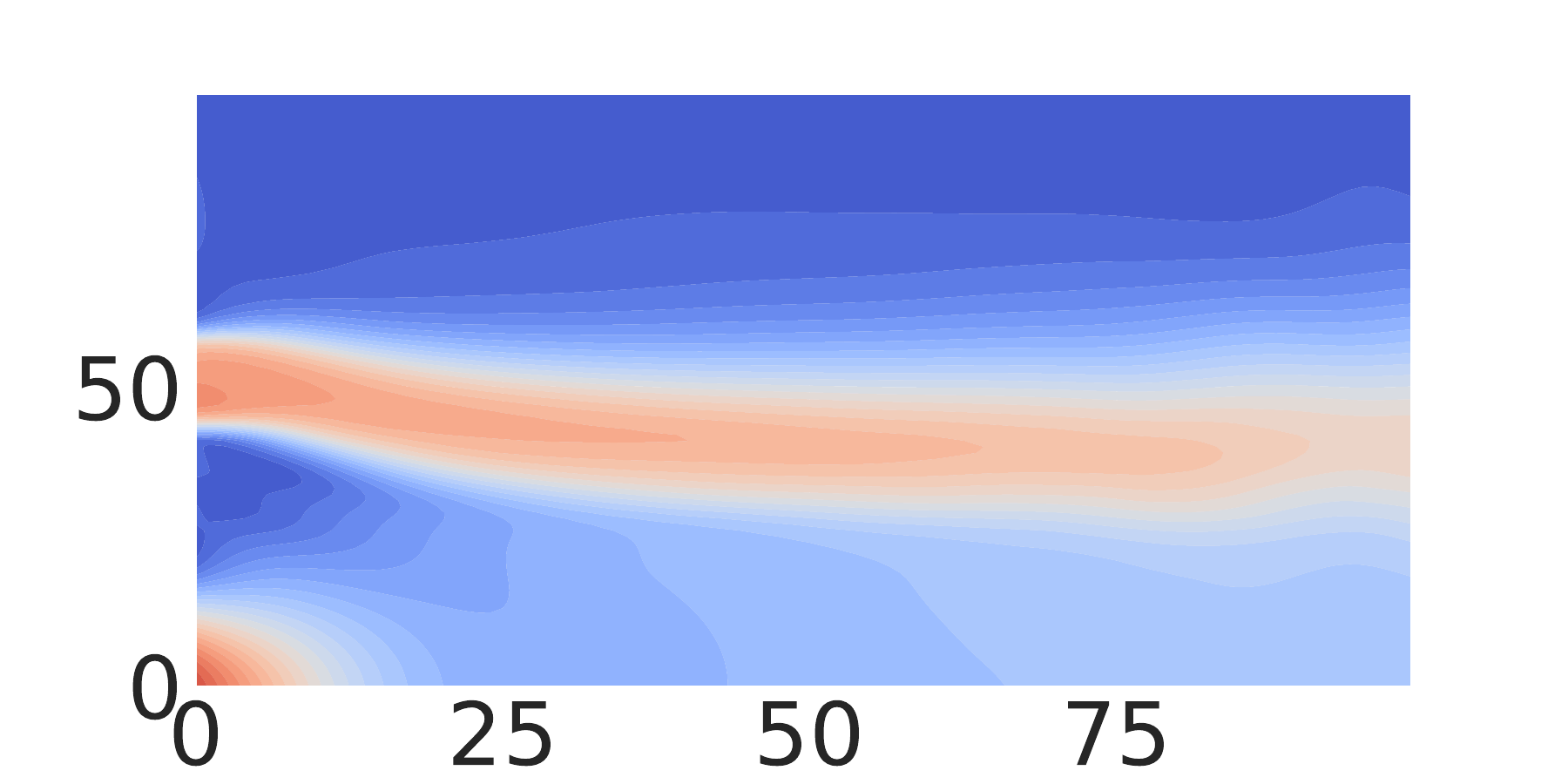}&
		\includegraphics[width=0.27\textwidth, angle=0]{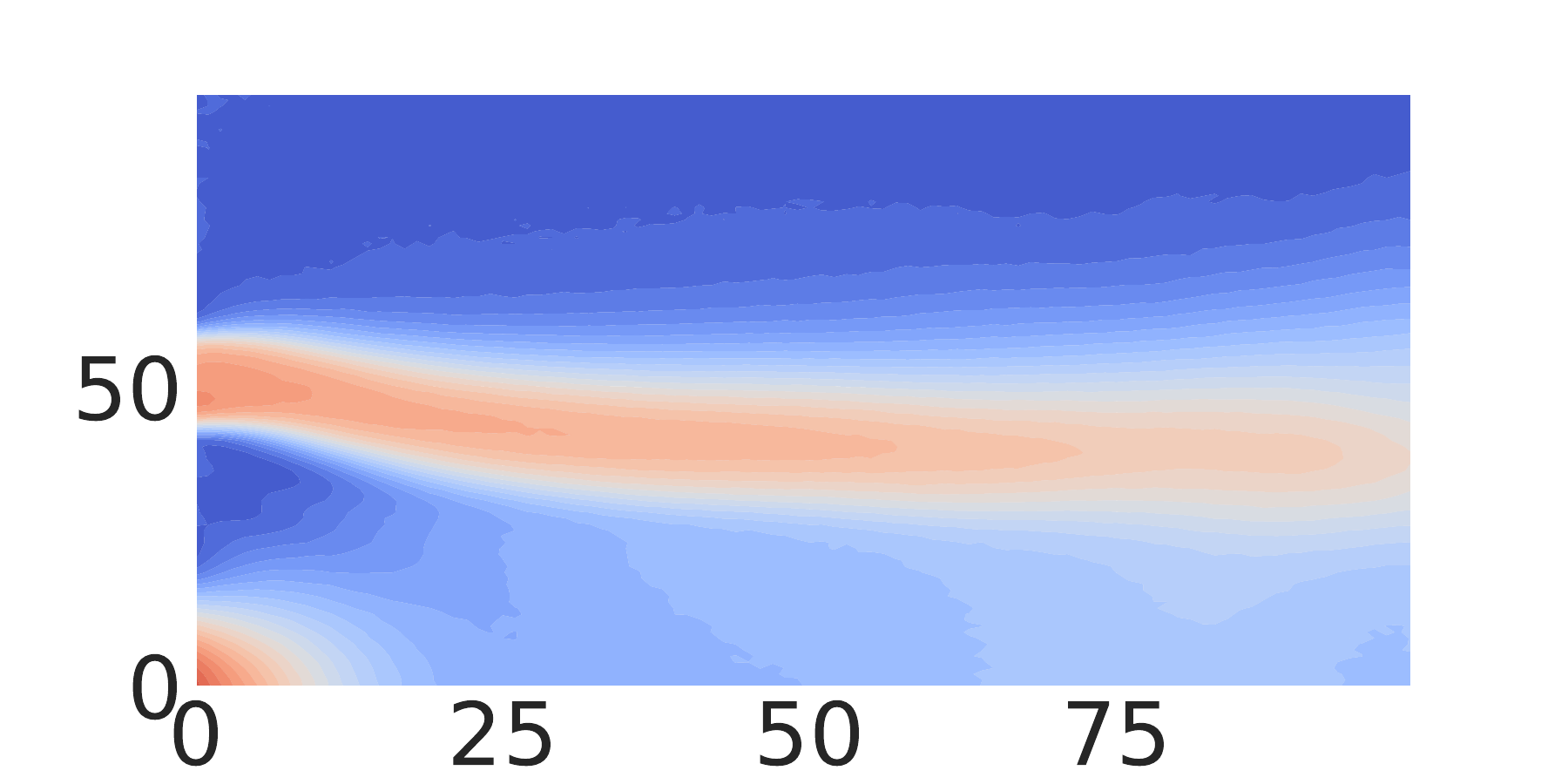}&
		\includegraphics[width=0.27\textwidth, angle=0]{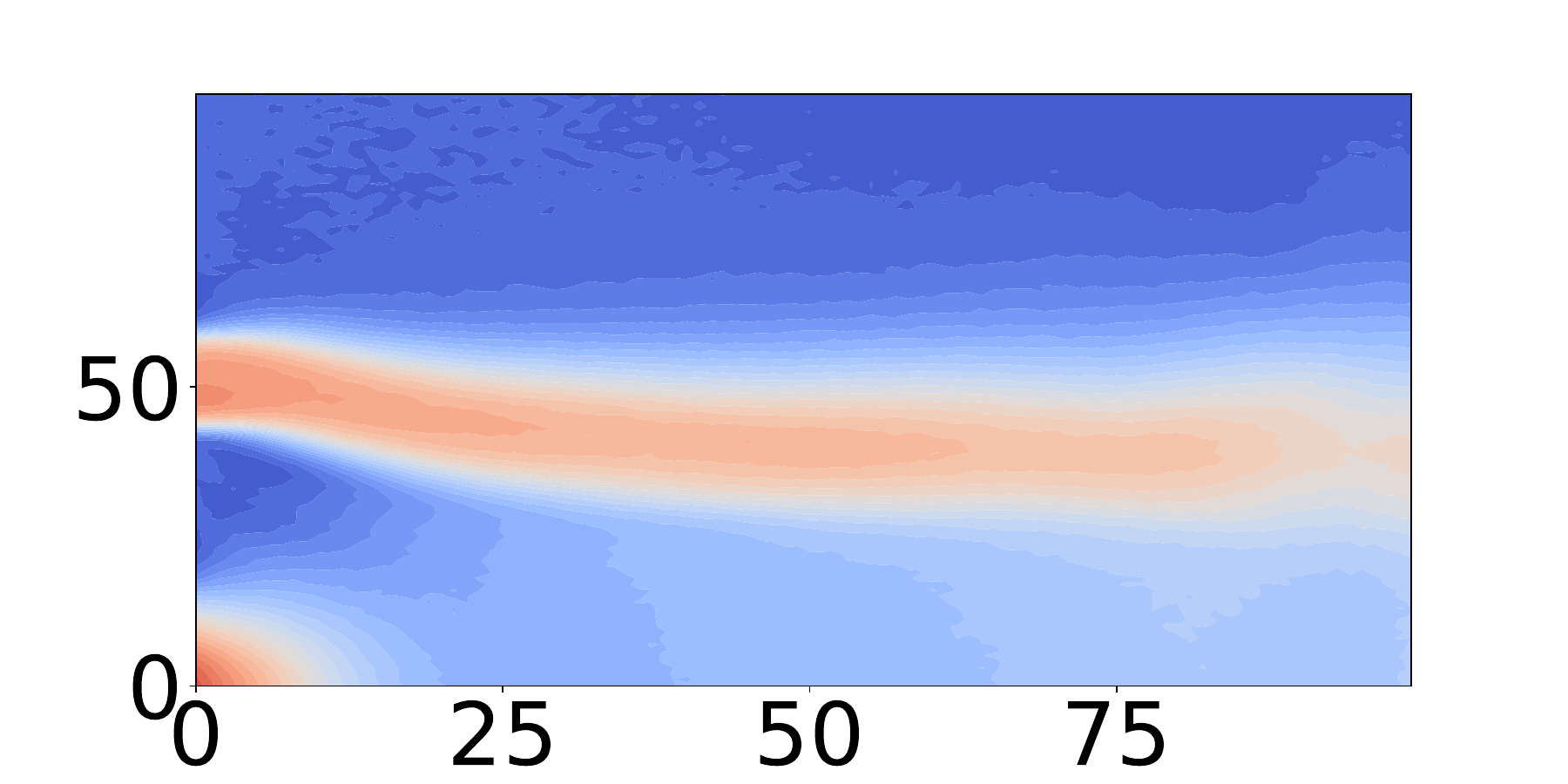}\\
	\end{tabular}
	\caption{Same as Figure \ref{fig: ann_bluff_sts_rom} for case M3$_{ROM}$.}
	\vspace{0.5 cm}
	\label{fig: ann_bluff_cts_rom}
\end{figure}

Again, the RRMSE is used to compute the prediction error performed by NN predictions. The corresponding errors for cases M2, M3, M2$_{ROM}$ and M3$_{ROM}$ are listed in Table \ref{tab: rrmse_m2_m3}.

\begin{table}[H]
	\centering
	\begin{tabular}{|c|c|c|c|c|}
		\hline
		& M2 & M2$_{ROM}$ & M3 & M3$_{ROM}$ \\ \hline
		RNN & $0.186$ & $0.085 \;\, \& \;\, 0.155$ & $0.115$ & $0.058 \;\, \& \;\, 0.091 $ \\ \hline
		CNN  & $0.061$ & $0.035 \;\, \& \;\, 0.155 $ & $0.034$ & $0.016 \;\, \& \;\, 0.091 $ \\ \hline
	\end{tabular}
	\caption{RRMSE from the predictions obtained using the RNN and CNN models, for cases M2, M2$_{ROM}$, M3 and M3$_{ROM}$. The sub-index $ROM$ represents reconstructed data sets through HODMD. Note that in both cases M2$_{ROM}$ and M3$_{ROM}$, we indicate the actual prediction error as well as the reconstruction error, listed in Table \ref{RRMSE reconstruccion}, also measured with RRMSE, where $0.155$ is the reconstruction error for case M2$_{ROM}$ and $0.091$ for case M3$_{ROM}$.}
	\label{tab: rrmse_m2_m3}
\end{table}

\subsection{Computation time of NNs for training and prediction}
The time required for these NN models to be trained is independent of the data set used for training. The same happened with the time required to compute predictions. These training and prediction times, for either RNN and CNN models, are listed in Table \ref{tab: training_testing_time}. Note that to generate a prediction both models need almost the same time. However, to train these models; CNN model needs less than one third of what RNN requires. This is mainly because of the training length of CNN (70 epochs) is half that of the RNN (140 epochs). And that the CNN model ($1,690,027$) has less trainable parameters than the RNN ($18,349,880$).

\begin{table}[H]
	\centering
	\begin{tabular}{|c|c|c|}
		\hline
		& Training & Prediction \\ \hline
		RNN  & 3 min. & 9 sec. \\ \hline
		CNN & 0.84 min. & 8 sec. \\ \hline
	\end{tabular}
	\caption{Computation time required to train deep learning models (in minutes) and to generate the prediction (in seconds) of the two future samples. Note that the models are trained only once. Thereafter, to generate new predictions it is only required to input the past $10$ samples in order to predict the $2$ following. It is not necessary to retrain the model each time we want to predict the two following samples.}
	\label{tab: training_testing_time}
\end{table}

\section{Conclusions}\label{sec: conclusions}
The development of a real-time prediction model to improve fuel chamber injectors' performance in turbofan engines, is a complex task very difficult to achieve with classical methods: using experimental data or numerical simulations. However, this work shows that forecasting models based on deep learning can be a very promising tool for achieving such a task. Here we have presented two deep learning models; RNN and CNN, where as shown in table \ref{tab: training_testing_time} the computational cost required is extremely low. The architecture of both models is the same regardless of the flow dynamics we are predicting, showing the generalization capabilities of the models presented, which have been tested suitable to predict the flow evolution from numerical databases, reducing the computational cost of the numerical simulations. We have also shown how modal decomposition techniques such as HODMD can be used to identify the main structures in a flow and use them to reconstruct the original database with a less complex dynamics, but at the same time maintaining its main attributes. This reduction of complexity in the data set also reduces the NN training complexity, as shown in Tables \ref{tab: rrmse_s2_s3} and \ref{tab: rrmse_m2_m3}. Further note from the results how the prediction error when using the CNN model is smaller than when using the RNN architecture. We believe the data set format could be one reason to justify this result. Recall, data sets used in this work are snapshots taken from direct numerical simulations of a two-phase flow, where each snapshot has a size $(100 \times 200 \times 1)$. Given that convolutional architectures were designed to deal with snapshots, we can pass them directly into the CNN model. However, the RNN model is mainly composed by an LSTM layer, which was not designed to deal with snapshots, but with vectors. Therefore, we need to flatten the original snapshots into vectors, before transferring them to the model. This reshaping could cause a loss of information in the spatial dimension and consequently the RNN model will not be able to achieve such a low prediction error as the CNN model does. The biggest advantage of our proposed models is their capability to generate two future predictions in each flow, with a low prediction error, without varying the architecture and using only $10$ previous snapshots.

\section*{Acknowledgements}
The authors would like to thank the support of the Comunidad de Madrid through the call Research Grants for Young Investigators from the Universidad Politécnica de Madrid and also acknowledge the grant PID2020-114173RB-I00 funded by the Spanish Ministry MCIN/AEI/10.13039/501100011033.

\bibliographystyle{elsarticle-harv} 
\bibliography{referencesClean}





\end{document}